\documentclass[a4paper,12pt,openany]{report}
\usepackage[inner=40mm,outer=20mm,top=30mm,bottom=20mm]{geometry}
\usepackage{epsfig}
\usepackage{subfigure}
\usepackage{amssymb}
\usepackage{amsmath}
\usepackage{multirow}
\newtheorem{theorem}{Theorem}[section]
\newtheorem{remark}{Remark}[section]
\newtheorem{assumption}{Assumption}[section]
\newtheorem{definition}{Definition}[section]
\newtheorem{notation}{Notation}[section]

\newcommand{\vect}[1]{\vec{\boldsymbol{#1}}}
\newcommand{\robots}{\boldsymbol{s}}

\title{\Large Sensor Network Based Collision-Free Navigation and Map Building for Mobile Robots}
\author{\bf \Large Hang Li}
\date{}

\begin{document}
\maketitle
\pagenumbering{roman}
\setcounter{page}{2}

\chapter*{Abstract}
\addcontentsline{toc}{chapter}{Abstract}
Safe robot navigation is a fundamental research field for autonomous robots including ground mobile robots and flying robots. The primary objective of a safe robot navigation algorithm is to guide an autonomous robot from its initial position to a target or along a desired path with obstacle avoidance. With the development of information technology and sensor technology, the implementations combining robotics with sensor network are focused on in the recent researches. One of the relevant implementations is the sensor network based robot navigation. Moreover, another important navigation problem of robotics is safe area search and map building.

In this report, a global collision-free path planning algorithm for ground mobile robots in dynamic environments is presented firstly. Considering the advantages of sensor network, the presented path planning algorithm is developed to a sensor network based navigation algorithm for ground mobile robots. The 2D range finder sensor network is used in the presented method to detect static and dynamic obstacles. The sensor network can guide each ground mobile robot in the detected safe area to the target. The computer simulations and experiments confirm the performance of the presented method. Furthermore, considering the implementations of small-sized flying robots in industry, the presented navigation algorithm is extended into 3D environments. In the presented method, a time-of-flight camera network is used to detect the static and moving obstacles. With the measurements of the sensor network, any flying robot in the workspace is navigated by the presented algorithm from the initial position to the target and avoids any obstacles in the workspace.

Moreover, in this report, another navigation problem, safe area search and map building for ground mobile robot, is studied and two algorithms are presented. In the first presented method, we consider a ground mobile robot equipped with a 2D range finder sensor searching a bounded 2D area without any collision and building a complete 2D map of the area. Furthermore, the first presented map building algorithm is extended to another algorithm for 3D map building.

\tableofcontents

\listoffigures
\addcontentsline{toc}{chapter}{\listfigurename}

\listoftables
\addcontentsline{toc}{chapter}{\listtablename}

\pagenumbering{arabic}
%%%%%%%%%%%%%%%%%%%%%%%%%%%%%%%%%%%%%%%%%%%%%%%%%%%%%%%%
\chapter{Introduction}
\chaptermark{Introduction}
Collision-free navigation is a fundamental problem of robotics for mobile robots. The collision-free navigation of a mobile robot is defined as the process of guiding a mobile robot to a target or along a desired path with obstacle avoidance. With the development of mathematics, electronics and communications, both the robot's computer performance and the control strategies are being improved dramatically during the previous years.

With the development of mechanical engineering, varieties of mobile robots are designed for different tasks in different environments. According to the environments where the robot works, the mobile robots are generally classified into three categories; i.e. ground mobile robots, underwater robots and flying robots. In this report, we focus on the navigation of ground mobile robots and flying robots.

The models we used in this report for both ground mobile robot and flying robot are non-holonomic models. We consider a ground mobile robot as a differential wheeled robot, which is a Dubin's car with a non-holonomic constraint \cite{WANG2014137,6935304,Rao2017,ABDESSEMED200431}. Comparing with an omnidirectional mobile robot, like \cite{6461945,7383260,7827337}, the non-holonomic ground mobile robot model is more practical. Moreover, we consider a flying mobile robot as an under-actuated non-holonomic mobile robot \cite{Matveev2017,Wang2016a}.

In this chapter, the previous research works in robot navigation and relevant topics are studied and reviewed as follows.

\section{Robot navigation Problems}
Robot navigation for mobile robots is a fundamental problem in robotics; see \cite{Lapierre2007,
Savkin2013,Savkin2014,
Savkin2013a,Teimoori2010,Matveev2011a,
Matveev2012,
Babinec2014,Savkin2016,
Wang2016a,
Hoy2015}. It is an important technology in engineering, military and commercial applications. Generally, the solutions of the robot navigation problem are classified into two classes, local and global navigation algorithms. A local navigation algorithm is to use the local information measured by the robot's sensors (e.g. range finder sensor, sonar sensor and visual sensor) to guide the robot to a target; see \cite{Pudics2015,Tatsukawa2015,Kim2015a}. However, with the development of computer science, microelectronics and sensor network technology, global navigation algorithms have become another significant type of solution in current research works and implementations.

In the local navigation algorithms, the mobile robot obtains the environment information by using the mounted sensors and determines the motion planning to navigate the robot to a target or along a desired path; see \cite{Teimoori2010a,7125610,6878733,6630630,6871831}. In the previous works in local navigation, there are different types of sensors used to obtain the environment information. In the works \cite{HIREMATH201441Laser,
4406749,5707420}, laser range finder is considered because it can provide accurate map of the local area. In other works \cite{Murray2000Using,
5649136,Royer2007Monocular}, camera is used to capture the environment image and generate the map for the local area. It involves image process, which is more complex than laser range finder. Sonar sensor is another common sensor used for small-sized and cheap robot; see \cite{6290366}. In the past few years, there are a lot of local navigation algorithms proposed. In the work \cite{SGORBISSA2012628Planning}, the authors propose a hybrid approach to finish the tasks by combining the environment information with local perceptions. In the work \cite{5565069}, a path planning algorithm is proposed based on the potential field, which is widely used in many other works, such as \cite{6027049,6451251,6058518}. In the work \cite{Matveev2011}, a sliding mode controller is proposed by using only distance from the robot to a moving target to drive the robot to a predefined distance to the target. In the work \cite{MATVEEV2013312problem}, the boundary following problem for a unicycle-like robot is addressed and a sliding mode control law is proposed to drive a unicycle-like mobile robot at a predefined distance from an obstacle's boundary while moving along the boundary. In another work \cite{Matveev2015}, a reactive navigation algorithm is proposed to guide a mobile robot to a target with obstacle avoidance. In this research work, the obstacles can be moving and deforming. The proposed control law is proved to be globally converging according to the mathematical analysis.

Another significant problem in robotics is robot formation building with multi-agent system \cite{4118472,1605401,6303906,6596518,1304923}. The aim of the robot formation is to drive multiple mobile robots to achieve prescribed constraints on their states \cite{OH2015424survey}. Generally the problem can be classified into shape producing problem and shape tracking problem. There are a lot of research works focusing on both of the two problems. In the work \cite{4543197}, the market-based coordination protocols are used to solve the dynamical task assignment problem in the distributed formation algorithm for shape producing, while another work \cite{Ji2006Role} which provides an off-line task assignment strategy. In the work \cite{Wu2007Consensus}, a model is proposed to describe the dynamic characteristic of the multi-robot system and a model-based control strategy is given to solve the formation and trajectory tracking problem. In the work \cite{Mastellone2008Formation}, firstly, a collision-free path tracking controller is designed for a single non-holonomic ground mobile robot. Then the proposed controller is used to solve the formation problem for multi-robot system. In another research work \cite{SAVKIN2016463}, the non-holonomic mobile robot is considered as the work \cite{Mastellone2008Formation}. In this work, a method for decentralized flocking and global formation building for multi non-holonomic mobile robot network is proposed. The robot model considered in this work has hard constraints on the robot's linear and angular velocities. The robots can be guide to move in a desired geometric pattern. In another work \cite{XIAO20092605}, the authors propose a formation framework to control a multi-agent robot system with a large number of robots in finite time by separating the formation information into local and global parts. This method can perform the navigation and formation control with less data exchange. In another work \cite{DONG201626Time}, the time-varying formation control problem is addressed for flying robots. The control method is proposed based on the Lyapunov approach combining with Riccati technique. A similar problem with \cite{DONG201626Time} is mentioned and solved in another work \cite{6851896} for linear time-invariant multi-agent systems. In this work, time delay is considered. In the work \cite{savkin_2015Distributed}, a distributed control algorithm for multiple non-holonomic robots is proposed. In the research work, robots only can see a part of the environment by camera and measure the range to other robots. The navigation algorithm drive the robots to encircle a given dynamic target with uniform distribution over the respective circle and collision avoidance.

\section{Sensor network based navigation algorithms}
As mentioned above, the sensor network based navigation problem becomes a new challenge in recent years with the development of the sensor network. In the past decade, the sensor network technology was developed significantly. There are a variety of sensor networks proposed for different environments and implementations; see \cite{5350445,6122042,5434384,
Karaboga2012Cluster,Xie2014}. The main advantage of the sensor network is that it can capture the environment information in a wide area rapidly rather than single sensor in a limited coverage. This advantage promote the cooperation of sensor network and mobile robots in a lot of fields, such as development using mobile robots, data collection, mobile robot localization and navigation; see \cite{Rawat2014,7294063,5658491}.

In the past five years, there are a lot of works proposed in the sensor network based navigation. Comparing with the local navigation algorithms, the sensor network based navigation algorithms use much more information captured by the sensor network to perform a more efficient navigation. The previously proposed research works in the sensor network based navigation can be generally classified into collision-free navigation methods and target-reaching navigation methods. The objective of a collision-free navigation algorithm is to use the measurements captured by the sensor network to guide a mobile robot to a target with obstacle avoidance. It takes the main advantage of the sensor network to detect the static and dynamic obstacles in the environments; see an example in \cite{WANG2013858}.

In the previous works of sensor network based collision-free navigation, camera sensor network is one of the common types of sensor networks used for obstacle detection; see \cite{Tian201688,cheng_jiang_hu_2014,6885782}. In the work \cite{Tian201688}, a harmonic navigation algorithm for multiple ground wheelchair robots is presented by using wireless sensor nodes, which are deployed in dynamic indoor environments. The sensor network used in \cite{Tian201688} consists of some cameras. The location of the wheelchairs and the occupied area are estimated by the image process. It can guide the wheelchairs avoiding the static obstacles and moving obstacles. In another work \cite{cheng_jiang_hu_2014}, the same camera sensor network is used to localize the robots and recognize the obstacles. The proposed algorithm in \cite{cheng_jiang_hu_2014} allows the mobile robot or a vehicle which has less intelligence to perform a sophisticated mobility without a large number of on-board computations. The camera deployed in the indoor environment is also used in another work \cite{6885782} to guide a mobile robot. However, in the proposed method in \cite{6885782}, only the single camera is used, not a sensor network. In addition to the camera sensor network, there are other types of sensor network used in the previous works. In the work \cite{ZhangY2013}, an ultrasonic sensor network is deployed on the ceiling of an indoor environment according to the square grid. Each ultrasonic sensor node measures the distance from the ceiling to the static obstacle below the sensor. According to the presented ultrasonic sensor network, a 3D map of the environment can be built approximately. Then, a D*Lite \cite{1435479} path planning algorithm is used to plan the robot path. One of the disadvantages of the method in \cite{ZhangY2013} is that the method requires a large number of sensor nodes densely deployed in an indoor environment. In the other two works \cite{6775785,Enriquez2013}, the RSSI (received signal strength indicator) based sensor network is used to perform a safe navigation for a ground mobile robot. In the work \cite{6775785}, the authors propose an RSSI-based localization and navigation method in a static environment. Similar as \cite{ZhangY2013}, the sensor nodes are deployed according to the grid. Then, the A* path searching algorithm is used to search a collision-free robot path. However, the authors do not indicate which approach the sensor network used to detect obstacles. In another RSSI-based method \cite{Enriquez2013}, RSSI-based sensor network combined with RFID (radio frequency identification) is used to guide a ground mobile robot in an static indoor environment and avoid obstacles. The RSSI-based sensor network indicates the reference heading for robot and the RFID tags around obstacles are used to perform the robot obstacle avoidance. The main difficulty of applying the RSSI-based navigation methods in practical implementations is that the sensor network involves a large number of sensors to cover the whole environment. The sensor nodes should be deployed manually, which is difficult in a larger area and is not economical. In addition to the sensor networks above, the infra-red (IR) sensors and 2D range finder sensors are used in the sensor network to detect obstacles; see \cite{7027292} and \cite{5766050}, respectively. In the work \cite{7027292}, a sensor network measures the environment data and uses the IR signals to couple the neighbour nodes and recognize the obstacles. Then, the data collection points are determined and an optimal collision-free data collection path is generated for the ground mobile robot. In another work \cite{5766050}, a 2D range finder sensor network is considered in the environments. According to the measurements captured by different sensor nodes, a partly detected map can be obtained and the PRM (probabilistic roadmap) algorithm is used to generate a safe robot path. The main advantage of the work \cite{5766050} is that the 2D range finder sensor network can obtain an accurate map indicating the detected obstacles and unoccupied area. It involves much less number of sensor nodes than RSSI-based sensor network to cover a same area. However, the authors of \cite{5766050} do not consider the motion control of any robot model in the work.

Target-reaching navigation is another topic in the sensor network based robot navigation. The main difference of the target-reaching navigation with collision-free navigation is that the collision and obstacles are generally not considered. The objective of target-reaching navigation is to guide the mobile robot from an initial position to a target sensor node or an unknown position estimated by the sensor network. In this field, the RSSI-based sensor network is widely used for the robot navigation. One of the disadvantages of an RSSI-based sensor network is that it cannot detect the obstacles and provide a collision-free navigation. Although there are some RSSI-based collision-free navigation algorithms discussed above, the obstacle detection is still not solved by the RSSI-based sensor network. In the collision-free navigation algorithm proposed in the work \cite{Enriquez2013}, the RSSI-based sensor network does not provide the obstacle avoidance. Therefore, the RFID tags are involved to indicate the possible collision with static obstacles. In another work \cite{6775785}, the approach of the obstacle detection in the RSSI-based sensor network is not discussed. In other works of the RSSI-based target-reaching navigation, some of the works are localization-free navigation, like \cite{Sun2010,6579651,6943217}, and some works requires the odometry of the mobile robot, like \cite{ZhengZhang2013,6423636}. In the work \cite{6579651}, the target is localized by the RSSI-based sensor network and a pseudogradient is proposed to navigate the mobile robot to the target. The robot's location is also estimated by the sensor network. In another work \cite{6943217}, a path generation strategy is proposed based on the pseudogradient proposed in \cite{6579651} to navigate the robot with a shorter trajectory. In the work \cite{Sun2010}, a localization-free and range-free navigation algorithm for mobile robot is proposed to guide the robot along the node-to-node path.

\section{Safe area search and map building algorithms}
Area search and map building is another important topic in robot navigation; \cite{1241772,Beeson2010428,Rastgoo201512,
6106751,4209183,5509993,5509803}. The objective of an are search and map building algorithm is to guide the mobile robot along a desired collision-free path while searching and mapping the environment by the robot's local sensors.

One of the primary requirements in this problem is collision-free robot navigation for area exploration on the bounded flat ground. In the past few years, there are a lot of local navigation algorithms which are presented. Generally there are two mathematical models used to describe the mobile robot, onmidirectional mobile robot and non-holonomic mobile robot. The work \cite{Yorozu2016} considers the obstacle avoidance with an onmidirectional mobile robot and the work \cite{Matveev2011a} considers the navigation of a non-holonomic mobile robot. In the work \cite{Yang2016}, an extended state observer and a nonlinear controller are designed and analysed for obstacle avoidance of a non-holonomic ground mobile robot. In another work \cite{Jin2017}, a switch-system approach to obstacle avoidance is proposed. In the works \cite{Yang2016,Jin2017}, the authors consider the control of both the linear velocity and angular velocity of a non-holonomic ground mobile robot with complex computation. In the work \cite{MWHMET2017}, the authors consider the particle swarm optimization to search the robot's collision-free path and in the work \cite{YANG2016A}, Q-learning is considered for robot obstacle avoidance navigation. The main disadvantage of the particle swarm optimization and Q-learning is that the performance of the algorithms cannot be proved mathematically. In another work \cite{Park2016}, the authors design an mobile robot obstacle avoidance algorithm by using fuzzy potential field and sonar sensors.

Map building is another fundamental requirement of the topic. There are different types of on-board sensors used to capture the environment information. The 2D range finder sensor is a common type used to scan the obstacles in the environments. It performs a more accurate map than a monocular camera or other vision sensors, such as \cite{Dewangan2016}. Although there are a number of researchers using time-of-flight (ToF) camera to capture depth image of the environment for map building like \cite{7831932,SCHMUCK2016230,7413791}, a 2D range finder sensor is much more economical than a ToF camera and has a larger scanning angle. It can be noticed that in the work \cite{7558972}, the authors used the same type of sensor, 2D range finder sensor, to scan a 3D environment with an accurate map, unlike the rough 2D map in \cite{7866323}. However, in the work \cite{7558972}, the 2D range finder sensor is connected with a motor, which can be controlled to adjust the pitch of the sensor. It makes the map building control complicate.

Moreover, varieties of area search and map building problems for mobile robots have attracted a lot of attention in the robotics community; see e.g. 
\cite{Matveev2012,Kim2015,Noh2014,
Ray2012,
Hsu2013,Liu2012,
Wang2015a,Almansa-Valverde2012,
Vallve2015,
Wattanavekin2013,OFlaherty2015}.
Recent publications in  this field present many achievements in both single robot mapping and multi-robot mapping. The frontier-based exploration is the most common method in single robot exploration and mapping; see e.g. \cite{AlDahak2013,Oswald2016,
Freda2005,Wang2015a}. For multi-robot systems, not only frontier-based algorithms (see e.g. \cite{AlKhawaldah2015,Colares2016}) but other approaches (see e.g. \cite{Hoog2010,Liu2015}) are  proposed. For single robot exploration, there are a lot of achievements.
The work \cite{AlDahak2013} provides an efficient and simple algorithm to explore an closed environment by using incremental triangulation. However, this algorithm requires the obstacles in the environments are polygons. Another work \cite{Oswald2016} presented a fast exploration algorithm based on a simple prior topological map of the environment. One of the disadvantages is that the algorithm of \cite{Oswald2016} cannot be implemented in a completely unknown environment. The works \cite{Freda2005,Wang2015a} provides two similar frontier-based exploration algorithms with search trees. The main difference of these two algorithms is the algorithm of \cite{Freda2005} selects a candidate randomly and the algorithm of \cite{Wang2015a} determines the candidate by a multi criteria decision method.

\section{Contributions}
In this report, we present three algorithms of sensor network based navigation in networked control systems; see examples \cite{1193753,MATVEEV200451problem,
SAVKIN200651Analysis,Matveev2009Estimation}. Moreover, two algorithms of robot area search and map building are presented based on switched control system, such as \cite{Matveev2000Estimation,Savkin2002Hybrid,
SAVKIN199969Robust,SKAFIDAS1999553Stability}.

In Chapter \ref{Chapter2}, we present a novel sensor network based global path planning algorithm for non-holonomic ground mobile robots. In this method, the motion of obstacles is considered as uniform linear motion. A sensor network is used to detect the static and dynamic obstacles in the 2D environments and generate the globally shortest candidate path for each mobile robot. Comparing with other works, the main feature of this method is that the generated path is the globally optimal path in dynamic environments. It is mathematically proved that when sampling time approaches zero, the generated path converges to the globally shortest path in the environment, whereas many other algorithms do not consider the optimal path planning
(e.g. \cite{Ko2015,
Nishitani2015,
Montiel2015,
Li2015,
cheng_jiang_hu_2014})
or do not consider dynamic environments
(e.g. \cite{Brass2015}).
Moreover, a non-holonomic constraint on the motion of the mobile robot model is considered in the proposed method unlike the most of other works. Moreover, the performance of our method is proved mathematically and other heuristic algorithms cannot be proved.
(e.g. \cite{Hossain2015,
Mo2015,
Contreras-Cruz2015,
Alam2015,
Yao2015}).
Additionally, our method only requires a low-level path tracking controller on the mobile robot. It saves more robot power in computation and sensors than other local navigation algorithms
(e.g. \cite{Hossain2015,Nishitani2015,Ko2015}).
Furthermore, the proposed algorithm can be easily implemented in multi-robot systems. The sensor network plans the path for each mobile robot at any time.

In Chapter \ref{Chapter3}, we present a sensor network based ground mobile robot real-time navigation algorithm. In this method, we take the advantage of the range finder sensor network to navigate all the ground mobile robots centrally in the workspace. The main feature of the proposed method is that the navigation tasks for all the mobile robots are completely transferred and integrated into the sensor network. Different types of robots can be navigated simultaneously in the workspace by the sensor network. Each robot is only required to have a low-level path tracking controller and some basic navigation sensors, like inertial navigation sensors or odometry sensors. It does not require any robot sensor for obstacle detection and any other extra navigation algorithm. Moreover, the sensor network based navigation is more flexible in configuration than local navigation in an industrial environment with different types of robots working cooperatively. New robots can be added into the workspace directly without any specialization in navigation. Additionally, a sensor network navigates robots according to the extensive measurements of the environment and performs a shorter and more efficient trajectory than local navigation algorithm. Therefore, this is an efficient, safe and economic navigation system for multiple robots in a dynamic industrial workspace. Furthermore, a practical non-holonomic industrial mobile robot model is considered in our method and dynamic environments with moving obstacles are supposed in our method, unlike other path planning algorithms e.g. \cite{Yuan2014,6775785}.

In Chapter \ref{Chapter4}, we present a sensor network based flying mobile robot navigation algorithm. In this method, we take the advantages of the 3D range finder sensor network to navigate the flying robots in 3D dynamic environment. The main feature of the presented method is that the navigation of all the flying robots is completely transferred and integrated into the sensor network, which is different from the local navigation algorithms; e.g. \cite{Liao2016,Kownacki2016}. Only a low-level path tracking controller is required for each robot. The robot is not equipped with any obstacle detection sensors and does not execute any complex algorithm. Different type of micro flying robot can be navigated directly by the sensor network without any specialization. In the previous works,  the navigation approaches for flying robot with sensor network are proposed in \cite{Bohd2015,Corke2005}. In the work \cite{Bohd2015}, RSSI-based sensor network is used to achieve the flying robot navigation. In another work \cite{Corke2005}, binary sensors are used to indicate the surrounding obstacles. Both of these two methods simplified the flying robot navigation as 2D planar navigation, unlike our work. The main difference between our work and these two works is that, in our method, the computation load of the navigation is completely transferred to the sensor network and in the other two works, the robot obtain the environment information from the sensor network and perform the navigation on the robot's computer. Moreover, in our work, the 3D range finder sensors are used to obtain sophisticate maps of the dynamic environment and the motion planning is considered, which achieves an accurate and efficient navigation for flying robots. Because of the large detection rage of the sensor network, the trajectories of the flying robots in our navigation algorithm are shorter than local navigation algorithms. Therefore, this is an efficient, safe and economical navigation system for multiple micro unmanned aerial vehicles (UAVs) in a dynamic environment. The proposed sensor network can be implemented as one of the fundamental units in smart factory with multiple micro flying robots working cooperatively. It also provides a centralized framework to manage and supervise the working of each micro UAV for high-level management system.

In Chapter \ref{Chapter5}, we present a safe area search and 2D map building algorithm for ground mobile robot in 2D environment. The main feature of our method is that it combines complete map building with obstacle avoidance for a unicycle model with constant speed, whereas some other papers do not solve the safe navigation problem for this model. For example, in the works \cite{Freda2005,Wang2015a}, different methods to select the candidate points and build a search tree are proposed. However, like the most of other frontier-based exploration methods, these methods do not consider the motion control from a candidate to next candidate for a special model. Therefore, safe navigation should be considered additionally for these methods with special robot models. Secondly, the proposed algorithm is relatively simple and computationally efficient. Therefore, the computational load of the proposed randomized algorithm is smaller than the most of frontier-based algorithms such as
e.g. \cite{Oswald2016,AlDahak2013,
Basilico2011}.
Moreover, the proposed algorithm is based on a 2D range finder sensor. Comparing with other types of sensors (see e.g. \cite{Almansa-Valverde2012,Pudics2015}), the range finder measurements are simpler to process and the map is built more accurately with a range finder sensor; see e.g. \cite{Wang2013,Wang2015a}.
Furthermore, a more realistic non-holonomic model of the mobile robot's motion is considered whereas many other papers
\cite{Wang2015a,
AlDahak2013,Oswald2016,
Wattanavekin2013,OFlaherty2015}
in the area do not take into account non-holonomic motion constraints. 
Additionally, unlike many other publications in this topic, a mathematically rigorous theoretical analysis of the developed algorithm is given.

In Chapter \ref{Chapter6}, we present an safe area search and 3D map building algorithm for ground mobile robot in 3D environment. Comparing with the previous works, the main feature of our method is that it combines the complete 3D map building and the collision-free area search navigation together in one method, unlike other works \cite{SCHMUCK2016230,7558972,7413791,Dewangan2016}. Furthermore, we consider a 2D range finder sensor to scan the 3D structure of the environment. It performs an accurate 3D map than a monocular camera or other vision sensors; see \cite{Dewangan2016}. Although there are a number of researchers using time-of-flight camera to capture depth image of the environment for map building like \cite{7831932,SCHMUCK2016230,7413791}, a 2D range finder sensor we used is much more economical than a ToF camera and has a larger scanning angle. It can be noticed that in the work \cite{7558972}, the authors used the same type of sensor, 2D range finder sensor, to scan the environment. However, in the work \cite{7558972}, the 2D range finder sensor is connected with a motor, which can be controlled to adjust the pitch of the sensor. It makes the map building control more complicate than our method. Moreover, our proposed method is relatively simple and, therefore, it can be used in small-sized robot with limited power supply and poor computer performance, like the robot used in \cite{7866323}, to build an accurate 3D map, unlike the rough 2D map in \cite{7866323}. Another feature of our method is that a non-holonomic model is considered, which is more practical than other works. According to the mathematical analysis, it is proved that with the probability $1$ the robot can complete the 3D map building in a finite time.

\section{Report outline}
This report is organised as follows: In Chapter \ref{Chapter2}, \ref{Chapter3} and \ref{Chapter4}, we study the sensor network based safe navigation problem and present three collision-free navigation algorithm. In Chapter \ref{Chapter2}, we present a global path planning algorithm for ground mobile robots by a 2D range finder sensor network. It is followed by Chapter \ref{Chapter3} that the path planning algorithm is developed to a real-time safe navigation algorithm for ground mobile robots by a 2D range finder sensor network. In Chapter \ref{Chapter4}, we extended the presented safe navigation algorithm into 3D environments and a safe navigation algorithm for flying robots by a 3D time-of-flight camera network is presented. The relevant computer simulations and experiments presented in Chapter \ref{Chapter2}, \ref{Chapter3} and \ref{Chapter4} confirm the performance of the presented methods. In Chapter \ref{Chapter5} and \ref{Chapter6}, we study the safe area search and map building problem. In Chapter \ref{Chapter5}, we present a 2D map building and collision-free area search algorithm for non-holonomic ground mobile robot with rigorous mathematical proof. Then the presented algorithm is extended to a 3D map building and safe area search algorithm in Chapter \ref{Chapter6}. The relevant computer simulations and experiments presented confirm the expected performance of the presented methods in Chapter \ref{Chapter5} and \ref{Chapter6}.
%%%%%%%%%%%%%%%%%%%%%%%%%%%%%%%%%%%%%%%%%%%%%%%%
\chapter{Global Path Planning for Ground Mobile Robots}
\chaptermark{Ground Mobile Robot Path Planning}
\label{Chapter2}
This chapter is based on the the publication \cite{7554216}. In this chapter, we present an artificial potential field based global path planning algorithm for a non-holonomic mobile robot. In our method, the motion of obstacles is considered as uniform linear motion. A sensor network is used to detect the obstacles. The velocity of each obstacle is estimated by the sensor network. In the proposed algorithm, robot path is approximately represented as a series of equally spaced points tracked by the non-holonomic mobile robot. Some candidate paths are generated and optimized by the presented method to search the shortest candidate path. The presented navigation framework is a networked control system that the environment measurements, control input and robot's states are exchanged through the sensor network; see examples \cite{1193753,MATVEEV200451problem,
SAVKIN200651Analysis,Matveev2009Estimation}.

\section{Problem description}
A planar mobile robot is modelled as an unicycle with a non-holonomic constraint in a planar environment. It is widely used to describe many ground robots, unmanned aerial vehicles and missile etc.
\cite{Vilca2015,
Manchester2006,
Matveev2011b,
Guerra2016}.
The robot travels with a constant speed $v_r$ and is controlled by angular velocity $u$. The model of the vehicle is described as follows (see Fig. \ref{fig:c2_1}):

\begin{figure}[!htb]
\centering
\epsfig{figure=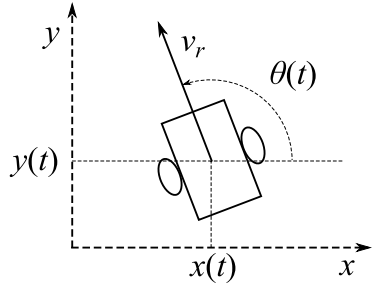,width=6cm}
\caption{
Unicycle model of the robot.
}
\label{fig:c2_1}
\end{figure}

\begin{equation}
\label{c2_1}
\left\{
\begin{array}{l}
\dot{x_r}(t) = v_r \cos \theta_r (t)
\\
\dot{y_r}(t) = v_r \sin \theta_r (t)
\\
\dot{\theta_r}(t) = u(t) \in [-u_M, u_M]
\end{array},
\begin{array}{l}
x_r(0) = x_{0}
\\
y_r(0) = y_{0}
\\
\theta_r(0) = \theta_{0}
\end{array}.
\right.
\end{equation}
In robot model (\ref{c2_1}), $(x_r,y_r)$ is the Cartesian coordinate of the vehicle and $\theta_r$ is the robot's heading at time $t$. The angular velocity $u$ satisfying the following non-holonomic constraint:

\begin{equation}
\vert u(t) \vert\leq u_M.
\end{equation}
This implies that the robot's minimum turning radius is

\begin{equation}
R_{\min}=\frac{v_r}{u_M}.
\end{equation}
The robot is equipped with an odometry sensor to help to obtain the position and heading direction relative to its starting location and heading. 

\begin{assumption}
\label{c2_AS8}
The robot's initial location $(x_0,y_0)$ and direction $\theta_0$ are known.
\end{assumption}

\begin{assumption}
The robot can adjust the heading direction quickly at the beginning.
\end{assumption}

In the planar environment, there are some disjoint static and moving obstacles $D_1,D_2,\ldots$. Notice that the shape of obstacles can be irregular and non-convex.

\begin{definition}
For any $i$, $D_i$ is a closed, bounded, and connected point set. 
\end{definition}

\begin{figure}[!htb]
\centering
\epsfig{figure=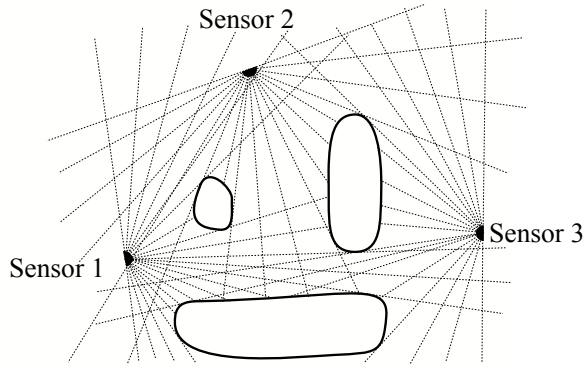,width=8cm}
\caption{
Sensor network in the planar environment.
}
\label{fig:c2_ad1}
\end{figure}

To detect the obstacles, a sensor network is deployed on the ground. The sensor network consists of several 2D range finder sensors that measure the distances to the nearest obstacles in different directions in the scanning ranges (see Fig. \ref{fig:c2_ad1}). A global map consisting of the obstacles' boundaries and unoccupied areas can be built by the sensor network.

\begin{assumption}
\label{c2_AS3}
Let $p$ be an arbitrary point on the boundaries of obstacles. At the initial time, there exists at least one sensor that the segment between the sensor and point $p$ does not intersect any obstacle. It means the point $p$ can be detected by at least one sensor.
\end{assumption}

\begin{notation}
\label{c2_adMINDIS}
Let $p$ be an arbitrary point. For any closed set $D$, the minimum distance $\rho(D,p)$ between $p$ and $D$ is
\begin{equation}
\rho(D,p):=\min_{q\in D}\Vert p-q\Vert.
\end{equation}
\end{notation}

\begin{figure}[!htb]
\centering
\epsfig{figure=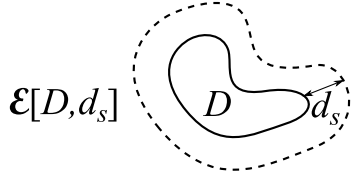,width=5cm}
\caption{
Enlarged region ${\cal E}[D,d_s]$.
}
\label{fig:c2_ad2}
\end{figure}

\begin{definition}
Let $d_s>0$ be a given safety margin. The robot should keep this safety margin to any obstacle while travelling. The $d_s$-enlarged region ${\cal E}[D,d_s]$ of a closed set $D$ is defined as follows (see Fig. \ref{fig:c2_ad2}): 
\begin{equation}
{\cal E}[D,d_s]:=\lbrace p:\rho(D,p)\leq d_s\rbrace.
\end{equation}
\end{definition}

\begin{assumption}
\label{c2_AS9}
The planar sets ${\cal E}[D_i,d_s]$, $i=1,2,\ldots$ are closed, bounded, connected and linearly connected sets.
\end{assumption}

\begin{assumption}
\label{c2_AS4}
For any $i\neq j$, ${\cal E}[D_i,d_s]$ does not overlap ${\cal E}[D_j,d_s]$ at any time.
\end{assumption}

\begin{assumption}
\label{c2_AS1}
The motion of any obstacle $D_i$, $i=1,2,\ldots$ is uniform linear motion with the speed $v_{D_i}\leq v_r$.
\end{assumption}

\begin{assumption}
\label{c2_AS2}
Let $\beta=\frac{v_{D_i}}{v_r}$. For any $i$, the boundary of ${\cal E}[D_i,d_s]$ is smooth and the curvature of the boundary is $<\frac{1}{(1+\beta)R_{\min}}$ at any point.
\end{assumption}

\begin{remark}
If Assumption \ref{c2_AS1} and \ref{c2_AS2} do not hold, the non-holonomic mobile robot may fail to patrol the $d_s$-enlarged boundaries of moving obstacles. In the proposed algorithm, patrolling the $d_s$-enlarged boundaries of obstacles is required.
\end{remark}

\begin{figure}[!htb]
\centering
\epsfig{figure=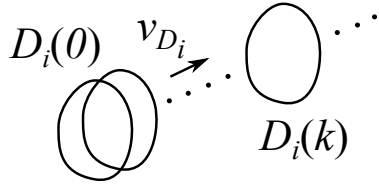,width=5cm}
\caption{
Position prediction of obstacles.
}
\label{fig:c2_ad14}
\end{figure}

With the above definitions and assumptions, the sensor network can detect the complete boundary of any obstacle $D_i$ at the initial time. Moreover, with a given sampling interval $\delta$, the velocity $v_{D_i}$ of each obstacle $D_i$ can be estimated by calculating the translation during the sampling interval $\delta$. For any obstacle $D_i$, let $D_i(k)$ denote the position of the obstacle at time $k\delta$. At the beginning, $D_i(k)$ can be predicted according to the detected initial position $D_i(0)$ and the estimated velocity $v_{D_i}$ (see Fig. \ref{fig:c2_ad14}):

\begin{equation}
D_i(k)=D_i(0)+v_{D_i}\cdot k\delta t.
\end{equation}

\begin{assumption}
\label{c2_AS6}
Give a target point ${\cal T}$. For any $i$ and $k$, $\rho(D_i(k),{\cal T})>d_s$.
\end{assumption}

\begin{assumption}
\label{c2_AS7}
Let $p=(x_{0},y_{0})$ be the initial position of mobile robot. For any $i$, $\rho(D_i(0),p)>d_s$.
\end{assumption}

\begin{definition}
Let $t_e>0$ be the travel time of the robot. For a robot path $P=\lbrace (x_r(t),y_r(t)) : t\in [0,t_e] \rbrace$, if the inequality
\begin{equation}
\rho(D_i(\frac{t}{\delta}),(x_r(t),y_r(t)))\geq d_s
\end{equation}
holds for any $t\in [0,t_e]$ and
\begin{equation}
(x_r(t_e),y_r(t_e))={\cal T},
\end{equation}
then the path $P$ is called a collision-free path.
\end{definition}

\begin{figure}[!htb]
\centering
\epsfig{figure=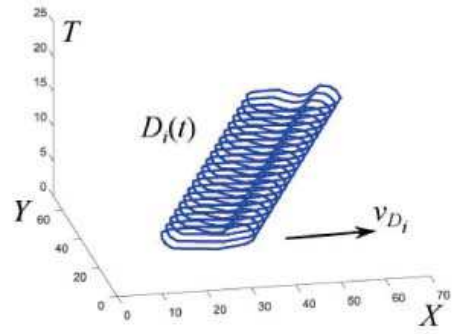,width=6cm}
\caption{
Set $D_i(t)$ in the X-Y-T spacetime.
}
\label{fig:c2_ad7}
\end{figure}

\begin{figure}[!htb]
\centering
\epsfig{figure=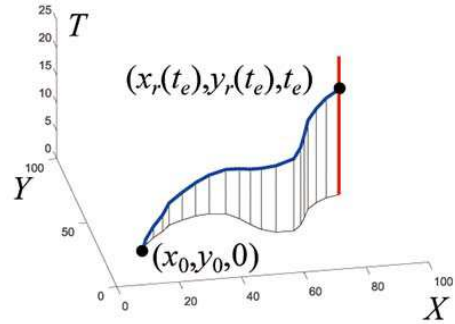,width=6cm}
\caption{
A robot path in the X-Y-T spacetime.
}
\label{fig:c2_ad8}
\end{figure}

Consider an X-Y-T spacetime for the path planning problem. For any collision-free path with a travel time $t_e>0$, the obstacles $D_1(t),D_2(t),\ldots$ are closed and bounded set in the X-Y-T spacetime during the period $[0,t_e]$ (see Fig. \ref{fig:c2_ad7}); and the collision-free path is a curver from the initial position $(x_0,y_0,0)$ to the target point $(x_r(t_e),y_r(t_e),t_e)$ (see Fig. \ref{fig:c2_ad8}). Then, we present the following definitions (see \cite{Roesmann2017}).

\begin{definition}
Let $P_1$ and $P_2$ be two collision-free paths in the X-Y-T spacetime. $P_1$ and $P_2$ are homotopic if and only if one can be continuously deformed into the other without intersecting the obstacles $D_1(t),D_2(t),\ldots$. The set of all paths that are homotopic to each other is denoted as homotopy class.
\end{definition}

\begin{definition}
Let $P$ be a collision-free path in the X-Y-T spacetime. If for any other paths which are homotopic to $P$, the travel time is larger than the travel time of $P$, then $P$ is called the locally shortest path in the corresponding homotopy class.
\end{definition}

\begin{definition}
Let $P$ be a locally shortest path, if for any homotopy class which $P$ does not belong to, the travel time of the locally shortest path is larger than the travel time of $P$, then $P$ is called the globally shortest path.
\end{definition}

The objective of the proposed algorithm is to generate a path which converges towards the globally shortest collision-free path with the minimum travel time $t_e$. The environment measurements, robot's position and heading, and robot's angular velocity are exchanged through the sensor network to construct a networked control system; see examples \cite{1193753,MATVEEV200451problem,
SAVKIN200651Analysis,Matveev2009Estimation}.

\section{Path planning algorithm}

\begin{figure}[!htb]
\centering
\epsfig{figure=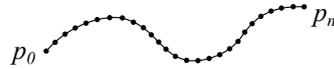,width=4.5cm}
\caption{
Robot path approximately represented by a series of equally spaced points.
}
\label{fig:c2_ad9}
\end{figure}

Here, the shortest path planning algorithm is proposed. In the proposed method, a robot path $P$ is approximately represented by finite equally spaced points $p_0,\ldots,p_n$ (see Fig. \ref{fig:c2_ad9}). The interval between any two successive points is

\begin{equation}
L=v_r\cdot \delta,
\end{equation}
where $\delta$ is a the sampling interval of the computer control system.

To search a path (denoted by $P^{*}$) which converges towards globally shortest path, several candidate paths $P_1,\ldots,P_m$ which converge towards different locally shortest paths are generated by an iterative search algorithm. Each candidate path is obtained by a proposed optimization algorithm with artificial potential fields. The path $P^{*}$ is proved to be the shortest path among the obtained candidate paths $P_1,\ldots,P_m$. 
The path $P^*$ can be tracked by a path tracking controller e.g. \cite{Chang2013}.

In the following parts, the path optimization algorithm and the iterative search algorithm are presented respectively. 

\subsection{Path optimization algorithm}

In the path optimization algorithm, a candidate path $P$ is prolonged gradually from the initial position $p_0=(x_{0},y_{0})$ to the target ${\cal T}$ by generating the path points $p_i$, $i=1,2,\ldots$, one by one in the artificial potential fields with the proposed rules.

Firstly, we define three vector fields $F_I$, $F_R$ and $F_P$ in the plane. For any point $p_i\in P$, $i\neq 0$, the resultant vector in the three vector fields is
\begin{equation}
\vec{F}(p_i)=\vec{F}_I(p_i)+\vec{F}_R(p_i)+\vec{F}_P(p_i).
\end{equation}
The path $P$ is optimized by moving any point $p_i$, $i\neq 0$ to an equilibrium point where $\vec{F}(p_i)=\vec{0}$. When $P$ is optimized and the last path point of $P$ does not reach the target ${\cal T}$, a new path point is added into $P$ next to the last path point to prolong $P$. With the new path point, $P$ needs to be optimized again since the equilibrium point might changes. The optimization can be presented as follows.

\begin{enumerate}
\item[\textbf{A1}:]
For each point $p_i$, $i\neq 0$, let $\vec{p}_i$ denote the position of $p_i$ and initialize a velocity vector $\vec{v}_i=0$ for $p_i$.

\item[\textbf{A2}:]
Start the following loop:

\begin{enumerate}
\item[\textbf{A2.1}:]
Calculate $\vec{F}(p_i)$ for each point $p_i$, $i\neq 0$.

\item[\textbf{A2.2}:]
Change the position and velocity of each $p_i$, $i\neq 0$, according to the following equations:

\begin{equation}
\left\{
\begin{array}{l}

\vec{p}_{i} \leftarrow \vec{p}_{i}+\vec{v}_{i}
\\
\vec{v}_{i} \leftarrow G_N\cdot \vec{v}_{i}+F(p_i)

\end{array},
\right.
\end{equation}
where $0<G_N<1$ is a tunable attenuation.
\end{enumerate}

\item[\textbf{A3}:]
Exit loop \textbf{A2} if $\Vert \vec{F}(p_i)\Vert<F_{th}$ for any $i\neq 0$, where $F_{th}>0$ is a given threshold.
\end{enumerate}

Now, we are here to give the definitions of the three vector fields $F_I$, $F_R$ and $F_P$.

\begin{figure}[!htb]
\centering
\epsfig{figure=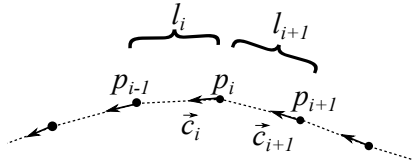,width=6cm}
\caption{
Vector $\vec{c}_i$ and distance $l_i$ .
}
\label{fig:c2_3}
\end{figure}

\textbf{Definition of} $\bf F_I$:
$F_I$ is a vector field that guarantees the interval between any two successive points of $P$ is approximately equal to $L$. For any $i\neq 0$, let distance $l_i=\Vert p_{i}-p_{i-1} \Vert$ and $\vec{c}_i$ be the unit vector pointing towards $p_{i-1}$ from $p_i$ (see Fig. \ref{fig:c2_3}). Then $F_I(p_{i})$ is defined as follows:

\begin{equation}
\vec{F}_{I}(p_{i})=\left\{
\begin{array}{ll}
G_I \cdot ((l_i-L) \cdot \vec{c}_i-(l_{i+1}-L) \cdot \vec{c}_{i+1}) &,i\neq n
\\
G_I \cdot ((l_i-L) \cdot \vec{c}_i) &,i=n
\end{array} ,
\right.
\end{equation}
where $G_I>0$ is a tunable gain.

\begin{figure}[!htb]
\centering
\epsfig{figure=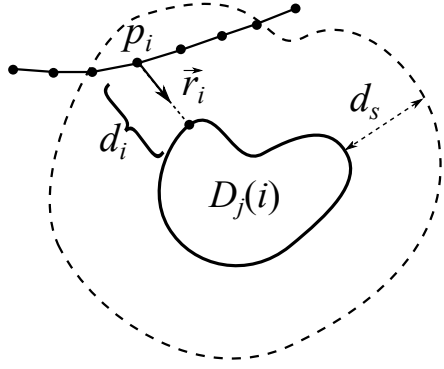,width=6cm}
\caption{
Vector $\vec{r}_i$ and distance $d_i$.
}
\label{fig:c2_4}
\end{figure}

\textbf{Definition of} $\bf F_R$:
$F_R$ is the repulsion around obstacles. It guarantees that, for any $i$, the minimum distance $d_i$ from point $p_i$ to all obstacles $D_j(i)$, $j=1,2,\ldots$ at time $i\delta$ is $\geq d_s$. Let $\vec{r}_i$ be the unit vector from $p_i$ to the closest obstacle at time $i\delta$ (see Fig. \ref{fig:c2_4}), then $F_R(p_{i})$ is defined as follows:

\begin{equation}
\vec{F}_R(p_{i})=\left\{
\begin{array}{ll}
G_R  \cdot (d_i-d_s) \cdot \vec{r}_i & ,d_i\leq d_s
\\
0 & ,d_i > d_s
\end{array},
\right.
\end{equation}
where $G_R>0$ is a tunable gain.

\begin{figure}[!htb]
\centering
\subfigure[$R1$]{
\epsfig{figure=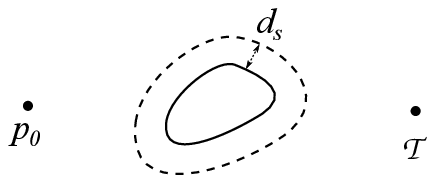,width=4cm}
}
\subfigure[$R1$]{
\epsfig{figure=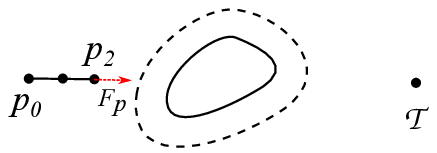,width=4cm}
}
\subfigure[$R1\rightarrow R2$]{
\epsfig{figure=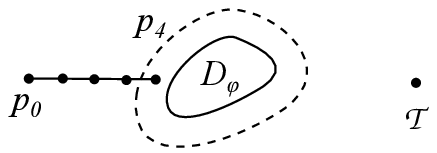,width=4cm}
}
\subfigure[$R2$]{
\epsfig{figure=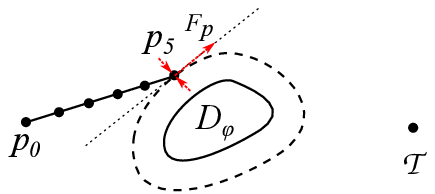,width=4cm}
}
\subfigure[$R2\rightarrow R1$]{
\epsfig{figure=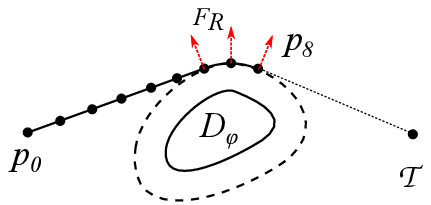,width=4cm}
}
\subfigure[$R1$]{
\epsfig{figure=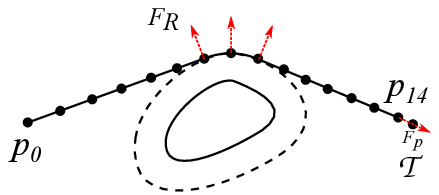,width=4cm}
}
\caption{
Mode transitions with a static obstacle.
}
\label{fig:c2ad3}
\end{figure}

\textbf{Definition of} $\bf F_P$:
$F_P$ is a vector field that only acts on the last point $p_n$. For other points $p_i$, $i\neq 0$ and $n$, $\vec{F}_P(p_i)=0$. The vector $\vec{F}_P(p_n)$ guides the prolonging of the path $P$ with two modes $R1$ and $R2$. The two modes and the mode transition rules are defined as follows (see Fig. \ref{fig:c2ad3}).

$\bf R1$:
When mode $R1$ is active, $\vec{F}_P(p_n)$ points towards the target ${\cal T}$ with a constant magnitude. Thus, the path $P$ is prolonged towards the target. Mode $R1$ is the initial mode at the beginning when $P$ only contains the first path point $p_0$.

$\bf R1\rightarrow R2$:
If there exists an obstacle $D_{\varphi}$ that makes $\rho(D_{\varphi}(n),p_{n})\leq d_s$ with mode $R1$, the mode will transition to $R2$ from $R1$.

$\bf R2$:
When mode $R2$ is active, $F_P(p_n)$ is a vector that can be decomposed into two mutually orthogonal components. The first component points towards the $d_s$-enlarged boundary radially with a linearly proportional magnitude to the difference of $\rho(D_{\varphi}(n),p_{n})$ and $d_s$. Another component only has a constant magnitude. Thus, the path $P$ is optimized and extended to bypass the obstacle $D_{\varphi}$ with the guidance of $F_P(p_n)$.

$\bf R2\rightarrow R1$:
If the line segment between $p_{n}$ and ${\cal T}$ does not intersect the $d_s$-enlargement of $D_{\varphi}(n)$ with mode $R2$, the mode will transition to $R1$ from $R2$.

\begin{figure}[!htb]
\centering
\epsfig{figure=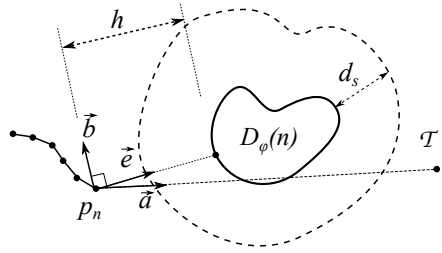,width=6cm}
\caption{
Vectors $\vec{a}$, $\vec{r}_n$, $\vec{c}$ and distance $h$.
}
\label{fig:c2_5}
\end{figure}

Let $\vec{a}$ be a unit vector pointing towards ${\cal T}$ from $p_{n}$. Let $h$ equals to $\rho(D_{\varphi}(n),p_{n})$ and $\vec{e}$ be the unit vector pointing towards $D_{\varphi}(n)$ radially. Let $\vec{b}$ be a unit vector perpendicular to $\vec{e}$ (see Fig. \ref{fig:c2_5}). For the last point $p_{n}$, $\vec{F}_P(p_{n})$ is defined as follows:

\begin{equation}
\vec{F}_P(p_{n})=\left\{
\begin{array}{lr}
G_P \cdot \vec{a} & ,R1
\\
G_R  \cdot ({h}-{d_s}) \cdot \vec{e}+\gamma \cdot G_P \cdot \vec{b} & ,R2
\end{array},
\right.
\end{equation}
where $G_P>0$ is a tunable gain. Notice that $\gamma\in\lbrace 1,-1\rbrace$ is a parameter that determines the direction (clockwise or anti-clockwise direction) in which the path $P$ bypasses obstacle $D_{\varphi}$ with the guidance of $F_P$.

Now, with the proposed path optimization algorithm \textbf{A1}-\textbf{A3}, the path prolonging algorithm is presented as follows:

\begin{enumerate}

\item[\textbf{B1}:]
Initialize a candidate path $P=\lbrace p_{0}(x_{0},y_{0})\rbrace$. Mode $R1$ is active.

\item[\textbf{B2}:]
Start the following loop:

\begin{enumerate}
\item[\textbf{B2.1}:]
Let $p_n$ denote the last point of $P$. Generate a new point $p_{n+1}\in P$ next to $p_n$ towards the target ${\cal T}$ with the distance $L$. $n\leftarrow n+1$.

\item[\textbf{B2.2}:]
Optimize $P$ by the path optimization algorithm \textbf{A1}-\textbf{A3}.

\item[\textbf{B2.3}:]
If $P$ meets the mode transition conditions, the active mode transitions to another mode. Additionally, if $R1\rightarrow R2$ occurs, the parameter $\gamma$ will be determined.
\end{enumerate}

\item[\textbf{B3}:]
Exit loop \textbf{B2} if $\Vert p_{n}-{\cal T}\Vert<L$.

\end{enumerate}

With the proposed algorithm \textbf{A1}-\textbf{A3} and \textbf{B1}-\textbf{B3}, a candidate path $P$ can be generated. $P$ is proved to converge towards the locally shortest path in the corresponding homotopy class.

\subsection{Iterative search algorithm}

In the algorithm \textbf{B1}-\textbf{B3}, a candidate path is generated by simultaneous prolonging and optimization. During this procedure, the parameter $\gamma$ is determined each time when $R1\rightarrow R2$ occurs. It is easy to know that with different combinations of the value choices of $\gamma$, the different paths belong to the different homotopy classes. Let ${\cal P}$ be a set of candidate paths Here we present the iterative algorithm as follows:

\begin{enumerate}

\item[\textbf{C1}:]
Initialize the first candidate path $P_1=\lbrace p_{0}(x_{0},y_{0})\rbrace$ and ${\cal P}=\lbrace P_1 \rbrace$.

\item[\textbf{C2}:]
Start the following loop:

\begin{enumerate}

\item[\textbf{C2.1}:]
For any candidate path $P_i\in {\cal P}$ which does not reach the target, prolong $P_i$ by the algorithm \textbf{B2}-\textbf{B3}. During the prolonging, at each time when $R1\rightarrow R2$ occurs, a new candidate path with duplicate path points of $P_i$ is generated into ${\cal P}$. The path $P_i$ and the new candidate path are with different value choice of the parameter $\gamma$.
\end{enumerate}

\item[\textbf{C3}:]
Exit loop \textbf{C2} if any candidate path $P_i\in {\cal P}$ reaches the target. Then the shortest candidate path is the path $P^{*}$.
\end{enumerate}

\section{Computer simulations}

In this section, computer simulations are carried out to confirm the performance of the proposed method in both static and dynamic environments. Moreover, the proposed method can be implemented with multi-robot systems. A computer simulation is carried out to show the performance of the proposed method with multiple robots.

\subsection{Simulations in static environments}

\begin{figure}[!htb]
\centering
\epsfig{figure=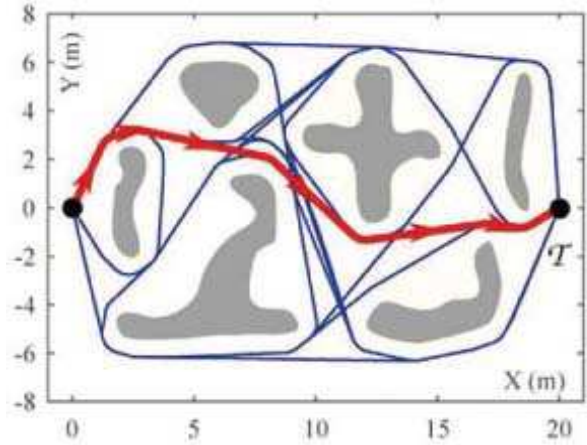,width=8cm}
\caption{
Result of the first simulation.
}
\label{fig:c2_14}
\end{figure}

\begin{figure}[!htb]
\centering
\epsfig{figure=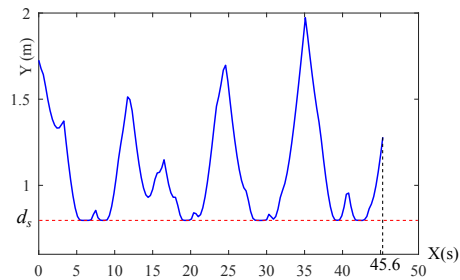,width=6cm}
\caption{
Minimum distance to obstacles in the first simulation.
}
\label{fig:c2_15}
\end{figure}

The first simulation (see Fig. \ref{fig:c2_14}) is carried out in a static environment. There are some static non-convex obstacles on the plane. With the proposed path planning algorithm, several candidate paths are generated and the shortest path $P^*$ is chosen. Table. \ref{c2_Simulation_Static} shows the parameters used in this simulation. According to the simulation result, it can be seen that the mobile robot reaches the target successfully and keeps a given safety margin $d_s$ to the obstacles (see Fig. \ref{fig:c2_15}).

\begin{table}[!htb]
\caption{Parameters of The First Simulation.}
\label{c2_Simulation_Static}
\begin{center}
\begin{tabular}{|c|c|c|}
\hline
Name & Symbol & Value\\
\hline
Robot's speed & $v_r$ & $0.5\text{m/s}$\\
\hline
Robot's maximum angular velocity & $u_M$ & $2\text{rad/s}$\\
\hline
Sampling interval & $\delta$ & $0.3\text{s}$\\
\hline
Safe margin & $d_s$ & $0.6\text{m}$\\
\hline
\end{tabular}
\end{center}
\end{table}

\subsection{Simulations in dynamic environments}

\begin{figure}[!htb]
\centering
\subfigure[$t=0\text{s}$]{
\epsfig{figure=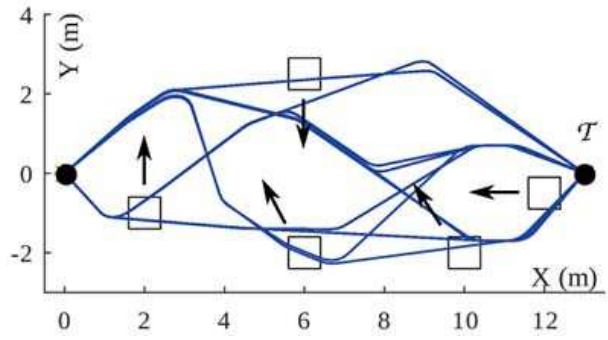,width=8cm}
}
\subfigure[$t=15\text{s}$]{
\epsfig{figure=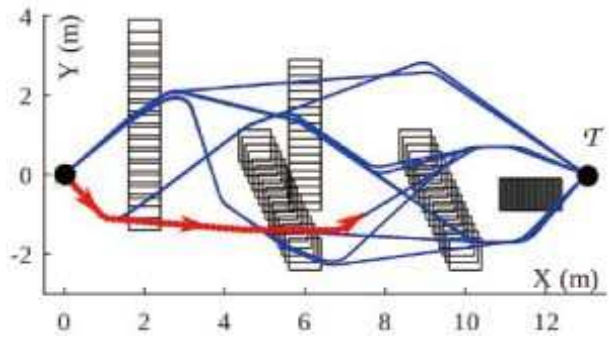,width=8cm}
}
\subfigure[$t=27.3\text{s}$]{
\epsfig{figure=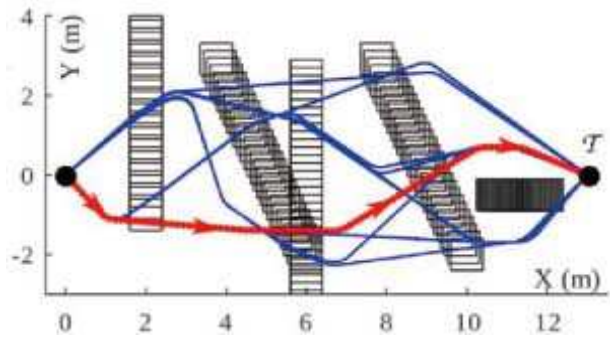,width=8cm}
}
\caption{
Result of the second simulation.
}
\label{fig:c2_17}
\end{figure}

\begin{figure}[!htb]
\centering
\epsfig{figure=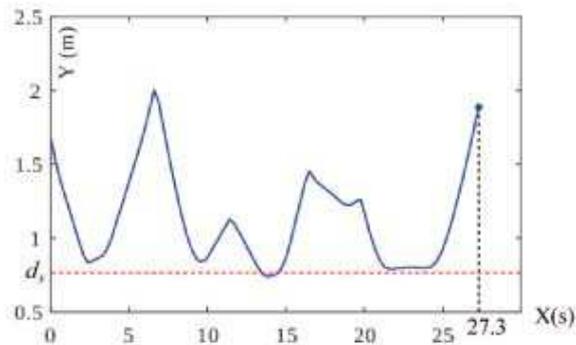,width=7.5cm}
\caption{
Minimum distance to obstacles in second simulation.
}
\label{fig:c2_21}
\end{figure}

The second simulation (see Fig. \ref{fig:c2_17}) is carried out in a dynamic environment with some moving obstacles. The motion of each obstacle is uniform linear motion. With the proposed algorithm, several robot candidate paths are generated like the first simulation. Each candidate path is a target-reaching path. According to the simulation result, the robot chooses the shortest candidate path to track and avoids all the moving obstacles with the safety margin. Table. \ref{c2_Simulation_Dynamic} shows the parameters used in this simulation. Fig. \ref{fig:c2_21} indicates that the mobile robot keeps the safety margin $d_s$ to obstacles while travelling.

\begin{table}[!htb]
\caption{Parameters of The Second Simulation.}
\label{c2_Simulation_Dynamic}
\begin{center}
\begin{tabular}{|c|c|c|}
\hline
Name & Symbol & Value\\
\hline
Robot's speed & $v_r$ & $0.5\text{m/s}$\\
\hline
Robot's maximum angular velocity & $u_M$ & $ 2\text{rad/s}$\\
\hline
Sampling interval & $\delta$ & $0.3\text{s}$\\
\hline
Safe margin & $d_s$ & $0.8\text{m}$\\
\hline
\end{tabular}
\end{center}
\end{table}

\begin{figure}[!htb]
\centering
\subfigure[$t=0\text{s}$]{
\epsfig{figure=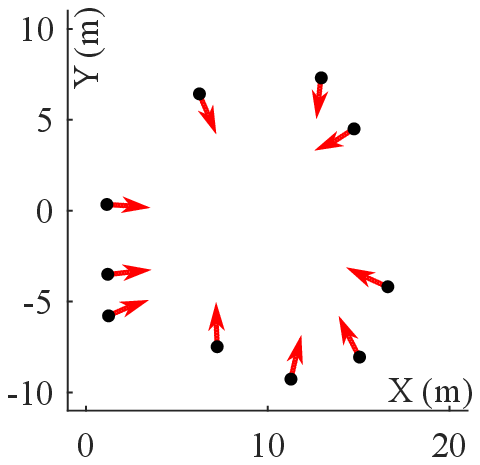,width=5cm}
}
\subfigure[$t=36.9\text{s}$]{
\epsfig{figure=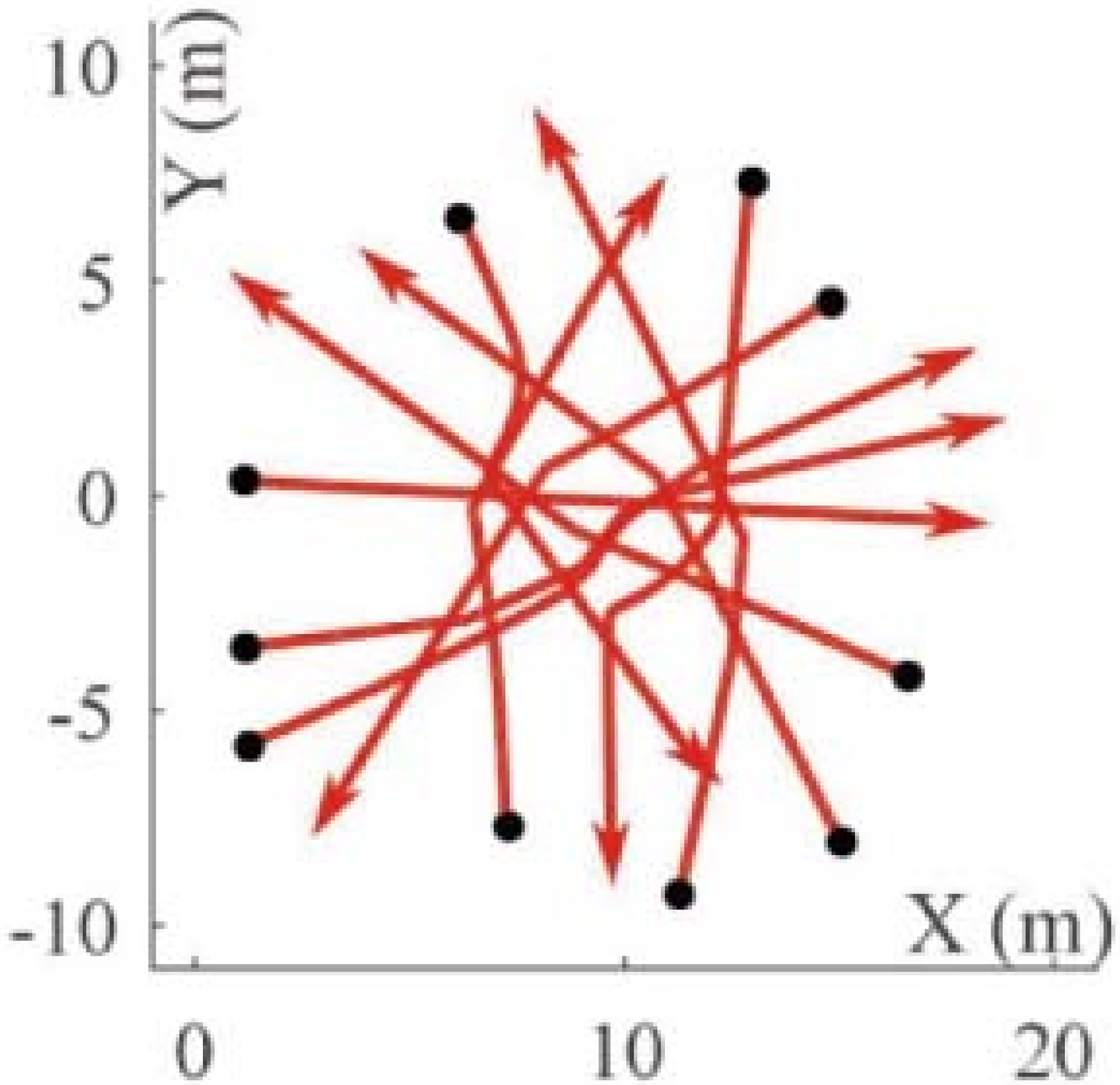,width=5cm}
}
\caption{
Result of the third simulation.
}
\label{fig:c2_simu3}
\end{figure}

\begin{figure}[!htb]
\centering
\epsfig{figure=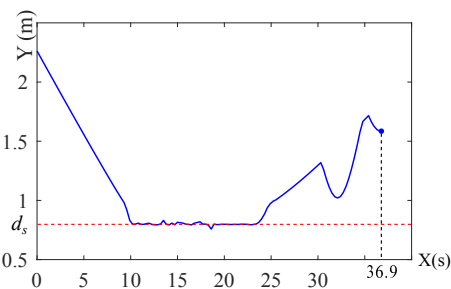,width=6cm}
\caption{
Minimum distance between any robots in the third simulation.
}
\label{fig:c2_simu3_2}
\end{figure}

In the third simulation, we consider the collision-free path planning for multiple robots by the proposed algorithm. The paths of the robots are planned successively. When plan the $i$-th path, the previous $i-1$ robots are considered as obstacles with known planned path. In this simulation, there are ten mobile robots in the planar environment with different initial positions and targets. The result of the simulation (see Fig. \ref{fig:c2_simu3}) shows that each mobile robot tracks the planned collision-free path to the target. The Fig. \ref{fig:c2_simu3_2} confirms that all the robots successfully avoid the collision with any other robot and keep the safety margin $d_s=0.8 \text{m}$ to each other.

\section{Summary}

A shortest path planning algorithm is proposed here for non-holonomic mobile robots in dynamic environments. The motion of the obstacles is considered as uniform linear motion. The sensor network measures obstacles and estimates the velocity of each obstacle. With the complete information of the obstacles, the proposed method searches a path converging towards the globally shortest path with obstacle avoidance. The performance of the algorithm is confirmed by the computer simulations.

Comparing with other works, the proposed method has some advantages. Firstly, the proposed path planning algorithm guides the robot with a globally shortest collision-free path in a dynamic environment. Secondly, the non-holonomic constraint on the motion of robot is considered. Furthermore, the proposed global path planning algorithm can be implemented in multi-robot systems.

%%%%%%%%%%%%%%%%%%%%%%%%%%%%%%%%%%%%%%%%%%%%%%%%
\chapter{Safe Navigation of Ground Mobile Robots in Dynamic Environments by a Sensor Network}
\chaptermark{Ground Mobile Robots Navigation}
\label{Chapter3}

This chapter is based on the the publications \cite{Hang_uj_2} and \cite{Hang_ccc2017_1}. In this chapter, we take the advantage of the range finder sensor network to navigate all the industrial mobile robots centrally in the workspace. Each range finder sensor node is deployed in dynamic industrial environments, such as factory floor, to detect walls, equipments, moving robots and walking people. Simultaneously, each robot measures its own real-time location and direction by localization, like odometry, and sends the measurements to the sensor network by the wireless communication. With the measurements of environment and robots' position, temporarily safe paths can be dynamically generated and the robots are navigated according to the generated path by the sensor network. The presented ground mobile robots navigation system is a networked control system; see examples \cite{1193753,MATVEEV200451problem,
SAVKIN200651Analysis,Matveev2009Estimation}. In the presented system, the environment measurements, control input and robots' states are exchanged through the sensor network.

\section{Problem description}

\begin{figure}[!htb]
\centering
\epsfig{figure=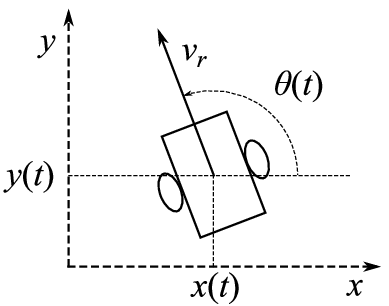,width=5cm}
\caption{
Unicycle model of the robot.
}
\label{fig:c3_1}
\end{figure}

A mobile robot is modelled as an unicycle with a non-holonomic constraint in a planar environment. It is widely used to describe many ground robots, unmanned aerial vehicles and missile etc.
\cite{Manchester2006,
Matveev2011b,
Savkin2016a}.
The robot travels with a constant speed $v_r$ and is controlled by angular velocity $u$. The model of the vehicle is described as follows (see Fig. \ref{fig:c3_1}):

\begin{equation}
\label{c3_1}
\left\{
\begin{array}{l}
\dot{x}(t) = v_r \cos \theta (t)
\\
\dot{y}(t) = v_r \sin \theta (t)
\\
\dot{\theta}(t) = u(t) \in [-u_M, u_M]
\end{array},
\begin{array}{l}
x(0) = x_{0}
\\
y(0) = y_{0}
\\
\theta(0) = \theta_{0}
\end{array}.
\right.
\end{equation}
In the robot model (\ref{c3_1}), $(x(t),y(t))$ is the Cartesian coordinates of the vehicle and $\theta(t)$ is the robot's heading at time $t$. The angular velocity $u$ satisfies the following non-holonomic constraint:

\begin{equation}
\label{c3_3}
\vert u(t) \vert\leq u_M.
\end{equation}
This implies that the robot's minimum turning radius is

\begin{equation}
R_{\min}=\frac{v_r}{u_M}.
\end{equation}

\begin{assumption}
\label{c3_AS8}
The robot measures its location and heading by some localization technologies, like odometry, etc.
\end{assumption}

\begin{figure}[!htb]
\centering
\epsfig{figure=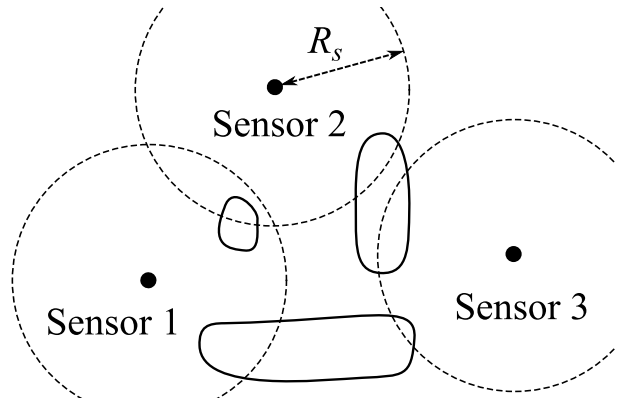,width=8cm}
\caption{
Sensor network in the planar environment.
}
\label{fig:c3_ad1}
\end{figure}

In planar environments, there are some static and moving obstacles, like walls, other robots and moving people. The obstacles can be non-convex and their velocity can be dynamic and unknown. To detect the obstacles, a sensor network is deployed in the workspace. The sensor network consists of some range finder sensor nodes. Each node measures the distances to the nearest obstacles in different directions within a measurement range denoted by $R_s$ (see Fig. \ref{fig:c3_ad1}). Each range finder is deployed higher than any mobile robots in the workspace. It means any mobile robots on the factory floor cannot be detected by the sensor network. Furthermore, a central computer node connects to the sensor network to obtain the real-time measurements from each sensor node. The central computer also can obtain the location and direction of any robots in the workspace by wireless communication. After obtaining the environment information and the location and direction of each robot, the sensor network dynamically generates a safe path for each robot according to the proposed navigation algorithm and send the paths to the robots for tracking.

\begin{figure}[!htb]
\centering
\epsfig{figure=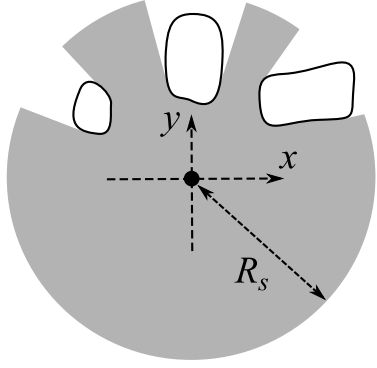,width=5cm}
\caption{
Detected area of a sensor node.
}
\label{fig:c3_ad25}
\end{figure}

For each sensor node, obstacles are detected partly in the measurement range and a detected area is obtained in the local coordinate system of the sensor node (see Fig. \ref{fig:c3_ad25}).

\begin{definition}
For each sensor node, the detected area is a closed, bounded, and connected point set.
\end{definition}

\begin{assumption}
For each sensor node, the location and direction where it is deployed are known.
\end{assumption}

\begin{figure}[!htb]
\centering
\epsfig{figure=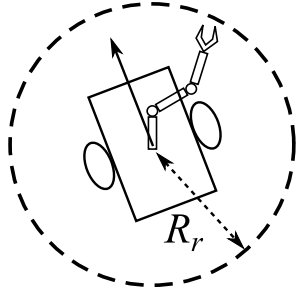,width=4cm}
\caption{
Physical size of a mobile robot.
}
\label{fig:c3_ad39}
\end{figure}

\begin{assumption}
\label{c3_AS6}
The physical size of any robot in the workspace can be covered by a disk with a given radius $R_r$ on the center of the robot (see Fig. \ref{fig:c3_ad39}). The disk called robot disk.
\end{assumption}

\begin{figure}[!htb]
\centering
\epsfig{figure=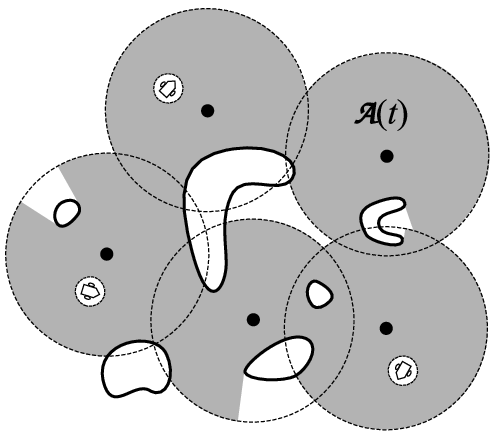,width=7cm}
\caption{
Unoccupied area ${\cal A}(t)$.
}
\label{fig:c3_ad20}
\end{figure}

With the help of the location and direction of each sensor node, the local detected areas of all the sensor nodes at any time $t$ can be converted to the global coordinate system, then a total detected area, which is the union of the local detected areas, is obtained. Moreover, the robot should avoid other mobile robots in the workspace. According to Assumption \ref{c3_AS6} and other robots' locations obtained by the central computer, the unoccupied area denoted by ${\cal A}(t)$ can be calculated, which equals to the relative complement of all the robot disks in the total detected area (see Fig. \ref{fig:c3_ad20}).

\begin{notation}
Let $p$ be an arbitrary point. For any closed set $D$, the minimum distance $\rho(D,p)$ between $p$ and $D$ is
\begin{equation}
\rho(D,p):=\min_{q\in D}\Vert p-q\Vert.
\end{equation}
\end{notation}

\begin{assumption}
\label{c3_AS3}
Let $p$ be an arbitrary point in the set ${\cal A}(t)$ and $\partial {\cal A}(t)$ denote the boundary of ${\cal A}(t)$. The derivative of the minimum distance $\rho(\partial {\cal A}(t),p)$ with respect to time $t$ is $\leq V_{\max}$, where $V_{\max}$ is a given constant which is $<v_r$.
\end{assumption}

\begin{remark}
If Assumption \ref{c3_AS3} does not hold, the robot may fail to avoid the dynamic undetected areas, which may contain obstacles.
\end{remark}

\begin{figure}[!htb]
\centering
\epsfig{figure=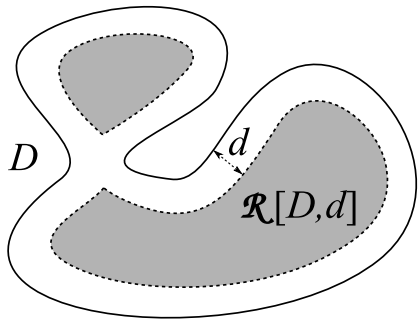,width=6cm}
\caption{
$d$-reduced region ${\cal R}[D,d]$ of a set $D$.
}
\label{fig:c3_ad2}
\end{figure}

\begin{definition}
Let $d>0$ be a constant. Let $\partial D$ denote the boundary of a closed set $D$. The $d$-reduction of the closed set $D$ is a set ${\cal R}[D,d]$ defined as follows (see Fig. \ref{fig:c3_ad2}): 
\begin{equation}
{\cal R}[D,d]:=\lbrace p\in D:\rho(\partial D,p)\geq d\rbrace.
\end{equation}
\end{definition}

\begin{definition}
The safety margin $d_s>0$ is a given constant that the robot should keep from the boundary $\partial{\cal A}(t)$ at any time $t$.
\end{definition}

\begin{figure}[!htb]
\centering
\epsfig{figure=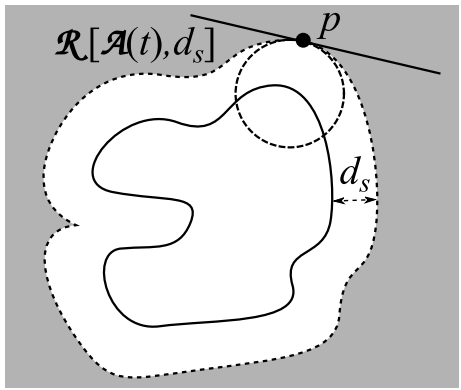,width=6cm}
\caption{
Osculating circle of the boundary of ${\cal R}[{\cal A}(t),d_s]$.
}
\label{fig:c3_ad32}
\end{figure}

\begin{assumption}
\label{c3_AS1}
Let $p$ be an arbitrary point on the boundary of ${\cal R}[{\cal A}(t),d_s]$. At any time $t$, if $p$ is not a singularity and the osculating circle at the point $p$ is to the different side of the tangent with ${\cal R}[{\cal A}(t),d_s]$, then the radius of the osculating circle is $\geq R_{\min}$ (see Fig. \ref{fig:c3_ad32}).
\end{assumption}

\begin{assumption}
The robot's initial position $(x_0,y_0)$ belongs to the initial unoccupied area ${\cal R}[{\cal A}(0),d_s]$.
\end{assumption}

The objective of the proposed algorithm is to drive the mobile robot to travel in the dynamic and deformable $d_s$-reduced area ${\cal R}[{\cal A}(t),d_s]$ and finally reach a target point denoted by ${\cal T}$ with a relatively short trajectory. In the presented method, the environment measurements, robot's position and heading, and robots' angular velocity are exchanged through the sensor network to construct a networked control system; see examples \cite{1193753,MATVEEV200451problem,
SAVKIN200651Analysis,Matveev2009Estimation}.

\begin{assumption}
The target ${\cal T}$ belongs to the set ${\cal R}[{\cal A}(t),d_s]$ at any time $t$.
\end{assumption}

\begin{assumption}
The set ${\cal R}[{\cal A}(t),d_s]$ is a connected set at any time $t$.
\end{assumption}

\begin{assumption}
\label{c3_AS2}
The distance between the target and the initial position is $>2R_{\min}$.
\end{assumption}

\section{Safe navigation algorithm}

In this section, we propose our sensor network based navigation algorithm. Firstly, we give the brief introduction of the algorithm as follows. In an environment, each node of a range finder sensor network is deployed to detect the surrounding obstacles. There is a central computer node connecting to the sensor network to collect the data from each node. Any robots in the workspace can upload the real-time locations and directions to the central computer node via the sensor network by wireless communication. Let $\delta$ be the given sampling interval. At any discrete time step $t=0,\delta,2\delta,\ldots$, the central computer node calculates the unoccupied area ${\cal A}(t)$ according to the obtained real-time information. Then it generates a relatively short safe path, denoted by $P^*$, from the robot's current position $(x(t),y(t))$ to the target ${\cal T}$. The path $P^*$ satisfies the non-holonomic constraint of the robot's motion. Let $T$ be a given time window, which is a positive integer. The path $P^*$ is proved to be safe over the time period $[t,t+T\delta]$. However, the robot only tracks $P^*$ over the next sampling interval $[t,t+\delta]$, then the algorithm repeats and updates the path $P^*$ at the next time step $t+\delta$. The algorithm repeats periodically at each time step and navigates the robot to the target without any collision.

Let $k=0,1,2,\ldots$. To guarantee the safety of the robot in the partly detected dynamic environment during the given time window $[t,t+T\delta]$, we consider any potentially unsafe areas in the environment and give the following definition.

\begin{definition}
\label{c3_DE2}
Let $t$ be the current time. According to Assumption \ref{c3_AS3}, the area which is obsoletely safe at time $t+k\delta$ is ${\cal R}[{\cal A}(t),k\delta V_{\max}]$. Therefore we define an area $\hat{\cal A}(t,k)$ as follows to help to generate a temporarily safe path:
\begin{equation}
\hat{\cal A}(t,k):=
\begin{cases}
{\cal R}[{\cal A}(t),k\delta V_{\max}] & \mathrm{if}\ k\leq T
\\
{\cal R}[{\cal A}(t),T\delta V_{\max}] & \mathrm{if}\ k> T
\end{cases}.
\end{equation}
\end{definition}

\begin{assumption}
\label{c3_AS4}
The target point ${\cal T}$ belongs to the set ${\cal R}[{\cal A}(t),d_s+T\delta V_{\max}]$ at any time $t$.
\end{assumption}

\begin{assumption}
\label{c3_AS5}
The set ${\cal R}[{\cal A}(t),d_s+T\delta V_{\max}]$ is a connected set at any time $t$.
\end{assumption}

\begin{figure}[!htb]
\centering
\epsfig{figure=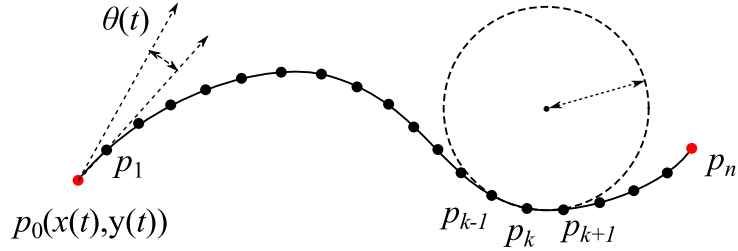,width=10cm}
\caption{
Robot path represented as some equally spaced points.
}
\label{fig:c3_ad13}
\end{figure}

Let $p_0$ denote the robot's current position $(x(t),y(t))$ at the current time $t$. A robot path can be approximately represented as some finite equally spaced points $p_0,p_1,\ldots,p_n$ (see Fig. \ref{fig:c3_ad13}). Each point $p_{k}$ represents the position of the robot at the future time $t+k\delta$. The interval between any two successive points is a constant $L$ which equals to $v_r\delta$. According to the non-holonomic constraint, the radius of the circumscribed circle of any three successive points $p_{k-1}$, $p_k$ and $p_{k+1}$ should be $\geq R_{\min}$ and the angle between the robot's current heading $\theta(t)$ and the vector from $p_{0}$ to $p_1$ should be $\leq \arcsin\frac{L}{2R_{\min}}$.

\begin{definition}
\label{c3_DE3}
Let $P=\lbrace p_0,\ldots,p_n\rbrace$ be a robot path at time $t$. For any $k$, if the point $p_k$ belongs to the set ${\cal R}[\hat{\cal A}(t,k),d_s]$, $P$ is called a temporarily safe path. It is guaranteed that the path $P$ is absolutely safe over the time period $[t,t+T\delta]$. 
\end{definition}

\begin{definition}
Let $P=\lbrace p_0,\ldots,p_n\rbrace$ be a robot path. If $\Vert {\cal T}-p_n\Vert\leq L$, the path $P$ is called a target-reaching path. It guides the robot to the target ${\cal T}$.
\end{definition}

Considering the minimum turning radius of the robot, we define two circles called initial circles \cite{Savkin2016} as follows.

\begin{figure}[!htb]
\centering
\epsfig{figure=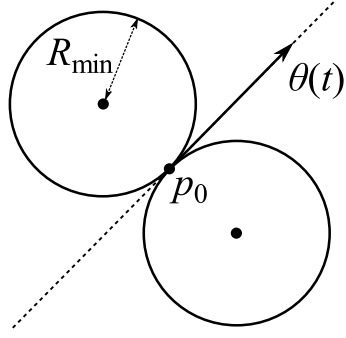,width=6cm}
\caption{
Two initial circles.
}
\label{fig:c3_ad30}
\end{figure}

\begin{definition}
\label{c3_DE1}
The two initial circles are tangent to the robot's current heading $\theta(t)$ and cross the robot's current position $p_0(x(t),y(t))$. The radius of each initial circle is equal to $R_{\min}$ (see Fig. \ref{fig:c3_ad30}).
\end{definition}

Moreover, according to \cite{Roesmann2017}, we give the following definition of homotopic paths.

\begin{definition}
Let $P_1$ and $P_2$ be two paths with same initial point and target. $P_1$ and $P_2$ are homotopic if and only if one can be continuously deformed into the other without intersecting the boundaries of the area ${\cal A}(t)$ and the initial circles. The set of all paths that are homotopic to each other is denoted as homotopy class.
\end{definition}

To generate the relatively short target-reaching path $P^*$, firstly, a path planning algorithm is proposed to adjust a candidate path to an approximate shortest temporarily safe path among the homotopy class of the given path. It is followed by a graph search algorithm that generates several appropriate candidate paths, which belong to different homotopy classes. Combining the graph search algorithm with the path planning algorithm, different candidate paths, which belong to different homotopy classes, are adjusted. Then, the path $P^*$ is selected as the shortest adjusted candidate paths.

\subsection{Path planning algorithm}

Let $P=\lbrace p_0,p_1,\ldots,p_n\rbrace$ be a given candidate path that needs to be adjusted. A path planning algorithm is proposed here to adjust the path $P$ to an approximate shortest temporarily safe path among the homotopy class of the path $P$. Any point of $P$ should be in the unoccupied area ${\cal A}(t)$.

We define four vector fields $F_I$, $F_R$, $F_P$ and $F_C$ in the plane. For any $k\neq 0$, the resultant vector at point $p_k$ is
\begin{equation}
\vec{F}(p_k)=\vec{F}_I(p_k)+\vec{F}_R(p_k)+\vec{F}_P(p_k)+\vec{F}_C(p_k).
\end{equation}
The path $P$ is adjusted by moving any $p_k$, $k\neq 0$ towards the direction of $\vec{F}(p_k)$ to an equilibrium point where $\vec{F}(p_k)=\vec{0}$. Moreover, while adjusting, some path points may be added or removed to prolong or shorten the length of $P$ until $\vec{F}(p_k)=\vec{0}$ for any $k$. Now, we are here to give the definitions of four vector fields.

\begin{figure}[!htb]
\centering
\epsfig{figure=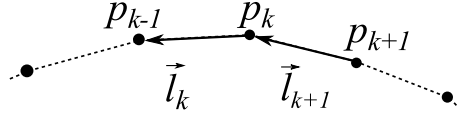,width=6cm}
\caption{
Vector $\vec{l}_k$.
}
\label{fig:c3_3}
\end{figure}

\textbf{Definition of} $\bf F_I$:
$F_I$ is a vector field which guarantees that the interval between any two successive points of $P$ is approximately equal to $L$. For any $k\neq 0$, let $\vec{l}_k$ be the vector from $p_k$ to $p_{k-1}$ (see Fig. \ref{fig:c3_3}), then $\vec{F}_I(p_{k})$ is defined as follows:

\begin{equation}
\vec{F}_{I}(p_{k}):=
\begin{cases}
G_I\cdot (b(k)\vec{l}_k-b(k+1)\vec{l}_{k+1}) & \mathrm{if}\ k\neq n
\\
G_I\cdot (b(k)\vec{l}_k) & \mathrm{if}\ k=n
\end{cases},
\end{equation}
where $G_I>0$ is a tunable gain and $b(k)$ is

\begin{equation}
b(k):=1-\frac{L}{\Vert \vec{l}_k \Vert}.
\end{equation}

\begin{figure}[!htb]
\centering
\epsfig{figure=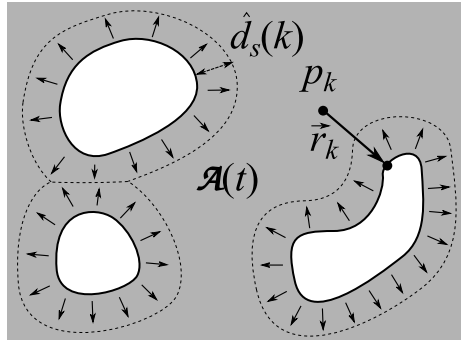,width=6cm}
\caption{
Vector field $F_R$ and vector $\vec{r}_k$.
}
\label{fig:c3_ad21}
\end{figure}

\textbf{Definition of} $\bf F_R$:
$F_R$ is a vector field which guarantees that the minimum distance from the point $p_k$ to the boundary of $\hat{\cal A}(t,k)$ is $\geq d_s$ for any $k\neq 0$. Let $\vec{r}_k$ be the shortest vector from $p_{k}$ to the boundary of ${\cal A}(t)$, then $\vec{F}_R(p_{k})$ is defined as follows (see Fig. \ref{fig:c3_ad21}):

\begin{equation}
\vec{F}_R(p_{k}):=
\begin{cases}
G_R\cdot (1-\frac{\hat{d}_s(k)}{\Vert\vec{r}_k\Vert}) \vec{r}_k & \mathrm{if}\ \Vert\vec{r}_k\Vert\leq \hat{d}_s(k)
\\
\vec{0} & \mathrm{if}\ \Vert\vec{r}_k\Vert > \hat{d}_s(k)
\end{cases},
\end{equation}
where $G_R>0$ is a tunable gain and $\hat{d}_s(k)$ is

\begin{equation}
\hat{d}_s(k):=
\begin{cases}
d_s+k\delta V_{\max} & \mathrm{if}\ k\leq T
\\
d_s+T\delta V_{\max} & \mathrm{if}\ k> T
\end{cases}.
\end{equation}

\begin{figure}[!htb]
\centering
\epsfig{figure=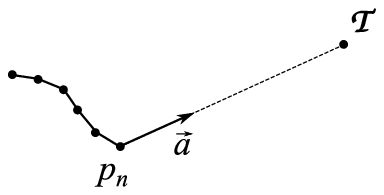,width=6cm}
\caption{
Vectors $\vec{a}$.
}
\label{fig:c3_5}
\end{figure}

\textbf{Definition of} $\bf F_P$:
$F_P$ is a vector field that only acts on the last point $p_n$. For any other $k\neq n$, $\vec{F}_P(p_k)=\vec{0}$. For the point $p_n$, let $\vec{a}$ be a unit vector pointing towards ${\cal T}$ from $p_{n}$ (see Fig. \ref{fig:c3_5}), then $\vec{F}_P(p_{n})$ is defined as follows:

\begin{equation}
\vec{F}_P(p_{n}):=G_P\cdot \vec{a},
\end{equation}
where $G_P>0$ is a tunable gain.

\begin{figure}[!htb]
\centering
\epsfig{figure=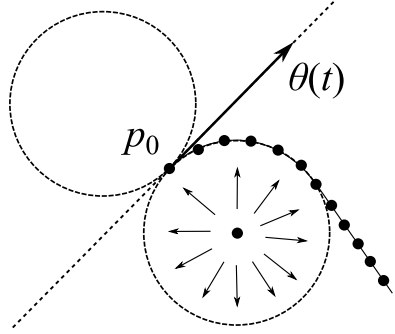,width=6cm}
\caption{
Vector field $F_C$.
}
\label{fig:c3_ad33}
\end{figure}

\textbf{Definition of} $\bf F_C$:
$F_C$ is a vector field which guarantees that the path satisfies the non-holonomic constraint at the beginning. For any $k$, let $\vec{h}_k$ be the vector from the point $p_k$ to the centre of the initial circle which is to the same side of the robot's heading direction with the point $p_1$ (see Fig. \ref{fig:c3_ad33}), then $\vec{F}_C(p_{k})$ is defined as follows:

\begin{equation}
\vec{F}_C(p_{k}):=
\begin{cases}
G_C\cdot (1-\frac{R_{\min}}{\Vert\vec{h}_k\Vert}) \vec{h}_k & \mathrm{if}\ \Vert\vec{h}_k\Vert\leq R_{\min}
\\
\vec{0} & \mathrm{if}\ \Vert\vec{h}_k\Vert > R_{\min}
\end{cases},
\end{equation}
where $G_C>0$ is a tunable gain.

Now, we are here to propose the path planning algorithm as follows to adjust the given candidate path $P$:

\begin{enumerate}
\item[\textbf{A1}:]
For each point $p_k\in P$, $k\neq 0$, let $\vec{p}_k$ denote the position vector of $p_k$ and initialize a velocity vector $\vec{v}_k=\vec{0}$ for $p_k$.

\item[\textbf{A2}:]
Start the following loop:

\begin{enumerate}
\item[\textbf{A2.1}:]
Calculate the resultant vector $\vec{F}(p_k)$ for any point $p_k$.

\item[\textbf{A2.2}:]
Change the position vector and velocity vector of each $p_k$ as follows:

\begin{equation}
\left\{
\begin{array}{l}

\vec{p}_{k} \leftarrow \vec{p}_{k}+\vec{v}_{k}
\\
\vec{v}_{k} \leftarrow G_N \cdot \vec{v}_{k}+\vec{F}(p_k)

\end{array},
\right.
\end{equation}
where $0<G_N<1$ is a tunable attenuation.

\item[\textbf{A2.3}:]
Let $p_\text{last}$ denote the last point of $P$. If $\Vert p_\text{last}-{\cal T}\Vert<\underline{L}$, remove $p_\text{last}$ from $P$. If $\Vert p_\text{last}-{\cal T}\Vert>\overline{L}$, add a new path point next to $p_\text{last}$ in $P$. Here $\underline{L}$ and $\overline{L}$ are two given constants, which satisfy the inequalities $0<\underline{L}<L$ and $L<\overline{L}<2L$.

\end{enumerate}

\item[\textbf{A3}:]
Exit loop \textbf{A2} if the inequality $\Vert \vec{F}(p_k)\Vert<F_{th}$ holds for any $k\neq 0$, where $F_{th}>0$ is a given threshold.
\end{enumerate}

\begin{figure}[!htb]
\centering
\subfigure[The path before adjustment]{

\epsfig{figure=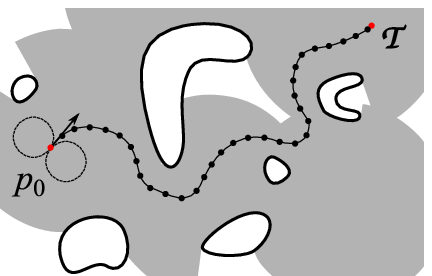,width=6cm}
}
\subfigure[The path after adjustment]{

\epsfig{figure=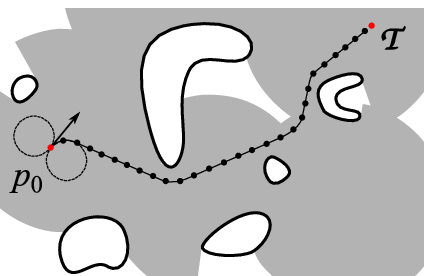,width=6cm}
}
\caption{
Adjustment of a given target-reaching path.
}
\label{fig:c3_ad31}
\end{figure}

According to the proposed path planning algorithm, an approximate shortest temporarily safe path among the homotopy class of the path $P$ can be obtained, which satisfies the criteria defined in Definition \ref{c3_DE3} and the non-holonomic constraint (\ref{c3_3}); e.g. see Fig. \ref{fig:c3_ad31}.

\subsection{Candidate paths generation}

A graph search algorithm is proposed here to generate some target-reaching candidate paths belonging to the different homotopy classes. Then, these candidate paths are adjusted according to the algorithm \textbf{A1}-\textbf{A3} to search the path $P^*$. Firstly we introduce a graph as follows.

The boundary of the $d_s$-reduced area ${\cal R}[{\cal A}(t),d_s]$ can be represented as the union of some simple closed curves denoted by $C_0,C_1,\ldots$. Let $C_0$ denote the largest curve which encircles other curves $C_1,C_2,\ldots$.

\begin{figure}[!htb]
\centering
\subfigure[$(AT)$-segment]{
\label{fig:c3_ad37-1}
\epsfig{figure=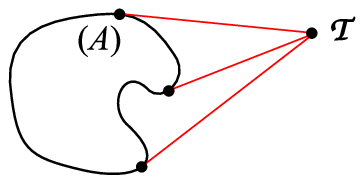,width=4cm}
}
\subfigure[$(AA')$-segment]{
\label{fig:c3_ad37-2}
\epsfig{figure=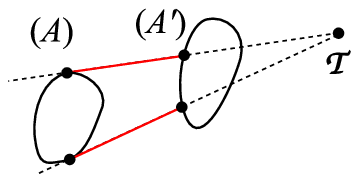,width=4cm}
}
\subfigure[$(S)$-point]{
\label{fig:c3_ad37-3}
\epsfig{figure=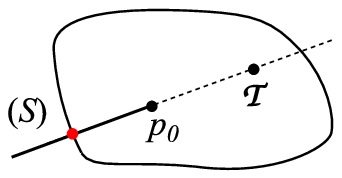,width=4cm}
}
\subfigure[$(BT)$-segment]{
\label{fig:c3_ad37-4}
\epsfig{figure=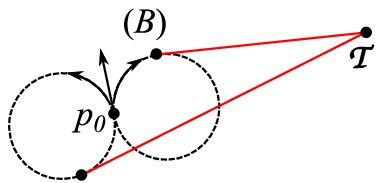,width=4cm}
}
\subfigure[$(BB')$-segment]{
\label{fig:c3_ad37-5}
\epsfig{figure=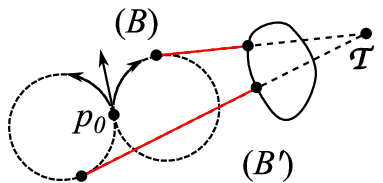,width=4cm}
}
\subfigure[$(V)$-point]{
\label{fig:c3_ad37-6}
\epsfig{figure=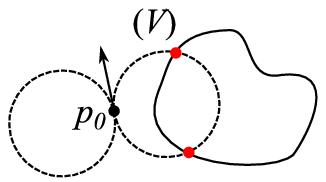,width=4cm}
}
\caption{
Different types of segments and points.
}
\label{fig:c3_ad37}
\end{figure}

\begin{definition}
For any $i$, if a line passing through the target ${\cal T}$ is tangent to $C_i$ and the segment between the tangent point and the target ${\cal T}$ does not intersect with any curve, the tangent point is called an $(A)$-point and the segment is called an $(AT)$-segment (see Fig. \ref{fig:c3_ad37-1}); if the segment between the tangent point and the target ${\cal T}$ intersects with any $C_j$, $i\neq j$, the tangent point is also called an $(A)$-point and the closest point of intersection to the $(A)$-point is called an $(A')$-point; the segment between a pair of $(A)$-point and $(A')$-point is called an $(AA')$-segment (see Fig. \ref{fig:c3_ad37-2}). Furthermore, a ray that its endpoint is the robot's position $p_0(x(t),y(t))$ and its direction is opposite to the target ${\cal T}$ intersects with the curve $C_0$ at some points which are called $(S)$-points (see Fig. \ref{fig:c3_ad37-3}).

According to Definition \ref{c3_DE1}, there exist two initial circles. For each initial circle, there exists one tangent line which passes through the target ${\cal T}$ and is able to be tracked by the robot to exit the initial circle from the tangent point. If the tangent point is not encircled by any curve $C_i$, $i\neq 0$, it is called a $(B)$-point. If the segment between the $(B)$-point and the target ${\cal T}$ does not intersect with any curve, the segment is called a $(BT)$-segment (see Fig. \ref{fig:c3_ad37-4}); otherwise the closest point of intersection to the $(B)$-point is called a $(B')$-point and the segment between a pair of $(B)$-point and $(B')$-point is called a $(BB')$-segment (see Fig. \ref{fig:c3_ad37-5}). Moreover, if the initial circles intersect with any curve, the points of intersection are called $(V)$-points (see Fig. \ref{fig:c3_ad37-6}).
\end{definition}

\begin{figure}[!htb]
\centering
\epsfig{figure=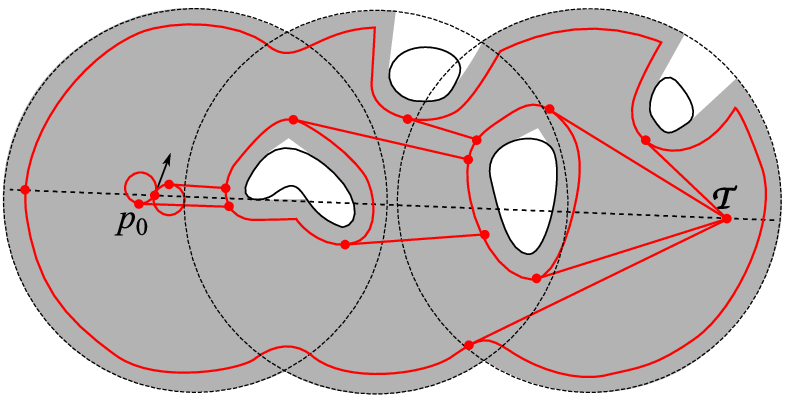,width=10cm}
\caption{
Graph ${\cal G}$.
}
\label{fig:c3_ad34}
\end{figure}

\begin{definition}
A graph denoted by ${\cal G}$ is introduced that its vertices are the target ${\cal T}$, robot's position $p_0(x(t),y(t))$, the points of $(A)$, $(A')$, $(B)$, $(B')$, $(S)$ and $(V)$ types. Its edges are the segments of the curves $C_0,C_1,\ldots$, the arc of the initial circles and the segments of $(AT)$, $(AA')$, $(BT)$ and $(BB')$ types (see Fig. \ref{fig:c3_ad34}).
\end{definition}

Now, we propose the candidate paths generation rules as follows to generate some candidate paths iteratively according to the graph ${\cal G}$:

\begin{enumerate}
\item
At the beginning, initialize two candidate paths, each of which only includes the point $p_0$. Generate new points along the different initial circle to prolong each candidate path respectively until the candidate path meets a $(B)$-point or a $(V)$-point.

\item
If a candidate path meets an $(A)$-point or a $(B)$-point, continue generating new points along the corresponding segment of $(AA')$, $(AT)$, $(BB')$ or $(BT)$ type until the candidate path meets an $(A')$-point, a $(B')$-point or the target ${\cal T}$.

\item
If a candidate path meets an $(A')$-point or a $(B')$-point, generate a new candidate path with the duplicate points of this candidate path, then continue generating new points of these two candidate paths along the curve in different directions until each candidate path meets an $(A)$-point or an $(S)$-point.

\item
If a candidate path meets a $(V)$-point, continue generating new points along the curve in the same direction with the initial circle (clockwise or anti-clockwise direction) until the candidate path meets an $(A)$-point or an $(S)$-point.

\item
If a candidate path meets an $(S)$-point, this candidate path is abandoned.

\item
If a candidate path meets the target ${\cal T}$, the generation of this candidate path is completed.
\end{enumerate}

\begin{figure}[!htb]
\centering
\subfigure[]{
\epsfig{figure=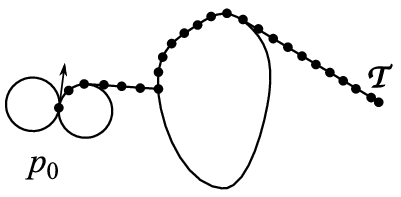,width=5cm}
}
\subfigure[]{
\epsfig{figure=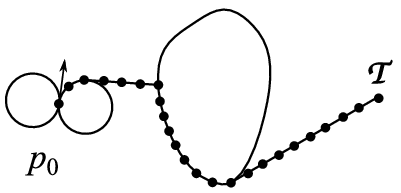,width=5cm}
}
\subfigure[]{
\epsfig{figure=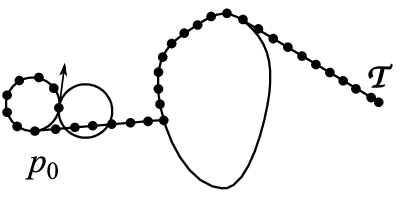,width=5cm}
}
\subfigure[]{
\epsfig{figure=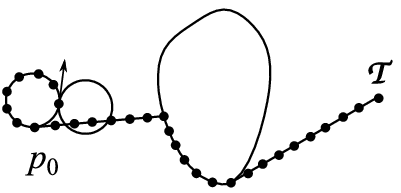,width=5cm}
}
\caption{
Target-reaching candidate paths generated by the graph ${\cal G}$.
}
\label{fig:c3_ad35}
\end{figure}

By following the paths generation rules, some target-reaching candidate paths can be generated iteratively according to the graph ${\cal G}$ (e.g. Fig. \ref{fig:c3_ad35}). Then, all the candidate paths are adjusted according to the proposed path planning algorithm \textbf{A1}-\textbf{A3}. Finally, the shortest adjusted path is selected as the path $P^*$, which should be tracked by the robot.

\begin{remark}
Instead of adjusting all the candidate paths, we can only select and adjust the shortest candidate path. It reduces the computation of the navigation algorithm in practical implementations.
\end{remark}

Notice that there exist two cases, in which the adjustment of a candidate path cannot be successful and the corresponding candidate path should be abandoned. The first case is that the vector field $F_C$ clashes with the field $F_R$. The second case is that the field $F_R$ clashes with itself.

\section{Computer simulations}

In this section, computer simulations are carried out to confirm the performance of the proposed navigation algorithm in static and dynamic environments. In the presented computer simulations, the practical scenes are simulated with static and moving obstacles. In the scenes, the static obstacles can be walls, machines and equipments and the moving obstacles can be walking people and other moving robots. The presented simulations focus on the navigation of a single robot to assess the performance of the proposed navigation algorithm. In general cases with multiple robots, any robots can be navigated simultaneously to the destination by the sensor network.

\begin{figure}[!htb]
\centering
\epsfig{figure=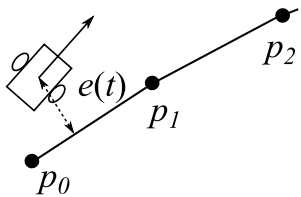,width=4cm}
\caption{
Minimum distance $e(t)$.
}
\label{fig:c3_ad27}
\end{figure}

Firstly, to drive the ground mobile robot to track the generated path $P^*$ during each sampling interval, we are here to give a simple control strategy, which is a modification of the control law proposed in \cite{Matveev2011a}. Let $e(t)$ be the minimum distance from the robot's position to the path $P^*$, which is a polygonal chain with vertices $p_0,p_1,\ldots,p_n$ belonging to $P^*$ (see Fig. \ref{fig:c3_ad27}), and make the minimum distance positive if the closest point on the polygonal chain to the robot's position is in the upper half-plane of the robot's local coordinate system and negative if in the lower half-plane. Let $\lambda>0$ and $\sigma>0$ be tunable constants. The angular velocity of the robot is controlled as follows:

\begin{equation}
\label{c3_2}
u(t)=u_M\cdot \text{sgn}[\dot{e}(t)+{\cal X}(e(t))].
\end{equation}
The saturation function ${\cal X}(z)$ is

\begin{equation}
{\cal X}(z):=
\begin{cases}
\lambda z & \mathrm{if}\ \vert z\vert\leq\sigma
\\
\lambda \sigma\text{sgn}(z) & \mathrm{if}\ \vert z\vert>\sigma
\end{cases}
\end{equation}
and the sign function $\text{sgn}(x)$ is

\begin{equation}
\text{sgn}(x)=
\begin{cases}
1 & \mathrm{if}\ x>0
\\
0 & \mathrm{if}\ x=0
\\
-1 & \mathrm{if}\ x<0
\end{cases}.
\end{equation}

Now, we are here to carry out four simulations in different scenes including static and dynamic environments. To reduce the computational load, only the shortest candidate path generated according to the graph ${\cal G}$ is selected and adjusted, instead of all the candidate paths.

\subsection{Simulations in static environments}

To confirm the performance of the proposed navigation algorithm in static industrial environments, we build two static scenes with same static obstacles but different deployment of the sensor nodes.

\begin{figure}[!htb]
\centering
\epsfig{figure=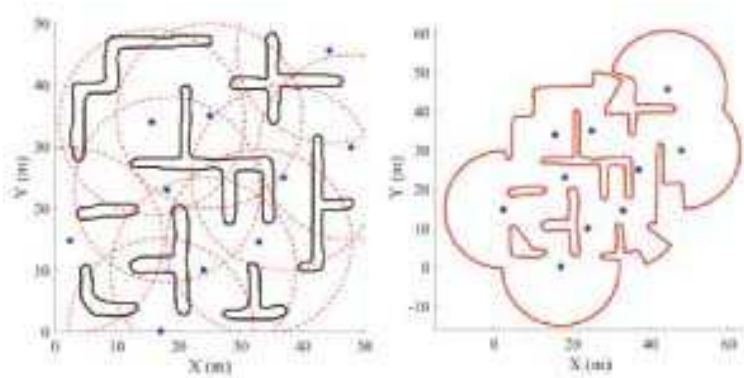,width=10cm}
\caption{
First scene and the boundaries of the unoccupied area. The asterisks are sensor nodes. The dashed lines are the measurement range of each sensor node.
}
\label{fig:c3_s1-1}
\end{figure}

\begin{figure}[!htb]
\centering
\epsfig{figure=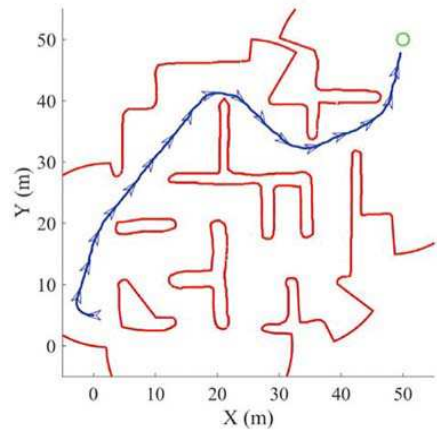,width=6cm}
\caption{
Simulation in the first scene.
}
\label{fig:c3_s1-3}
\end{figure}

\begin{figure}[!htb]
\centering
\epsfig{figure=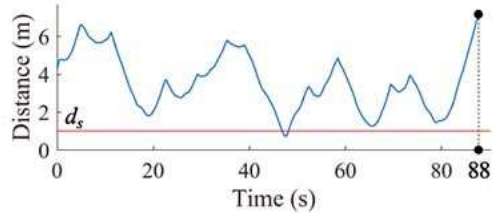,width=7cm}
\caption{
Minimum distance to the undetected areas in the first scene.
}
\label{fig:c3_s1-4}
\end{figure}

In the first scene, there are some static, non-convex and irregular-shaped obstacles. To detect these obstacles, there are some range finder sensors deployed in the scene (see Fig. \ref{fig:c3_s1-1}). The main parameters in this simulation are indicated in Table \ref{tb:c3_1}. The parameters $V_{\max}$ and $T$ are $0.5\text{m/s}$ and $4$. In this scene, a robot moves from the initial position to a target (see Fig. \ref{fig:c3_s1-3}). It can be seen that the robot's trajectory is relatively short. During the travelling, the robot avoids all the obstacles and undetected areas successfully and keeps the given safety margin (see Fig. \ref{fig:c3_s1-4}).

\begin{table}[!htb]
\centering
\caption{Main Parameters for the First Scene}
\label{tb:c3_1}
\begin{tabular}{c|c|c}
\hline
Measurement range & $R_s$ & $15\text{m}$\\
\hline
Speed of robot & $v_r$ & $1\text{m/s}$\\
\hline
Maximum angular velocity & $u_{M}$ & $0.5\text{rad/s}$\\
\hline
Safety margin & $d_s$ & $1\text{m}$\\
\hline
Sampling interval & $\delta$ & $1\text{s}$\\
\hline
\end{tabular}
\end{table}

\begin{figure}[!htb]
\centering
\epsfig{figure=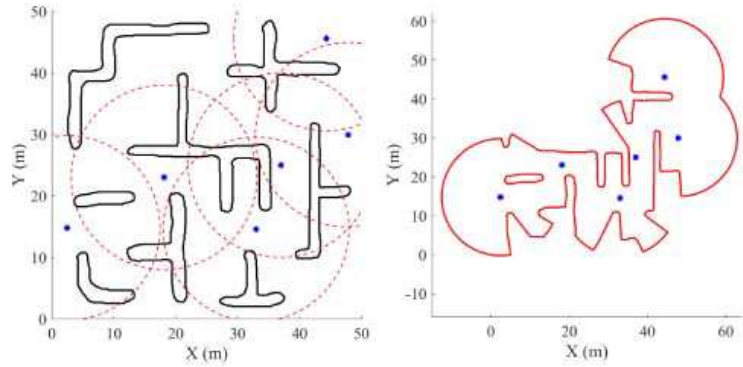,width=10cm}
\caption{
Second scene and the boundaries of the unoccupied area. The asterisks are sensor nodes. The dashed lines are the measurement range of sensor nodes.
}
\label{fig:c3_s2-1}
\end{figure}

\begin{figure}[!htb]
\centering
\epsfig{figure=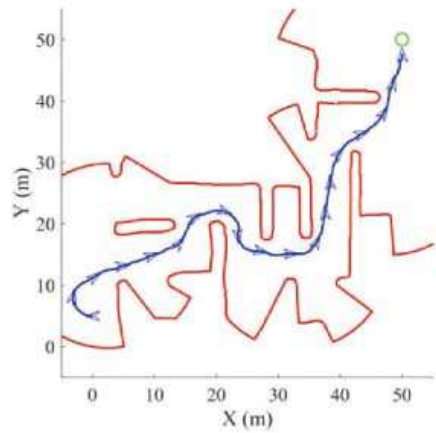,width=6cm}
\caption{
Simulation in the second scene.
}
\label{fig:c3_s2-3}
\end{figure}

\begin{figure}[!htb]
\centering
\epsfig{figure=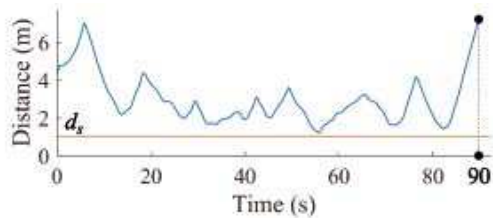,width=7cm}
\caption{
Minimum distance to the undetected areas in the second scene.
}
\label{fig:c3_s2-4}
\end{figure}

In the second scene, there are the same obstacles with the first scene. Comparing with the first scene, less range finder sensors are deployed (see Fig. \ref{fig:c3_s2-1}). All the parameters, robot's initial position and target are same as the first simulation. In this scene, a robot successfully reaches the target with a safe and relatively short trajectory in the unoccupied area (see Fig. \ref{fig:c3_s2-3}). During the travelling, the robot keeps the given safety margin to the undetected areas (see Fig. \ref{fig:c3_s2-4}).

\subsection{Simulations in dynamic environments}

To confirm the performance of the proposed navigation algorithm in dynamic industrial environments, we build another two scenes with moving obstacles. The moving obstacles can be walking people and other robots in the factory.

\begin{figure}[!htb]
\centering
\epsfig{figure=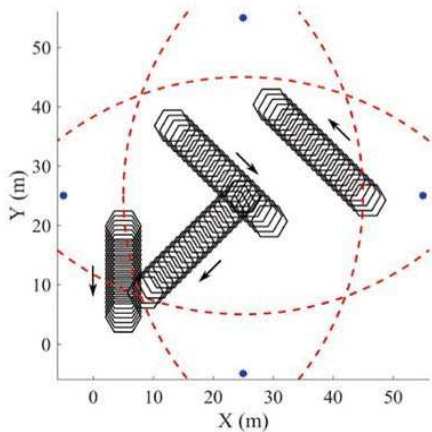,width=6cm}
\caption{
Third scene. The asterisks are sensor nodes. The dashed lines are the measurement range of sensor nodes.
}
\label{fig:c3_s3-1}
\end{figure}

\begin{figure}[!htb]
\centering
\subfigure[$t=20\text{s}$]{
\epsfig{figure=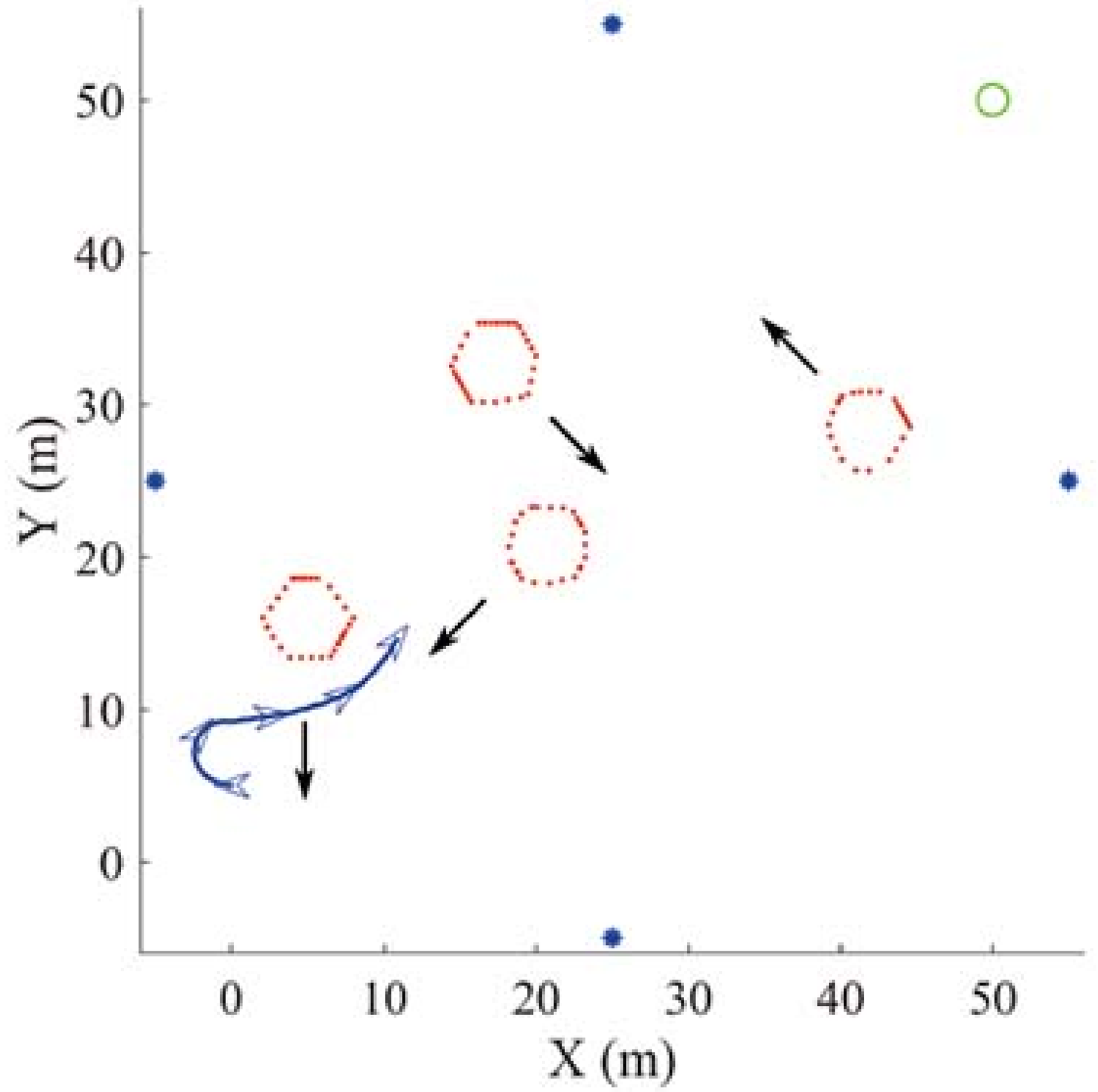,width=5cm}
}
\subfigure[$t=40\text{s}$]{
\epsfig{figure=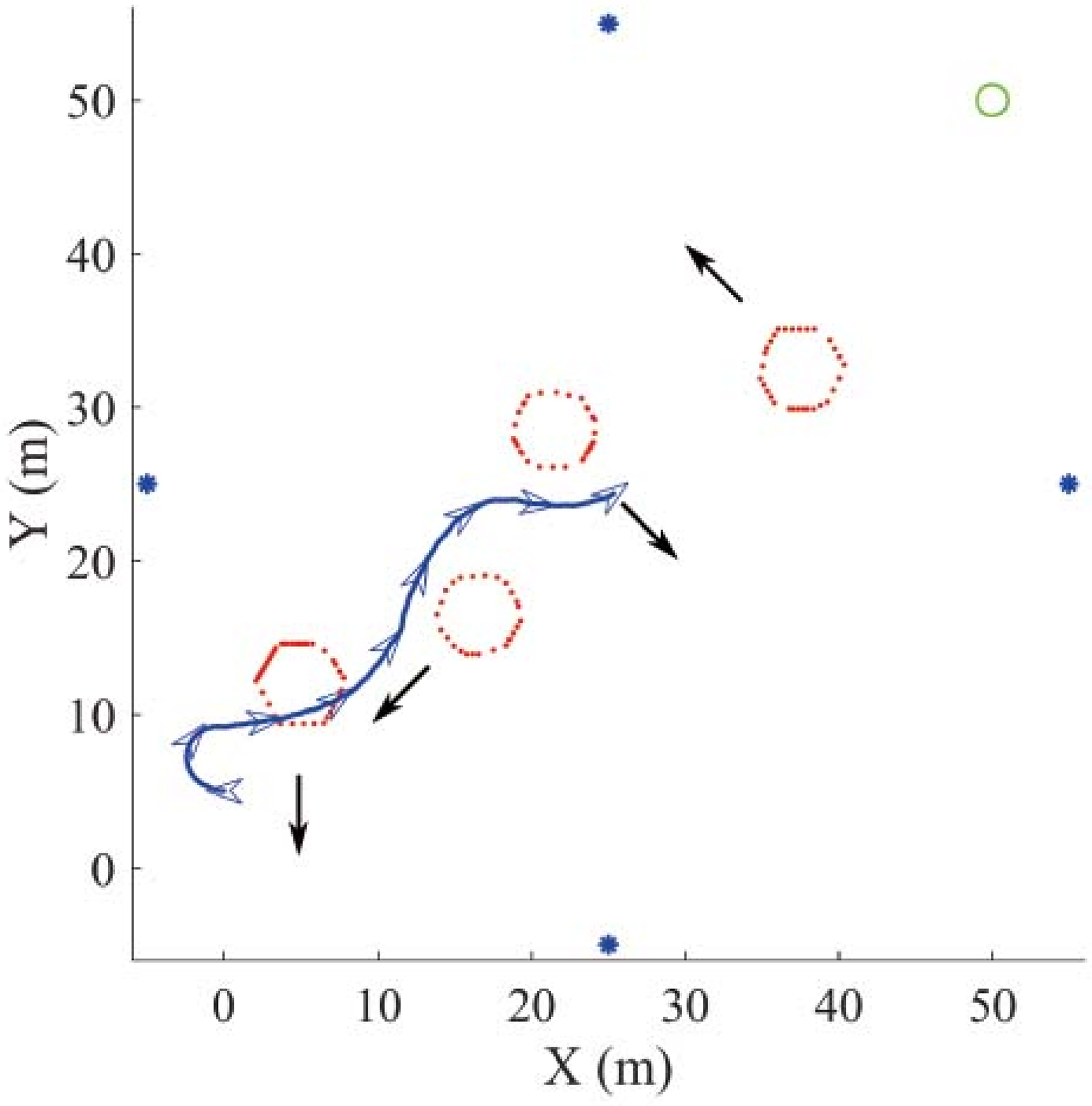,width=5cm}
}
\subfigure[$t=50\text{s}$]{
\epsfig{figure=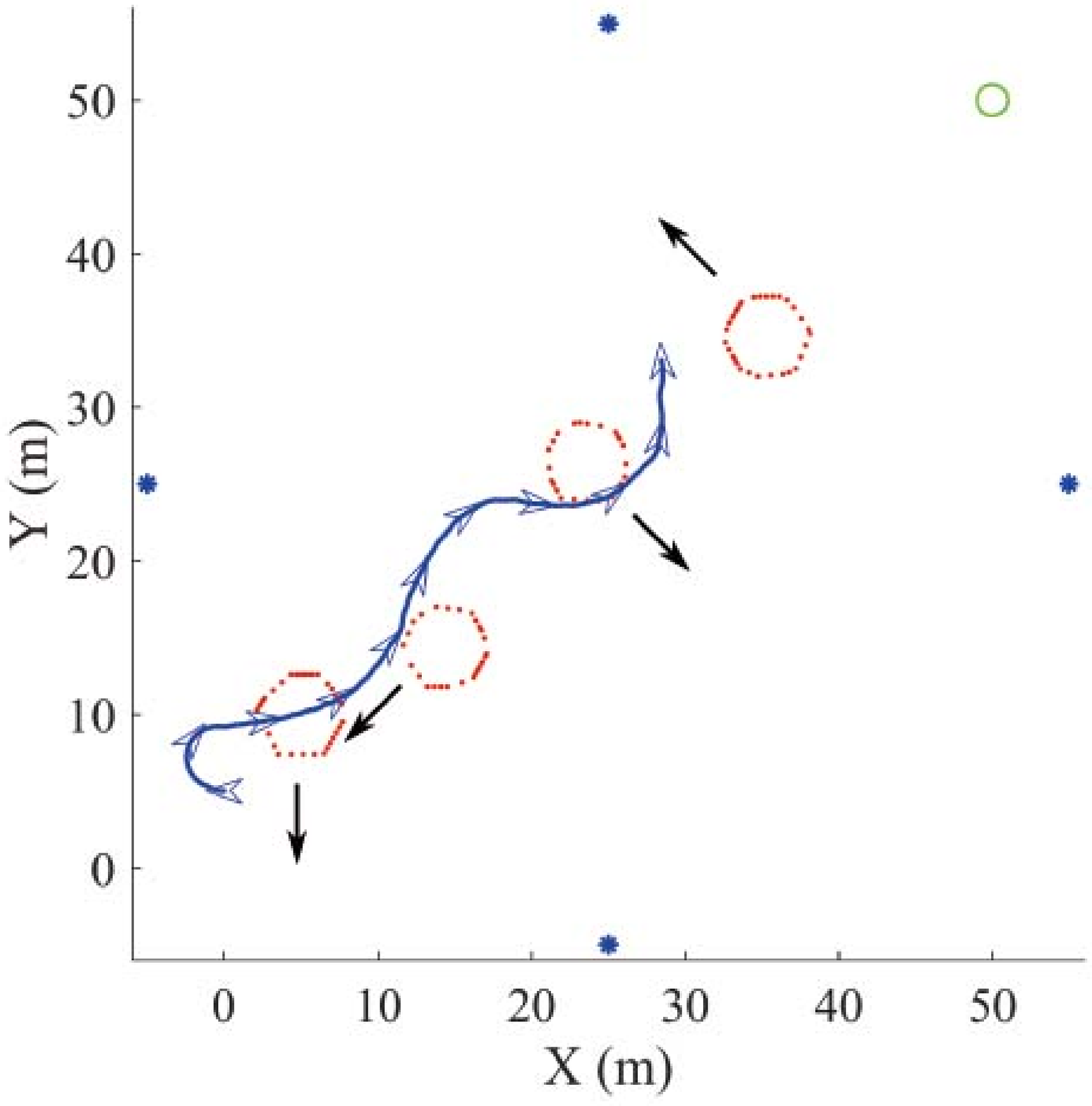,width=5cm}
}
\subfigure[$t=78\text{s}$]{
\epsfig{figure=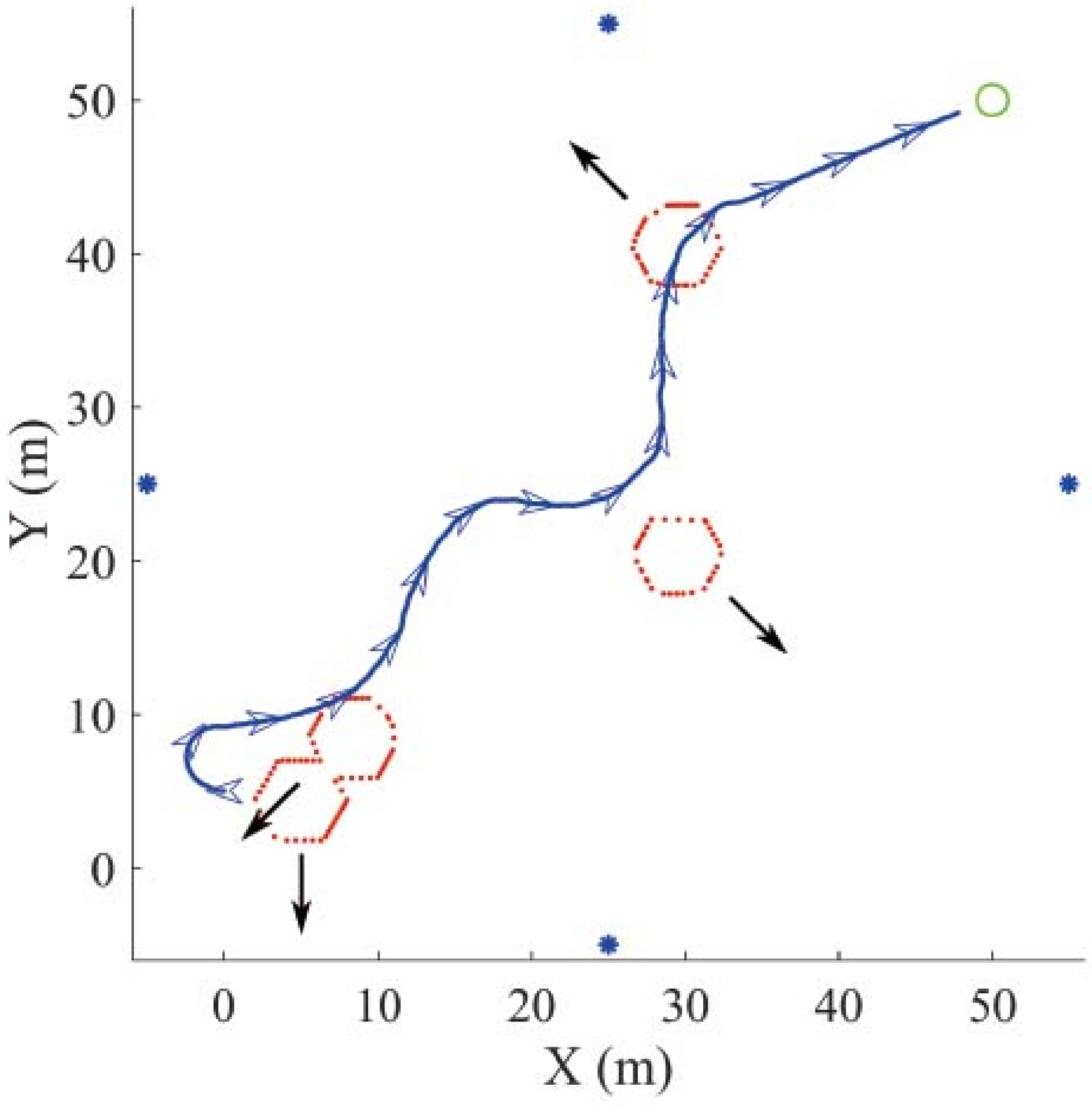,width=5cm}
}
\caption{
Simulation in the third scene.
}
\label{fig:c3_s3-2}
\end{figure}

\begin{figure}[!htb]
\centering
\epsfig{figure=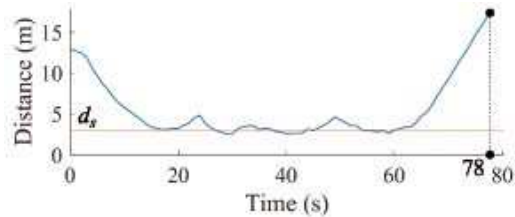,width=7cm}
\caption{
Minimum distance to the undetected areas in the third scene.
}
\label{fig:c3_s3-3}
\end{figure}

In the third scene, there are four obstacles moving in the plane. To detect these obstacles, there are four range finder sensors deployed in the scene (see Fig. \ref{fig:c3_s3-1}). The safety margin $d_s$, the measurement range $R_s$ of the sensor nodes, the parameter $T$ and the maximum speed of obstacle $V_{\max}$ are $3\text{m}$, $50\text{m}$, $7$ and $0.5\text{m/s}$ in this simulation. The other parameters are the same as Table. \ref{tb:c3_1}. In this scene, these obstacles can be detected completely by the sensor network. A robot moves towards a given target in this scene and avoids the moving obstacles (see Fig. \ref{fig:c3_s3-2}). During the travelling, the robot keeps the given safety margin to the obstacles as we expect (see Fig. \ref{fig:c3_s3-3}).

\begin{figure}[!htb]
\centering
\epsfig{figure=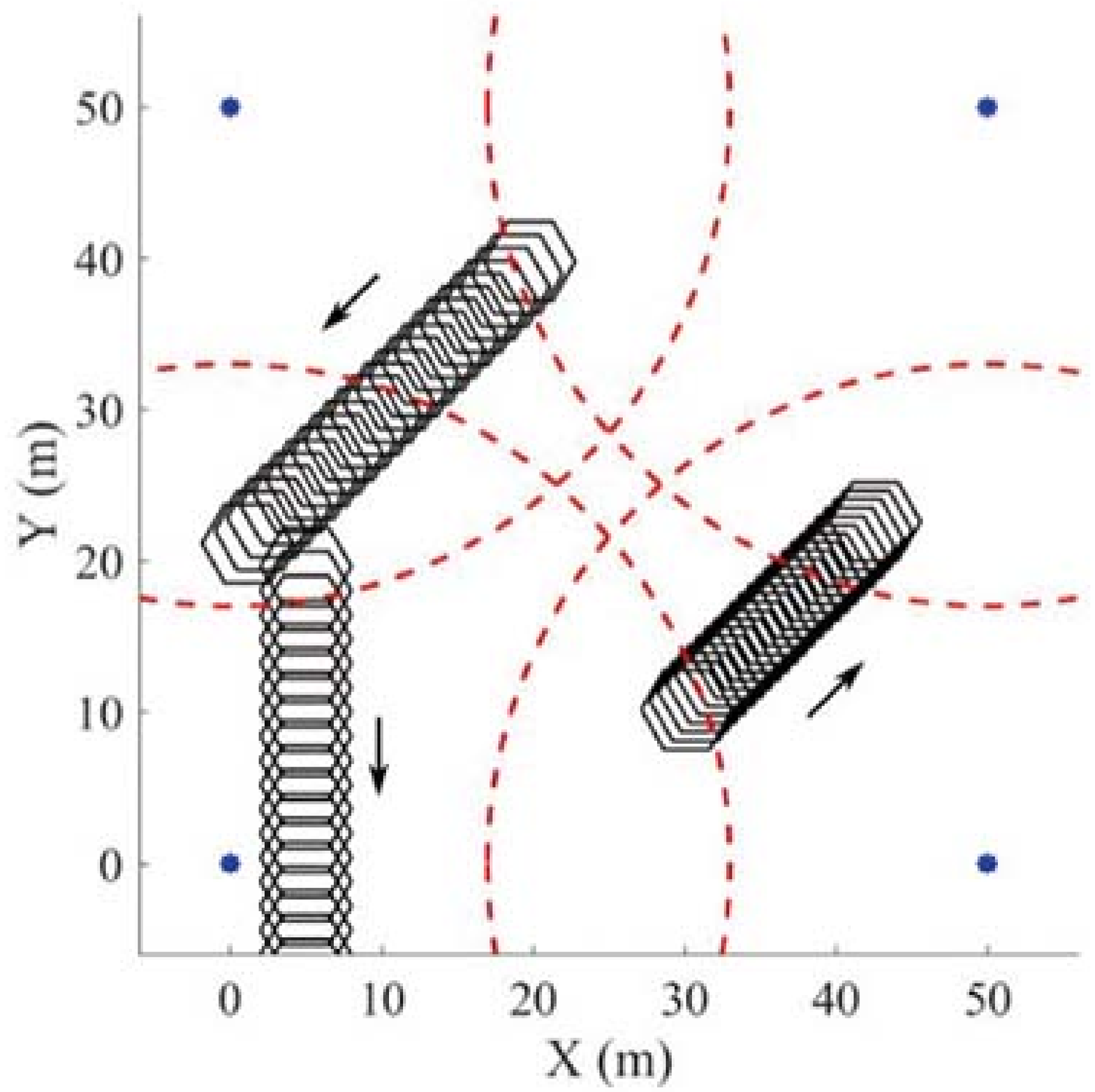,width=6cm}
\caption{
Fourth scene. The asterisks are sensor nodes. The dashed lines are the measurement range of sensor nodes.
}
\label{fig:c3_s4-1}
\end{figure}

\begin{figure}[!htb]
\centering
\subfigure[$t=5\text{s}$]{
\epsfig{figure=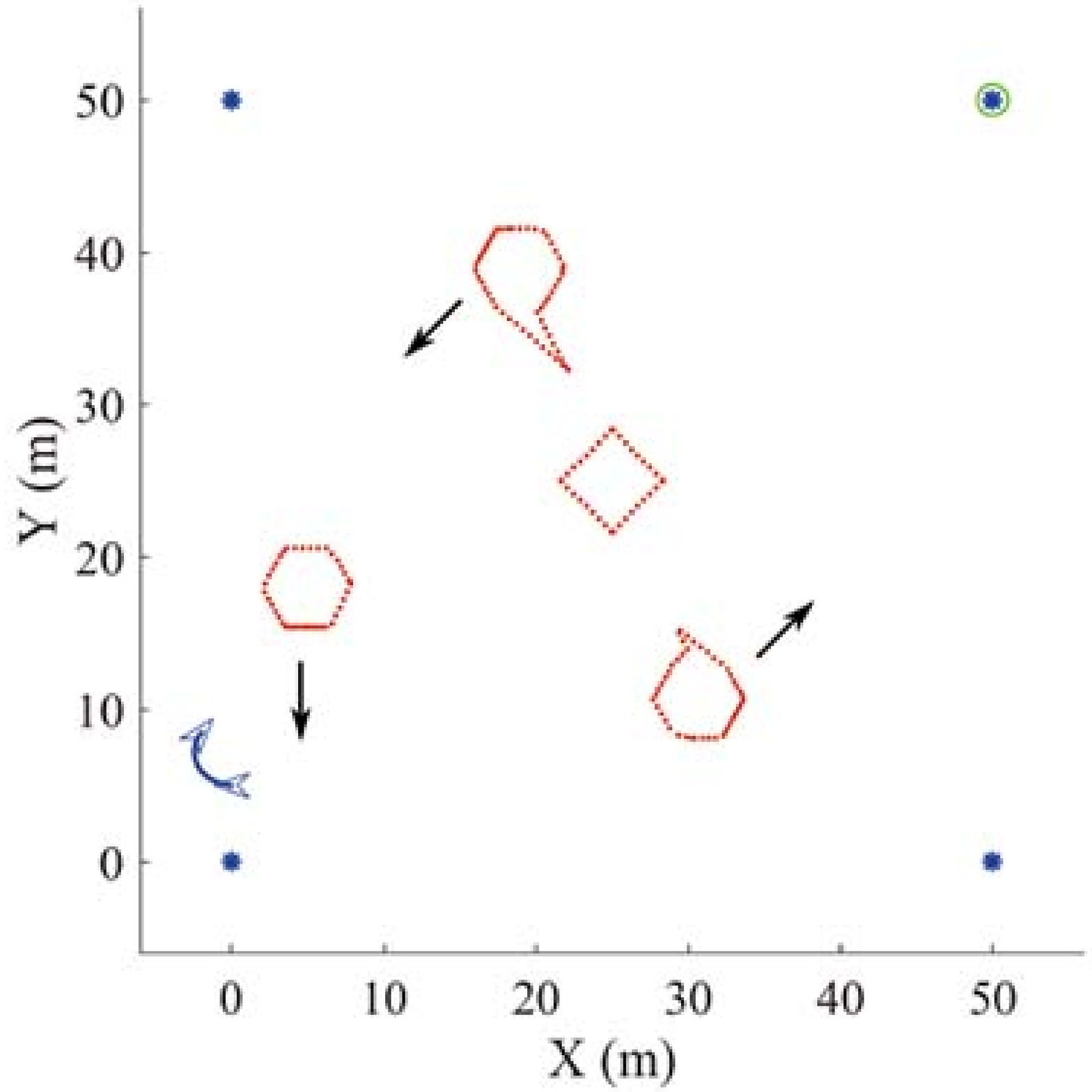,width=5cm}
}
\subfigure[$t=20\text{s}$]{
\epsfig{figure=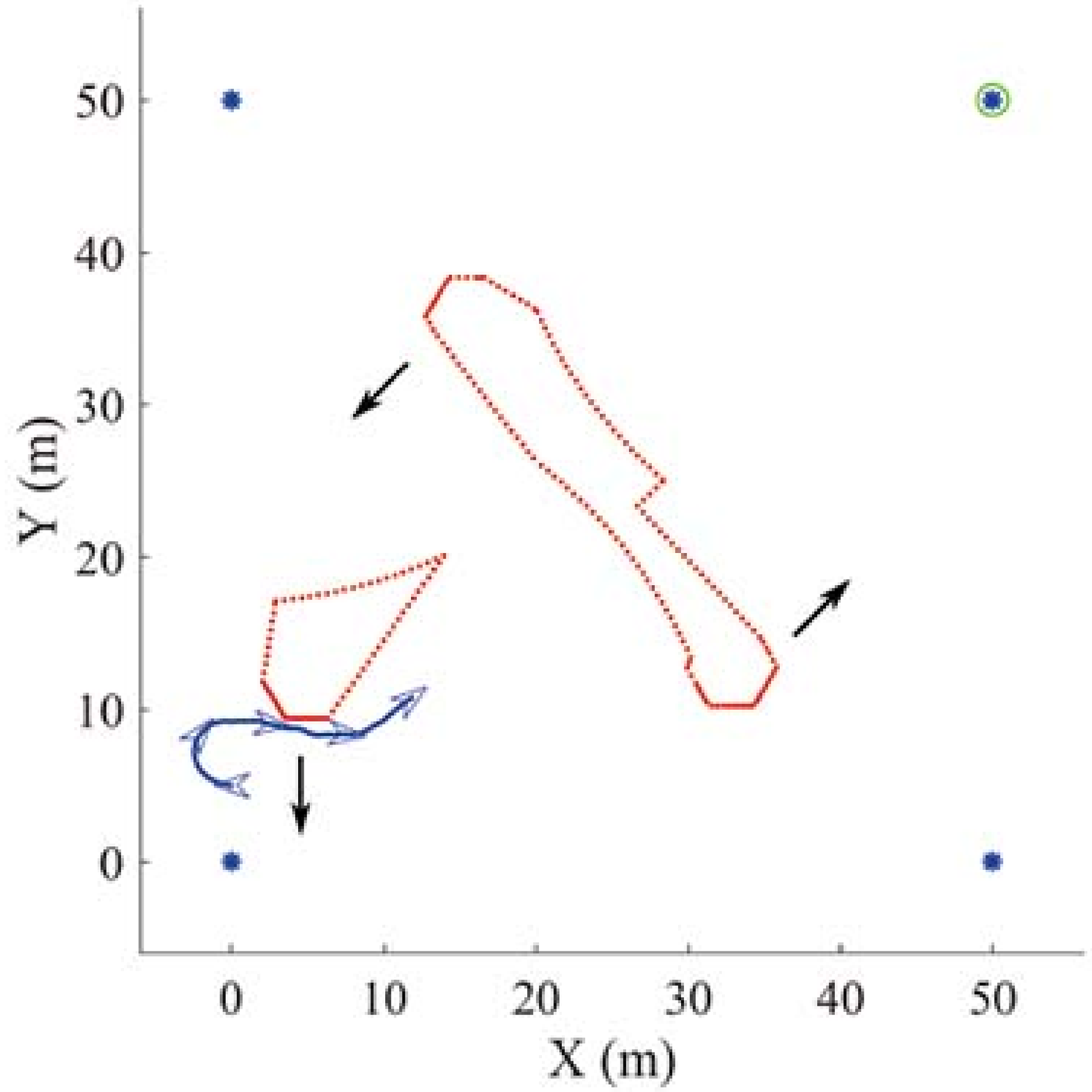,width=5cm}
}
\subfigure[$t=30\text{s}$]{
\epsfig{figure=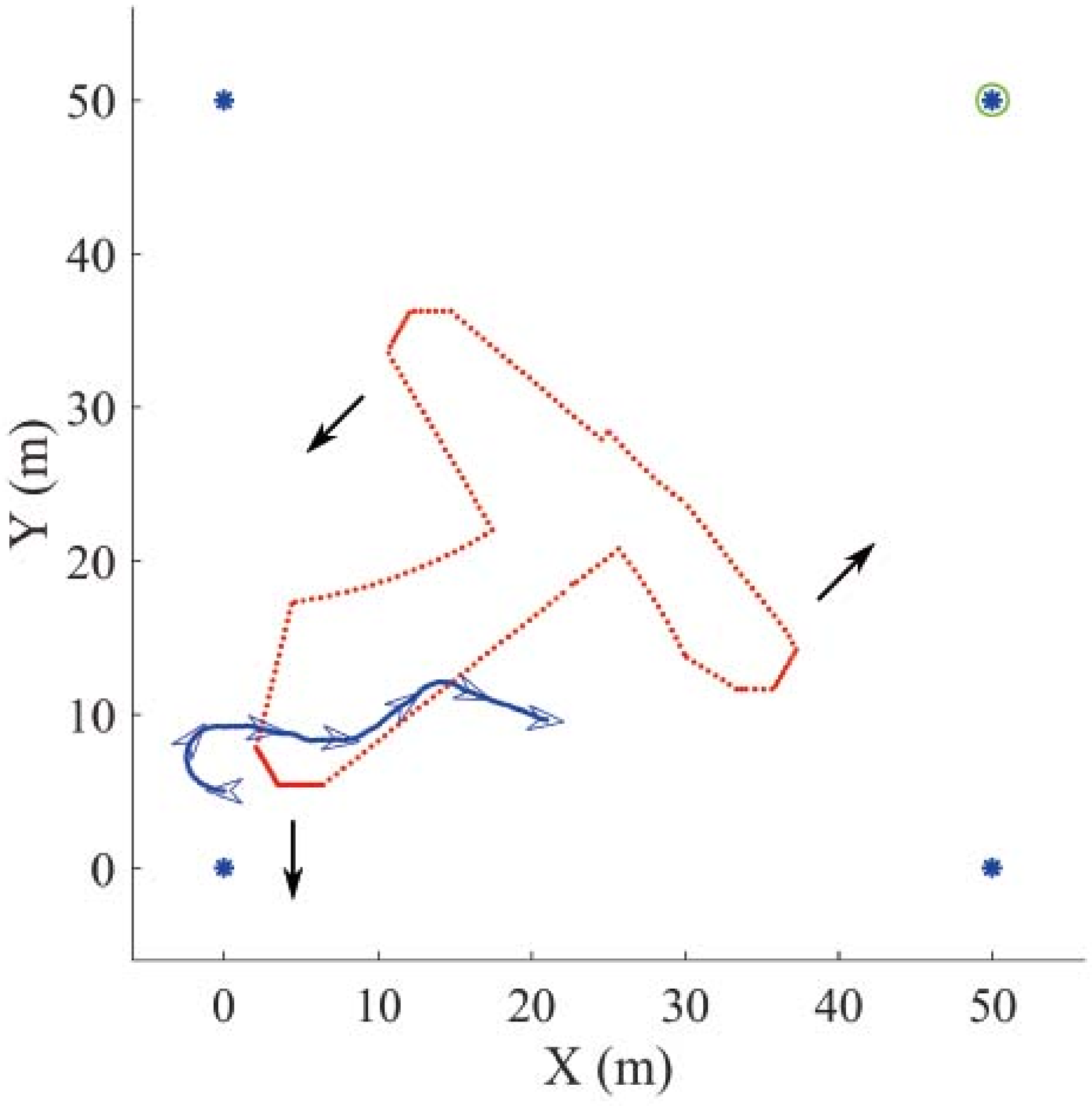,width=5cm}
}
\subfigure[$t=40\text{s}$]{
\epsfig{figure=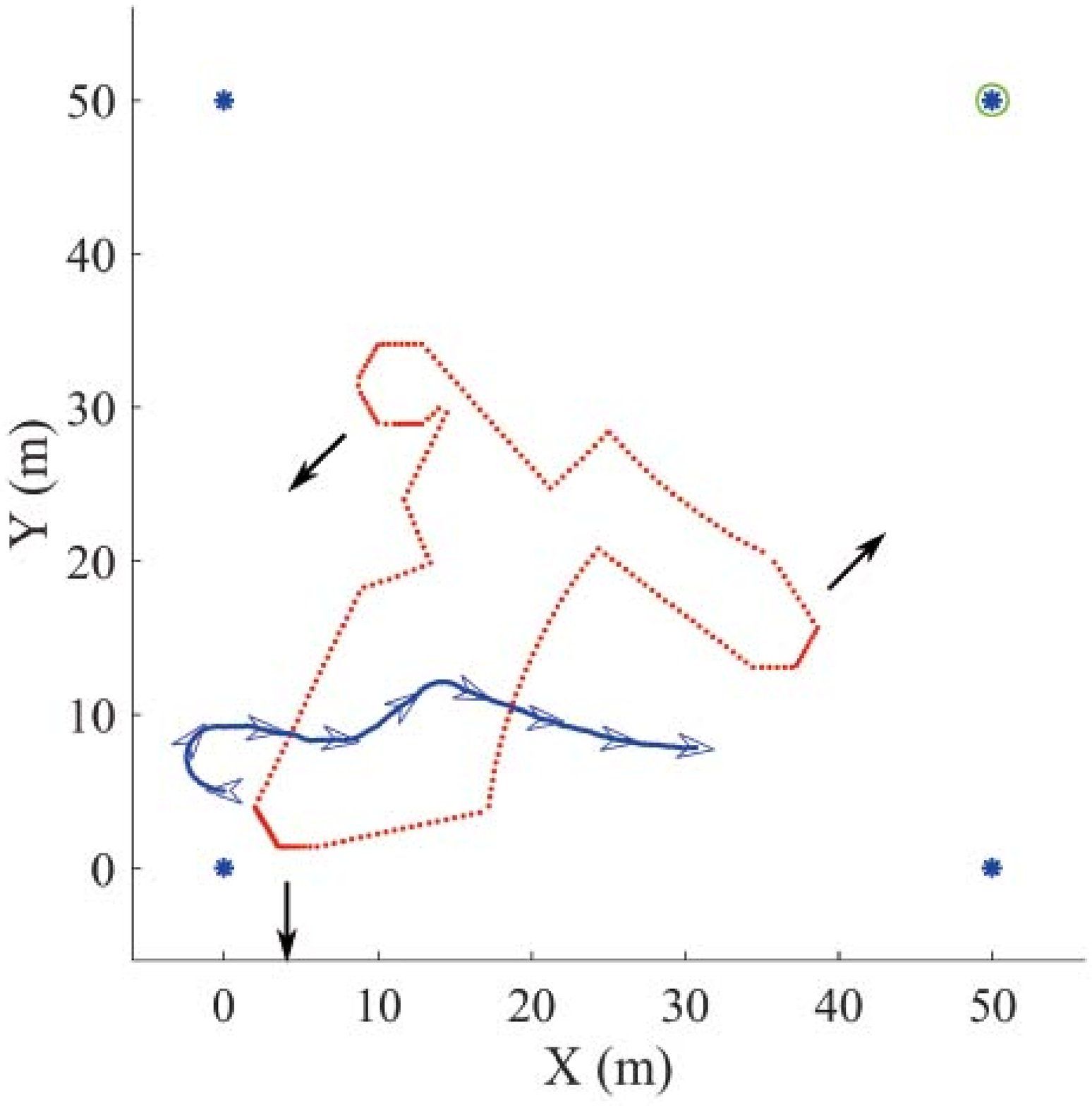,width=5cm}
}
\subfigure[$t=60\text{s}$]{
\epsfig{figure=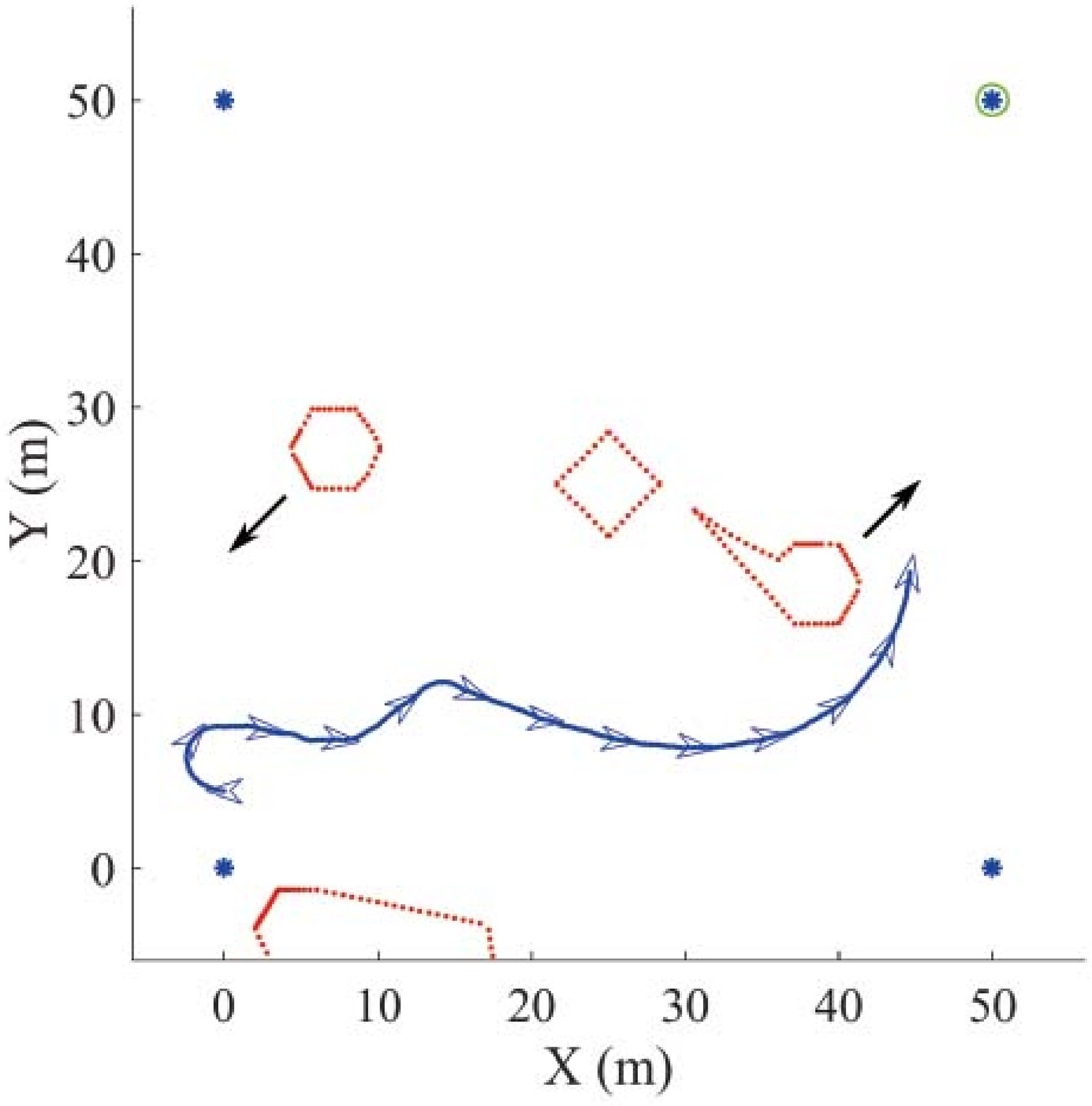,width=5cm}
}
\subfigure[$t=89\text{s}$]{
\epsfig{figure=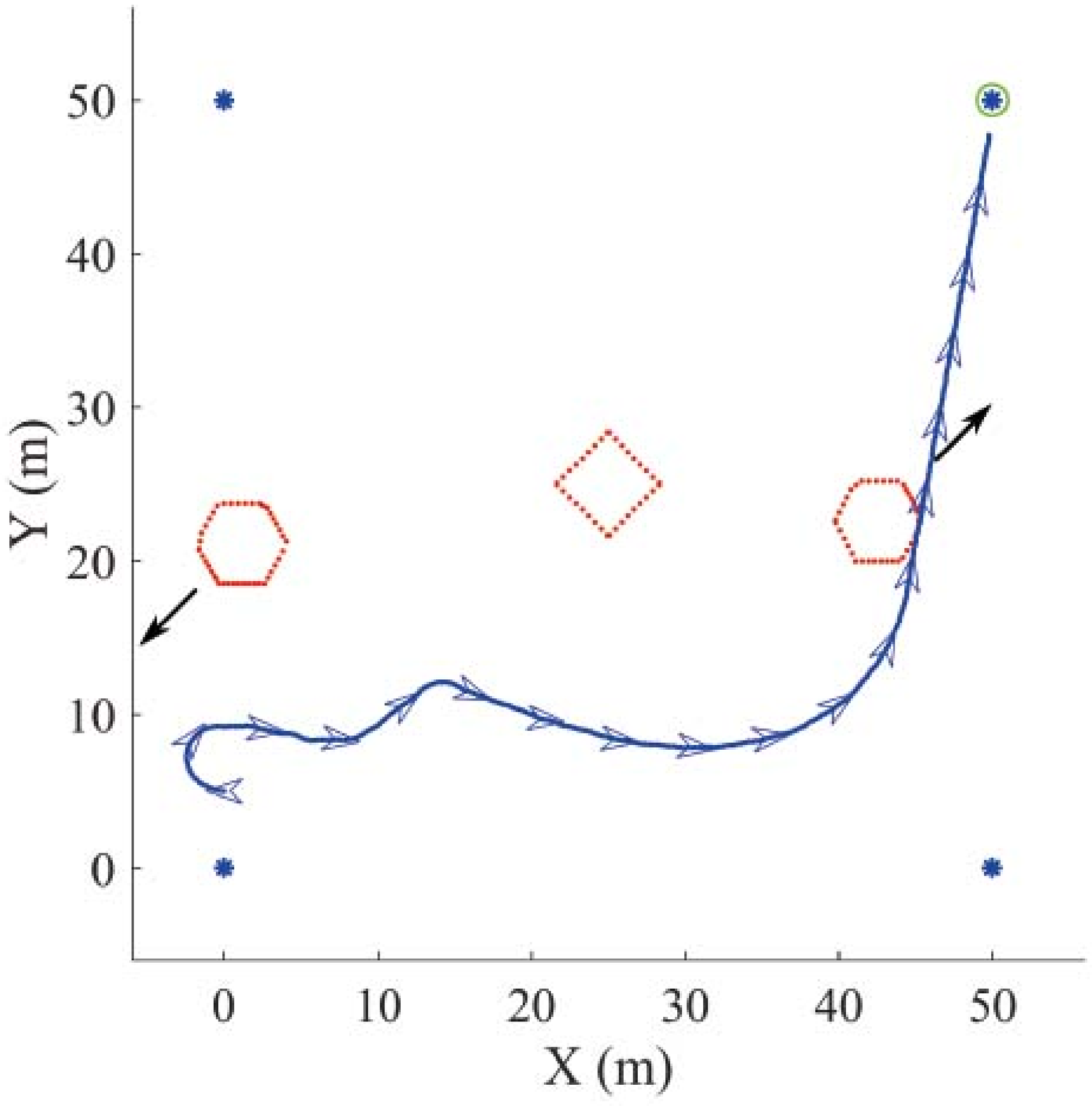,width=5cm}
}
\caption{
Simulation in the fourth scene.
}
\label{fig:c3_s4-2}
\end{figure}

\begin{figure}[!htb]
\centering
\epsfig{figure=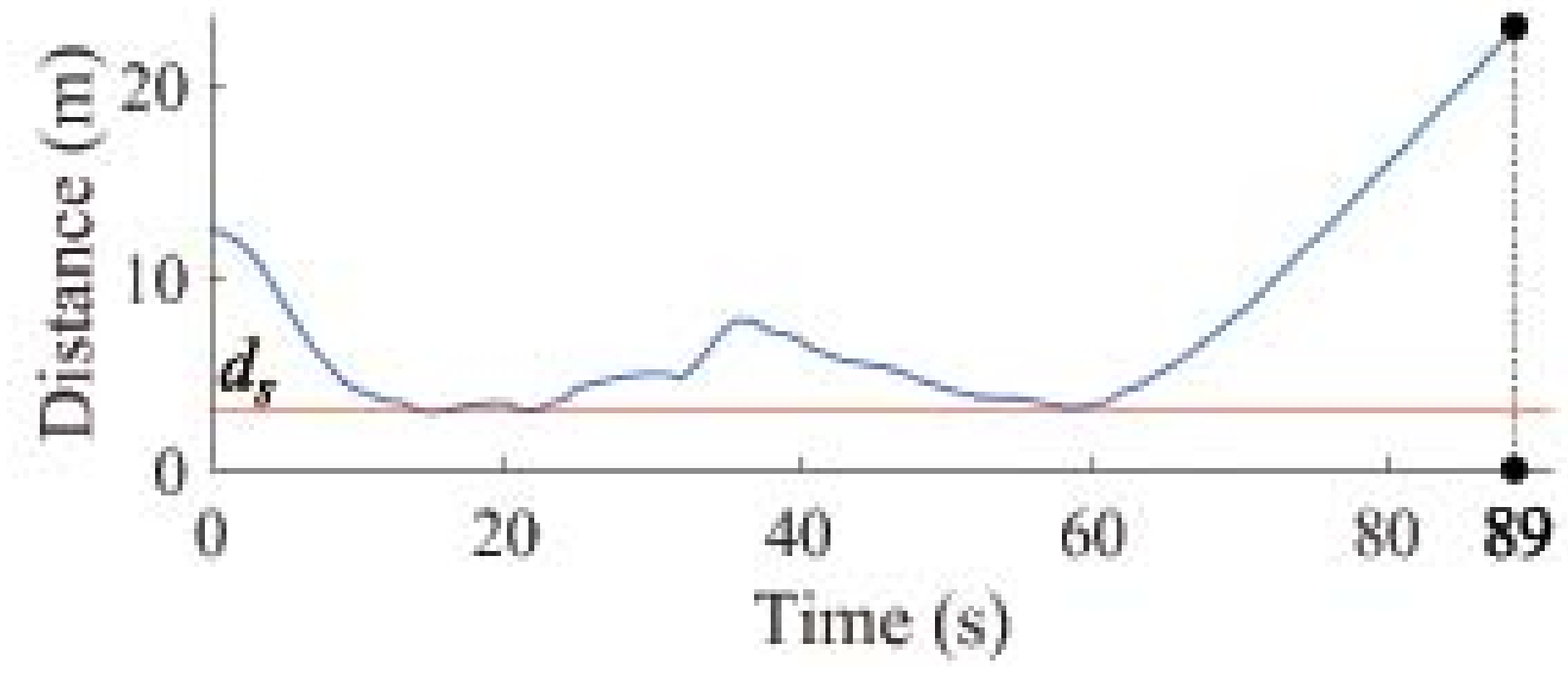,width=7cm}
\caption{
Minimum distance to the undetected areas in the fourth scene.
}
\label{fig:c3_s4-3}
\end{figure}

In the fourth scene, there are three moving obstacles arranged in the environment. The measurement range $R_s$ of the sensor nodes is $50\text{m}$. Therefore, the obstacles in this scene cannot be detected completely (see Fig. \ref{fig:c3_s4-1}). Other parameters in this simulation are the same as the third simulation. In this scene, a robot moves with a safe and relatively short trajectory and avoids the obstacles and the dynamic deformed undetected areas (see Fig. \ref{fig:c3_s4-2}). It can be seen that the robot keeps the given safety margin to the dynamic undetected areas (see Fig. \ref{fig:c3_s4-3}). This simulation indicates that the mobile robots can be navigated to avoid any possible obstacles which are not detected by the sensor network.

\section{Experiments with real mobile robot}

To evaluate and confirm the performance of the proposed navigation system in practical environments, three experiments with a range finder sensor network and a real mobile robot are carried out in this section. In the presented experiments, the similar scenes to practical environments are arranged with some static obstacles and moving people. The presented experiments focus on the navigation of a single robot. In general cases with multiple robots, any robots can be navigated respectively and simultaneously to the targets by the sensor network.

\begin{figure}[!htb]
\centering
\subfigure[SICK LMS-200 laser range finder]{
\epsfig{figure=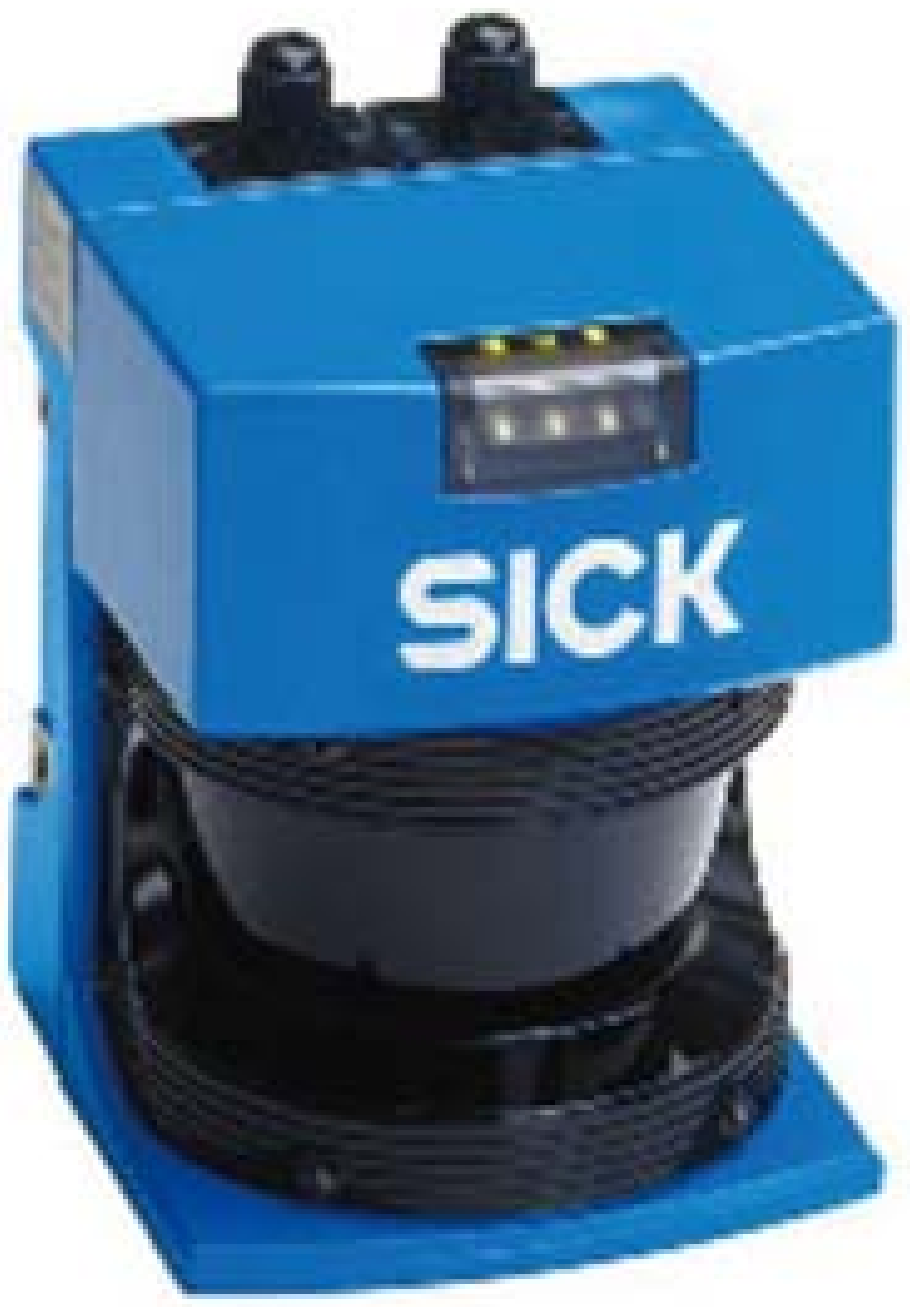,width=6cm}
\label{fig:c3_2-1}
}
\subfigure[Pioneer3-DX robot]{
\epsfig{figure=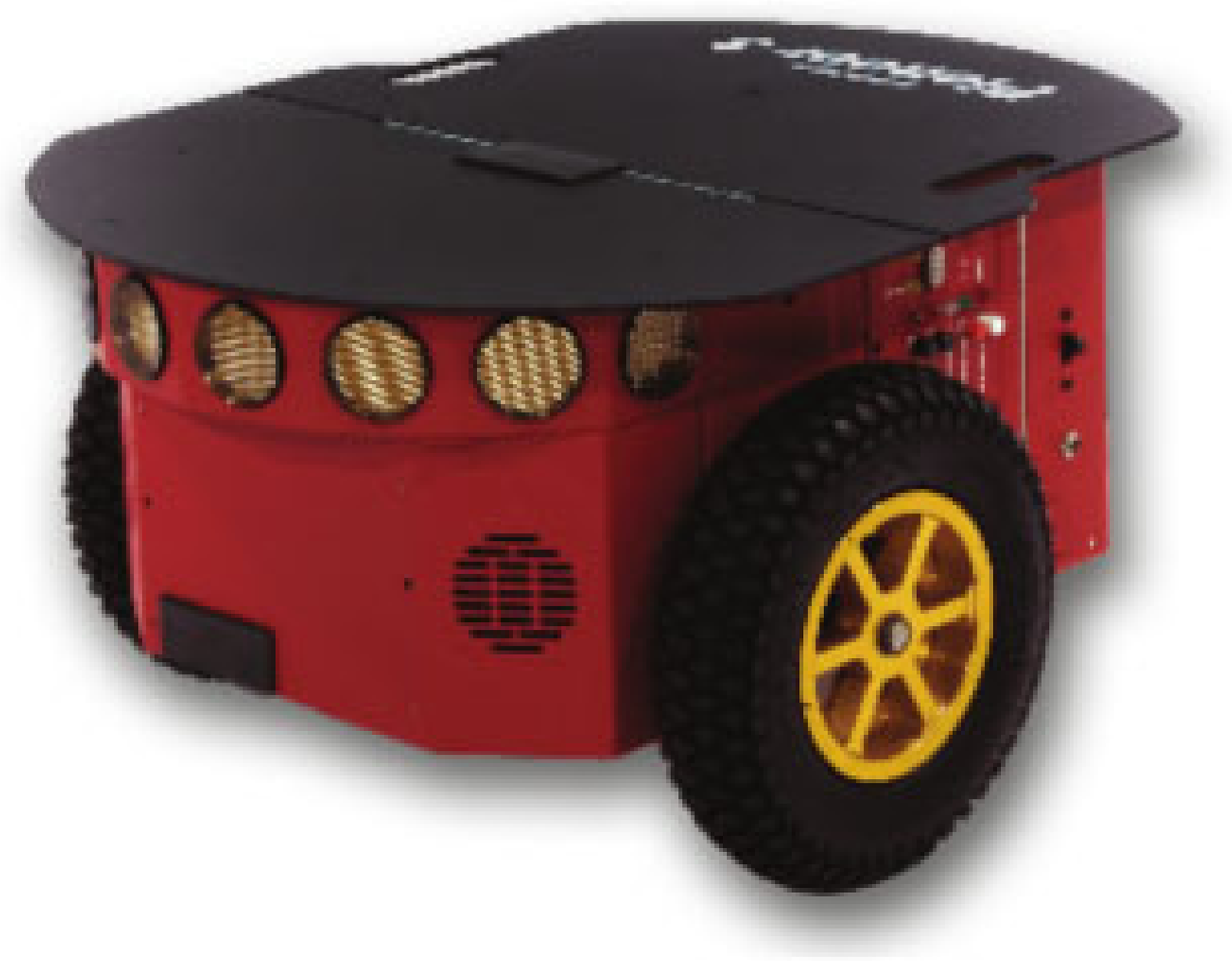,width=6cm}
\label{fig:c3_2-2}
}
\caption{
Laser range finder and mobile robot used in the experiments.
}
\end{figure}

In the experiments, a real sensor network is deployed on the floor of an indoor environment. The sensor network consists of three SICK LMS-200 laser range finders (see Fig. \ref{fig:c3_2-1}), which are connected to a central computer node. Each laser range finder's scanning range is $180^\circ$. The location and direction of each laser range finder are known previously. The central computer is programmed with the proposed navigation algorithm. It collects measurements of the environment and obstacles from each sensor node, obtains the location and heading of the mobile robot from the robot. Then, the computer calculates the safe path $P^*$ according to the obtained information and sends the path $P^*$ to the mobile robot by wireless communication. The target coordinates is predetermined. The real mobile robot used in the experiments is a Pioneer3-DX mobile robot (see Fig. \ref{fig:c3_2-2}). It has an on-board computer and a wireless network device. It can upload the location and heading direction to the central computer and receive the path $P^*$ from the central computer. The mobile robot is programmed with the proposed controller (\ref{c3_2}) to track the received path $P^*$. The mobile robot measures its location and heading direction by odometry. The robot's initial location is $(0,0)$ and the initial heading is $0^\circ$. The odometry error can be ignored in the experiments because the scenes in the experiments are not large. In larger and practical factories, other approaches can be used to reduce the odometry error e.g. \cite{Doh2003,
Doh2006,
Sauer2001}.
The inertial navigation sensors also can be used to increase the accuracy of the localization.

\subsection{Experiments in static environments}

There were two experiments made in static indoor environments. Some folding cartons were arranged as static obstacles. The sensor network was deployed in the environment to detect the obstacles. The main parameters in the experiments are indicated in Table \ref{tb:c3_2}. Notice that the safety margin $d_s$ in the second experiment was changed to $0.6\text{m}$.

\begin{figure}[!htb]
\centering
\subfigure[$t=0\text{s}$]{
\epsfig{figure=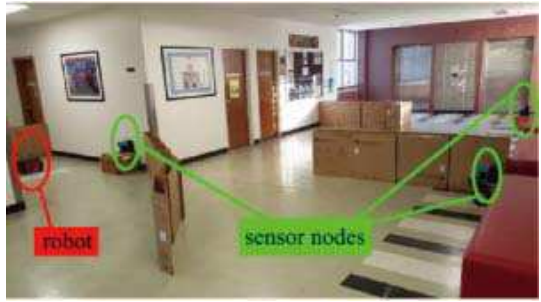,width=7cm}
}
\subfigure[$t=34\text{s}$]{
\epsfig{figure=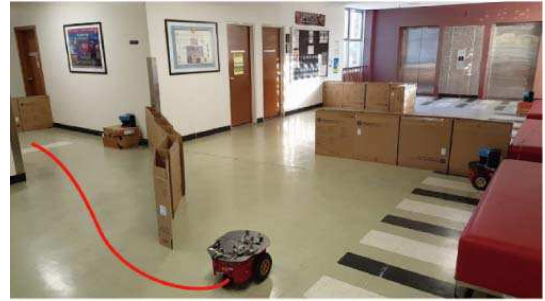,width=7cm}
}
\subfigure[$t=49\text{s}$]{
\epsfig{figure=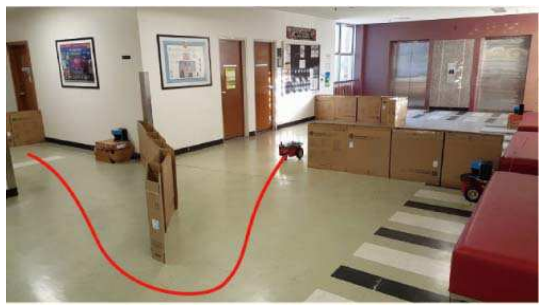,width=7cm}
}
\subfigure[$t=66\text{s}$]{
\epsfig{figure=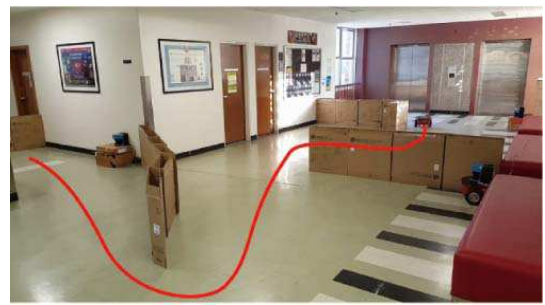,width=7cm}
}
\caption{
Pictures of the first experiment.
}
\label{fig:c3_e1-3}
\end{figure}

\begin{figure}[!htb]
\centering
\epsfig{figure=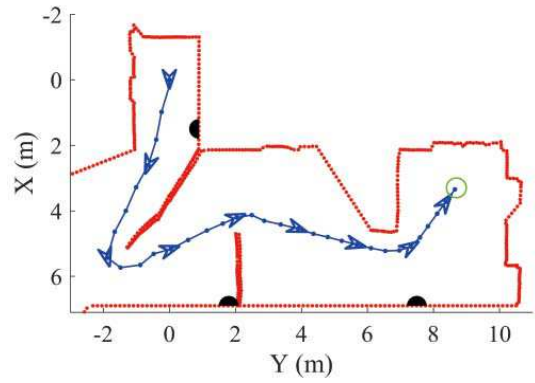,width=7cm}
\caption{
Experimental result of the first experiment.
}
\label{fig:c3_e1-1}
\end{figure}

\begin{figure}[!htb]
\centering
\epsfig{figure=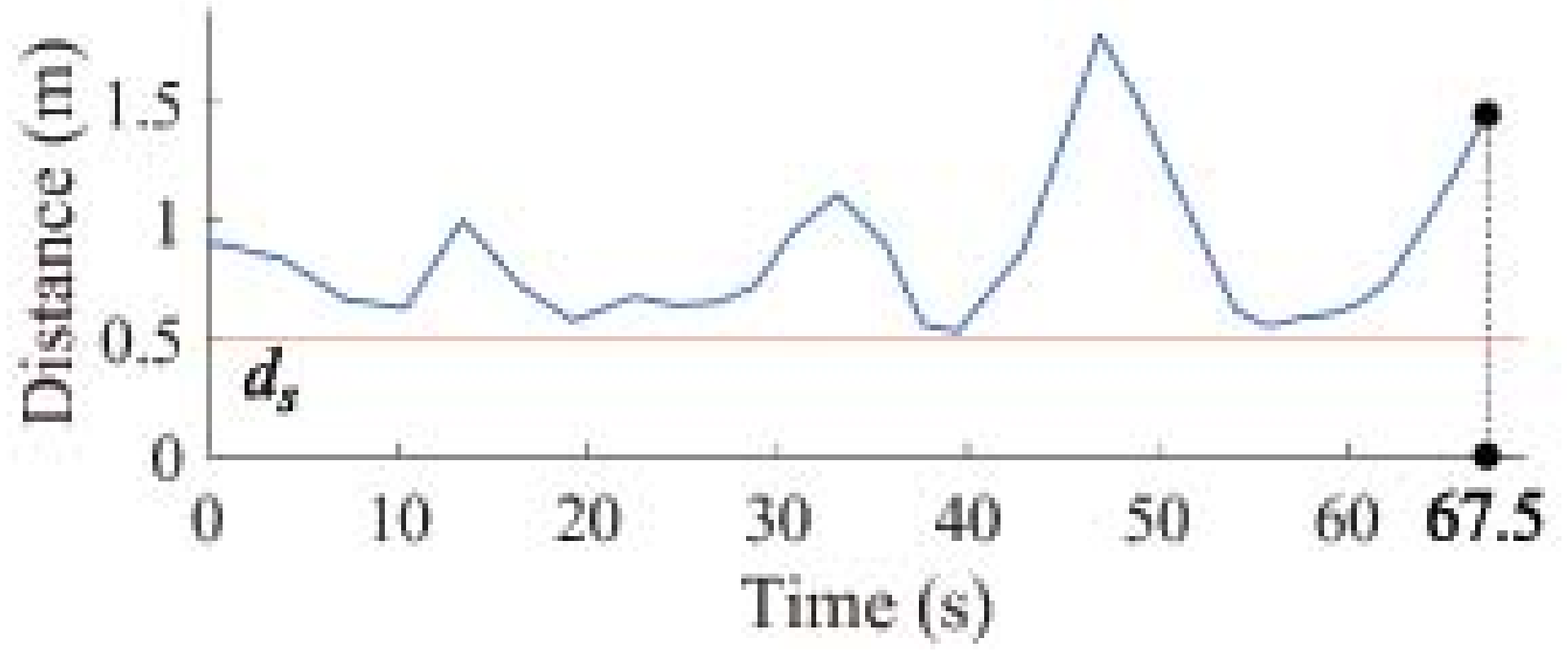,width=8cm}
\caption{
Minimum distance to the undetected areas in the first experiment.
}
\label{fig:c3_e1-2}
\end{figure}

In the first experiment (see Fig. \ref{fig:c3_e1-3}), it can be seen that three laser range finders detected different parts of the environment and the central computer combine the measurements from each sensor node. Then, it navigated the mobile robot from the initial position to the predetermined target without collision. The experimental result (see Fig. \ref{fig:c3_e1-1}) shows that the sensor network mapped the environment correctly and navigated the mobile robot accurately with a relatively short trajectory. According to Fig. \ref{fig:c3_e1-2}, it can be seen that the distance from the obstacles and undetected areas to the mobile robot were larger than the safety margin $d_s$ while the robot was travelling.

\begin{table}[!htb]
\centering
\caption{Main Parameters for the First Experiment}
\label{tb:c3_2}
\begin{tabular}{c|c|c}
\hline
Measurement range & $R_s$ & $8\text{m}$\\
\hline
Speed of robot & $v_r$ & $0.3\text{m/s}$\\
\hline
Maximum angular velocity & $u_{M}$ & $1.2\text{rad/s}$\\
\hline
Safety margin & $d_s$ & $0.5\text{m}$\\
\hline
Sampling interval & $\delta$ & $1\text{s}$\\
\hline
\end{tabular}
\end{table}

\begin{figure}[!htb]
\centering
\subfigure[$t=0\text{s}$]{
\epsfig{figure=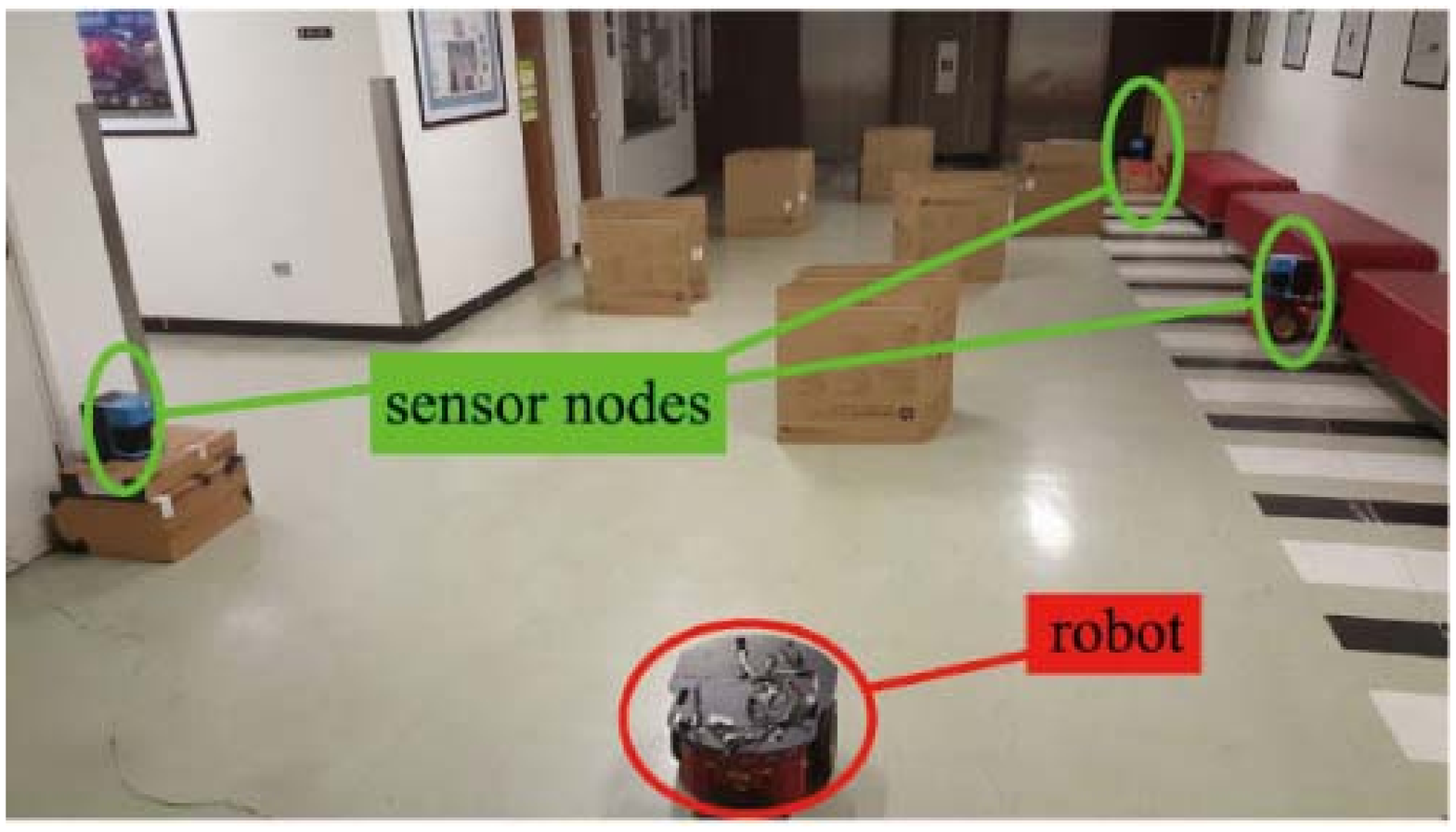,width=7cm}
}
\subfigure[$t=14\text{s}$]{
\epsfig{figure=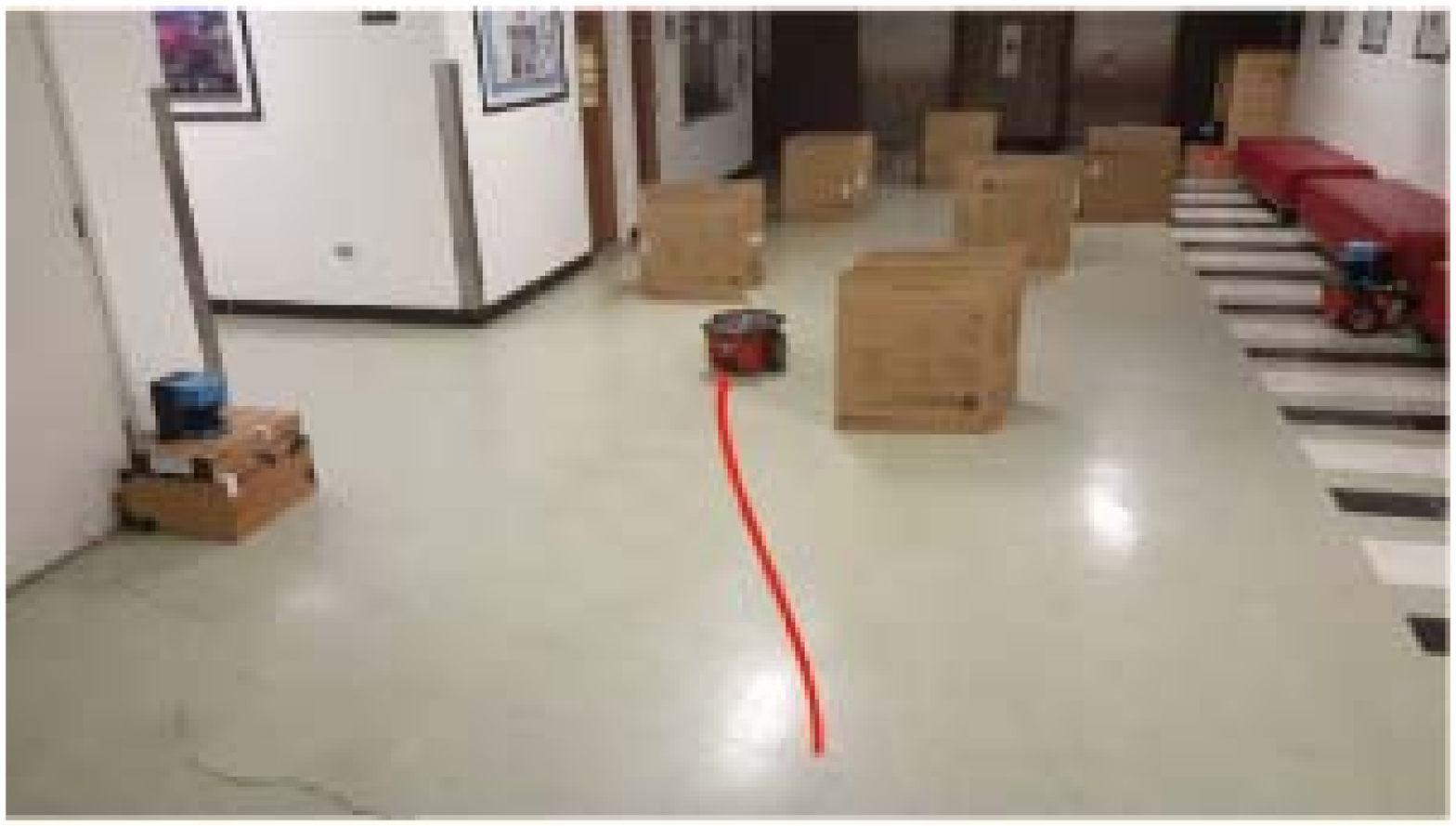,width=7cm}
}
\subfigure[$t=30\text{s}$]{
\epsfig{figure=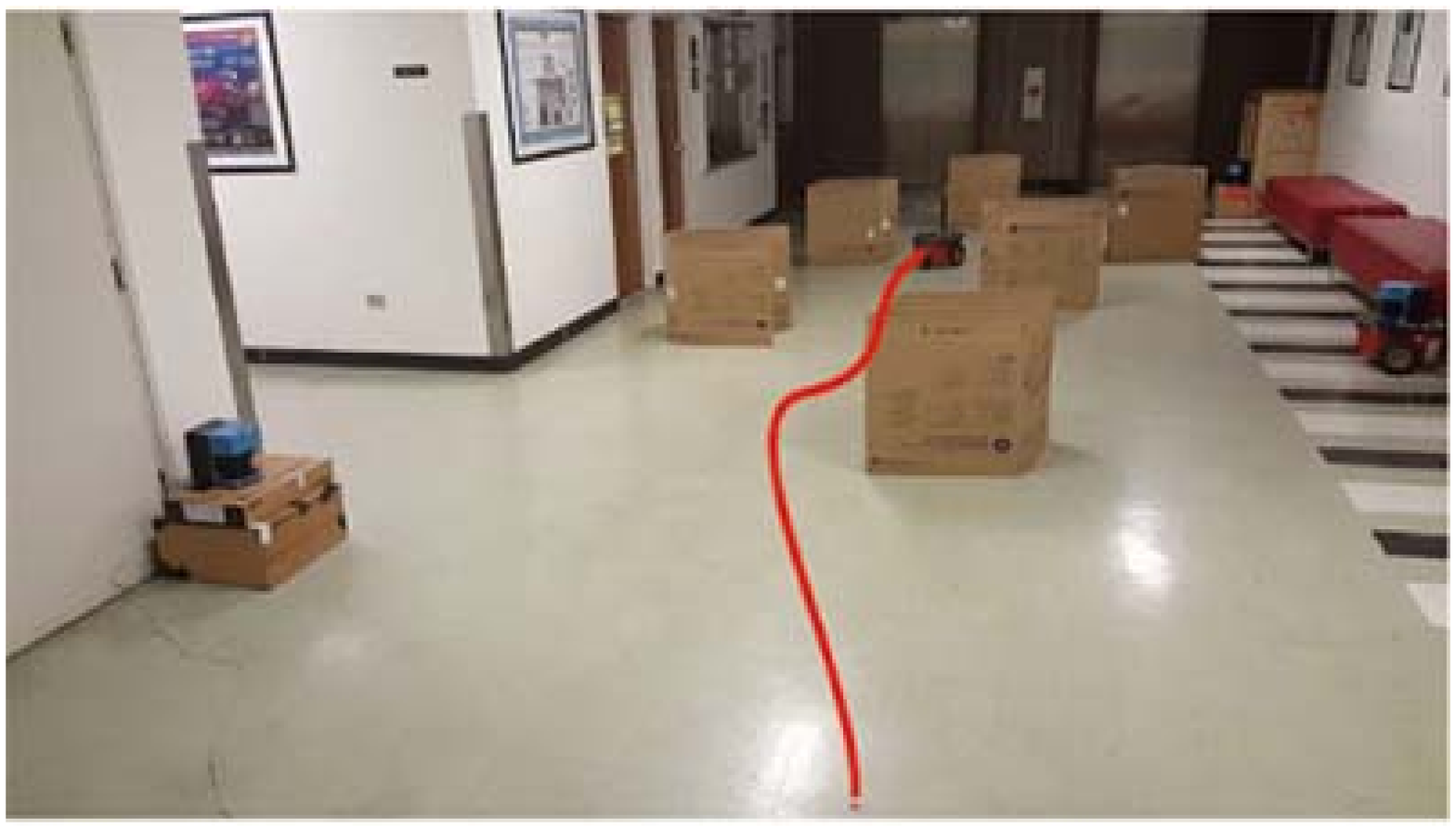,width=7cm}
}
\subfigure[$t=42\text{s}$]{
\epsfig{figure=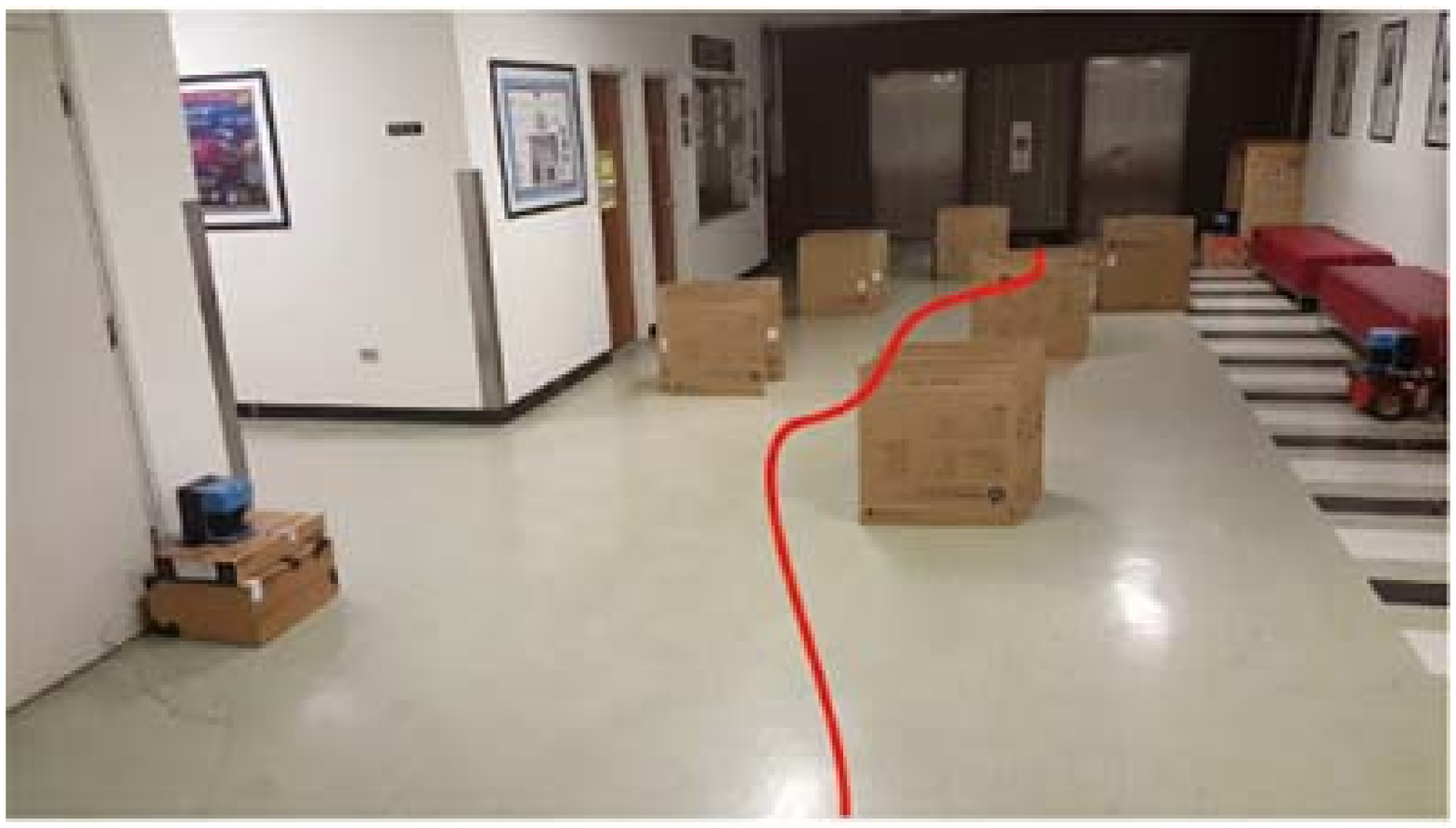,width=7cm}
}
\caption{
Pictures of the second experiment.
}
\label{fig:c3_e3-3}
\end{figure}

\begin{figure}[!htb]
\centering
\epsfig{figure=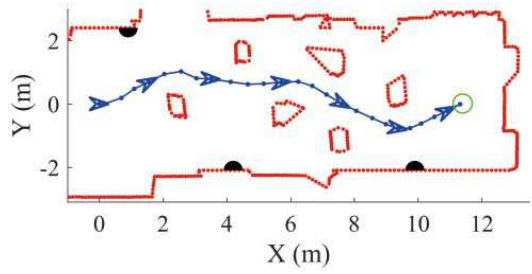,width=7cm}
\caption{
Experimental result of the second experiment.
}
\label{fig:c3_e3-1}
\end{figure}

\begin{figure}[!htb]
\centering
\epsfig{figure=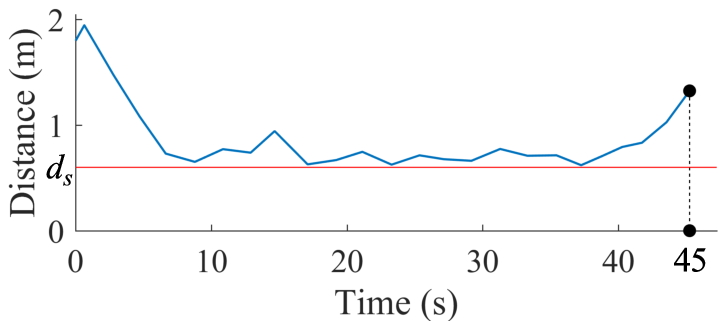,width=8cm}
\caption{
Minimum distance to the undetected areas in the second experiment.
}
\label{fig:c3_e3-2}
\end{figure}

Similarly, in the second experiment (see Fig. \ref{fig:c3_e3-3}), the sensor network navigated the mobile robot to avoid any obstacle on the floor and reach the target. The experimental result (see Fig. \ref{fig:c3_e3-1}) shows the trajectory of the robot and the map built by the sensor network. According to Fig. \ref{fig:c3_e3-2}, it can be seen that the mobile robot was keeping the safety margin $d_s$ while travelling.

\subsection{Experiments in dynamic environments}

\begin{figure}[!htb]
\centering
\subfigure[$t=0\text{s}$]{
\epsfig{figure=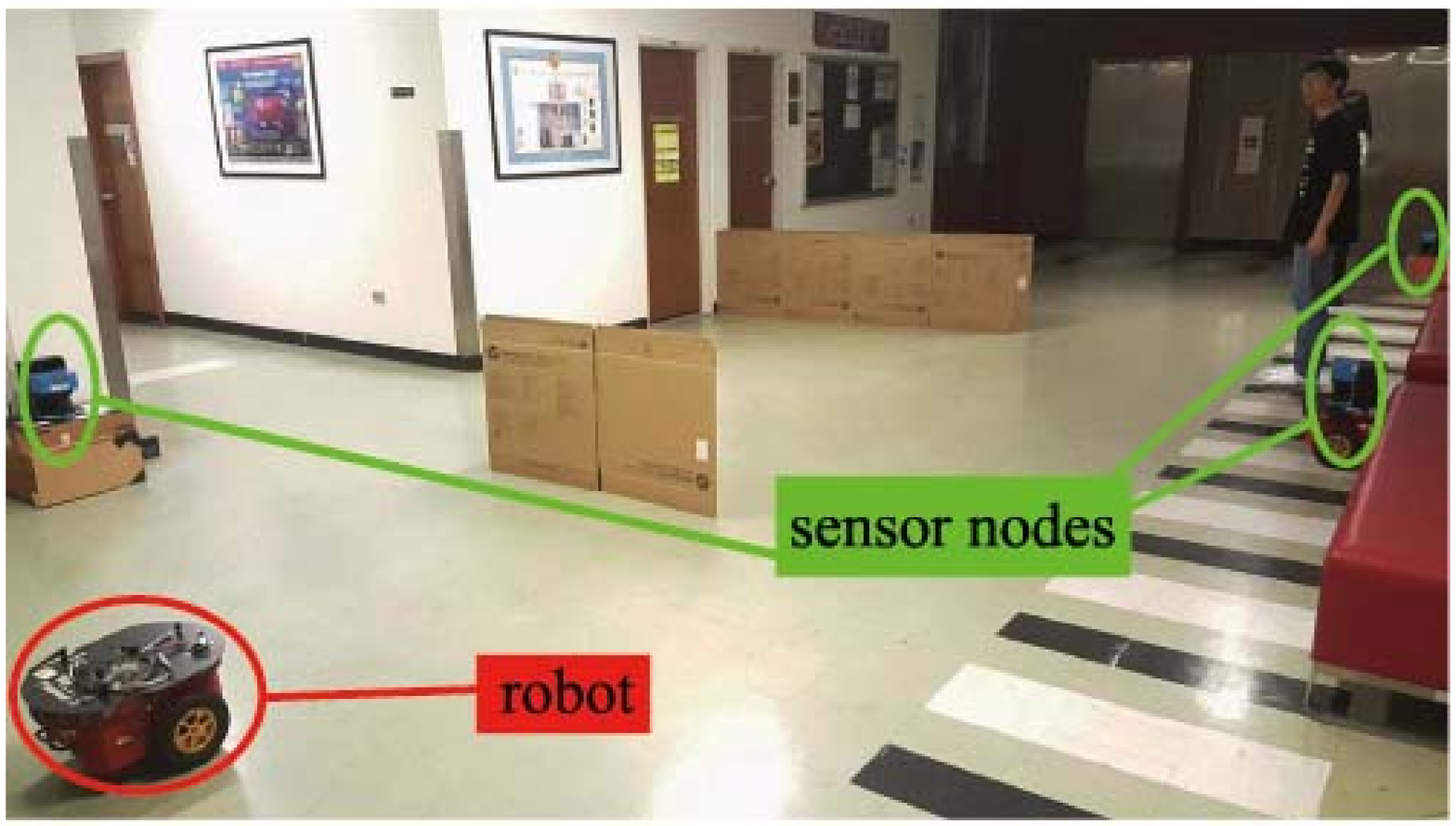,width=7cm}
}
\subfigure[$t=23\text{s}$]{
\epsfig{figure=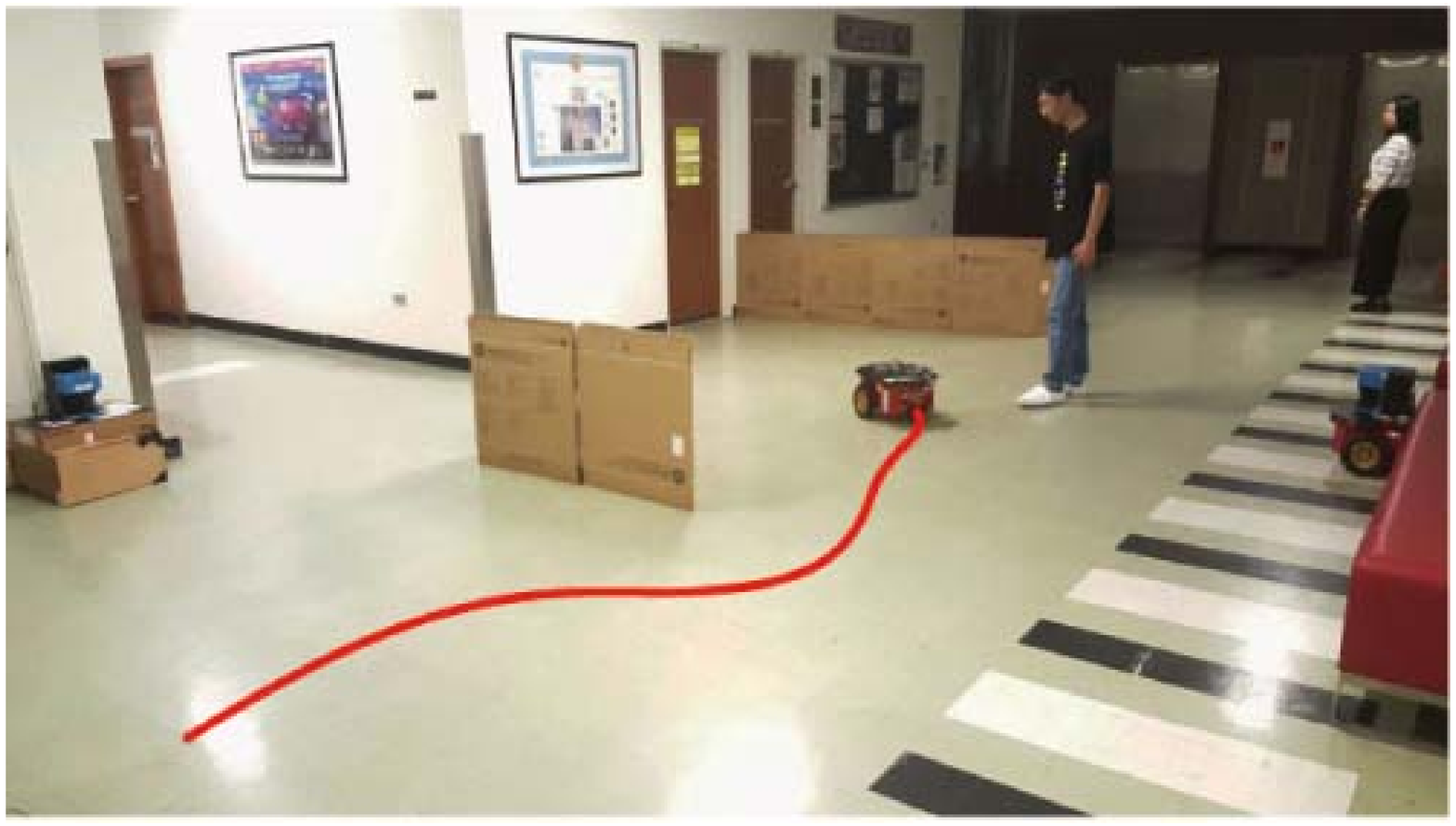,width=7cm}
}
\subfigure[$t=30\text{s}$]{
\epsfig{figure=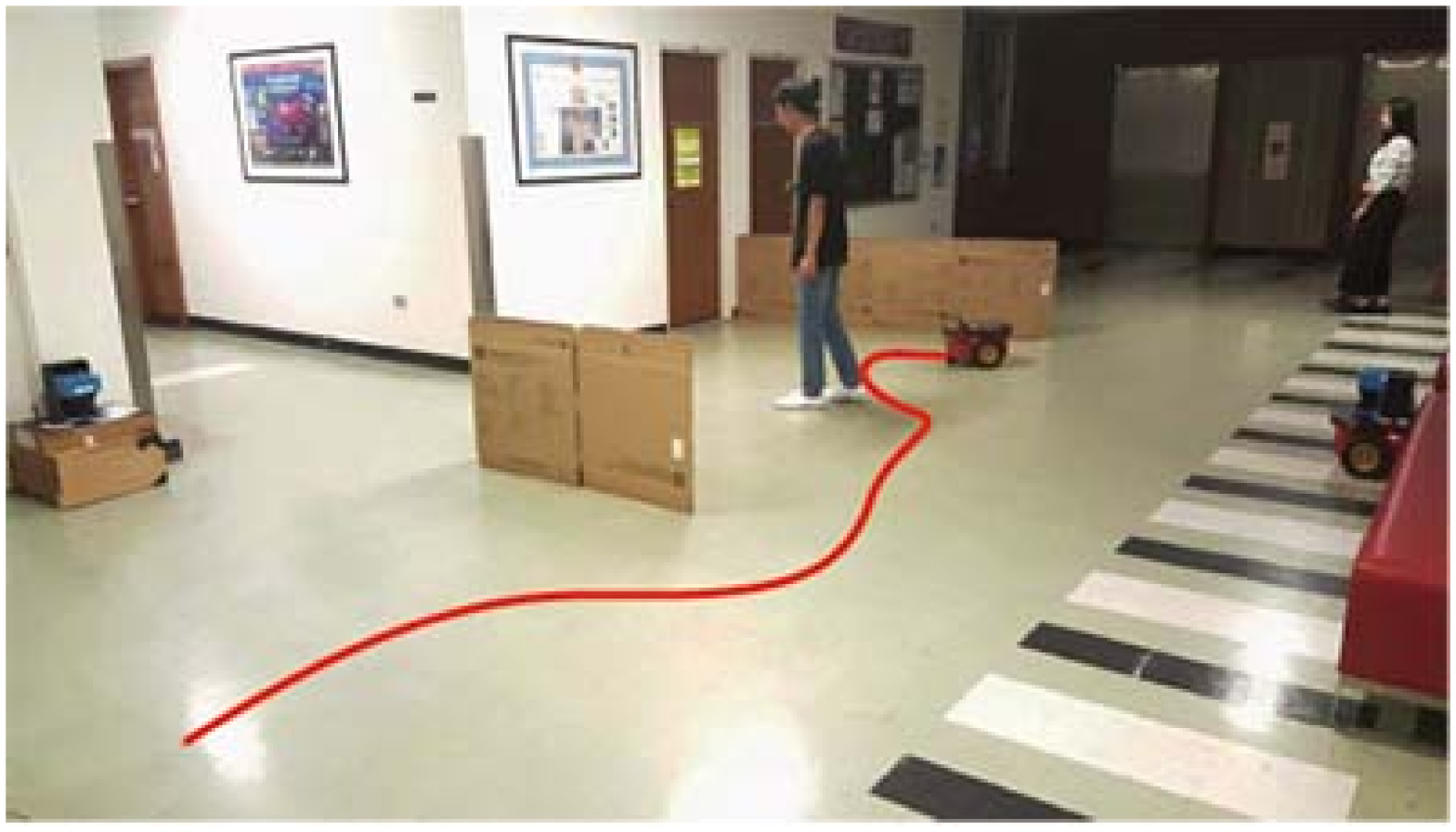,width=7cm}
}
\subfigure[$t=45\text{s}$]{
\epsfig{figure=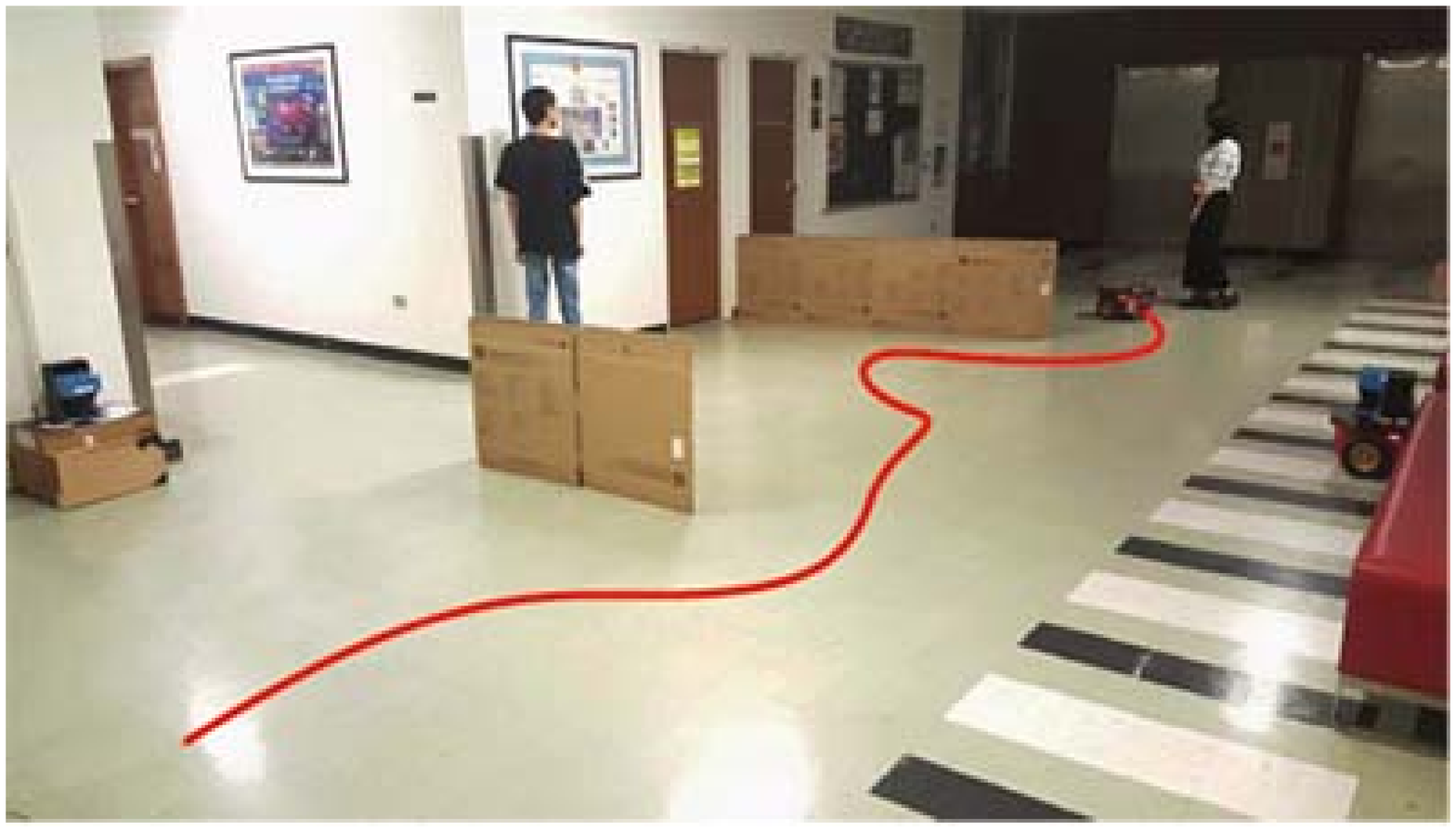,width=7cm}
}
\subfigure[$t=49\text{s}$]{
\epsfig{figure=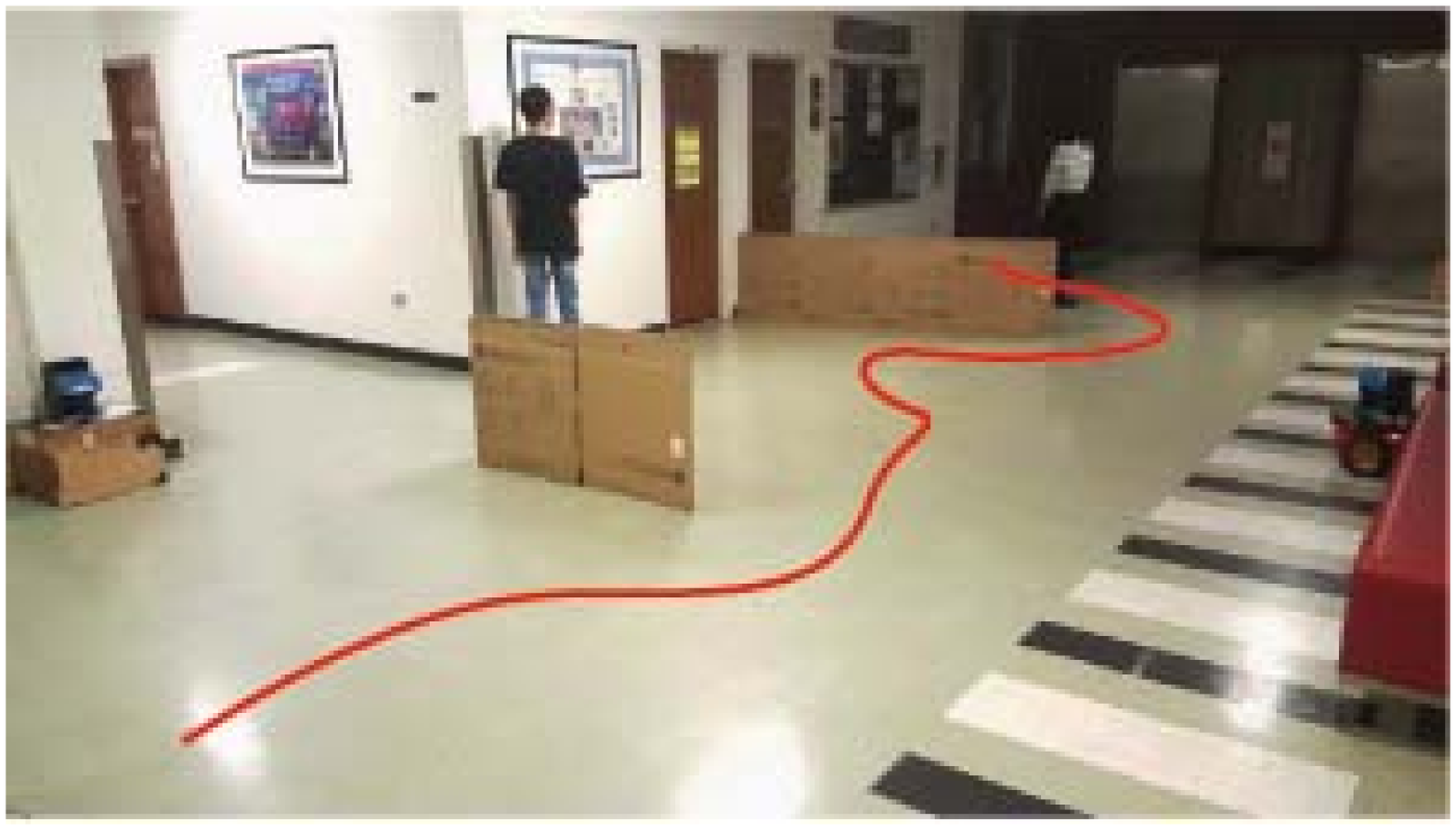,width=7cm}
}
\subfigure[$t=54\text{s}$]{
\epsfig{figure=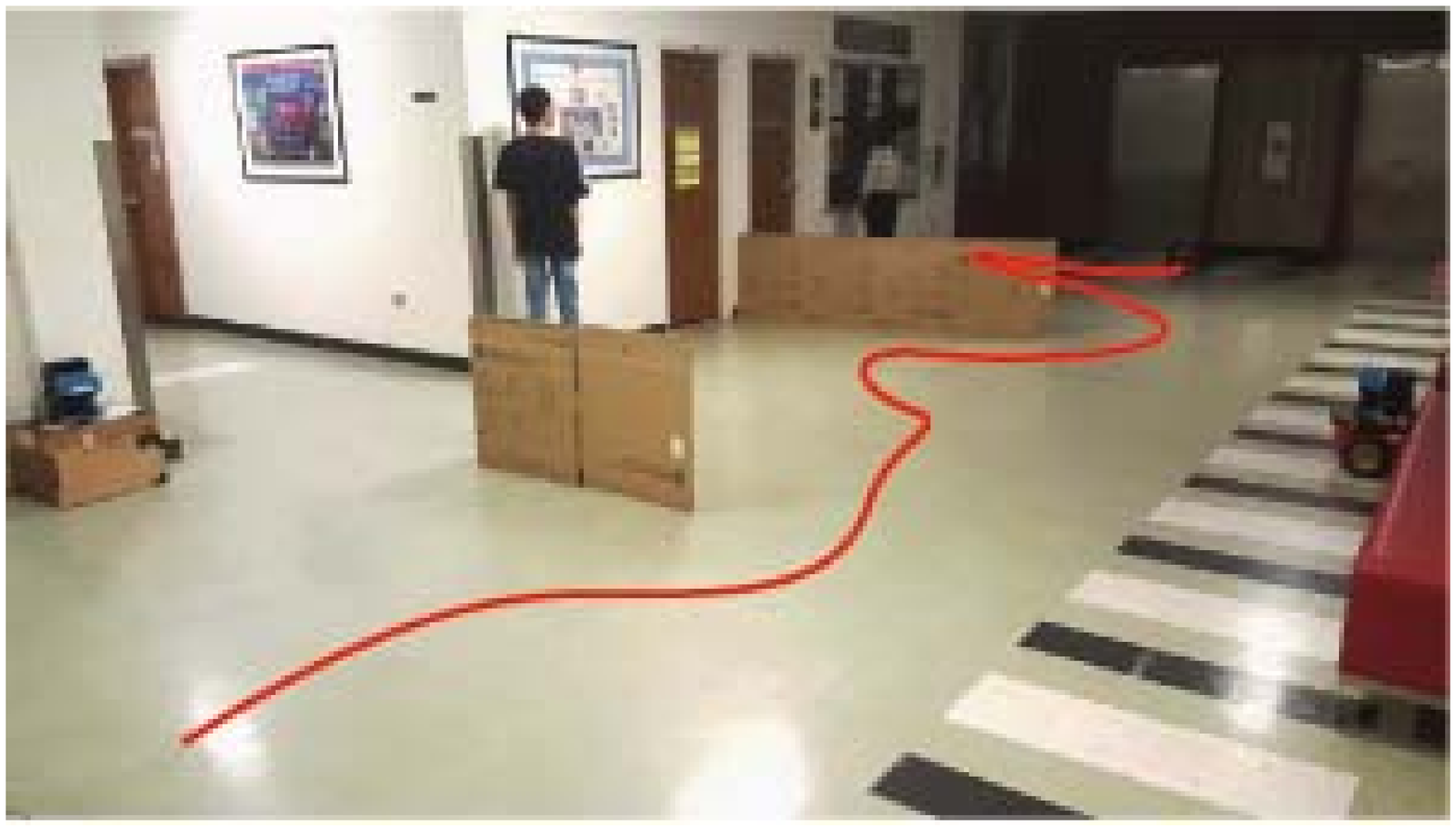,width=7cm}
}
\caption{
Pictures of the third experiment.
}
\label{fig:c3_e4-3}
\end{figure}

\begin{figure}[!htb]
\centering
\subfigure[$t=11\text{s}$]{
\epsfig{figure=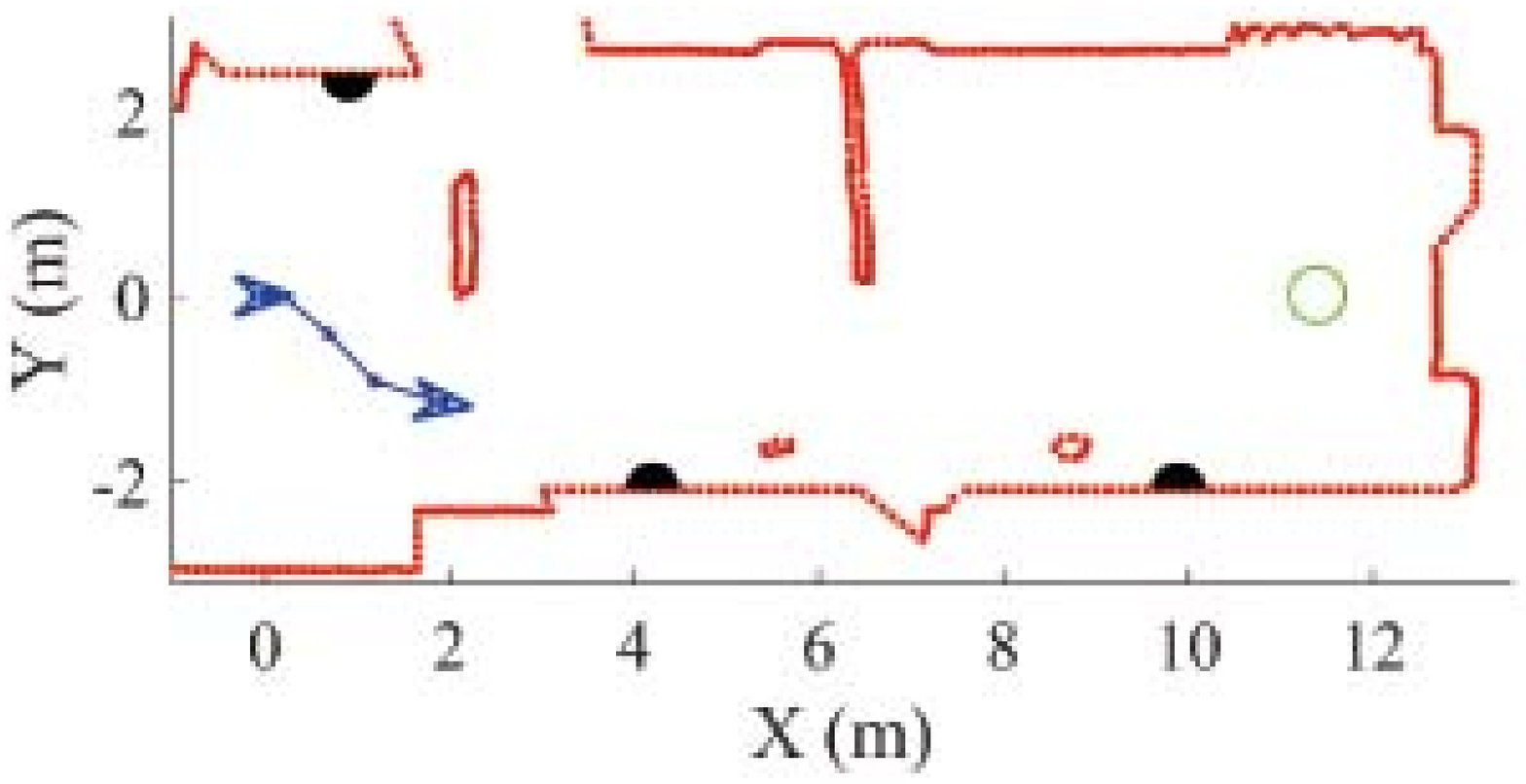,width=7cm}
}
\subfigure[$t=23\text{s}$]{
\epsfig{figure=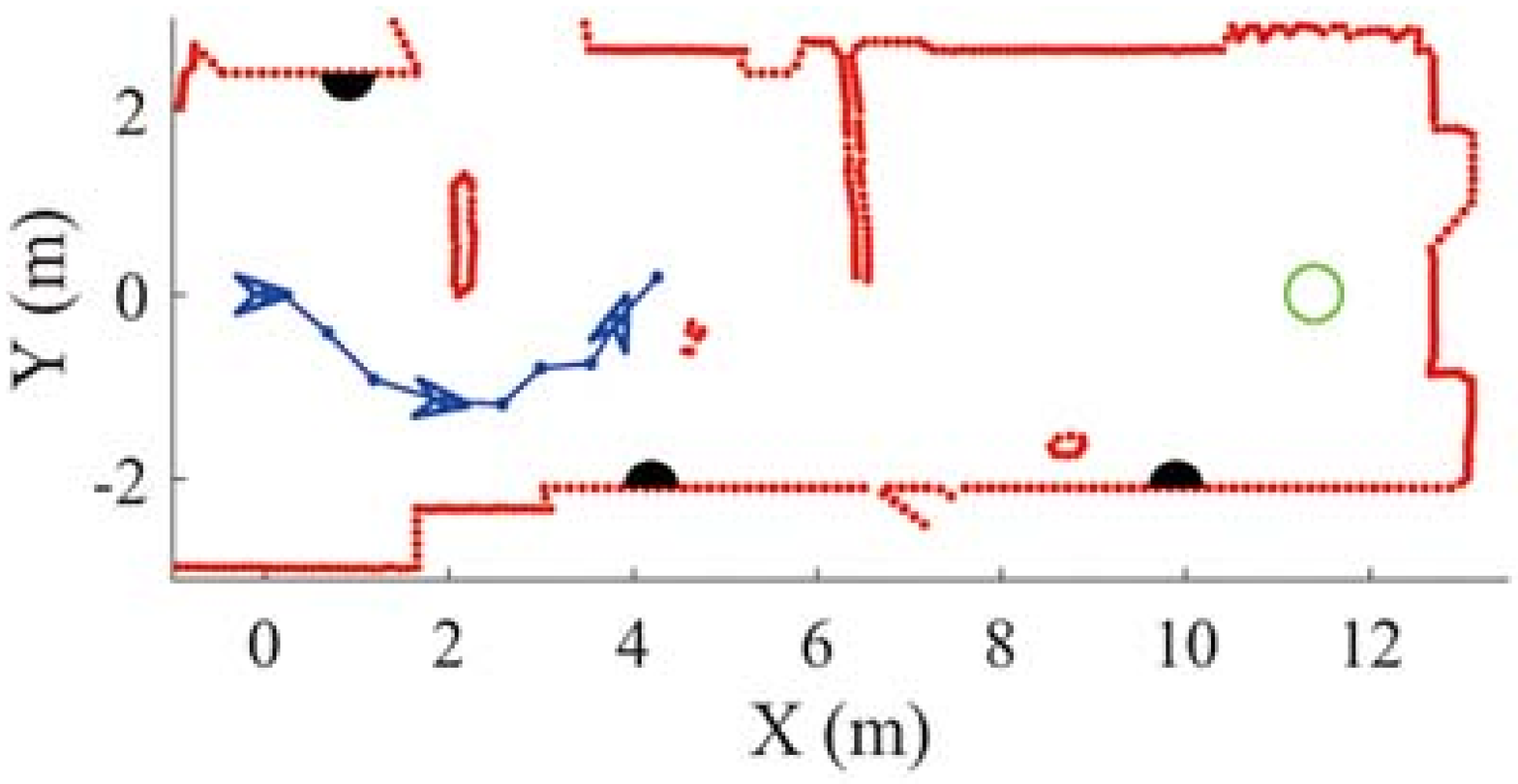,width=7cm}
}
\subfigure[$t=30\text{s}$]{
\epsfig{figure=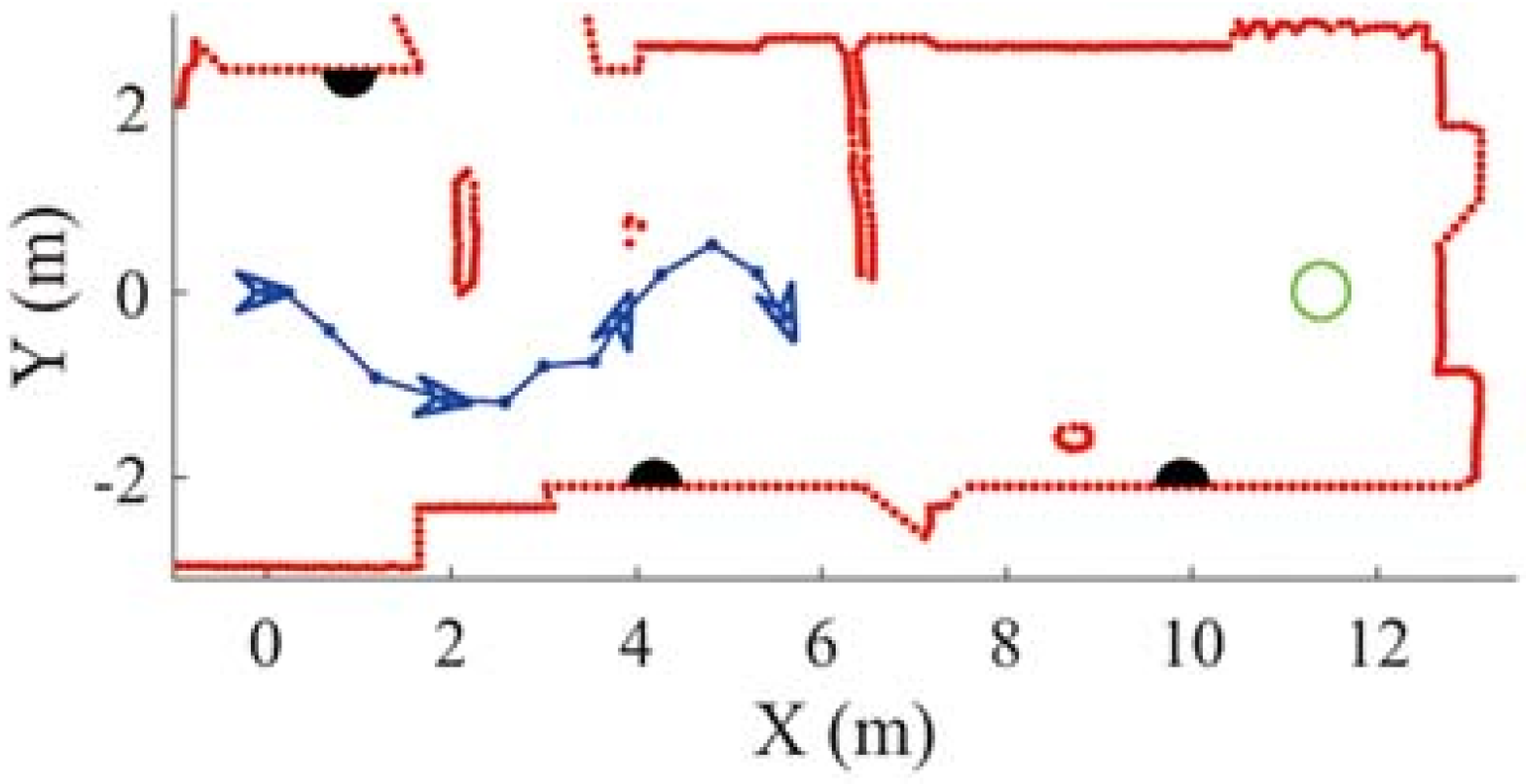,width=7cm}
}
\subfigure[$t=43\text{s}$]{
\epsfig{figure=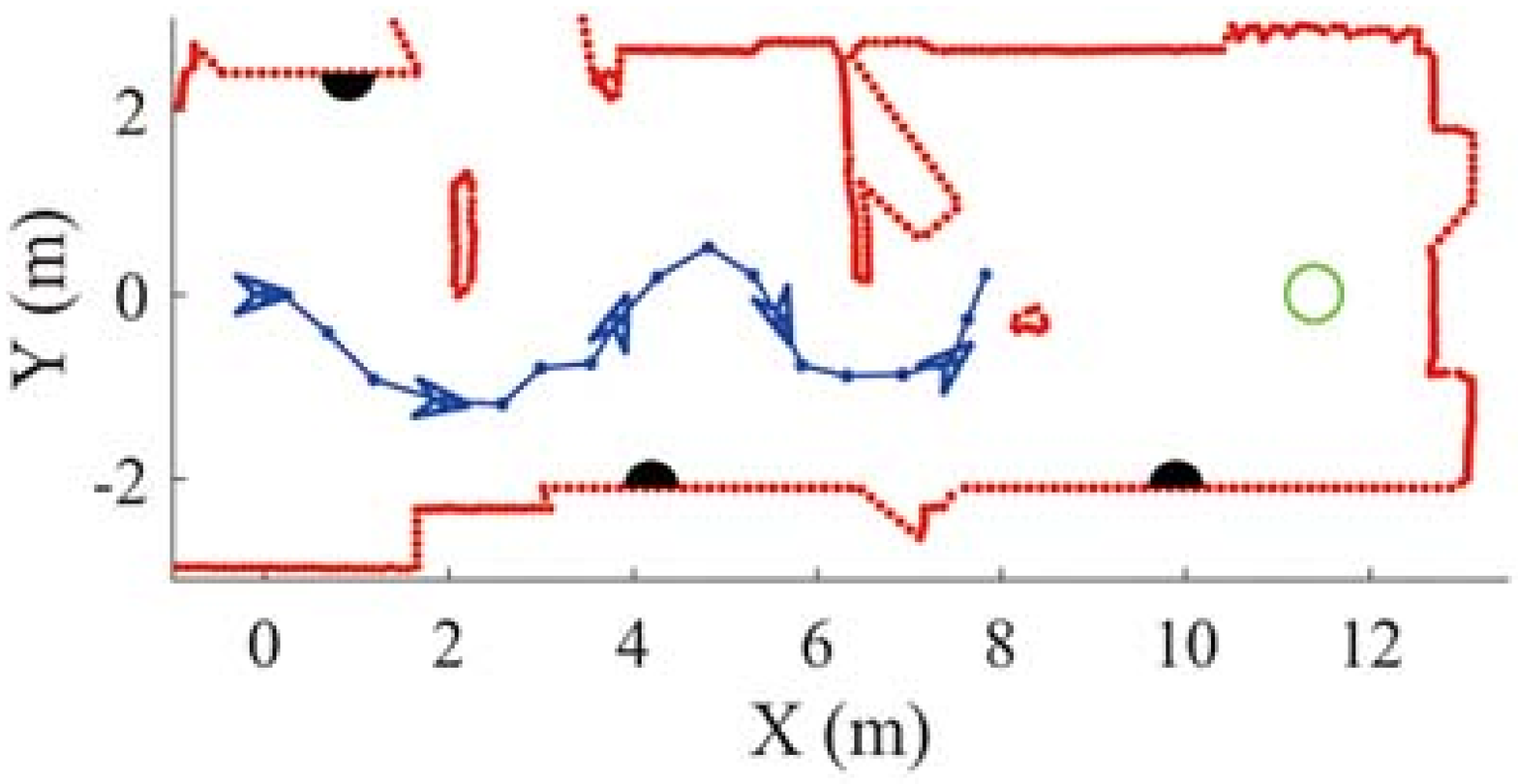,width=7cm}
}
\subfigure[$t=48\text{s}$]{
\epsfig{figure=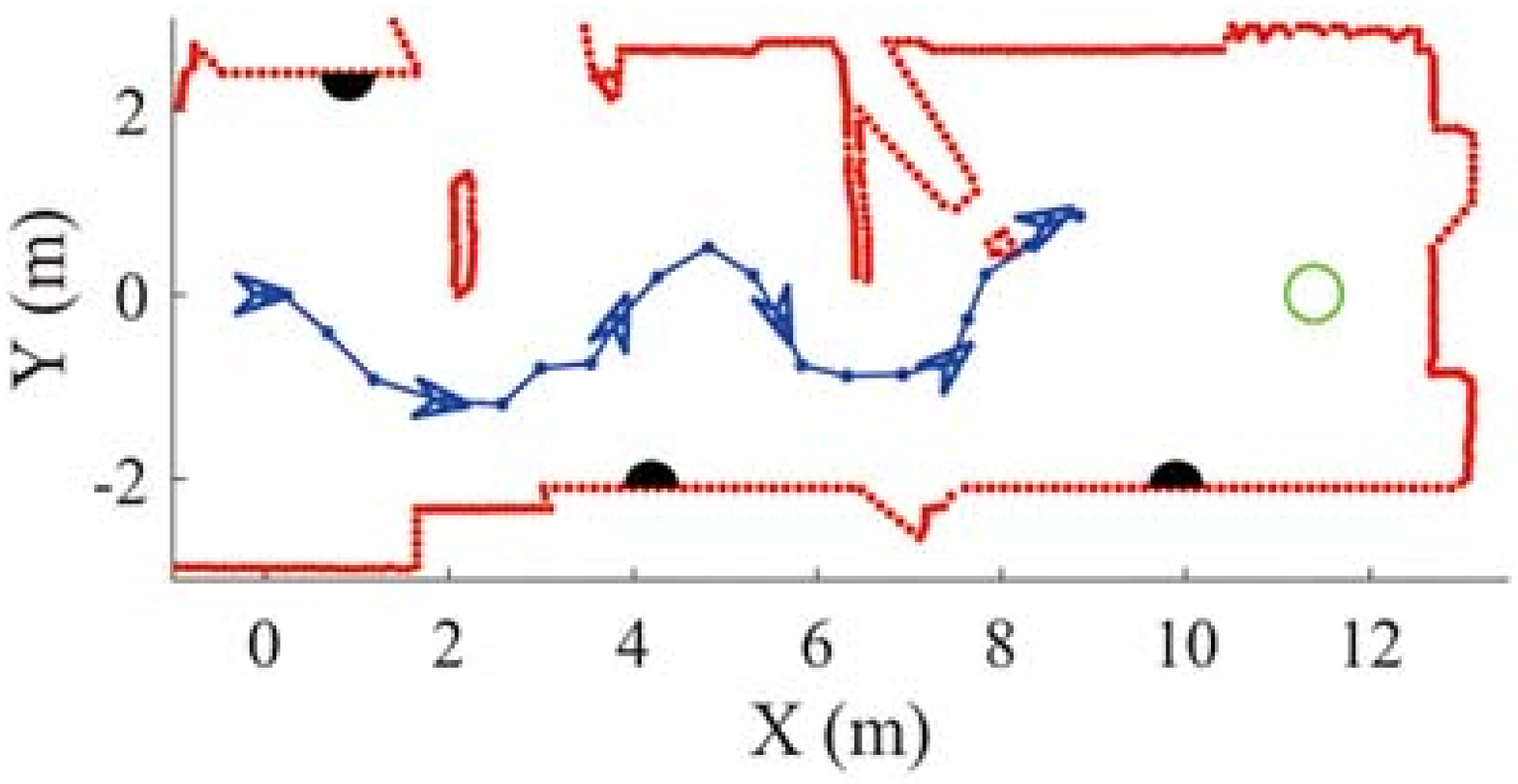,width=7cm}
}
\subfigure[$t=54\text{s}$]{
\epsfig{figure=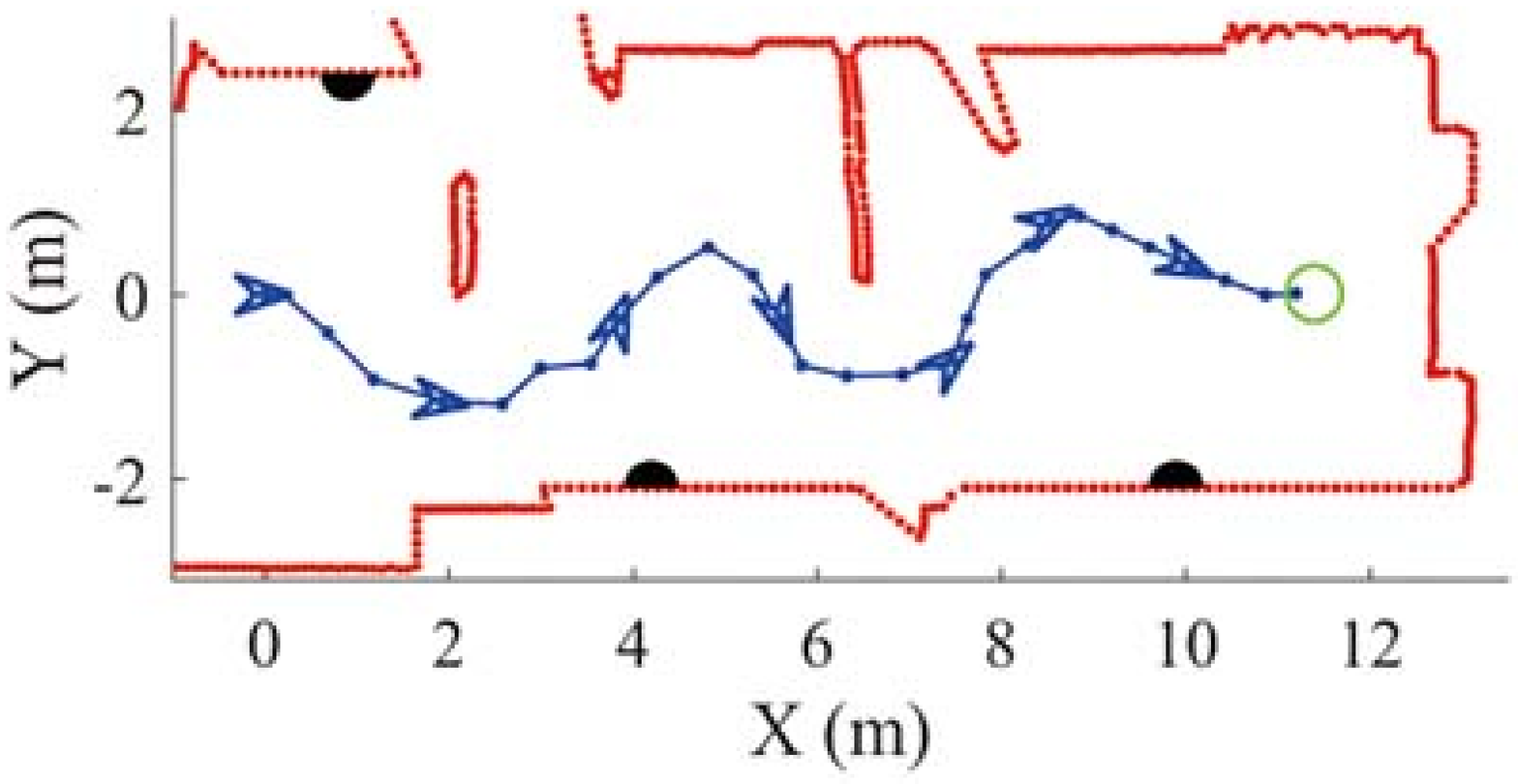,width=7cm}
}
\caption{
Experimental result of the third experiment.
}
\label{fig:c3_e4-1}
\end{figure}

\begin{figure}[!htb]
\centering
\epsfig{figure=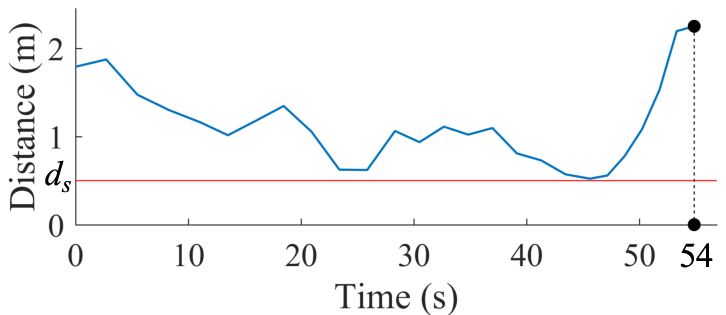,width=8cm}
\caption{
Minimum distance to the undetected areas in the third experiment.
}
\label{fig:c3_e4-2}
\end{figure}

In the third experiment (see Fig. \ref{fig:c3_e4-3}), we tested the proposed navigation algorithm in a dynamic environment. In this scene, some folding cartons were arranged as two static obstacles and two volunteers were walking in this scene. The volunteers' speed were smaller than the given maximum speed $V_{\max}$, which is $0.4\text{m/s}$. The parameter $T$ is determined as $2$. Other parameters for this experiment are indicated in Table \ref{tb:c3_3}. According to the experimental result in Fig. \ref{fig:c3_e4-1}, it can be seen that the sensor network built the real-time map of the dynamic environment. The mobile robot successfully avoided both the static obstacles and the moving obstacles. Fig. \ref{fig:c3_e4-2} shows that the mobile robot was keeping the safety margin $d_s$ while travelling.

\begin{table}[!htb]
\centering
\caption{Main Parameters for the Third Experiment}
\label{tb:c3_3}
\begin{tabular}{c|c|c}
\hline
Measurement range & $R_s$ & $8\text{m}$\\
\hline
Speed of robot & $v_r$ & $0.5\text{m/s}$\\
\hline
Maximum angular velocity & $u_{M}$ & $2\text{rad/s}$\\
\hline
Safety margin & $d_s$ & $0.5\text{m}$\\
\hline
Sampling interval & $\delta$ & $1\text{s}$\\
\hline
\end{tabular}
\end{table}

\section{Summary}

The safe navigation is a fundamental problem of robotics. In the proposed method, we take the advantage of the range finder sensor network and propose a sensor network based navigation algorithm for any types of mobile robots working in the environments. Comparing with other types of sensor network, the range finder sensor network can obtain accuracy environment maps to help perform a faster and more reliable navigation than RSSI-based sensor networks. Moreover, It involves simpler data processing than visual sensor networks.

The proposed sensor network is deployed in dynamic industrial environments to detect the static and dynamic obstacles. Simultaneously, each robot measures and uploads its own location and direction by the wireless communication. Then, the sensor network navigate each mobile robot respectively according to the environment and robots information. The computer simulations and experiments confirmed the expected performance of the proposed algorithm.

The main advantage of the proposed method is that the navigation tasks for any mobile robots are integrated into the sensor network. The sensor network can be used to navigate different types of robots simultaneously. The proposed navigation method does not require any robot sensor for obstacle detection and any other extra navigation algorithm. Moreover, the proposed method is flexible in configuration for multiple robots. Each robot is only required for a low-level path tracking controller. Therefore, this is an efficient, integrated and economic navigation system for multiple robots in practical dynamic environments.

The proposed navigation framework can be implemented as one of the most fundamental units in smart factory with various robots working cooperatively. With the arranged sensor network, manufacturers can purchase any types of robots without considering the navigation and obstacle avoidance. This system also provides a centralized framework to manage and supervise the working of all the robots. The sensor network collects each robot's status and sensor measurements, such as battery, system errors and radiant intensity alarm, and sends the information to the manager or high-level management system. On the other hand, the manufacturers of industrial mobile robots can remove the obstacle detection sensors and navigation algorithms from the robots to decrease the robots' cost and price.
%%%%%%%%%%%%%%%%%%%%%%%%%%%%%%%%%%%%%%%%%%%%%%%%
\chapter{Safe Navigation of Flying Robots in Dynamic Environments by a Sensor Network}
\chaptermark{Flying Robots Navigation}
\label{Chapter4}

This chapter is based on the the publications \cite{Hang_uj_3} and \cite{Hang_ccc2017_2}. In this chapter, we take the advantages of the wireless sensor network to navigate the flying robots. The sensor network consists of some 3D range finder sensors, such as time-of-flight cameras. Each sensor node is deployed in the dynamic industrial environments, such as floors and walls, to detect the obstacles like walls, equipments and walking people. Simultaneously, each flying robot measures the real-time location and direction by localization technology, like odometry, and sends the measurements to the central computer via the wireless sensor network. With the measurements of the environment and each robot's position, instant safe paths are dynamically generated from each robot's current position to the destinations by the central computer. The instant safe paths are updated at each time step and the flying robots keep tracking the generated path to the targets. The presented flying robot navigation framework is a networked control system that the sensor data, control input and flying robots' states are exchanged via the sensor network; see examples \cite{1193753,MATVEEV200451problem,
SAVKIN200651Analysis,Matveev2009Estimation}.

\section{Problem description}

\begin{figure}[!htb]
\centering
\epsfig{figure=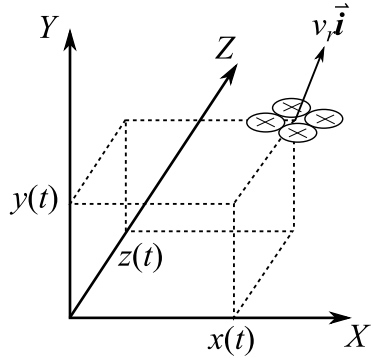,width=5cm}
\caption{
Flying robot model.
}
\label{fig:c4_1}
\end{figure}

We consider any micro flying robot as a three-dimensional under-actuated
non-holonomic autonomous vehicle, which is widely used to describe flying robots and missile; e.g. \cite{Wang2016a,Matveev2017}. It is equipped with the inertial measurement unit (IMU) and a wireless communication device. The mathematical model of the flying robot is as follows. Let

\begin{equation}
\robots(t):=[x(t),y(t),z(t)]
\end{equation}
be the Cartesian coordinates of the robot in the 3D space. For multiple flying robots, let $\robots_i$ denote the $i$-th robot's position. Then, the motion of the robot can be described by the following equations (see Fig. \ref{fig:c4_1}):

\begin{equation}
\label{c4_EQ2}
\left\{
\begin{array}{l}
\dot{\robots} = v_r \vect{i}
\\
\dot{\vect{i}} = \vect{u}
\end{array},
\right.
\end{equation}
where $v_r$ is a constant speed, $\vect{i}\in \mathbb{R}^3$ is the unit vector indicating the direction of the robot's velocity. $\vect{u}\in \mathbb{R}^3$ is a two degree of freedom control input, which is subject to

\begin{equation}
\label{c4_EQ1}
\left\{
\begin{array}{l}
\Vert \vect{u} \Vert\leq U_M
\\
\vect{i} \perp \vect{u}
\end{array}.
\right.
\end{equation}
Here $\perp$ denotes that two vectors are orthogonal. The robot's location $\robots(t)$ and the direction of velocity $\vect{i}(t)$ can be obtained by the robot's IMU and odometry. The constraint of the control input $\vect{u}$ implies that the minimum turning radius of the flying robot is

\begin{equation}
R_{\text{min}}=\frac{v_r}{U_M}.
\end{equation}

\begin{notation}
Let $p$ be an arbitrary point. For any closed set $D$, the minimum distance $\rho(D,p)$ between $p$ and $D$ is
\begin{equation}
\rho(D,p):=\min_{q\in D}\Vert p-q\Vert.
\end{equation}
\end{notation}

Considering the minimum turning radius, we define a torus called initial torus, which is a extension of the initial circle proposed in \cite{Savkin2016}; also see \cite{Masehian2015}.

\begin{figure}[!htb]
\centering
\epsfig{figure=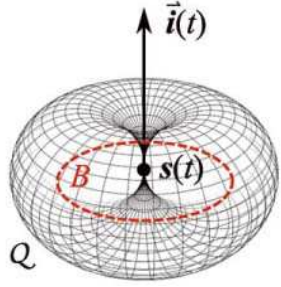,width=5cm}
\caption{
Initial torus.
}
\label{fig:c4_11}
\end{figure}

\begin{definition}
Let $B$ be a circle described as follows:
\begin{equation}
B=\lbrace p\in\mathbb{R}^3 : (p-\robots)\perp \vect{i}, \Vert p-\robots \Vert=R_{\text{min}} \rbrace.
\end{equation}
Then, an initial torus ${\cal Q}$ can be defined as follows (see Fig. \ref{fig:c4_11}):
\begin{equation}
{\cal Q}:=\lbrace p\in\mathbb{R}^3 : \rho(B,p)=R_{\text{min}}\rbrace.
\end{equation}
\end{definition}

\begin{figure}[!htb]
\centering
\epsfig{figure=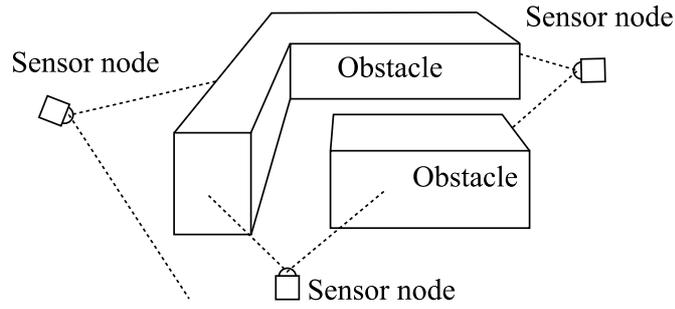,width=9cm}
\caption{
A 3D range finder WSN in the environment.
}
\label{fig:c4_2}
\end{figure}

\begin{figure}[!htb]
\centering
\epsfig{figure=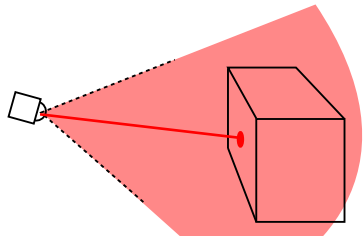,width=5cm}
\caption{
A 3D range finder.
}
\label{fig:c4_3}
\end{figure}

In an practical environment, there are static and moving obstacles (e.g. walls, walking people, etc.) with irregular shape and dynamic velocity. To detect the obstacles extensively, we consider a WSN deployed in the workspace (see Fig. \ref{fig:c4_2}). Each sensor node is a 3D range finder, which measures the distances to the nearest obstacles in different directions within the field of view (see Fig. \ref{fig:c4_3}). There are two types of 3D range finder that are commonly used in many industrial applications. The first type is spinning 2D range finder, see \cite{Olivka2016}, which provides omnidirectional measurements of distance. The second type is ToF camera that only measures the distance in a limited field of view like a normal camera; such as \cite{Alenya2014}. However, a ToF camera has a better performance in dynamic environments than a spinning 2D range finder. Both of these two types of 3D range finder can be used in the proposed method.

\begin{assumption}
For any sensor node, the location and attitude, in which it is deployed, are static and known.
\end{assumption}

\begin{figure}[!htb]
\centering
\epsfig{figure=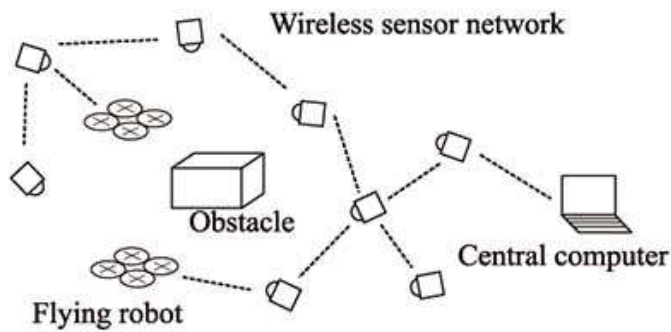,width=9cm}
\caption{
Framework of the proposed method.
}
\label{fig:c4_17}
\end{figure}

In the presented method, a central computer is involved to execute the proposed navigate algorithm for the micro flying robots. It connects to one of the sensor nodes and accesses all the sensor nodes via the WSN to collect the real-time measurements of the obstacles from each 3D range finder. Then, the central computer combine the measurements to build a 3D map of the unoccupied area. In addition, the flying robots in the workspace can connect to the WSN. Thus, the central computer can communicate with any flying robot via the WSN as well. In practical implementations in industrial environments, the micro flying robots can be equipped with some light sensors for some tasks and upload the sensor data to the central computer. For instance, radiation detectors can be mounted on the flying robots to monitor the radiant intensity in nuclear power stations. The framework of this navigation system is shown in Fig. \ref{fig:c4_17}.

\begin{figure}[!htb]
\centering
\epsfig{figure=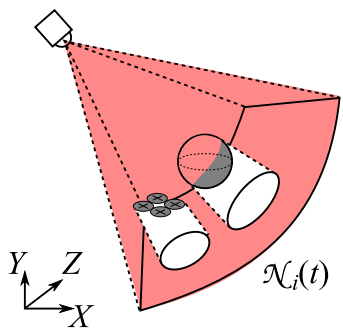,width=5cm}
\caption{
Detected region ${\cal N}_i(t)$ of the $i$-th sensor node.
}
\label{fig:c4_4}
\end{figure}

In the industrial environment, the obstacles and flying robots can be detected by the sensor nodes. For any $i$ and time $t$, the $i$-th sensor node detects objects partly in its field of view and obtains a 3D detected region, denoted by ${\cal N}_i(t)$. The region ${\cal N}_i(t)$ can be obtained by converting the sensor's measurements to the global coordinate system from its local coordinate system with the help of the its location and attitude (see Fig. \ref{fig:c4_4}).

\begin{definition}
For any $i$ and $t$, the detected region ${\cal N}_i(t)$ is a closed, bounded, and connected point set in $\mathbb{R}^3$.
\end{definition}

\begin{figure}[!htb]
\centering
\epsfig{figure=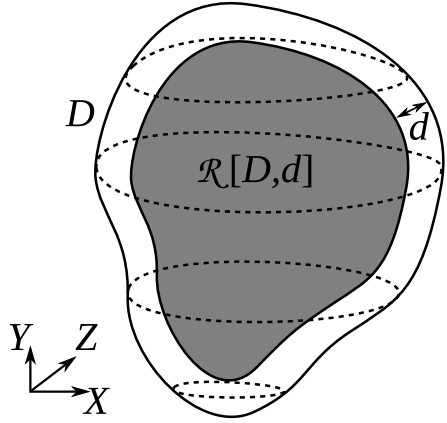,width=6cm}
\caption{
$d$-reduced region ${\cal R}[D,d]$ of a set $D$.
}
\label{fig:c4_6}
\end{figure}

\begin{definition}
Let $d>0$ be a constant. Let $\partial D$ denote the boundary of a closed set $D$. The $d$-reduction of the closed set $D$ is a set ${\cal R}[D,d]$ defined as follows (see Fig. \ref{fig:c4_6}): 
\begin{equation}
{\cal R}[D,d]:=\lbrace p\in D:\rho(\partial D,p)\geq d\rbrace.
\end{equation}
\end{definition}

\begin{figure}[!htb]
\centering
\epsfig{figure=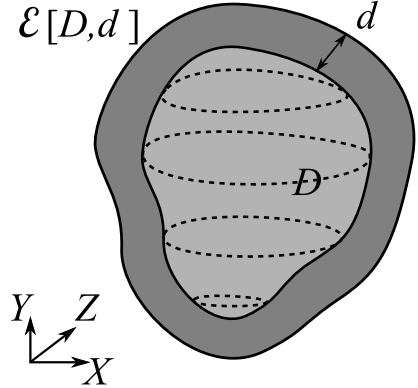,width=6cm}
\caption{
$d$-enlargement region ${\cal E}[D,d]$ of a set $D$.
}
\label{fig:c4_20}
\end{figure}

\begin{definition}
Let $d>0$ be a constant. The $d$-enlargement of the closed set $D$ is a set ${\cal E}[D,d]$ defined as follows (see Fig. \ref{fig:c4_20}): 
\begin{equation}
{\cal E}[D,d]:=\lbrace p\in \mathbb{R}^3:\rho(D,p)\leq d\rbrace.
\end{equation}
\end{definition}

\begin{figure}[!htb]
\centering
\epsfig{figure=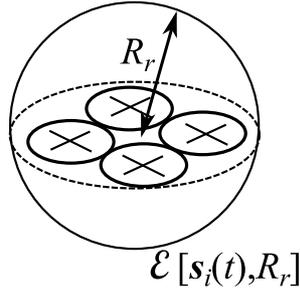,width=4cm}
\caption{
Physical size of a flying robot.
}
\label{fig:c4_18}
\end{figure}

\begin{assumption}
\label{c4_AS2}
The physical size of any flying robot in the workspace can be covered by a sphere with a given radius $R_r$ on the center of the robot (see Fig. \ref{fig:c4_18}). The sphere is called robot sphere; i.e. the $i$-th robot sphere is the set ${\cal E}[\robots_i(t),R_r]$.
\end{assumption}

It is noticed that each micro flying robot's location is uploaded to the central computer. At the same time, it is also detected by the 3D range finder sensors. Therefore, we suppose that error correction for the robot's odometry is applied. There are a lot of methods in the error correction of odometry by 3D range finder, such as \cite{Held2016}.

\begin{figure}[!htb]
\centering
\epsfig{figure=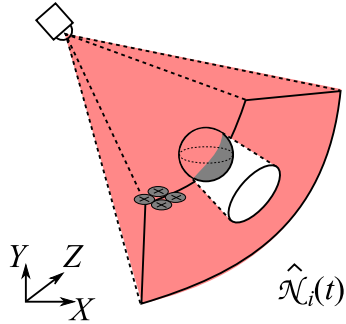,width=5cm}
\caption{
Estimated region $\hat{\cal N}_i(t)$ without the effect of micro flying robots for the $i$-th sensor node.
}
\label{fig:c4_5}
\end{figure}

Moreover, because the location of each robot is known and the size of each robot is micro, a filter should be involved to remove the effect of any robots on each 3D region ${\cal N}_i(t)$, $i=1,2,\ldots$; i.e. the 3D region which is detected without any micro flying robot in the field of view by the $i$-th sensor node should be estimated according to both the detected region ${\cal N}_i(t)$ and the positions of all the flying robots. Here, let $\hat{\cal N}_i(t)$ denote the estimated region of ${\cal N}_i(t)$ by the filter (see Fig. \ref{fig:c4_5}).

\begin{figure}[!htb]
\centering
\epsfig{figure=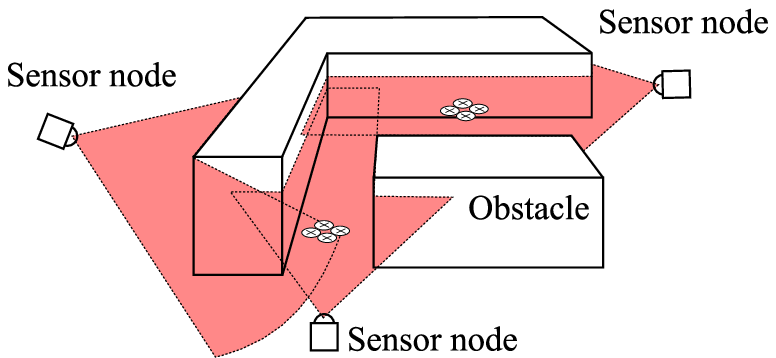,width=9cm}
\caption{
Unoccupied area ${\cal A}(t)$.
}
\label{fig:c4_19}
\end{figure}

Now, an unoccupied area, denoted by ${\cal A}(t)$, can be calculated as follows (see Fig. \ref{fig:c4_19}):

\begin{equation}
{\cal A}(t)=\bigcap_{i=1}^{m}\hat{\cal N}_i(t),
\end{equation}
where $m$ is the number of the sensor nodes.

\begin{assumption}
\label{c4_AS1}
Let $p$ be an arbitrary point in the set ${\cal A}(t)$ and $\partial {\cal A}(t)$ denote the boundary of ${\cal A}(t)$. The derivative of the minimum distance $\rho(\partial {\cal A}(t),p)$ with respect to time $t$ is $\leq V_{\max}$, where $V_{\max}$ is a given constant, which is $<v_r$.
\end{assumption}

\begin{remark}
If Assumption \ref{c4_AS1} does not hold, the robot may fail to avoid the dynamic obstacles and undetected regions, which may contain undetected obstacles.
\end{remark}

\begin{definition}
The safety margin $d_s>0$ is a given constant that the flying robots should keep from the boundary $\partial{\cal A}(t)$ at any time $t$. It satisfies the following inequality:
\begin{equation}
d_s\geq R_{\text{min}}.
\end{equation}
\end{definition}

\begin{assumption}
The robot's initial position $\robots(0)$ belongs to the set ${\cal R}[{\cal A}(0),d_s]$ and is far away from the boundary of ${\cal A}(0)$.
\end{assumption}

\begin{assumption}
The set ${\cal R}[{\cal A}(t),d_s]$ is a connected set at any time $t$.
\end{assumption}

The objective of the proposed algorithm is to drive a micro flying robots to travel in the dynamic and deformable $d_s$-reduced region ${\cal R}[{\cal A}(t),d_s]$, avoid any other flying robot and reach a target, which is denoted by ${\cal T}$, with a relatively short trajectory. In the presented method, the environment measurements, robot's position and heading, and robots' control signals velocity are exchanged through the sensor network to construct a networked control system; see examples \cite{1193753,MATVEEV200451problem,
SAVKIN200651Analysis,Matveev2009Estimation}.

\begin{assumption}
The target ${\cal T}$ belongs to the set ${\cal R}[{\cal A}(t),d_s]$ at any time $t$.
\end{assumption}

\begin{assumption}
The target ${\cal T}$ is far away from the robot's initial position $\robots(0)$.
\end{assumption}

\section{Safe navigation algorithm}

In this section, we propose a discrete-time navigation algorithm for the micro flying robots in the workspace based on the presented WSN. Each flying robot tracks a relatively short and temporarily safe path denoted by $P^*$, which is generated by the central computer according to the proposed navigation algorithm.

\begin{figure}[!htb]
\centering
\epsfig{figure=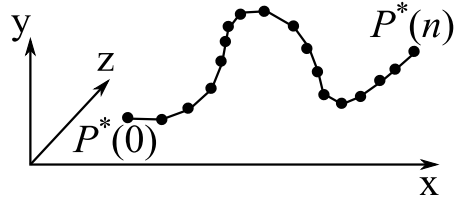,width=6cm}
\caption{
A 3D robot path represented as some equally spaced points.
}
\label{fig:c4_7}
\end{figure}

The path $P^*$ is represented as some successive and finite equally spaced points denoted by $P^*(0),P^*(1),\ldots,P^*(n)$ (see Fig. \ref{fig:c4_7}), where $n$ is the length of the path. $P^*(0)$ is the flying robot's current position $\robots(t)$. For any $k$, $P^*(k)$ represents the possible position of the robot at the future time $t+k\delta$. The distance between any two successive points is a constant $L$, which is equal to $v_r\delta$. $P^*$ should approximately satisfy the non-holonomic constraint of the flying robots; i.e. the radius of the circumscribed circle of any three successive points should be $\geq R_{\text{min}}$ and the point $P^*(1)$ should not be within the initial torus ${\cal Q}$.

In the proposed method, we consider two conditions of a micro flying robot. In the first condition, a flying robot is steady at its current position before a target is arranged or after it reach a target. In this case, $P^*$ has only one point, i.e. $n=0$. In the second condition, the flying robot is travelling to an arranged target with a generated path $P^*$, i.e. $n>0$.

Let $\delta$ be the sampling interval. At each time step $t=0,\delta,2\delta,\ldots$, the central computer calculates the unoccupied area ${\cal A}(t)$ and obtains the flying robots' real-time positions $\robots_1(t),\robots_2(t),\ldots$ via the WSN. Then, at each time step, the paths $P^*_1,P^*_2,\ldots$ for all the robots are updated successively according to the real-time unoccupied area and the robots' positions.

Let $T$ be a given time window ,which is a positive integer. The generated path $P^*$ has the following properties. Firstly, $P^*$ is a path from the robot's current position $\robots(t)$ to the target ${\cal T}$. Secondly, $P^*$ satisfies the non-holonomic constraint of the robot's motion. Thirdly, $P^*$ is guaranteed to be collision-free to the obstacles and other robots over the time period $[t,t+T\delta]$. Because $P^*$ is updated at each time step with interval $\delta$, the robot finally can reach the target successful without any collisions. 

To guarantee the safety of each robot in the partly detected dynamic environment, we consider any potentially unsafe regions excluding robot spheres in the environment and give the following definitions and assumptions.

\begin{definition}
Let $k=0,1,2,\ldots$ and $t$ be the current time. Over the time window $T$, the region which is obsoletely safe at time $t+k\delta$, $k\leq T$ is

\begin{equation}
\bar{\cal A}(t,k):={\cal R}[{\cal A}(t),k\delta V_{\max}].
\end{equation}
Then we extend it as follows to help to generate a temporarily safe path in the time window $T$:

\begin{equation}
\begin{array}{ll}
\bar{\cal A}(t,k):={\cal R}[{\cal A}(t),T\delta V_{\max}] & ,k> T.
\end{array}
\end{equation}
\end{definition}

\begin{definition}
Let $k=0,1,2,\ldots$ and $t$ be the current time. For the $i$-th micro flying robot, considering the collisions with other robots, a valid area $\hat{\cal A}(t,k)$ at any future time $t+k\delta$ is defined as follows:
\begin{equation}
\hat{\cal A}(t,k):=\bar{\cal A}(t,k)\setminus \bigcup_{j\neq i}{\cal E}[P^*_j(k),R_r].
\end{equation}
Here we extend any robot's path $P^*_j$ as follows
\begin{equation}
\begin{array}{ll}
P^*_j(k):=P^*_j(n) & ,k>n
\end{array}
\end{equation}
and $\setminus$ denotes the set difference.
\end{definition}

\begin{assumption}
For any $t$ and time step $k$, the target point ${\cal T}$ belongs to the set ${\cal R}[\hat{\cal A}(t,k),d_s]$.
\end{assumption}

\begin{assumption}
For any $t$ and $k$, the set ${\cal R}[\hat{\cal A}(t,k),d_s]$ is a connected set.
\end{assumption}

\begin{definition}
\label{c4_DE1}
The path $P^*$ is called a temporarily safe path, if for any $k$ and $t$, the point $P^*(k)$ belongs to the set ${\cal R}[\hat{\cal A}(t,k),d_s]$.
\end{definition}

\begin{definition}
\label{c4_DE2}
The path $P^*$ is called a target-reaching path if $\Vert {\cal T}-P^*(n)\Vert\leq L$.
\end{definition}

Now, we are here to briefly introduce the two parts of the proposed navigation algorithm. The first part of the algorithm is rough path generation. It is followed by the second part, which is path planning. At each time step, for any flying robot, if a target is arranged and the robot needs to be navigated to the target, an initial rough path $\bar{P}$ need to be generated firstly by the probabilistic roadmap algorithm. Then the generated rough path $\bar{P}$ is adjusted to be the path $P^*$ by the proposed path planning algorithm. With the presented algorithm, any micro flying robot navigated by the WSN can avoid any obstacles and other flying robots in the workspace and reach the target successfully.

\subsection{Rough path generation}

To initialize a temporarily safe rough path for a micro flying robot with an arranged target ${\cal T}$, the probabilistic roadmap method is used to generate a relatively short rough path $\bar{P}$ in the area ${\cal R}[{\cal A}(t),d_s+T\delta V_{\max}]$. The probabilistic roadmap method proposed in \cite{Kavraki1996} is widely used in robotics to solve the motion planning problem; e.g. \cite{Roesmann2017}.

In the probabilistic roadmap method, a graph ${\cal G}$ is construct to search the shortest path from a starting point to a target point in the graph. The vertices of the graph ${\cal G}$ are the random samples of the configuration space of the robot, the starting point and the target point. Each edge of the graph ${\cal G}$ represents a valid path connecting the corresponding pair of vertices. Then, a shortest path search algorithm is used to find the shortest path from the starting point to the target point in the graph ${\cal G}$. In our method, the vertices of ${\cal G}$, excluding the starting point and the target point, are randomly taken from the area ${\cal R}[{\cal A}(t),d_s+T\delta V_{\max}]$ with uniform distribution. Then, we give a definition and the algorithm description as follows.

\begin{figure}[!htb]
\centering
\epsfig{figure=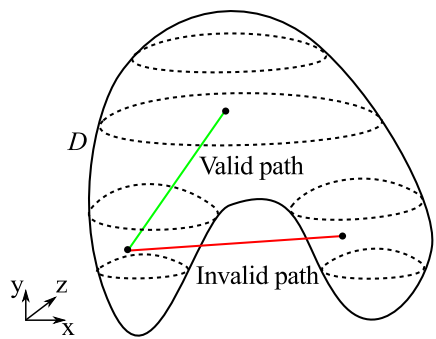,width=6cm}
\caption{
A valid path between two points.
}
\label{fig:c4_14}
\end{figure}

\begin{definition}
Let $D$ be a closed point set. A line segment path between any two points $a$, $b\in D$ is called a valid path if the line segment between $a$ and $b$ does not intersect with the boundary of $D$ (see Fig. \ref{fig:c4_14}).
\end{definition}

\begin{figure}[!htb]
\centering
\epsfig{figure=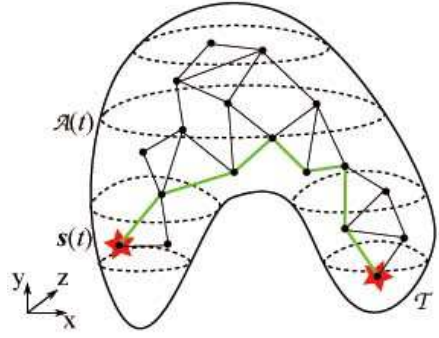,width=6cm}
\caption{
Rough path generation by probabilistic roadmap method.
}
\label{fig:c4_12}
\end{figure}

Now, we are here to give the rough path generation algorithm as follows (see Fig. \ref{fig:c4_12}):

\begin{enumerate}
\item[\textbf{A1}:]
Initialize a graph ${\cal G}=(V,E)$, where the set of vertices $V$ and the set of edges $E$ are empty sets.

\item[\textbf{A2}:]
Let $N_s>0$ be a given integer. Then, select $N_s$ points $\lbrace \xi_1,\xi_2,\ldots,\xi_{N_s} \rbrace$ randomly from the point set ${\cal R}[{\cal A}(t),d_s+T\delta V_{\max}]$ with uniform distribution. Then, add these selected points into the set $V$ as follows:
\begin{equation}
V\leftarrow V\cup \lbrace \xi_1,\xi_2,\ldots,\xi_{N_s} \rbrace.
\end{equation}

\item[\textbf{A3}:]
Let $N_c>0$ be a given integer that is smaller than $N_s$. For any point $\xi_i\in V$, $i=1,2,\ldots,N_s$, choose the $N_c$ closest vertices, denoted by $\lbrace\xi'_{i,1},\xi'_{i,2},\ldots,\xi'_{i,N_c}\rbrace$, in $V$ according to the distance. For each chosen vertex $\xi'_{i,j}\in V$, $j=1,2,\ldots,N_c$, if the line segment between $\xi_i$ and $\xi'_{i,j}$ is a valid path in ${\cal R}[{\cal A}(t),d_s+T\delta V_{\max}]$, add the corresponding edge into the set $E$ as follows:
\begin{equation}
E\leftarrow E\cup \lbrace (\xi_i,\xi'_{i,j}) \rbrace.
\end{equation}

\item[\textbf{A4}:]
Let $\xi_{\text{init}}$ be the flying robot's current position $\robots(t)$ and add it into the set $V$ as follows:
\begin{equation}
V\leftarrow V\cup \lbrace \xi_{\text{init}} \rbrace.
\end{equation}
For the starting vertex $\xi_{\text{init}}\in V$, choose the $N_c$ closest vertices, denoted by $\lbrace\xi'_{\text{init},1},\xi'_{\text{init},2},\ldots,\xi'_{\text{init},N_c}\rbrace$ in $V$ according to the distance. Considering the constraint of the control input (\ref{c4_EQ1}), for each chosen vertex $\xi'_{\text{init},j}\in V$, $j=1,2,\ldots,N_c$, if the line segment between $\xi_{\text{init}}$ and $\xi'_{\text{init},j}$ is a valid path in ${\cal R}[{\cal A}(t),d_s+T\delta V_{\max}]$ and the following inequality holds:
\begin{equation}
\arccos(\frac{a(t)\cdot \zeta_{j}}{\Vert a(t)\Vert\cdot\Vert\zeta_{j}\Vert})\leq \arcsin(\frac{L\cdot U_M}{2 v_r}),
\end{equation}
where $\zeta_{j}=\xi'_{\text{init},j}-\xi_{\text{init}}$, then add the corresponding edge into the set $E$ as follows:
\begin{equation}
E\leftarrow E\cup \lbrace (\xi_{\text{init}},\xi'_{\text{init},j}) \rbrace.
\end{equation}

\item[\textbf{A5}:]
Let $\xi_{\text{goal}}$ be the target point ${\cal T}$ and add it into the set $V$ as follows:
\begin{equation}
V\leftarrow V\cup \lbrace \xi_{\text{goal}} \rbrace.
\end{equation}
For the goal vertex $\xi_{\text{goal}}\in V$, choose the $N_c$ closest vertices, denoted by $\lbrace\xi'_{\text{goal},1},\xi'_{\text{goal},2},\ldots,\xi'_{\text{goal},N_c}\rbrace$ in $V$ according to the distance. For each chosen vertex $\xi'_{\text{goal},j}\in V$, $j=1,2,\ldots,N_c$, if the line segment between $\xi_{\text{goal}}$ and $\xi'_{\text{goal},j}$ is a valid path in ${\cal R}[{\cal A}(t),d_s+T\delta V_{\max}]$, then add the corresponding edge into the set $E$ as follows:
\begin{equation}
E\leftarrow E\cup \lbrace (\xi_{\text{goal}},\xi'_{\text{goal},j}) \rbrace.
\end{equation}

\item[\textbf{A6}:]
Search the shortest path from the starting vertex $\xi_{\text{init}}$ to the goal vertex $\xi_{\text{goal}}$ in the graph ${\cal G}$. Generally, Dijkstra's algorithm can be used to search the shortest path in a graph.

\item[\textbf{A7}:]
Convert the selected shortest path to the rough path $\bar{P}$, which consists of equally spaced points with interval $L$.
\end{enumerate}

\subsection{Path planing}

Now, we are here to propose the path planning algorithm that adjusts a given flying robot path to be an temporarily safe path $P^*$, which is relatively short in the valid area ${\cal R}[\hat{\cal A}(t,k),d_s]$. Let $P$ denote the given path that needs to be adjusted. It can be the initialized rough path $\bar{P}$ or the last temporarily safe $P^*$ at the last time step.

Here, let $p_0,p_1,\ldots,p_n$ denote the path points of the given path $P$. Then, we define four vector fields $F_I$, $F_R$, $F_P$ and $F_C$ in $\mathbb{R}^3$ to help adjust the path points of $P$. For any $k\neq 0$, the resultant vector of these four fields at point $p_k$ is
\begin{equation}
\vect{F}(p_k)=\vect{F}_I(p_k)+\vect{F}_R(p_k)+\vect{F}_P(p_k)+\vect{F}_C(p_k).
\end{equation}
The path $P$ is adjusted by moving any $p_k$, $k\neq 0$ towards the direction of $\vect{F}(p_k)$ to an equilibrium point where $\vect{F}(p_k)=\vect{0}$. Moreover, while adjusting, some points would be added or removed to prolong or shorten the length of $P$ until $\vect{F}(p_k)=\vec{0}$ for any $k\neq 0$. Now, we are here to give the definitions of four vector fields.

\begin{figure}[!htb]
\centering
\epsfig{figure=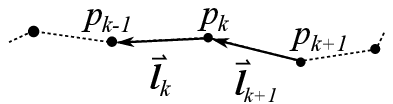,width=5cm}
\caption{
Vector $\vect{l}_k$.
}
\label{fig:c4_8}
\end{figure}

\textbf{Definition of} $\bf F_I$:
$F_I$ is a vector field which guarantees that the interval between any two successive points of $P$ is approximately equal to $L$. For any $k\neq 0$, let $\vect{l}_k$ be the vector from $p_k$ to $p_{k-1}$ (see Fig. \ref{fig:c4_8}), then $\vect{F}_I(p_{k})$ is defined as follows:

\begin{equation}
\vect{F}_{I}(p_{k}):=
\begin{cases}
G_I\cdot (b(k)\vect{l}_k-b(k+1)\vect{l}_{k+1}) & \mathrm{if}\ k\neq n
\\
G_I\cdot (b(k)\vect{l}_k) & \mathrm{if}\ k=n
\end{cases},
\end{equation}
where $G_I>0$ is a tunable gain and $b(k)$ is

\begin{equation}
b(k):=1-\frac{L}{\Vert \vect{l}_k \Vert}.
\end{equation}

\begin{figure}[!htb]
\centering
\epsfig{figure=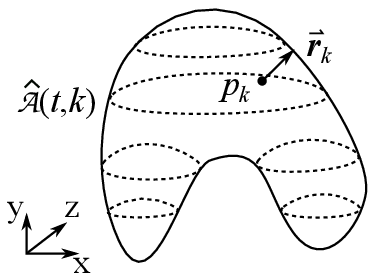,width=6cm}
\caption{
Vector $\vect{r}_k$.
}
\label{fig:c4_9}
\end{figure}

\textbf{Definition of} $\bf F_R$:
$F_R$ is a vector field which guarantees that the minimum distance from the point $p_k$ to the boundary of the valid area $\hat{\cal A}(t,k)$ is $\geq d_s$ for any $k\neq 0$. Let $\vect{r}_k$ be the shortest vector from $p_{k}$ to the boundary of $\hat{\cal A}(t,k)$, then $\vect{F}_R(p_{k})$ is defined as follows (see Fig. \ref{fig:c4_9}):

\begin{equation}
\vect{F}_R(p_{k}):=
\begin{cases}
G_R\cdot (1-\frac{d_s}{\Vert\vect{r}_k\Vert}) \vect{r}_k & \mathrm{if}\ \Vert\vect{r}_k\Vert\leq d_s
\\
\vect{0} & \mathrm{if}\ \Vert\vect{r}_k\Vert > d_s
\end{cases},
\end{equation}
where $G_R>0$ is a tunable gain.

\begin{figure}[!htb]
\centering
\epsfig{figure=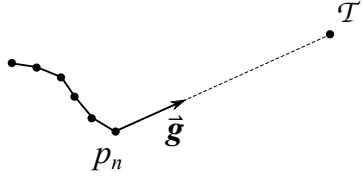,width=5cm}
\caption{
Direction vector $\vect{g}$.
}
\label{fig:c4_10}
\end{figure}

\textbf{Definition of} $\bf F_P$:
$F_P$ is a vector field pointing towards the target ${\cal T}$, which only attracts the last point $p_n$ of the path $P$. For any other $k\neq n$, $\vect{F}_P(p_k)=\vect{0}$. For the point $p_n$, let $\vect{g}$ be a unit vector pointing towards ${\cal T}$ from $p_{n}$ (see Fig. \ref{fig:c4_10}), then $\vect{F}_P(p_{n})$ is defined as follows:

\begin{equation}
\vect{F}_P(p_{n}):=G_P\cdot \vect{g},
\end{equation}
where $G_P>0$ is a tunable gain.

\textbf{Definition of} $\bf F_C$:
$F_C$ is a vector field which guarantees that the path satisfies the non-holonomic constraint and does not intersect with the initial torus ${\cal Q}$ at the beginning. For any $k\neq 0$, let $\vect{h}_k$ be the vector from the point $p_k$ to the closest point at the circle $B$, then $\vect{F}_C(p_{k})$ is defined as follows:

\begin{equation}
\vect{F}_C(p_{k}):=
\begin{cases}
G_C\cdot (1-\frac{R_{\min}}{\Vert\vect{h}_k\Vert}) \vect{h}_k & \mathrm{if}\ \Vert\vect{h}_k\Vert\leq R_{\min} 
\\
\vect{0} & \mathrm{if}\ \Vert\vect{h}_k\Vert > R_{\min} 
\end{cases},
\end{equation}
where $G_C>0$ is a tunable gain.

Now, we are here to propose the path planing algorithm as follows to adjust the given path $P$:

\begin{enumerate}
\item[\textbf{B1}:]
For any $p_k$, $k\neq 0$, let $\vect{p}_k$ denote the position vector of $p_k$, then initialize a velocity vector $\vect{v}_k=\vect{0}$ for $p_k$.

\item[\textbf{B2}:]
Start the following loop:

\begin{enumerate}
\item[\textbf{B2.1}:]
Change the position vector and velocity vector of each $p_k$, $k\neq 0$ as follows:

\begin{equation}
\left\{
\begin{array}{l}

\vect{p}_{k} \leftarrow \vect{p}_{k}+\vect{v}_{k}
\\
\vect{v}_{k} \leftarrow G_N \cdot \vect{v}_{k}+\vect{F}(p_k)

\end{array},
\right.
\end{equation}
where $0<G_N<1$ is a tunable attenuation.

\item[\textbf{B2.2}:]
If $\Vert p_n-{\cal T}\Vert<\underline{L}$, remove $p_n$ from $P$ and $n\leftarrow n-1$. If $\Vert p_n-{\cal T}\Vert>\overline{L}$, add a new point next to $p_n$ into $P$ and $n\leftarrow n+1$. The parameters $\underline{L}<L$ and $\overline{L}>L$ are two given thresholds.

\end{enumerate}

\item[\textbf{B3}:]
Exit loop \textbf{B2} if the inequality $\Vert \vec{F}(p_k)\Vert<F_{th}$ holds for any $k\neq 0$, where $F_{th}>0$ is a given threshold.
\end{enumerate}

\begin{figure}[!htb]
\centering
\subfigure[The given path $P$]{

\epsfig{figure=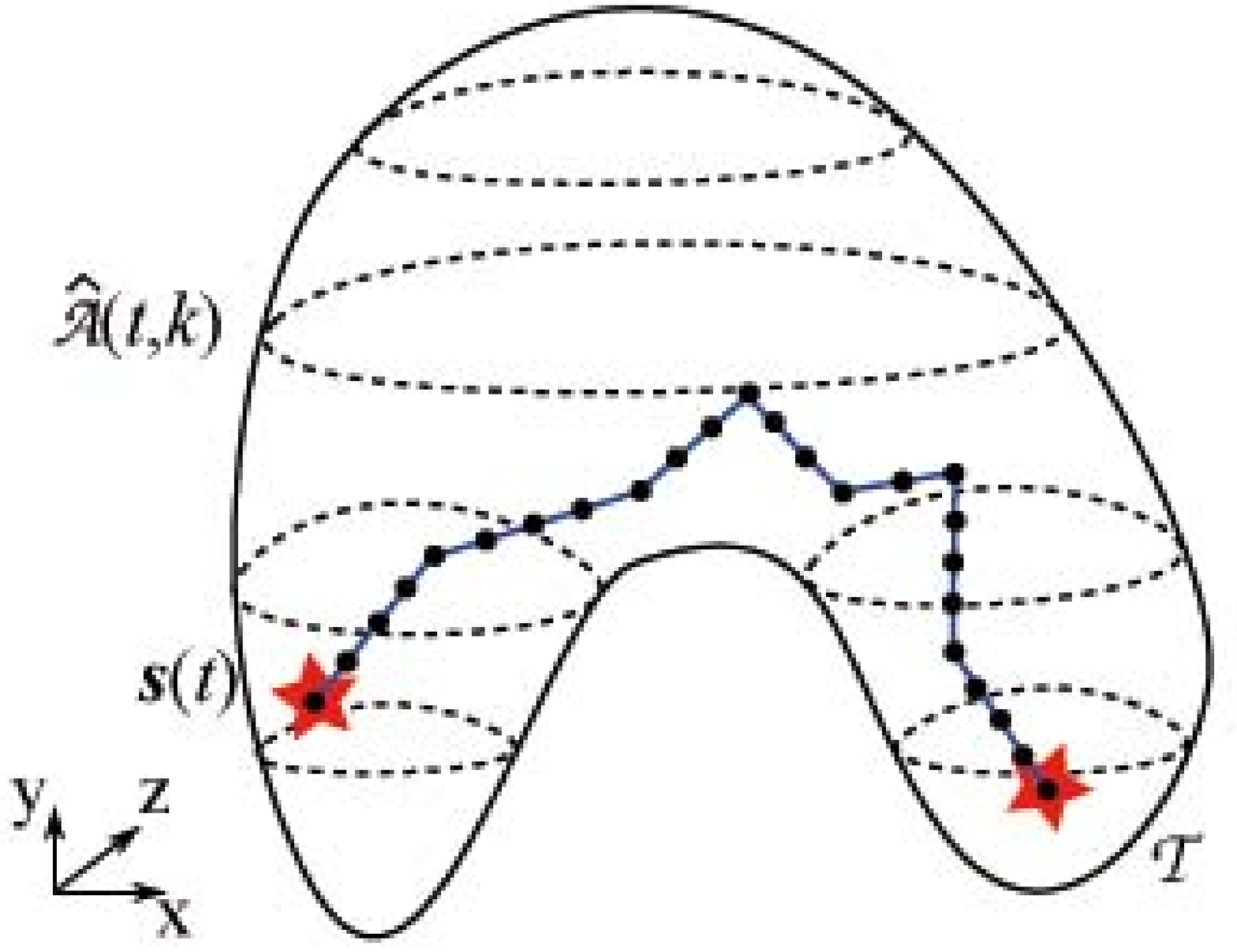,width=6cm}
}
\subfigure[The temporarily safe path $P^*$]{

\epsfig{figure=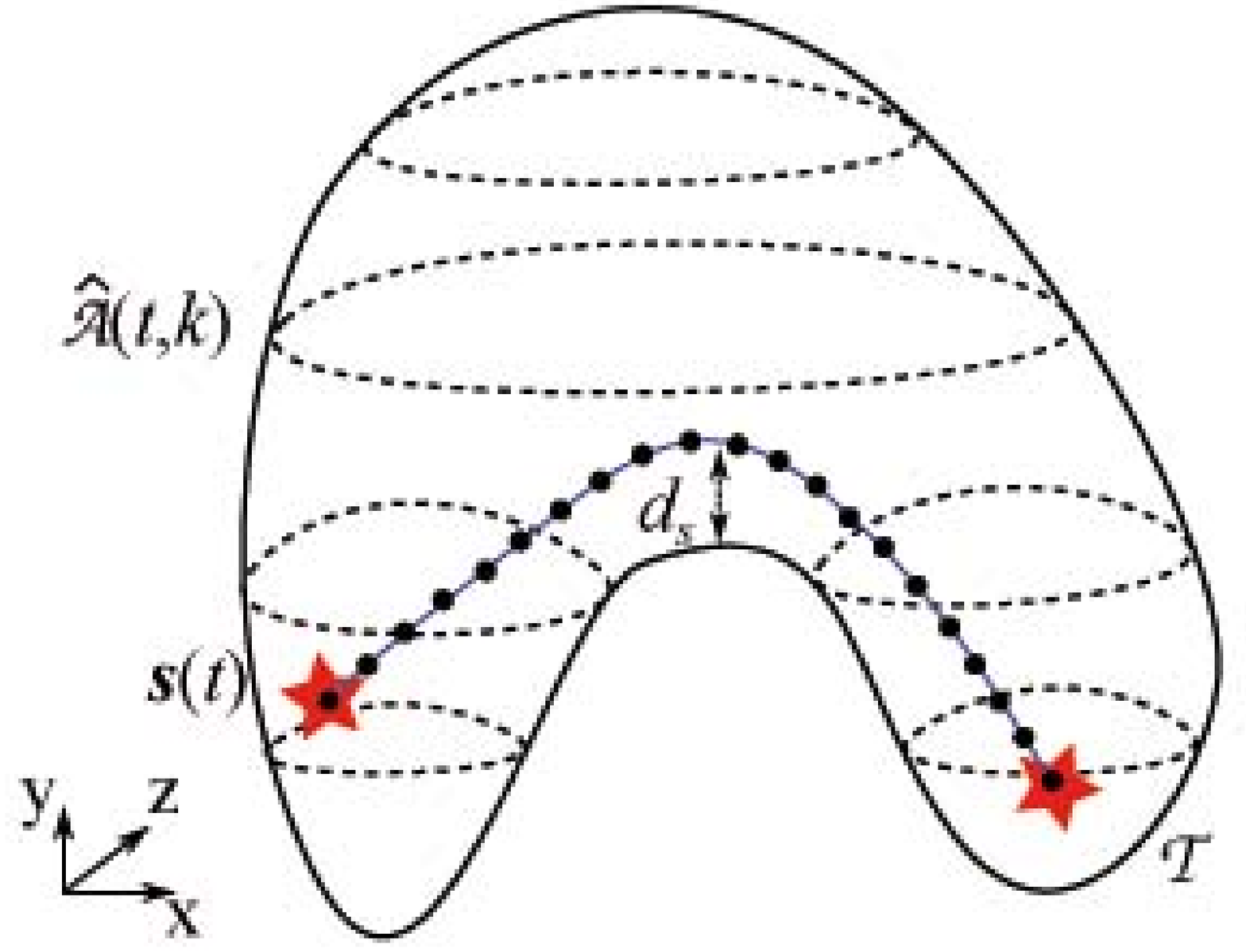,width=6cm}
}
\caption{
Adjustment of the given path $P$.
}
\label{fig:c4_13}
\end{figure}

With the proposed path planning algorithm, the temporarily safe path $P^*$ can be obtained for current time step. It satisfies the safety criteria defined in Definition \ref{c4_DE1}, \ref{c4_DE2} and the non-holonomic constraint (\ref{c4_EQ1}) (see Fig. \ref{fig:c4_13}).

\section{Computer simulations}

In this section, computer simulations are carried out to confirm the performance of the proposed navigation algorithm in an practical environment. Moreover, a path tracking controller for the micro flying robot is proposed firstly.

\subsection{Path tracking controller}
Firstly, to drive the flying robot to track the generated path $P^*$ during each sampling interval, we are here to propose a simple control strategy for flying robot path tracking. The proposed strategy is a modification of the control law which is proposed in \cite{Matveev2011a}.

\begin{figure}[!htb]
\centering
\epsfig{figure=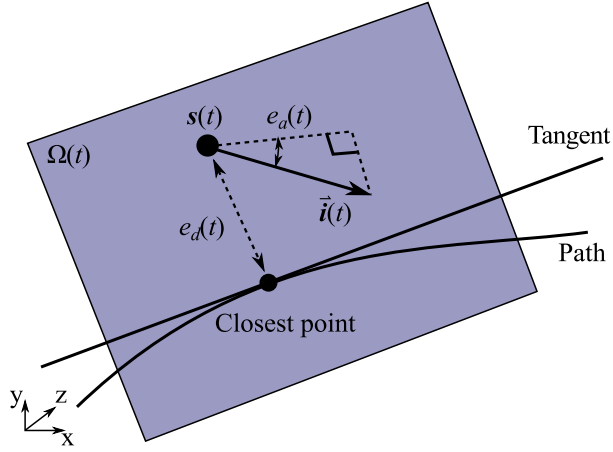,width=8cm}
\caption{
Minimum distance $e_d(t)$ and the angle $e_a(t)$.
}
\label{fig:c4_15}
\end{figure}

In 3D flying robot path tracking, there are two errors which should be minimized. The first error denoted by $e_d(t)$ is the minimum distance from the robot's real-time position $\robots(t)$ to the generated path. The second error denoted by $e_a(t)$ is the angle between the robot's velocity direction vector $\vect{i}(t)$ and the plane $\Omega(t)$, which is determined by the robot's position $\robots(t)$ and the tangent line of the path at the closest point to $\robots(t)$ (see Fig. \ref{fig:c4_15}).

\begin{notation}
The inner product of two vectors is denoted by $\bullet$ and the cross product of two vectors is denoted by $\times$.
\end{notation}

In the proposed control strategy, two sliding mode controllers are used to minimize the distance error $e_d(t)$ and the orientation error $e_a(t)$ respectively. Let $\vect{\nu}_a(t)$ be a normal vector of the plane $\Omega(t)$. The orientation error $e_a(t)$ can be obtained by the following equation:

\begin{equation}
e_a(t)=\arccos(\frac{\vect{i}(t)\bullet \vect{\nu}_a(t)}{\Vert \robots(t) \Vert\cdot\Vert \vect{\nu}_a(t) \Vert})-\frac{\pi}{2}.
\end{equation}

\begin{figure}[!htb]
\centering
\epsfig{figure=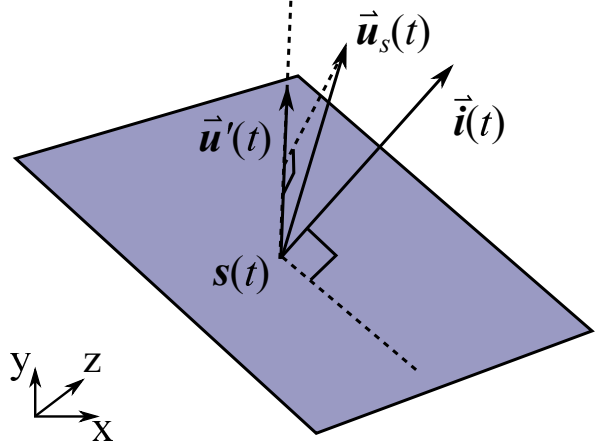,width=8cm}
\caption{
Control strategy of the flying robot path tracking.
}
\label{fig:c4_16}
\end{figure}

Let two tunable constants $\lambda_d>0$ and $\sigma_d>0$ be the parameters of the controller for the distance error $e_d(t)$; let another two tunable constants $\lambda_a>0$ and $\sigma_a>0$ be the parameters of the controller for the orientation error $e_a(t)$. Then, the two controllers for the two errors are designed as follows:

\begin{equation}
\left\{
\begin{array}{l}
u_d(t)=\text{sgn}[\dot{e}_d(t)+{\cal X}(e_d(t),\lambda_d,\sigma_d)]
\\
u_a(t)=\text{sgn}[\dot{e}_a(t)+{\cal X}(e_a(t),\lambda_a,\sigma_a)]
\end{array}.
\right.
\end{equation}
The saturation function ${\cal X}(z,\lambda,\sigma)$ is

\begin{equation}
{\cal X}(z,\lambda,\sigma):=
\begin{cases}
\lambda z & \mathrm{if}\ \vert z\vert\leq\sigma
\\
\lambda \sigma\text{sgn}(z) & \mathrm{if}\ \vert z\vert>\sigma
\end{cases}
\end{equation}
and the sign function $\text{sgn}(x)$ is

\begin{equation}
\text{sgn}(x)=
\begin{cases}
1 & \mathrm{if}\ x>0
\\
0 & \mathrm{if}\ x=0
\\
-1 & \mathrm{if}\ x<0
\end{cases}.
\end{equation}
Let $\vect{\nu}_d$ be the unit vector from the robot's current position $\robots(t)$ towards the closest point on the path. To minimize the two errors with one control input, we consider the sum of the two controller's outputs $u_d(t)$ and $u_a(t)$ in different directions $\vect{\nu}_d$ and $\vect{\nu}_a$ with the tunable weights $w_d>0$ and $w_a>0$ as follows:

\begin{equation}
\vect{u}_s(t)=w_d\cdot u_d(t)\cdot \vect{\nu}_d+w_a\cdot u_a(t)\cdot  \vect{\nu}_a.
\end{equation}
Moreover, considering the constraint of the robot's control input (\ref{c4_EQ1}), the direction vector of the orthogonal projection of the vector $\vect{u}_s(t)$ onto the plane, of which the normal vector is $\vect{i}(t)$, is calculated as follows (see Fig. \ref{fig:c4_16}):

\begin{equation}
\vect{u}'(t)=\vect{i}(t)\times \frac{\vect{u}_s(t)}{\Vert \vect{u}_s(t) \Vert}\times \vect{i}(t).
\end{equation}
Finally, the control input of the flying robot can is obtained as follows:

\begin{equation}
\vect{u}(t)=\vect{u}'(t)\cdot U_M.
\end{equation}

\begin{figure}[!htb]
\centering
\epsfig{figure=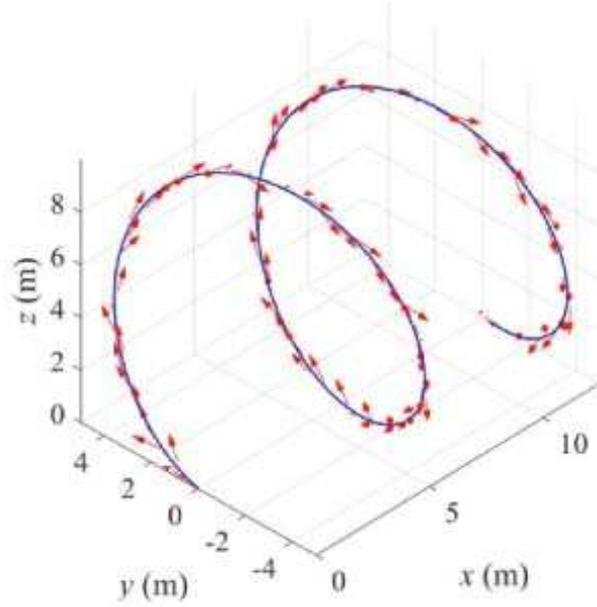,width=8cm}
\caption{
Result of the path tracking simulation. The blue solid line is the given 3D robot path and the red arrows are the trajectory and orientation of the flying robot.
}
\label{fig:c4_s_t_1}
\end{figure}

\begin{figure}[!htb]
\centering
\subfigure[]{
\epsfig{figure=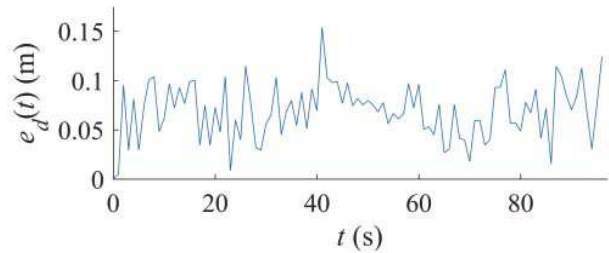,width=8cm}
}
\subfigure[]{
\epsfig{figure=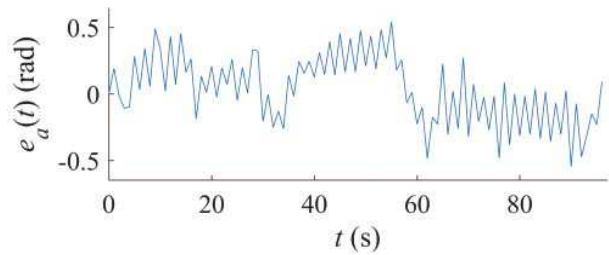,width=8cm}
}
\caption{
Distance error $e_d$ and the orientation error $e_a$.
}
\label{fig:c4_s_t_23}
\end{figure}

To confirm the performance of the proposed path tracking controller, we carry out a computer simulation. In this simulation, a 3D robot path is given as a smooth curve from the robot's initial position to a target. The flying robot's positions $\robots(t)$ and velocity direction vector $\vect{i}$ are known to calculate the distance error $e_d(t)$ and the orientation error $e_a(t)$. The parameters used in this simulation are indicated in Table \ref{c4_TB1}. The result of the simulation is shown in Fig. \ref{fig:c4_s_t_1}. It can be seen that the flying robot tracked the given path and reached the target. The distance error $e_d$ and the orientation error $e_a$ in this simulation are shown in Fig. \ref{fig:c4_s_t_23}, which indicates that the flying robot can be controlled accurately by the proposed path tracking controller.

\begin{table}[!htb]
\centering
\caption{Parameters for the Path Tracking Simulation}
\label{c4_TB1}
\begin{tabular}{c|c|c}
\hline
Speed of robot & $v_r$ & $0.7\text{m/s}$\\
\hline
Maximum angular velocity & $u_{M}$ & $2\text{rad/s}$\\
\hline
\multirow{6}{*}{Parameters of the controller}
 & $\lambda_d$ & $2$\\
\cline{2-3}
 & $\sigma_d$ & $1$\\
\cline{2-3}
 & $\lambda_a$ & $3$\\
\cline{2-3}
 & $\sigma_a$ & $1$\\
\cline{2-3}
 & $w_d$ & $1$\\
\cline{2-3}
 & $w_a$ & $1$\\ 
\hline
Sampling interval & $\delta$ & $0.2\text{s}$\\
\hline
\end{tabular}
\end{table}

\subsection{Time-of-flight camera}
\begin{figure}[!htb]
\centering
\epsfig{figure=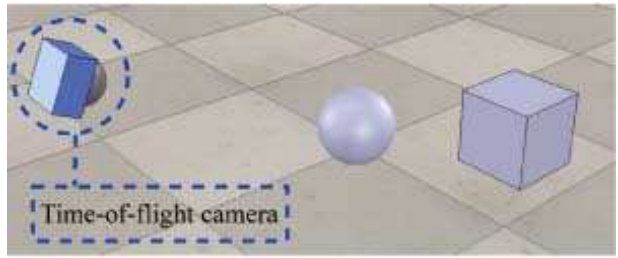,width=8cm}
\caption{
A ToF camera detects objects.
}
\label{fig:c4_s_i_1}
\end{figure}

\begin{figure}[!htb]
\centering
\epsfig{figure=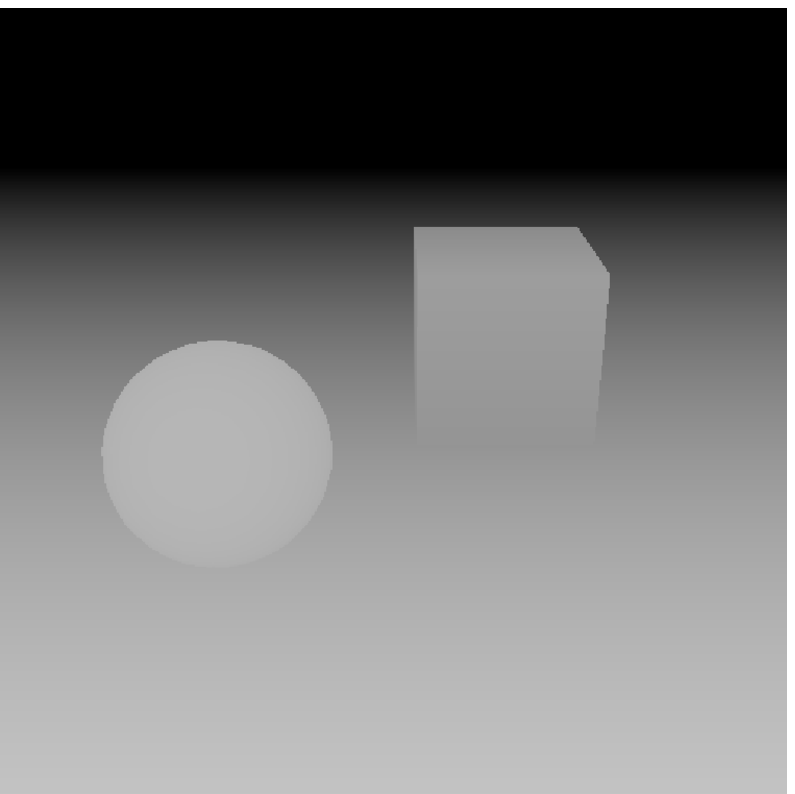,width=6cm}
\caption{
A depth image.
}
\label{fig:c4_s_i_2}
\end{figure}

\begin{figure}[!htb]
\centering
\epsfig{figure=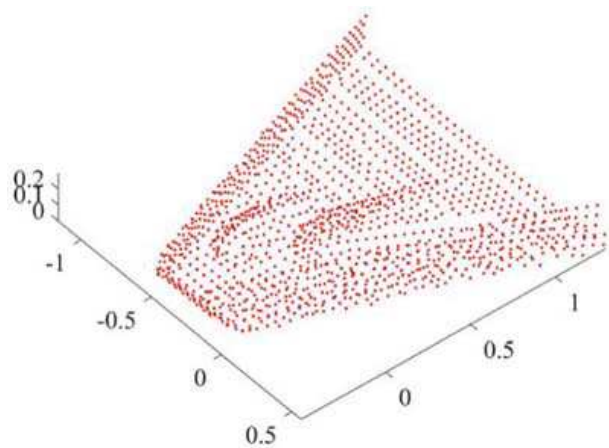,width=8cm}
\caption{
A point cloud of the ToF camera measurements.
}
\label{fig:c4_s_i_3}
\end{figure}

In our computer simulations, ToF camera are used to detect objects in the environments (see Fig. \ref{fig:c4_s_i_1}). It provides a depth image in the field of view with high frame rate, which perform a fast detection in dynamic environments. Fig. \ref{fig:c4_s_i_2} shows an example of the depth image, which is a grayscale image. According to the camera projection, the depth image can be converted to a point cloud to represent the objects in the environment (Fig. \ref{fig:c4_s_i_3}). The main parameters of the ToF camera is indicated in Table. \ref{c4_TB2}.

\begin{table}[!htb]
\centering
\caption{Parameters of the ToF Camera}
\label{c4_TB2}
\begin{tabular}{c|c}
\hline
Working range & $0\text{m to }9\text{m}$\\
\hline
Lens & $135^{\circ}\text{h}\times 135^{\circ}\text{v}$\\
\hline
Resolution & $128 \times 128$\\
\hline
\end{tabular}
\end{table}

\subsection{Simulations in static environments}

\begin{figure}[!htb]
\centering
\subfigure[]{
\epsfig{figure=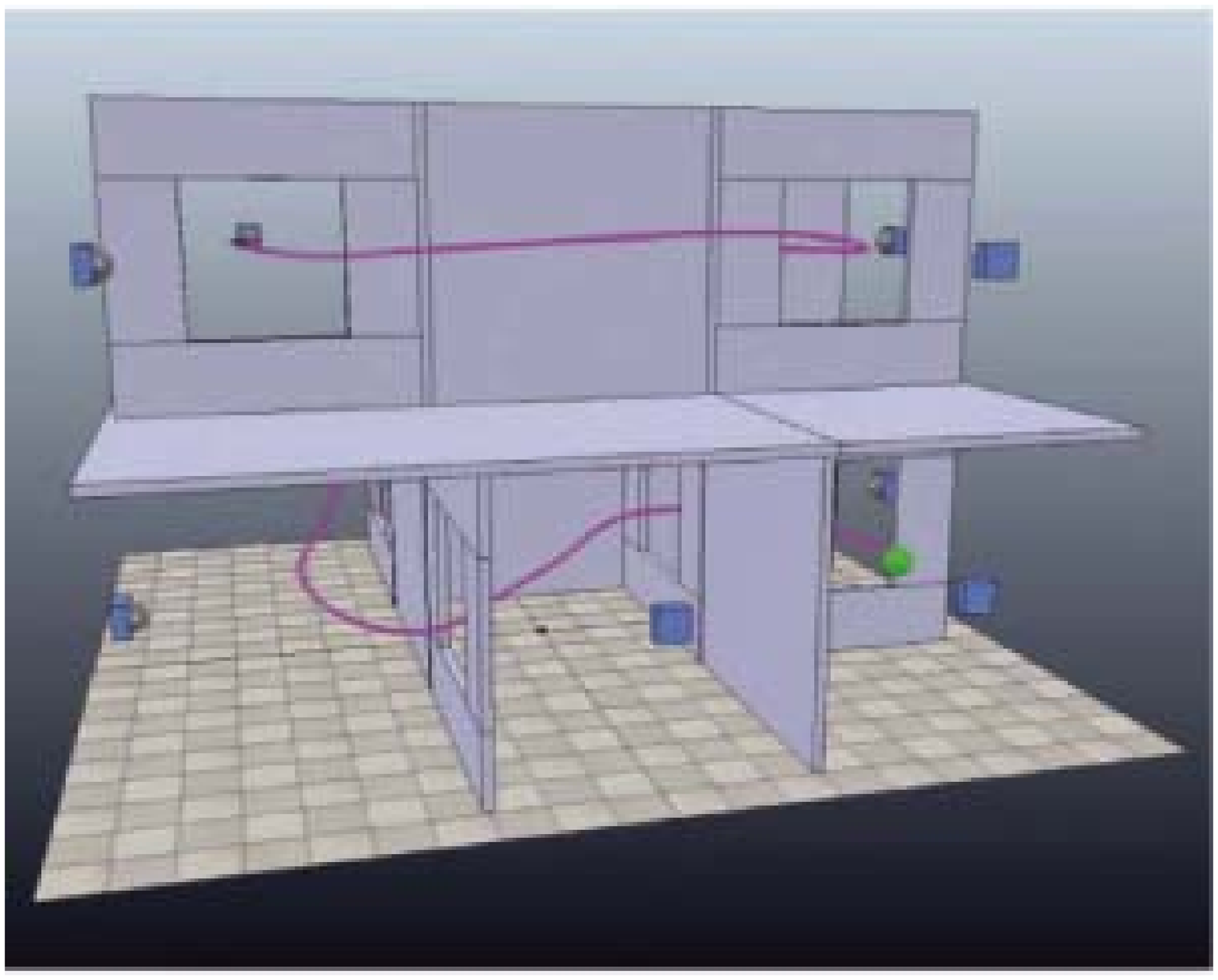,width=5cm}
}
\subfigure[]{
\epsfig{figure=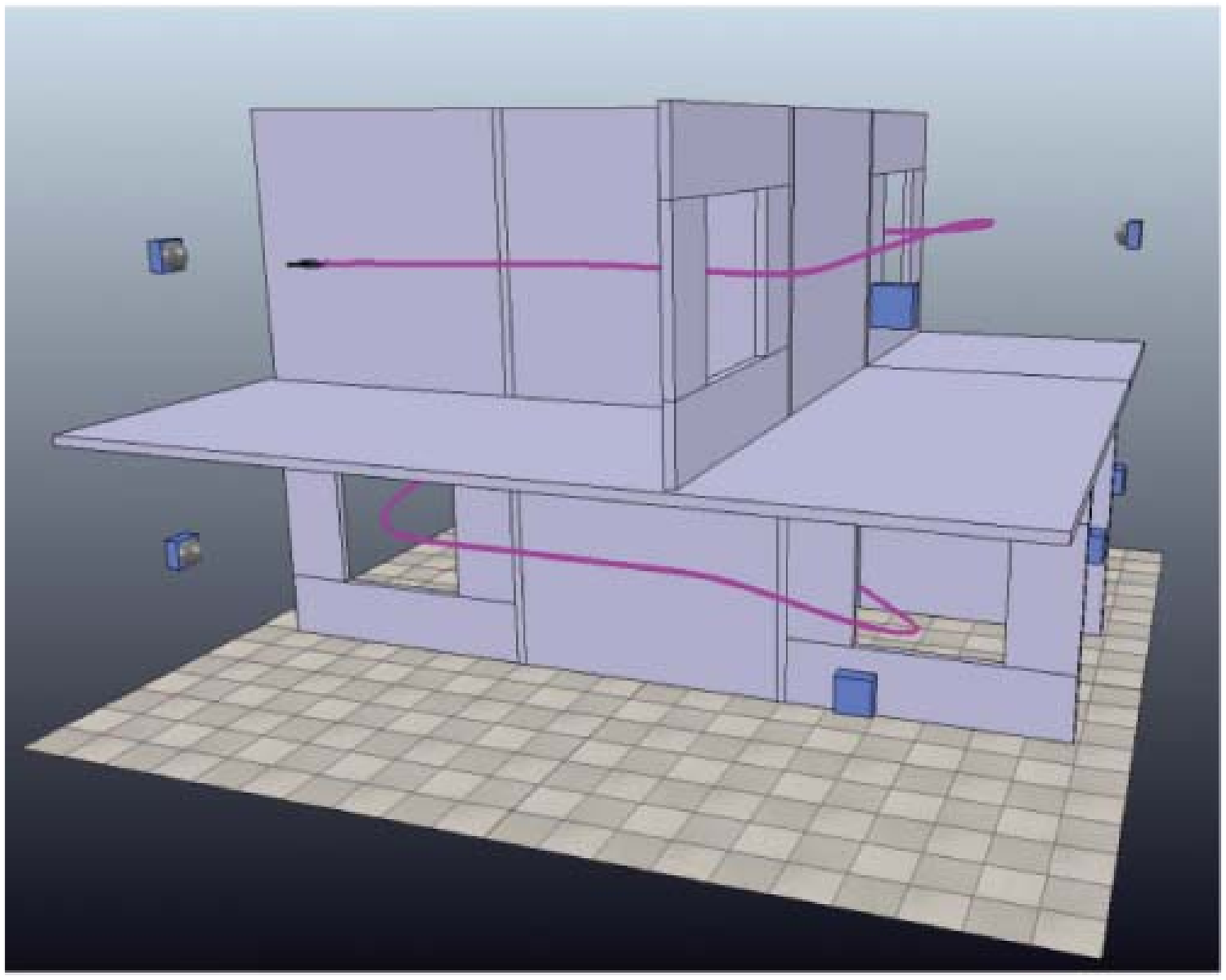,width=5cm}
}
\subfigure[]{
\epsfig{figure=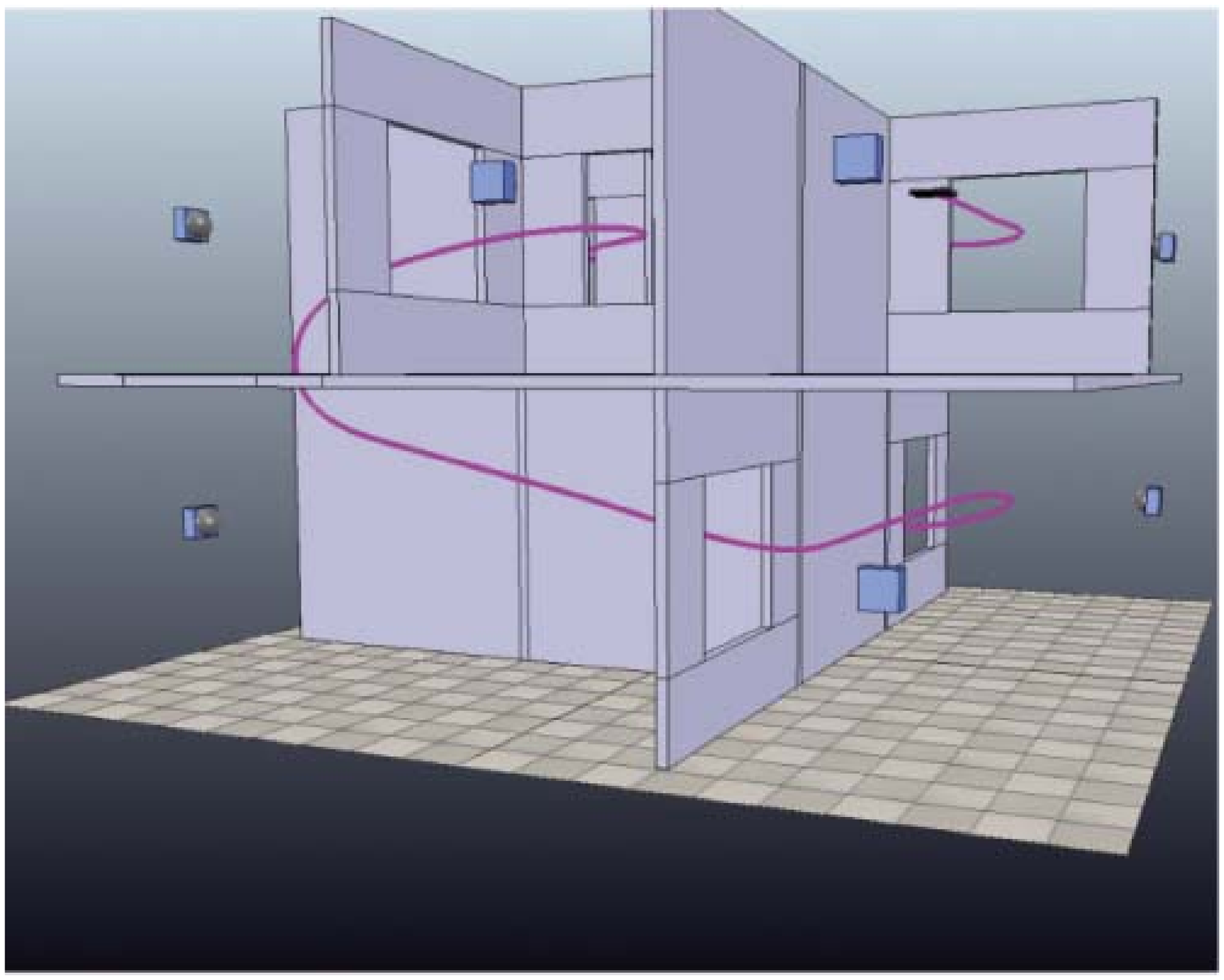,width=5cm}
}
\subfigure[]{
\epsfig{figure=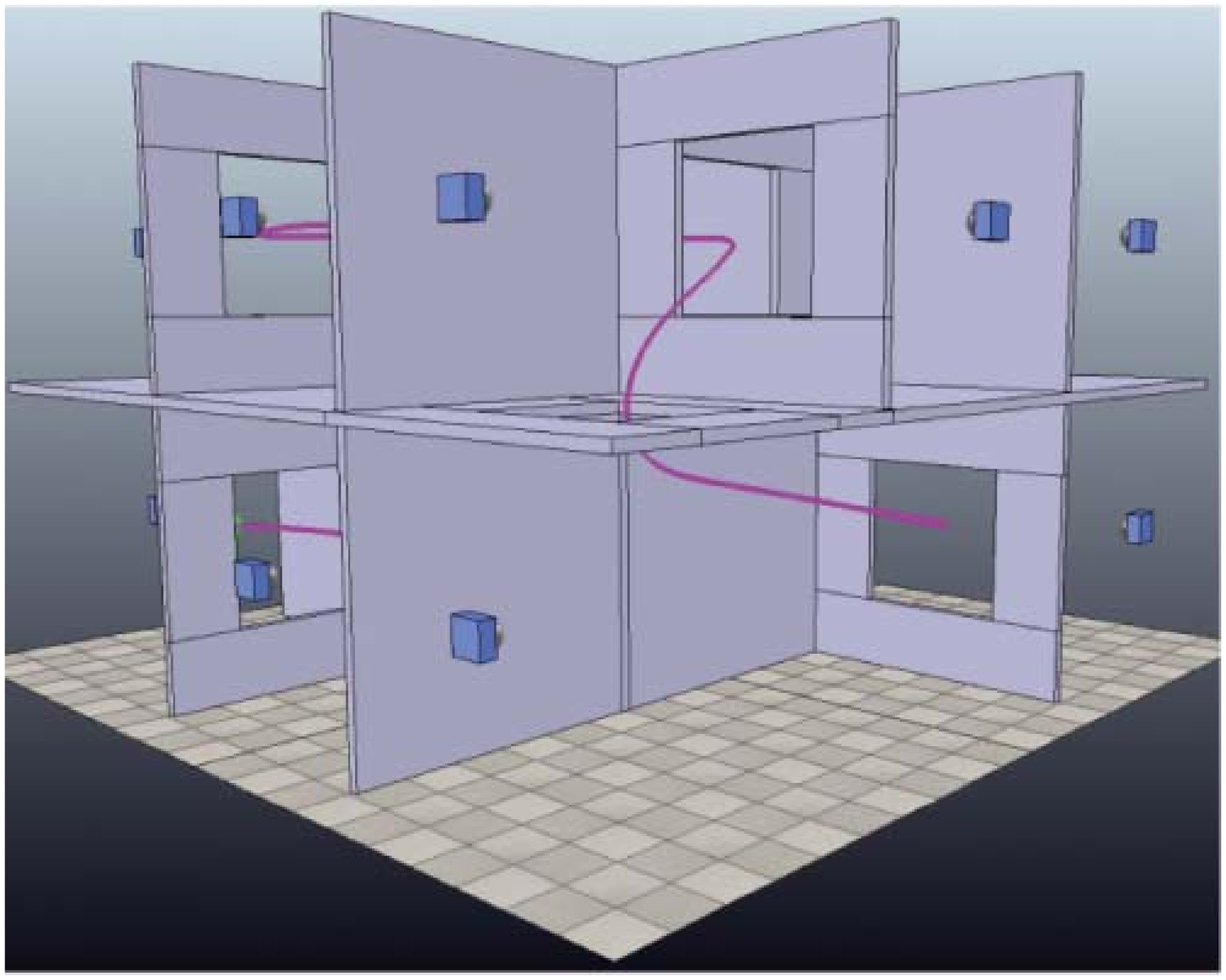,width=5cm}
}
\subfigure[]{
\epsfig{figure=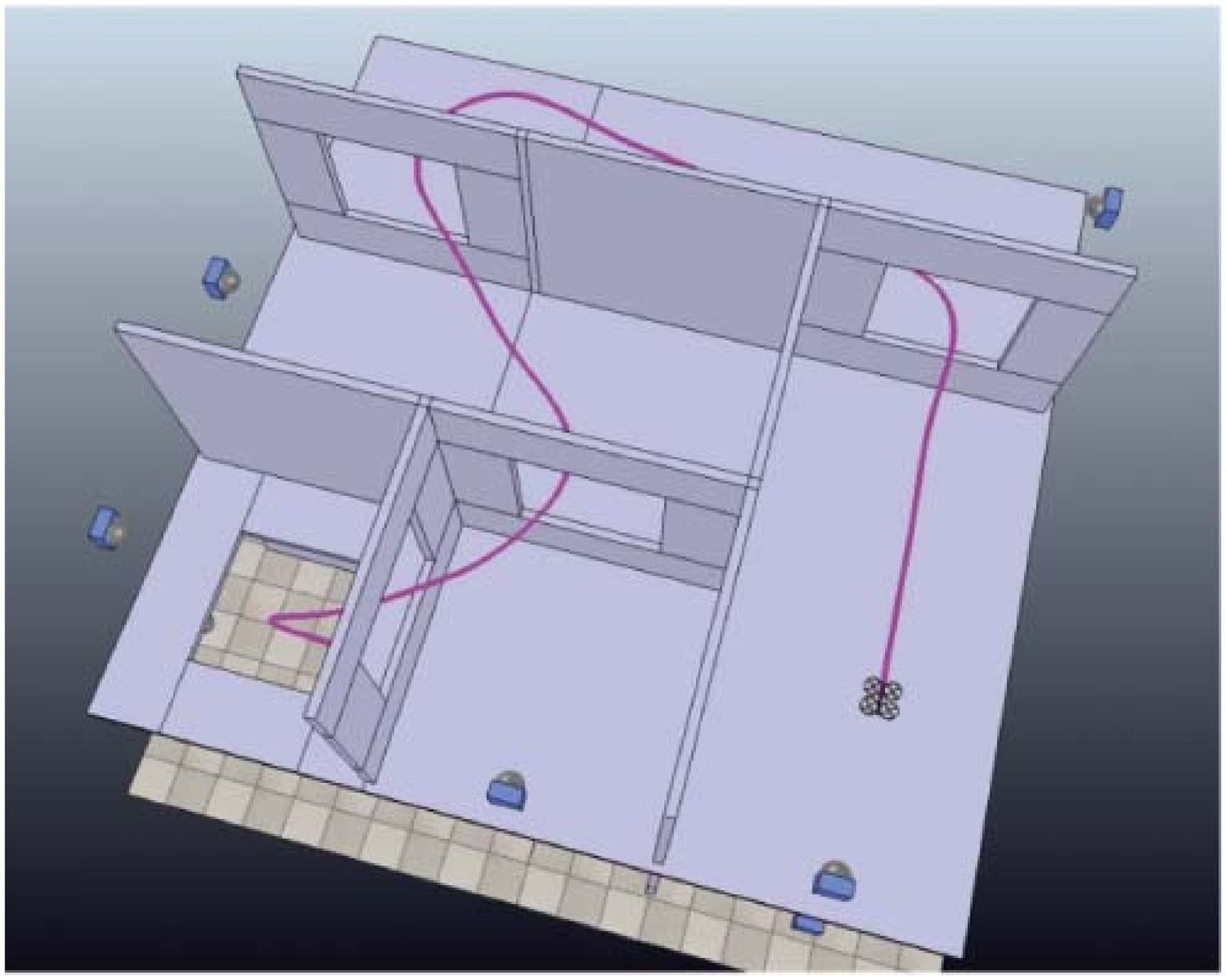,width=5cm}
}
\subfigure[]{
\epsfig{figure=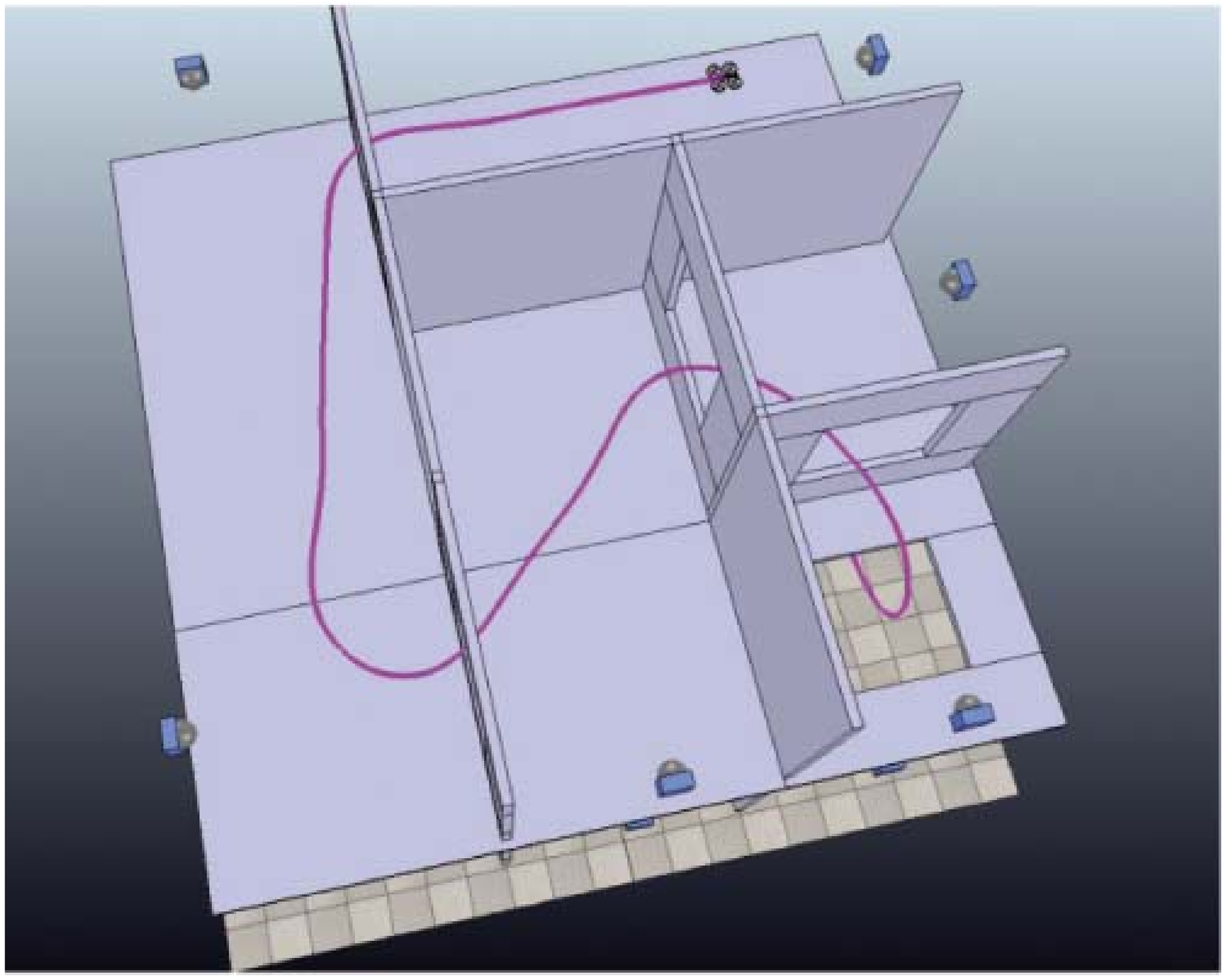,width=5cm}
}
\caption{
Results of the simulation in the static scene at different viewpoints. The top ceiling and exterior walls are not visible. The Green sphere indicates the robot's initial position. The magenta curve is the robot's trajectory.
}
\label{fig:c4_s_s_123456}
\end{figure}

\begin{figure}[!htb]
\centering
\epsfig{figure=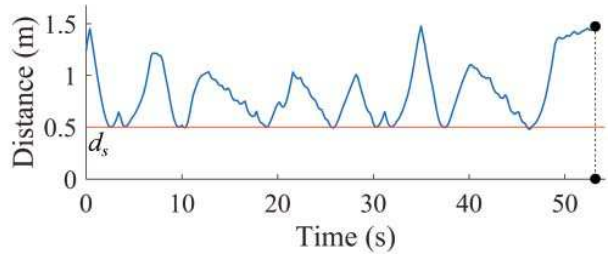,width=8cm}
\caption{
Minimum distance from the robot to the obstacles in the static scene.
}
\label{fig:c4_s_s_7}
\end{figure}

To confirm the performance of the proposed navigation algorithm \textbf{A1}-\textbf{A7}, \textbf{B1}-\textbf{B3} in static environments, we built a static scene, which is a closed indoor environment with some rooms. There are twelve ToF cameras deployed in this scene to detect the obstacles and construct a WSN. The size of the room is $9\text{m}\times 9\text{m} \times 6\text{m}$ with two floors. There are some holds on the walls to connect different rooms. The safety margin $d_s$ is $0.5\text{m}$ in this simulation. Other parameters are the same as Table \ref{c4_TB1}. A flying robot is navigated from an initial position to a given target by the proposed algorithm. The simulation results shown in Fig. \ref{fig:c4_s_s_123456} indicates that the flying robot flew though each rooms and reached the target successfully without any collision. The minimum distance from the robot to the obstacles is indicated in Fig. \ref{fig:c4_s_s_7}. It can be seen that the robot kept the safety margin $d_s$ during the travelling.

\subsection{Simulations in dynamic environments}

\begin{figure}[!htb]
\centering
%\subfigure[]{
%\epsfig{figure=./eps/s_d_1.eps,width=4cm}
%}
\subfigure[]{
\epsfig{figure=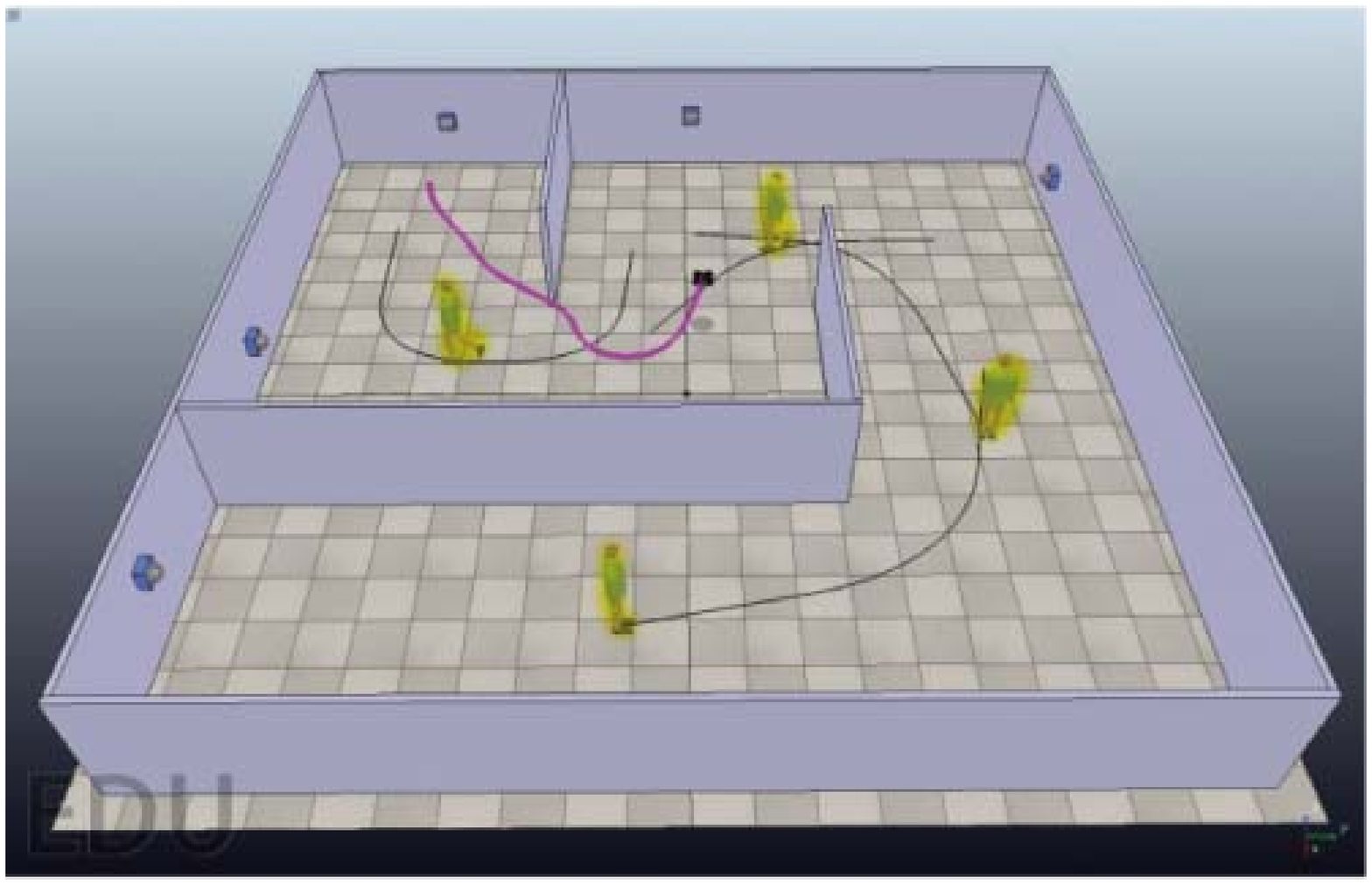,width=7cm}
}
%\subfigure[]{
%\epsfig{figure=./eps/s_d_3.eps,width=4cm}
%}
\subfigure[]{
\epsfig{figure=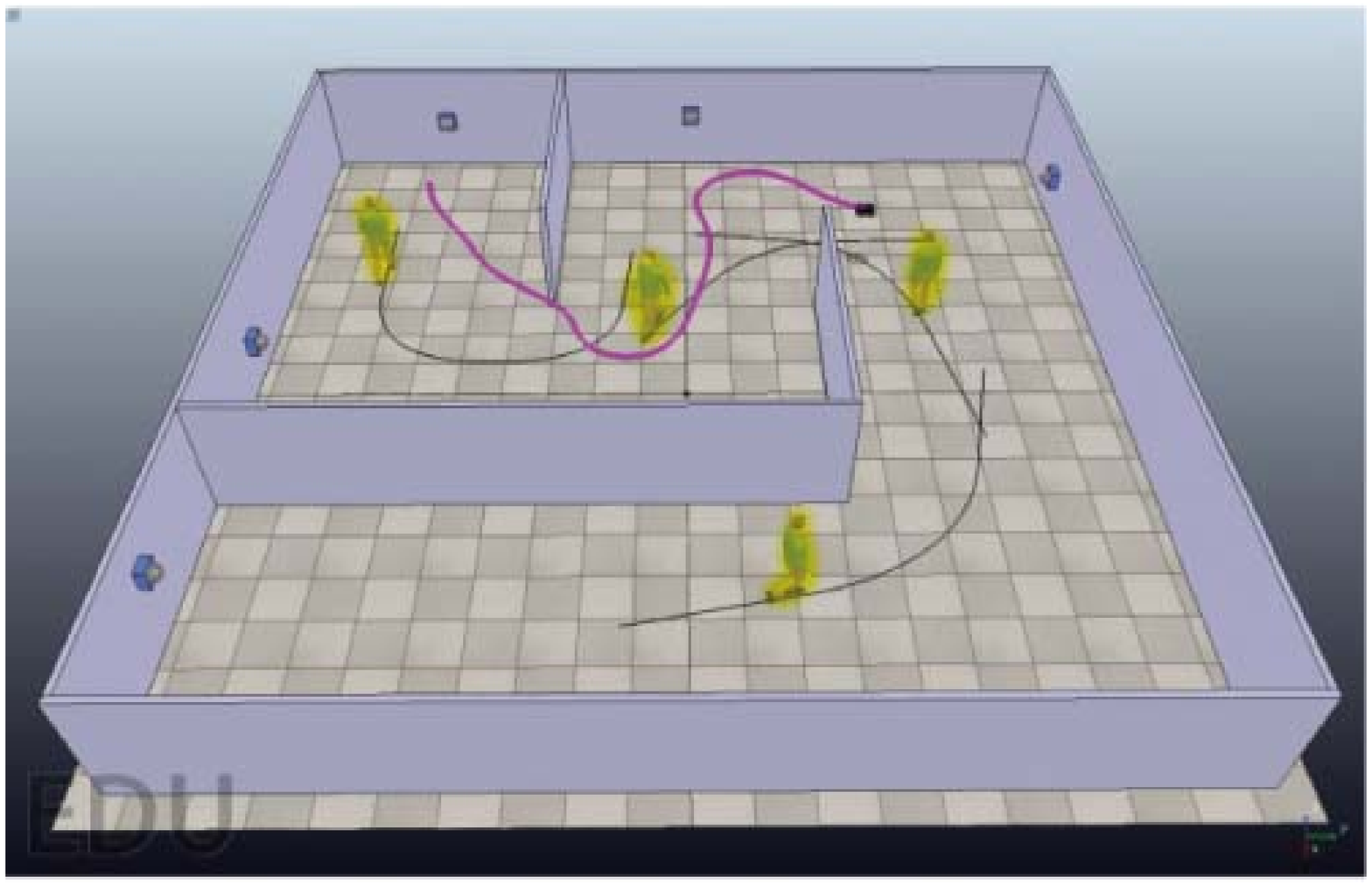,width=7cm}
}
%\subfigure[]{
%\epsfig{figure=./eps/s_d_5.eps,width=4cm}
%}
\subfigure[]{
\epsfig{figure=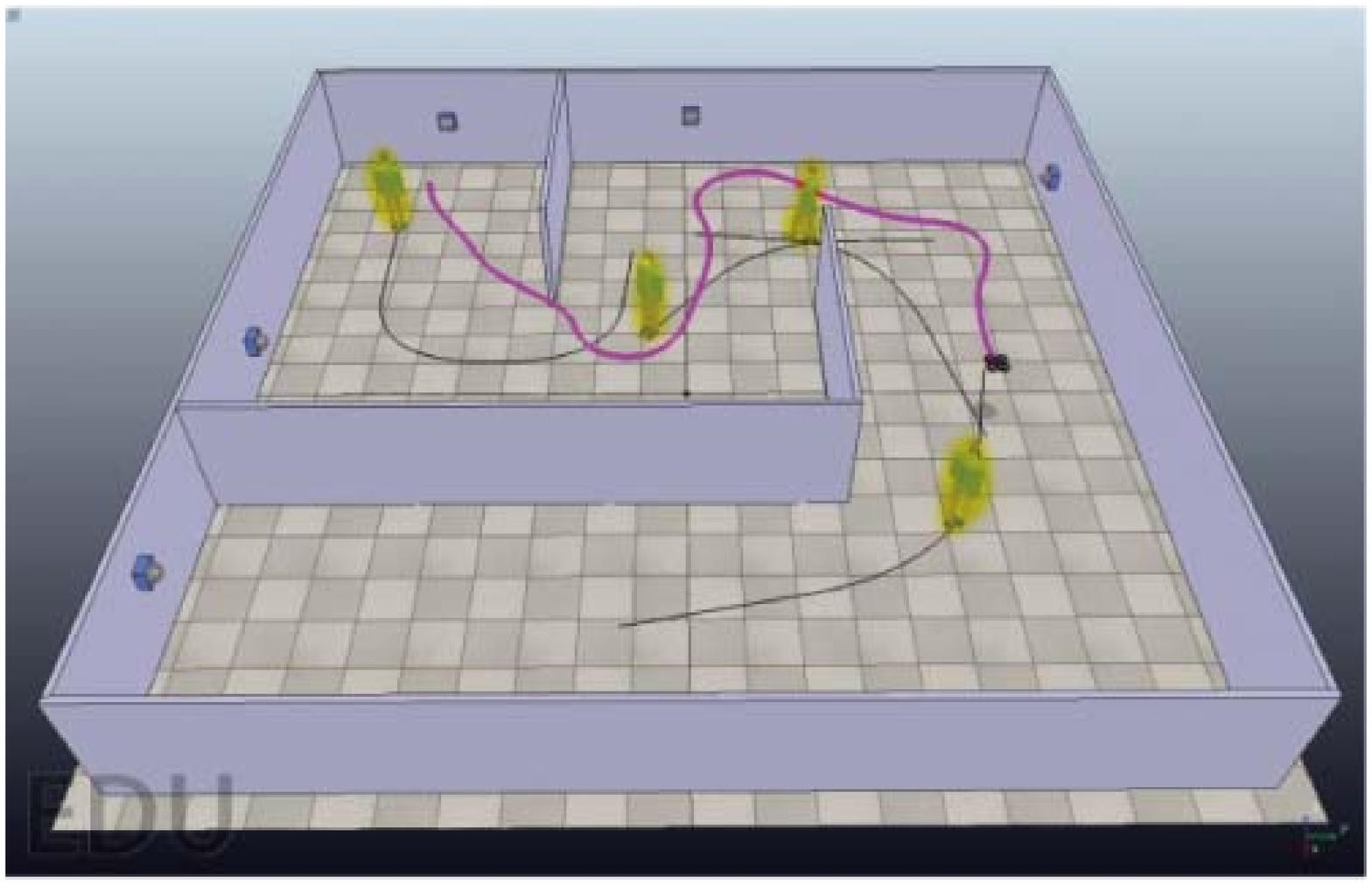,width=7cm}
}
%\subfigure[]{
%\epsfig{figure=./eps/s_d_7.eps,width=4cm}
%}
\subfigure[]{
\epsfig{figure=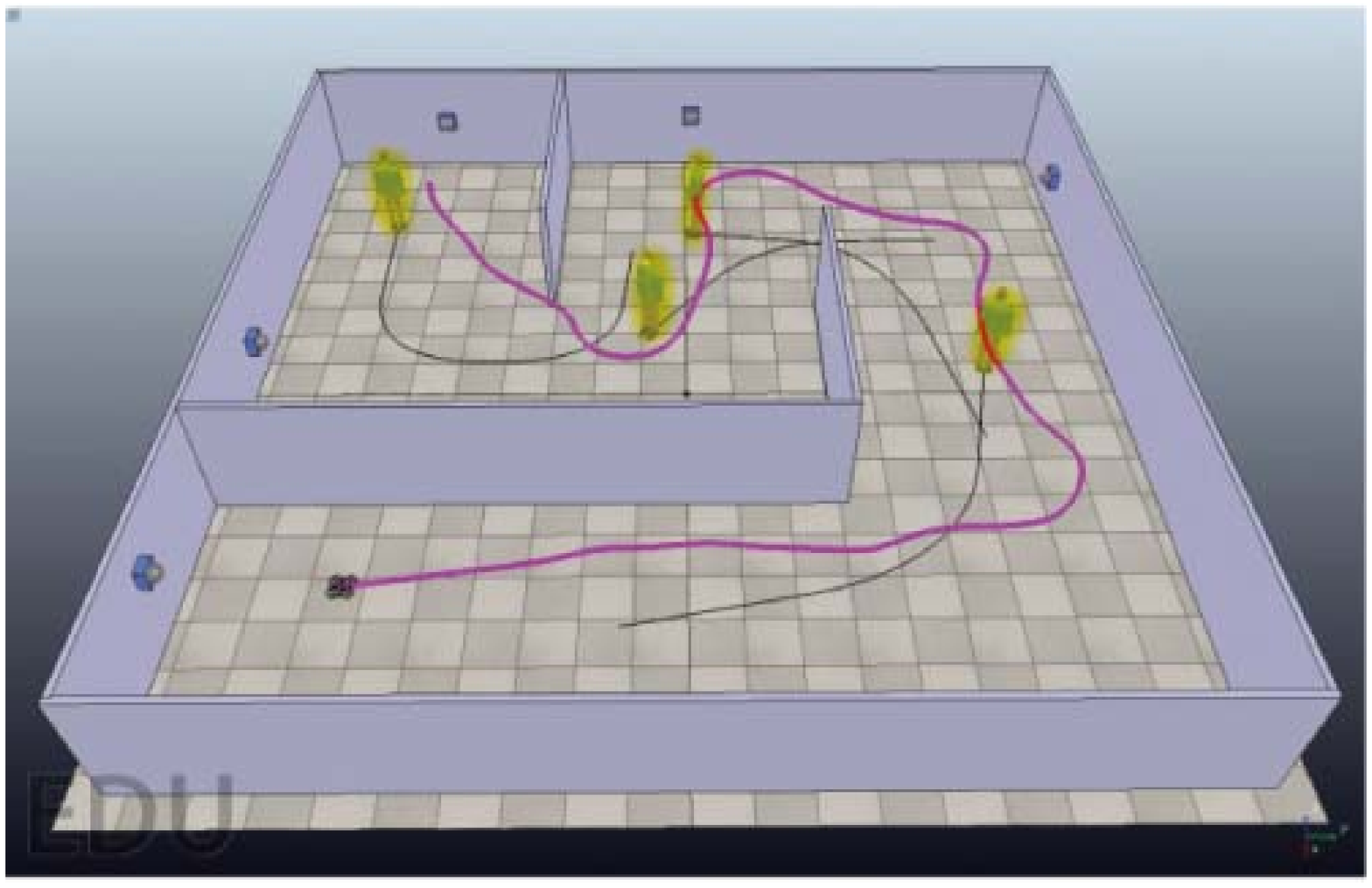,width=7cm}
}
\caption{
Results of the simulation in the dynamic scene at different time. The top ceiling is not visible. The magenta curve is the robot's trajectory. The yellow objects are walking people.
}
\label{fig:c4_s_d_12345678}
\end{figure}

\begin{figure}[!htb]
\centering
\epsfig{figure=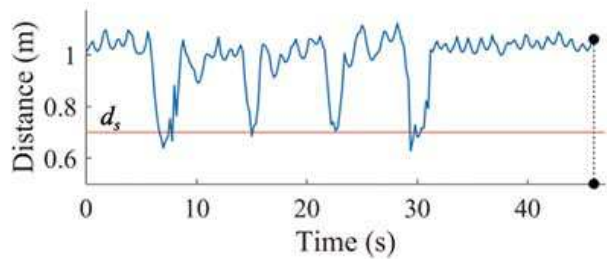,width=8cm}
\caption{
Minimum distance from the robot to the obstacles in the dynamic scene.
}
\label{fig:c4_s_d_9}
\end{figure}

To confirm the performance of the proposed navigation algorithm in dynamic environments, we present another computer simulation in a dynamic scene. In the dynamic scene, a closed indoor environment are constructed by some walls and ceiling. There are four people walking in the room with the maximum speed $V_{\max}=0.26\text{m/s}$. The size of the room is $14\text{m}\times 14\text{m} \times 2\text{m}$ with one floor. There are six ToF cameras deployed on the walls to detect the obstacles and walking people. The safety margin $d_s$ is $0.7\text{m}$ in this simulation. The time window $T$ is $6$. Other parameters are the same as Table \ref{c4_TB1}. According to the Fig. \ref{fig:c4_s_d_12345678}, it can be seen that the flying robot was moving under the sensor network's navigation. It started from the initial position and avoided the walls and any moving people. Finally, the flying robot was navigated to the target successfully by keeping the given safety margin $d_s$ (see Fig. \ref{fig:c4_s_d_9}).

\subsection{Simulations with multiple robots}

\begin{figure}[!htb]
\centering
\subfigure[]{
\epsfig{figure=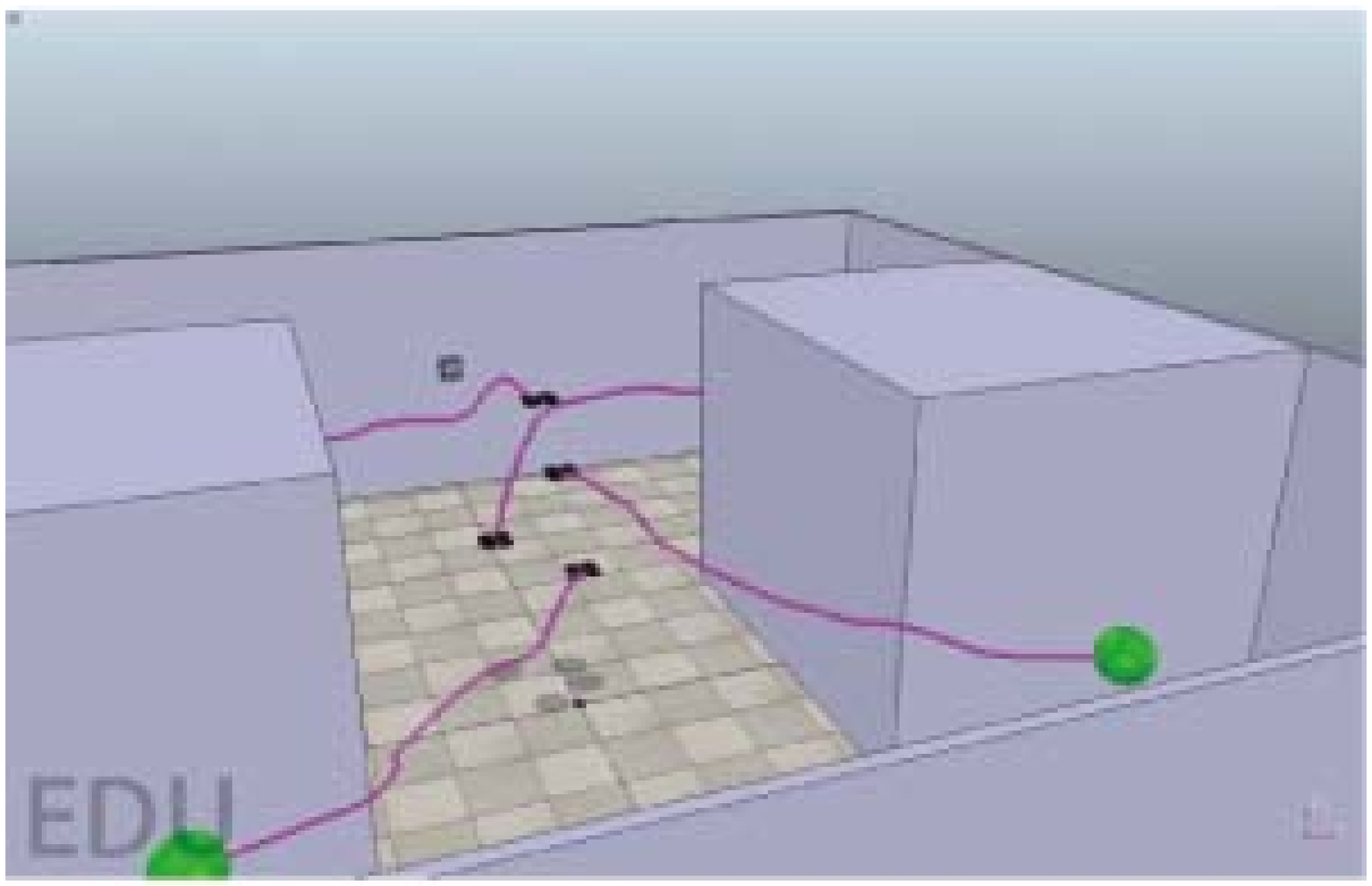,width=7cm}
}
\subfigure[]{
\epsfig{figure=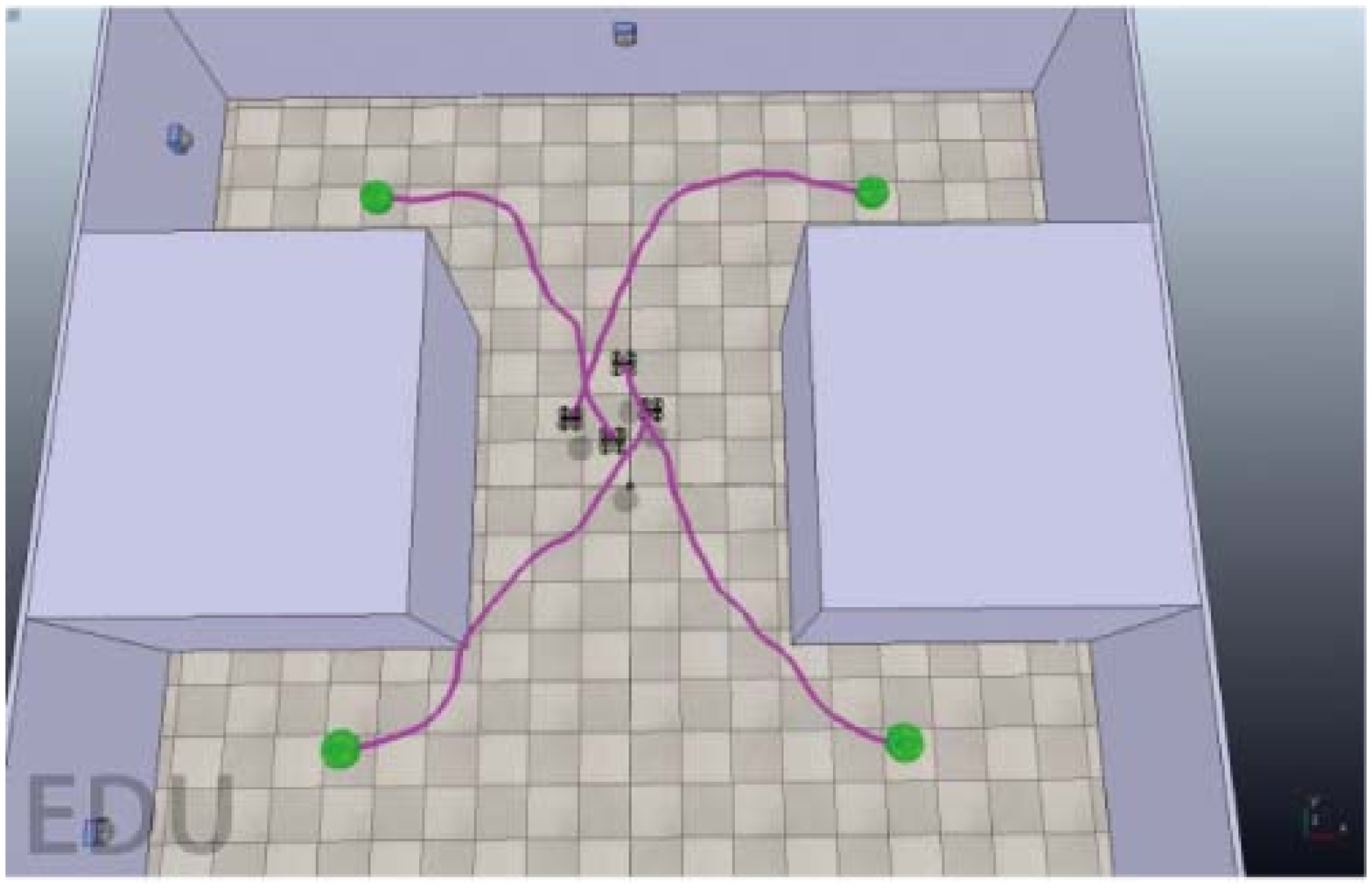,width=7cm}
}
\subfigure[]{
\epsfig{figure=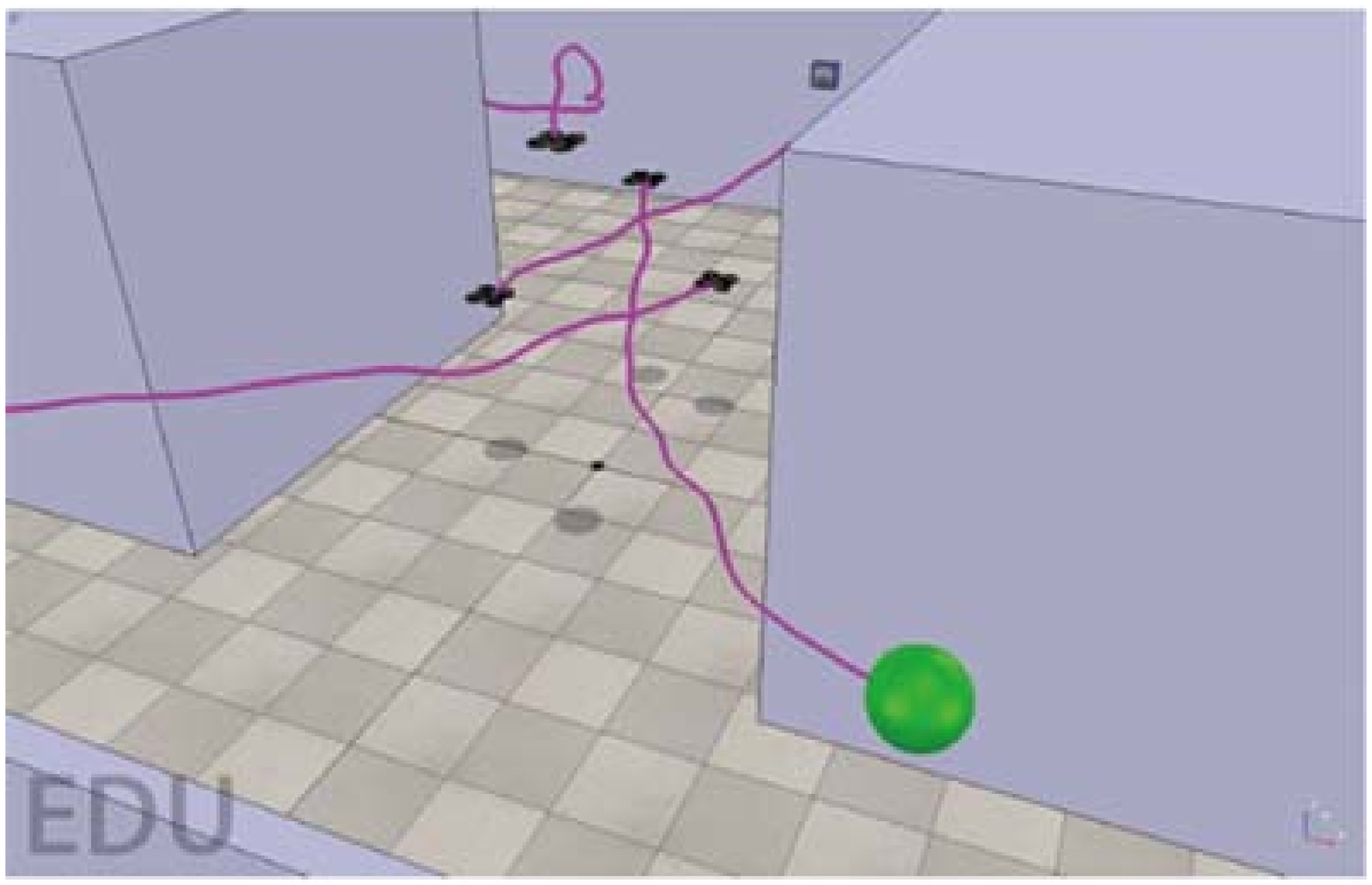,width=7cm}
}
\subfigure[]{
\epsfig{figure=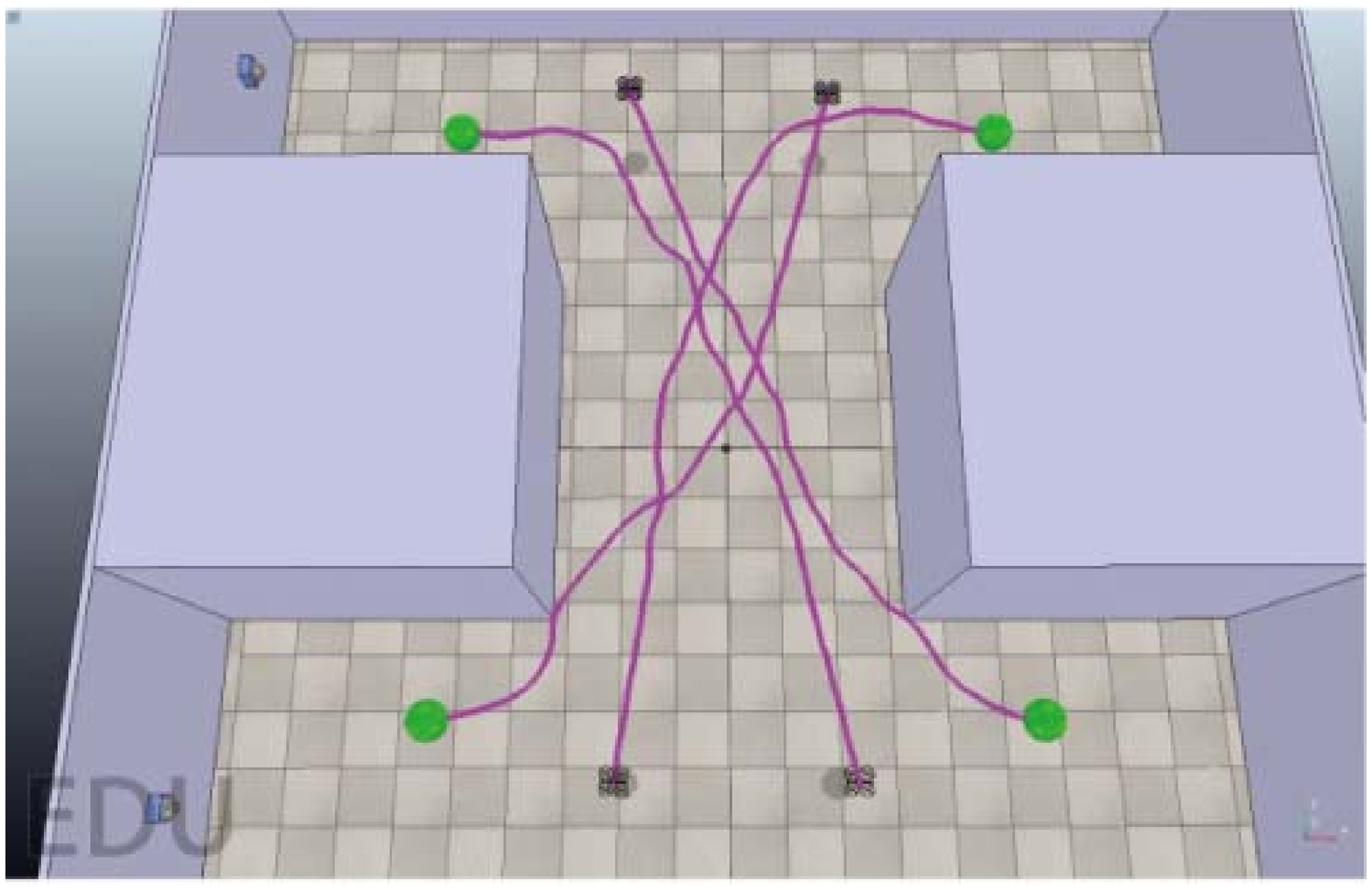,width=7cm}
}
\caption{
Results of the simulation with multiple micro flying robots. The top ceiling is not visible. The magenta curves are the robots' trajectories. The Green spheres indicate the robots' initial positions.
}
\label{fig:c4_s_m_1234}
\end{figure}

\begin{figure}[!htb]
\centering
\subfigure[]{
\epsfig{figure=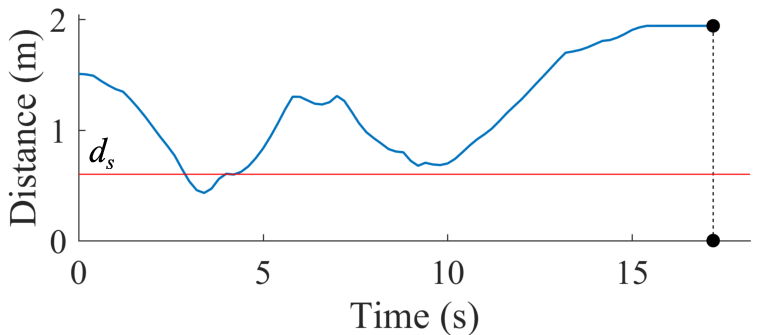,width=7cm}
}
\subfigure[]{
\epsfig{figure=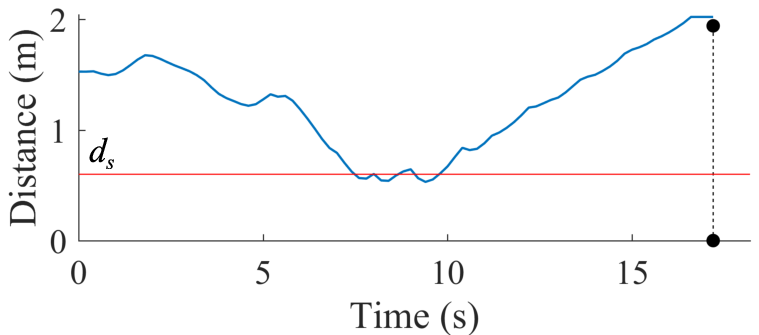,width=7cm}
}
\subfigure[]{
\epsfig{figure=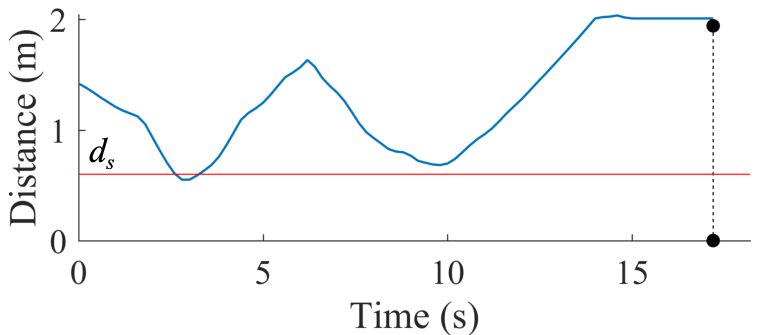,width=7cm}
}
\subfigure[]{
\epsfig{figure=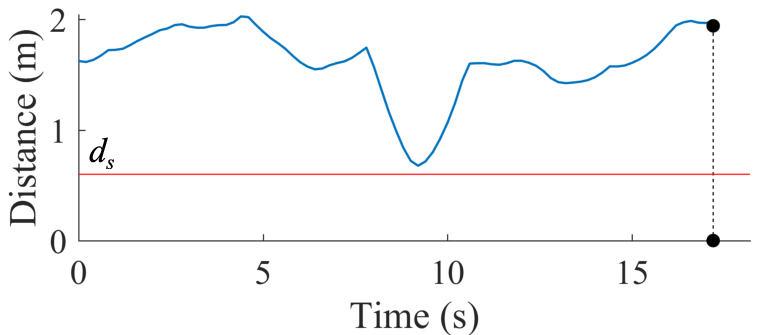,width=7cm}
}
\caption{
Minimum distances from each robot to the obstacles and other robots.
}
\label{fig:c4_s_m_5678}
\end{figure}

In the third computer simulation, we confirm the performance of the proposed navigation algorithm in a multiple robot system. In the simulation scene, we construct a closed indoor environment with static obstacles; e.g. walls. There are four micro flying robots working in the environment. The size of the room is $14\text{m}\times 14\text{m} \times 4\text{m}$ with one floor. To detect the environment, three time of flight cameras are deployed in the workspace. The safety margin $d_s$ is $0.6\text{m}$ in this simulation. Other parameters are the same as Table \ref{c4_TB1}. In this simulation, four micro flying robots are navigated simultaneously by the sensor network. The robots moved from the initial positions to the targets and avoided each other with the safety margin $d_s$; see the simulation result in Fig. \ref{fig:c4_s_m_1234}. It can be seen in Fig. \ref{fig:c4_s_m_5678} that the flying robots kept the given safety margin $d_s$ to both the obstacles and other flying robots successfully.

\section{Summary}

We proposed a collision-free navigation method for micro UAVs in dynamic environments by a wireless sensor network. A micro UAV cannot be equipped with heavy obstacle detection sensors. Therefore, to solve the navigation problem, a wireless sensor network consisting of 3D rang finders is involved to detect the static and dynamic obstacles in the workspace. With the navigation of the sensor network, only a path tracking controller is required for each micro flying robot. The robot can be navigated directly in the industrial workspace without any specialization. The computer simulations were carried out to confirm the performance of the proposed method.

The proposed method can be implemented in many industrial fields, such as the radiant intensity alarm in nuclear power stations. In the future work, a distributed navigation algorithm can be developed and other types of sensors in the sensor network can be used to make navigation system more economic.
%%%%%%%%%%%%%%%%%%%%%%%%%%%%%%%%%%%%%%%%%%%%%%%%
\chapter{Safe Area Search and 2D Map Building of a Ground Mobile Robot in Bounded Area}
\chaptermark{2D Map Building}
\label{Chapter5}

This chapter is based on the the publications \cite{savkin_li_2017} and \cite{Savkin2016}. In this chapter, we combine these two fundamental problems of modern robotics together and propose a robot navigation algorithm for exploring a complex unknown environment with obstacles and building a binary map while avoiding collisions with obstacles. The task is performed by a wheeled mobile robot equipped with a range finder sensor. We assume that the search area is cluttered with a number of possibly non-convex obstacles with unknown shapes and locations. We develop an algorithm for robot collision free navigation and map building that results in generating a complete map of the searched area based on the range finder sensor measurements. The proposed algorithm is randomized and we prove that with probability $1$ the task of the area search and map building will be completed  in a finite time. The control law we present in this method consists of switching between different control strategies with different conditions to construct a switched control system; see \cite{Matveev2000Estimation,Savkin2002Hybrid,
SAVKIN199969Robust,SKAFIDAS1999553Stability}.

\section{Problem description}

\begin{figure}[thpb]
\centering
\epsfig{figure=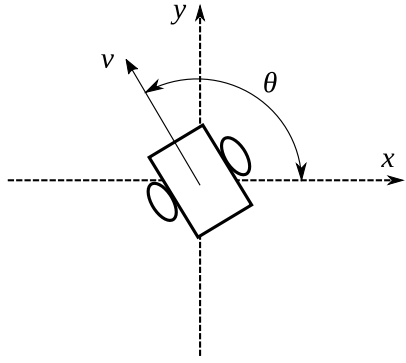,width=8cm}
\caption{
Unicycle model of a wheeled mobile robot.
}
\label{c5_veh.fig}
\end{figure}

In this chapter, a planar vehicle or wheeled mobile robot is modelled as
a Dubins car \cite{Dubins1957}. It travels with a constant speed and is controlled by the
angular velocity limited by a given constant. 
The model of the vehicle is as follows (see
Fig. \ref{c5_veh.fig}):

\begin{equation}
\label{c5_1}
\begin{array}{l}
\dot{x}(t) = v \cos \theta (t)
\\
\dot{y}(t) = v \sin \theta (t)
\\
\dot{\theta}(t) = u(t) \in [-u_M, u_M]
\end{array}, \qquad
\begin{array}{l}
x(0) = x_0
\\
y(0) = y_0
\\
\theta(0) = \theta_0.
\end{array} 
\end{equation}

Here $(x,y)$ is the vector of the vehicle's
Cartesian coordinates, $\theta$ gives its orientation, $v$ and $u$
are the speed and angular velocity, respectively. The maximal
angular velocity $u_M$ is given. 
The robot satisfies the standard non-holonomic constraint 
\begin{equation}
\label{c5_nonh}
|u(t)|\leq u_M.
\end{equation} 
This obviously implies that the minimum turning radius of
the robot is
\begin{equation}
\label{c5_Rmin} R_{min}= \frac{v}{u_M}.
\end{equation}
The non-holonomic model (\ref{c5_1}) is commonly used to describe planar motion of ground robots, unmanned aerial vehicles and 
missiles; see e.g. \cite{Manchester2006,Savkin2010,Savkin2013,Matveev2015a}.

The wheeled robot is moving in a bounded planar unknown area 
${\cal A}$ with  several disjoint obstacles
$D_1,\ldots,D_k$. Let the safety margin $d_0>0$ be given. 
The requirement is to drive the autonomous robot through the obstacle-free part of the area ${\cal A}$ while keeping the safety margin $d_0$ from both the obstacles and the boundary of area ${\cal A}$.

\begin{figure}[thpb]
\centering
\epsfig{figure=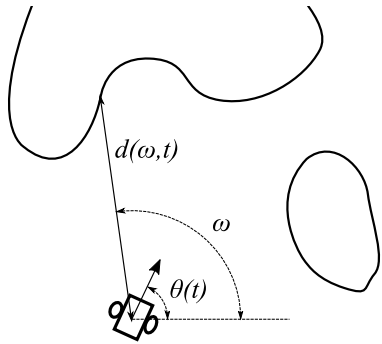,width=8cm}
\caption{
Measurements of a range finder sensor.
}
\label{c5_AM}
\end{figure}

{\bf Available Measurements:} The robot is
equipped with a range finder sensor that at any time $t\geq 0$ measures the distance
to the nearest object in all directions as follows: for any
angle $0\leq \omega<2\pi$, the robot measures the distance
$d(\omega,t)$ to the nearest boundary of an obstacle or the area ${\cal A}$ in the direction given by $\omega$ (see
Fig. \ref{c5_AM}). Furthermore , the robot is equipped with an odometry type sensor which measures the robot's position and heading relative to its starting location and heading.

The goal of the robot is to build a complete map of the unknown cluttered area 
${\cal A}$ while keeping a safe distance from the obstacles and the area boundary.

\begin{definition}
Let ${\cal M}$ be a planar set and ${\cal F}$ be a binary map
${\cal M}\rightarrow \left\{0,1\right\}$. The pair 
$\left( {\cal M, F}\right)$ is said to be the complete map
of the area ${\cal A}$ if ${\cal M}$ and ${\cal A}$ are congruent, i.e. one  can be transformed into the other by an isometry, and for any point $p\in {\cal M}$, ${\cal F}(p)=1$
if and only if $p$ under such isometry corresponds to  a point of either the boundary of ${\cal A}$ or the boundary of one of the obstacles
$D_1,\ldots,D_k$.
\end{definition}

\begin{figure}[thpb]
\centering
\epsfig{figure=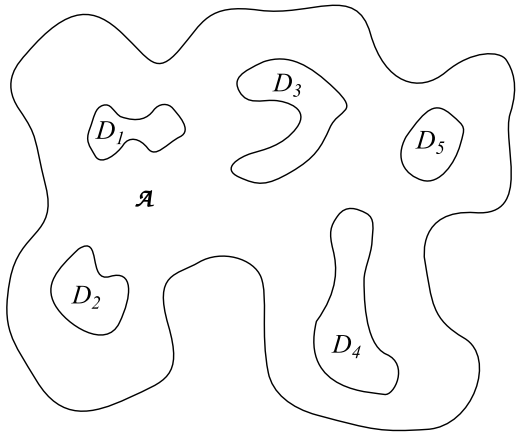,width=8cm}
\caption{
An example of a complete map.
}
\label{c5_CM}
\end{figure}

In other words, a complete map means that the geometry of the area ${\cal A}$ is precisely known with all the obstacles 
$D_1,\ldots,D_k$ that belong to ${\cal A}$; see e.g. Fig. \ref{c5_CM}. A topological (binary) map is used to represent the environment. This map indicates the parts of the environment where the robot cannot go and the parts where it can go. The motivation to use such a map is to get a simplest representation of the environment  to make our algorithm as computationally efficient as possible.

For any robot's trajectory, the following map building algorithm is proposed.

\begin{figure}[thpb]
\centering
\epsfig{figure=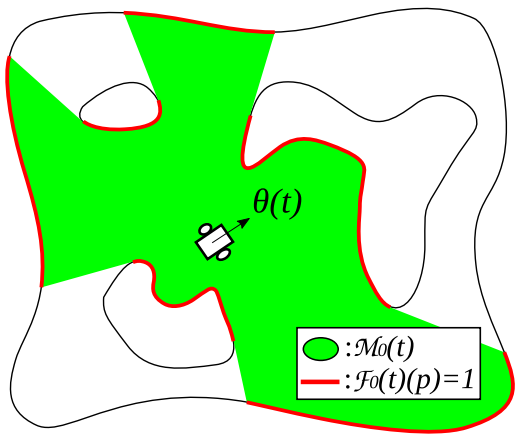,width=8cm}
\caption{
Planar set ${\cal M}_0(t)$ and binary map ${\cal F}_0(t)$.
}
\label{c5_M1}
\end{figure}

{\bf M1:} At any time $t\geq 0$, the robot builds a 
planar set ${\cal M}_0(t)$ and a binary map ${\cal F}_0(t)$
as follows: the set ${\cal M}_0(t)$ consists of all points 
$p$ of the plane that are visible by the range finder sensor at time $t$ and ${\cal F}_0(t)(p)=1$ if p belongs to either 
obstacle boundary or area boundary, and ${\cal F}_0(t)(p)=0$
otherwise; see Fig. \ref{c5_M1}. The pair 
$\left( {\cal M}_0(t),{\cal F}_0(t)\right)$ is called the instant map at time $t$.

{\bf M2:} At any time $t\geq 0$, the robot builds a 
planar set ${\cal M}(t)$ and a binary map ${\cal F}(t)$ by fusing with the help of odometry sensor all the instant 
maps $\left( {\cal M}_0(\tau),{\cal F}_0(\tau)\right)$
for $\tau\in [0,t]$. The pair 
$\left( {\cal M}(t),{\cal F}(t)\right)$ is called the total map at time $t$.

Now some definitions are given as follows.

\begin{figure}[thpb]
\centering
\epsfig{figure=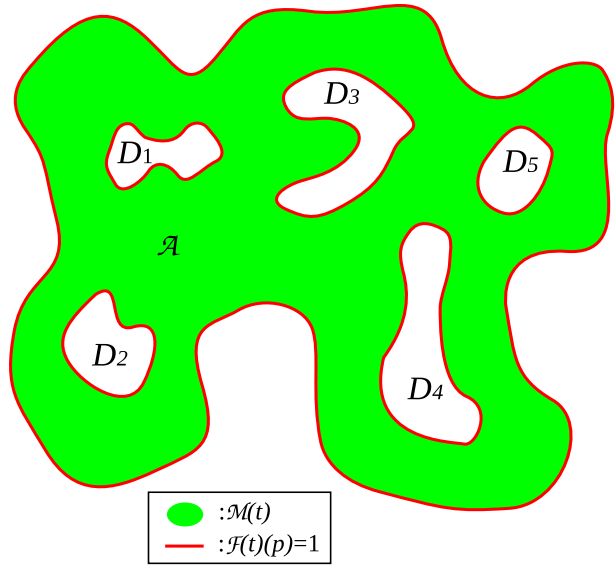,width=8cm}
\caption{
An example of ${\cal M}(t_f)$ and ${\cal F}(t_f)$ which satisfy the Definition 2.2.
}
\label{c5_D2.2}
\end{figure}

\begin{definition}
\label{c5_De2.2}
Consider a robot trajectory and the map building algorithm
{\bf M1--M2}. A time $t_f\geq 0$ is said to be a map building completing time if the total map $\left( {\cal M}(t_f),{\cal F}(t_f)\right)$ at time $t_f$ satisfies the following property:
the boundary of the set ${\cal M}(t_f)$ consists of points
$p$ such that ${\cal F}(t_f)(p)=1$ and does not include any point $p$ such that ${\cal F}(t_f)(p)=0$; see Fig \ref{c5_D2.2}.
\end{definition}

\begin{remark}
\label{c5_Rem}
If $t_f$ is a map building completing time, it can be seen that the total map $\left( {\cal M}(t_f),{\cal F}(t_f)\right)$ at the time $t_f$ is the complete map
of the area ${\cal A}$. Moreover,
if the robot has built by the time $t_f$ the complete map
of the area ${\cal A}$, then each point of the area  outside 
of the obstacles was seen by the robot's sensor at some time.
\end{remark}

\begin{definition}
A trajectory $p(t)=(x(t),y(t)), t\in [0,t_f]$ of the robot (\ref{c5_1}) is said to be {\em complete map building
with obstacle avoidance} if $t_f$ is a map completing time
and the distance between the robot's position  $p(t)$ 
and the boundary of the area ${\cal A}$ and the obstacles is
no less than $d_0$ 
for all $t\in [0,t_f]$.
\end{definition}

\begin{notation}
Let $D$ be any closed set, $p$ be a point in the plane.
Introduce the distance $\rho(D,p)$ as
\[
\rho(D,p):=\min_{q\in D}\|p-q\|.
\]
Notice that $\min$ is
achieved since $D$ is closed. Also, $\rho(D,p)=0$ if $p\in D$.
\end{notation}

\begin{definition}
For $d_0>0$, the $d_0-${\em enlargement} of the domain $D \subset {\bf
R}^2$ is the set ${\cal E}[D,d_0]$ formed by all points at the distance
$\leq d_0$ from $D$, i.e.,
${\cal E}[D,d_0]:=\left\{p  \in {\bf R}^2: \rho(D,p)\leq d_0 \right\}.
$ On the other hand, the $d_0-${\em reduction} of the domain $D \subset {\bf
R}^2$ is the set ${\cal R}[D,d_0]$ formed by all points of $D$ at the distance
$\geq d_0$ from the boundary $\partial D$ of $D$, i.e.,
$
{\cal R}[D,d_0]:=\left\{p  \in D: \rho(\partial D,p)\geq d_0 \right\}.$
\end{definition}

\begin{assumption}
\label{c5_Aso} The planar sets ${\cal R}[{\cal A},d_0]$ and  
${\cal E}[D_i,d_0], ~i=1,\ldots,k$ are  closed, bounded, connected and
linearly connected  
sets.
\end{assumption}

\begin{assumption}
\label{c5_Aso1} The sets  ${\cal E}[D_i,d_0]$ and 
${\cal E}[D_j,d_0]$
do
not overlap for any $i\neq j$. Any set ${\cal E}[D_i,d_0]$
is a subset of ${\cal R}[{\cal A},d_0]$.
\end{assumption}

\begin{figure}[thpb]
\centering
\epsfig{figure=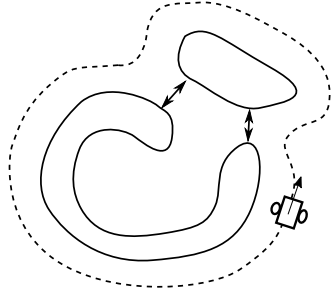,width=8cm}
\caption{
Robot cannot reach and map the area which is isolated by overlapped ${\cal E}[D_i,d_0]$.
}
\label{c5_R2.2}
\end{figure}

\begin{remark}
If Assumption {\ref{c5_Aso1}} does not hold, it can be seen that map building
with obstacle avoidance may be impossible (see e.g. Fig. \ref{c5_R2.2}). 
\end{remark}

\begin{assumption}
\label{c5_AsoN} Any point of the boundaries of ${\cal A}$ or
$D_i$ can be connected by a straight line segment non-crossing any obstacle of the region's boundary with a point of the boundary of ${\cal R}[{\cal A},d_0]$ or 
${\cal E}[D_i,d_0]$, correspondingly. 
\end{assumption}

\begin{figure}[thpb]
\centering
\epsfig{figure=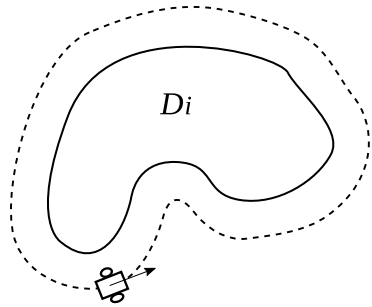,width=8cm}
\caption{
Robot can see any point of the boundary of ${\cal A}$ or $D_i$.
}
\label{c5_NN1}
\end{figure}

\begin{figure}[thpb]
\centering
\epsfig{figure=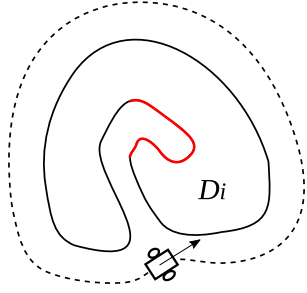,width=8cm}
\caption{
There is a segment on the boundary of $D_i$ which cannot be seen by the robot.
}
\label{c5_NN2}
\end{figure}

\begin{remark}
Assumption {\ref{c5_AsoN}} means that any point of the boundaries of ${\cal A}$ or
$D_i$ can be seen by the robot travelling along the boundaries of ${\cal R}[{\cal A},d_0]$ or 
${\cal E}[D_i,d_0]$, correspondingly, see e.g. Fig. \ref{c5_NN1}.
If Assumption {\ref{c5_AsoN}} does not hold, map building
with obstacle avoidance may be impossible since the boundaries of the  obstacles or the region can contain some points which are impossible to see from a safe distance (see e.g. Fig. \ref{c5_NN2}). 
\end{remark}

\begin{notation}
Let $\partial D_i(d_0)$ denote the boundary 
of the set ${\cal E}[D_i,d_0]$, and 
$\partial {\cal A}(d_0)$ denote the boundary 
of the set 
${\cal R}[{\cal A},d_0]$.
\end{notation}

\begin{assumption}
\label{c5_Aso2}
For all $i$,  the boundary $\partial D_i(d_0)$  is a non-self-intersecting, closed 
smooth curve with curvature $k_i(p)$ at any point $p$ satisfying $|k_i(p)|\leq \frac{1}{R_{min}}$. Moreover,
the boundary $\partial {\cal A}(d_0)$  is also a closed, non-self-intersecting 
smooth curve with curvature $k(p)$ at any point $p$ satisfying $|k(p)|\leq \frac{1}{R_{min}}$.
\end{assumption}

\begin{figure}[thpb]
\centering
\epsfig{figure=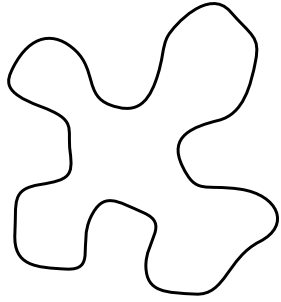,width=8cm}
\caption{
There may be negative curvature at some points on ${\cal R}[{\cal A},d_0]$ and 
${\cal E}[D_i,d_0]$.
}
\label{c5_R2.3}
\end{figure}

\begin{remark}
Notice that the sets 
${\cal R}[{\cal A},d_0]$ and 
${\cal E}[D_i,d_0]$ are not assumed to be convex.
Therefore,  their  boundaries  may have negative curvature at some
points; see e.g. Fig. \ref{c5_R2.3}. The standard definition
of curvature from differential geometry is used here (see e.g. \cite{Klingenberg2013}). 
\end{remark}

\begin{definition}
There are two  circles with the radius $R_{min}$ that cross the initial robot
position  $p(0)$ and
tangent to the robot initial heading $\theta(0)$. The two circles are called initial circles.
\end{definition}

These initial circles are assumed to be far enough from the obstacles and the boundary of ${\cal A}$.

\begin{assumption}
\label{c5_As3} Both the initial circles lie in the set 
${\cal R}[{\cal A},d_0]$ and do not intersect the sets 
${\cal E}[D_i,d_0]$. 
\end{assumption}

\section{Safe navigation algorithm}

In this section, the proposed  algorithm 
of safe navigation with map building is described. First, a number of definitions are introduced.

\begin{definition}
A straight line $L$ is said to be a tangent line if one of the following
conditions holds:
\begin{enumerate}
\item
The line $L$ is simultaneously tangent to two boundaries $\partial D_i(d_0)$ and 
$\partial D_j(d_0)$ where $i\neq j$.
\item
The line $L$ is simultaneously tangent to the boundary $\partial D_i(d_0)$ and
an initial circle.
\item
The line $L$ is simultaneously tangent to the boundary 
$\partial D_i(d_0)$ and 
the boundary $\partial {\cal A}(d_0)$
\item
The line $L$ is simultaneously tangent to the boundary 
$\partial {\cal A}(d_0)$ and
an initial circle.
\end{enumerate}
Points of the boundaries $\partial {\cal A}(d_0)$, $\partial D_i(d_0)$ and the initial circles belonging to the tangent lines
are called tangent points. 
\end{definition}

For the case of simplicity, the following assumptions are introduced.

\begin{assumption}
\label{c5_Aso5}
Any tangent point belongs to only one tangent line.
\end{assumption}

\begin{assumption}
\label{c5_Aso6}
If the set ${\cal A}(d_0)$ is non-convex, then
there exists a tangent point belonging to $\partial {\cal A}(d_0)$.
\end{assumption}

\begin{figure}[thpb]
\centering
\epsfig{figure=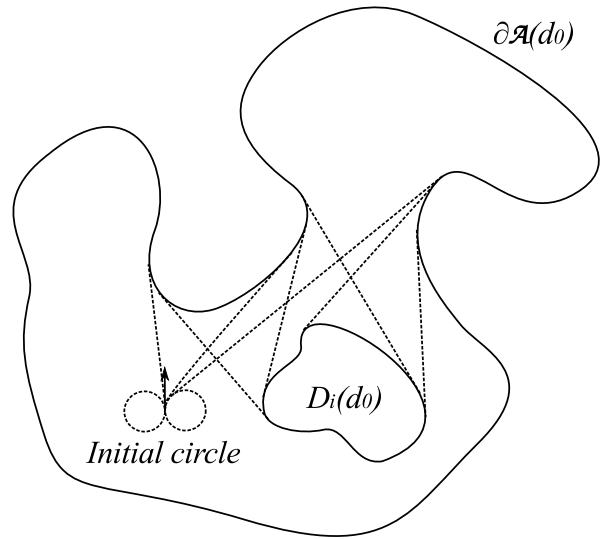,width=8cm}
\caption{
Tangent lines connecting the boundary $\partial {\cal A}(d_0)$
with either a boundary $\partial D_i(d_0)$ or an
initial circle.
}
\label{c5_DOB1}
\end{figure}

\begin{remark}
If the set ${\cal A}(d_0)$ is convex, then it can be seen
that the boundary $\partial {\cal A}(d_0)$ does not contain tangent points. If
the set ${\cal A}(d_0)$ is non-convex,
the boundary  $\partial {\cal A}(d_0)$ may have
tangent points, see Fig. \ref{c5_DOB1}.
\end{remark}

\begin{figure}[thpb]
\centering
\epsfig{figure=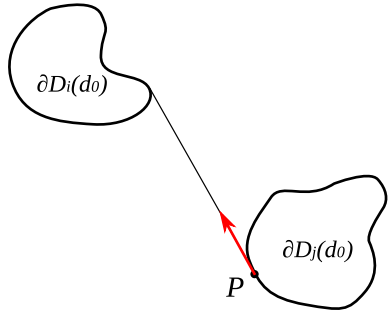,width=8cm}
\caption{
Exit direction of the tangent point $P$.
}
\label{c5_D3.2}
\end{figure}

\begin{definition}
Segments of tangent lines between
two tangent points are called tangent segments.
For any tangent point $P$, the direction of the corresponding tangent segment from this tangent point to the another tangent 
point of this segment is called the exit direction of the 
tangent point $P$; see Fig. \ref{c5_D3.2}.
\end{definition}

\begin{figure}[thpb]
\centering
\epsfig{figure=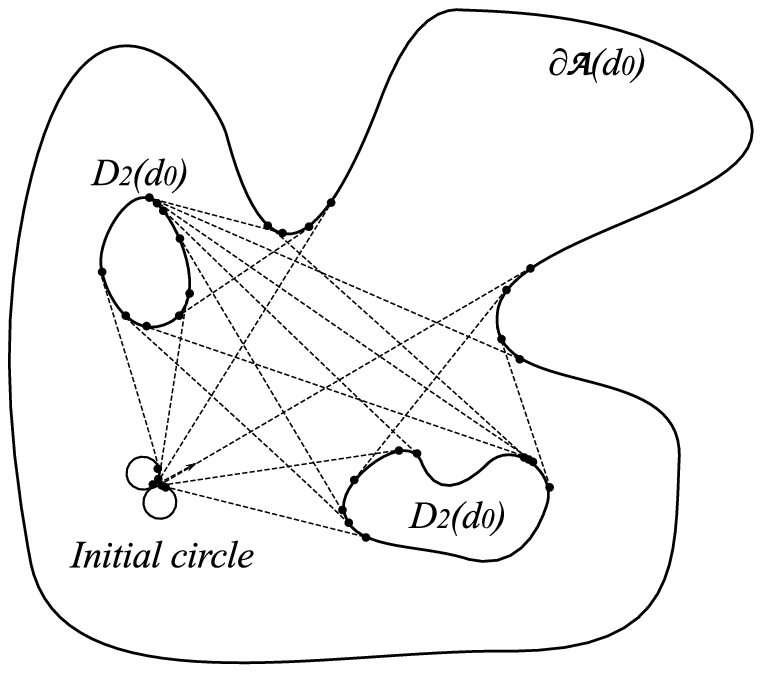,width=8cm}
\caption{
Graph ${\cal G}$.
}
\label{c5_FFF3}
\end{figure}

\begin{figure}[thpb]
\centering
\epsfig{figure=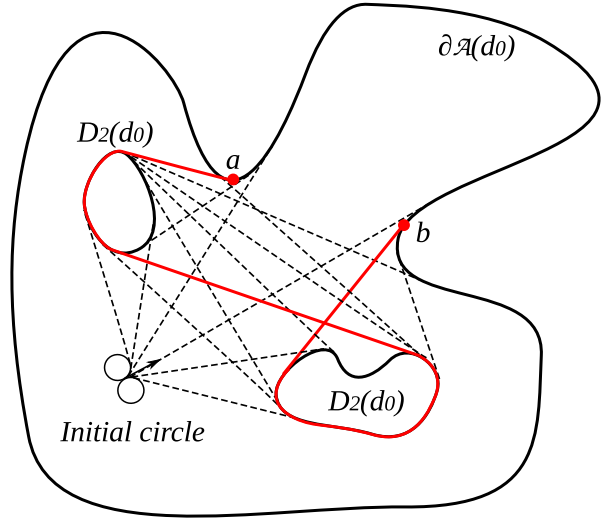,width=8cm}
\caption{
A viable path connecting the vertices $a$ and $b$ in the graph ${\cal G}$.
}
\label{c5_FFF4}
\end{figure}

\begin{definition}
The graph ${\cal G}$ is introduced that its 
vertices are the robot's initial position $p(0)$ and the tangent points,
and its edges are the tangent segments,  arcs of
the initial circles and the segments of the boundaries of the extended obstacles and the reduced area 
that connect the vertices of the graph. The graph
${\cal G}$ is called the extreme graph (see e.g. Fig. \ref{c5_FFF3}). A
path on this graph connecting two vertices  is said to be viable if the heading at the end of each edge of the path is
equal to the heading at the beginning of the next edge of
the path (see Fig. \ref{c5_FFF4})
\end{definition}

\begin{assumption}
\label{c5_Aso7}
Let $P_1$ and $P_2$ be any two tangent points
from any  boundaries $\partial D_i(d_0)$ or $\partial {\cal A}(d_0)$.
Then there exists a viable path from $P_1$ to $P_2$ on the extreme graph ${\cal G}$.
\end{assumption}

\begin{definition}
A tangent point on  the boundary 
 $\partial D_i(d_0)$ or $\partial {\cal A}(d_0)$ 
 is said to be non-circular if the other tangent point on
 the corresponding tangent segment does not belong to
 the initial circles.
\end{definition}

Let $0<q_0<1$ be a given number. To search and map the closed area ${\cal A}$ completely, the following probabilistic navigation algorithm is proposed:

{\bf A1:} The robot starts to move along any of two initial circles.

{\bf A2:} When the robot moving along an initial circle reaches a tangent point with the exit direction coinciding with
the robot's current heading, 
it starts to move along the corresponding tangent segment.

{\bf A3:} When the robot moves  along the boundary $\partial D_i(d_0)$
  or $\partial {\cal A}(d_0)$ 
  and reaches a non-circular  tangent point with the exit direction coinciding with
the robot's current heading, with probability $q_0$
it starts to move along the corresponding tangent segment, and
with probability $(1-q_0)$ it continues to move along  the boundary.

{\bf A4:} When the robot moves along a tangent segment and reaches
a tangent point on the boundary $\partial D_i(d_0)$ or  
$\partial {\cal A}(d_0)$, it starts to move along this
boundary.

The navigation algorithm {\bf A1--A4} is based
on switching between following the initial circles, tangent
lines and segments of boundaries of the extended obstacles or the reduced search area.
In the step {\bf A1}, the proposed algorithm makes the robot either move along the corresponding tangent segment
(with probability $q_0$) or continue to move along  the boundary (with probability $(1-q_0)$. Hence, the algorithm
{\bf A1-A4}  employs a degree of randomness as part of its logic and therefore belongs to the class of randomized algorithms.

Now the main theoretical result of this chapter is presented as follows.

\begin{theorem}
\label{c5_T2} 
Suppose that Assumptions~{\ref{c5_Aso} -- \ref{c5_As3}}, {\ref{c5_Aso5} -- \ref{c5_Aso7}}
hold and the robot is navigated by the algorithm {\bf A1--A4}
with some $q_0\in (0,1)$.  Then for  any initial position and heading of the robot, with probability $1$ there exists a time $t_f\geq 0$ such that the trajectory $p(t)=(x(t),y(t)), t\in [0,t_f]$ of the robot (\ref{c5_1}) is  complete map building
with obstacle avoidance.
\end{theorem}

Assumptions~{\ref{c5_Aso} -- \ref{c5_As3}}, {\ref{c5_Aso5} -- \ref{c5_Aso7}} are needed for mathematically rigorous proof of the main theoretical result (Theorem \ref{c5_T2}). In practice, the proposed algorithm often performs well even when some of them do not hold.

{\bf The proof of Theorem \ref{c5_T2}} The first step in the proof of Theorem \ref{c5_T2} is to prove  that with 
probability $1$ there exists a time $t_f\geq 0$ which is
 a map building completing time
(see Definition \ref{c5_De2.2}).
Indeed, there are the  following two possible cases:

{\bf Case 1:} The set ${\cal R}[{\cal A},d_0]$ is non-convex.

{\bf Case 2:} The set ${\cal R}[{\cal A},d_0]$ is convex.

In {\bf Case 1}, Assumption \ref{c5_Aso6} implies that the algorithm {\bf A1--A4} includes switching
between tangent lines and the boundaries $\partial {\cal A}(d_0)$ and $\partial D_i(d_0)$ with some non-zero probabilities. Therefore, it follows from
Assumption \ref{c5_Aso7}  that with probability $1$ there exists a time $t_f\geq 0$ such that by this time the robot's trajectory includes all the points of the boundaries $\partial {\cal A}(d_0)$ and $\partial D_i(d_0)$ for all $i$
(notice that the robot can travel along the boundaries in either clockwise or counter-clockwise direction).
Now it immediately follows from Assumption \ref{c5_AsoN} that
the robot by the time $t_f$ has seen every point of the boundaries of the region and the obstacles. Furthermore,
the navigation algorithm {\bf A1--A4}  consists of moving along the  boundaries $\partial D_i(d_0)$,
$\partial {\cal A}(d_0)$, two initial circles and tangent lines between them. According to Assumption 2.5,  the distance between both initial circles and obstacles and the boundary of the region is greater or equals $d_0$. Therefore, under the navigation algorithm {\bf A1--A4}, the robot always moves along lines such that the distance between these lines  and the obstacles and the boundary of the region is greater or equals $d_0$. Hence,
the distance between the robot and the obstacles and the boundary of the region is greater or equals $d_0$.

\begin{figure}[thpb]
\centering
\epsfig{figure=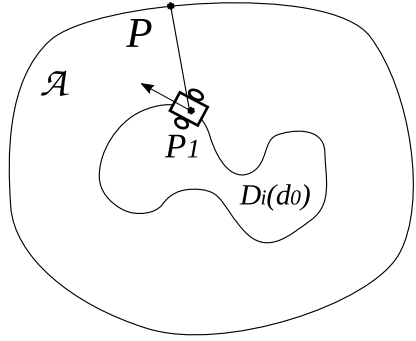,width=8cm}
\caption{
A straight line segment connecting $P$ with
a point $P_1$ of a boundary $\partial D_i(d_0)$.
}
\label{c5_DOB2}
\end{figure}

\begin{figure}[thpb]
\centering
\epsfig{figure=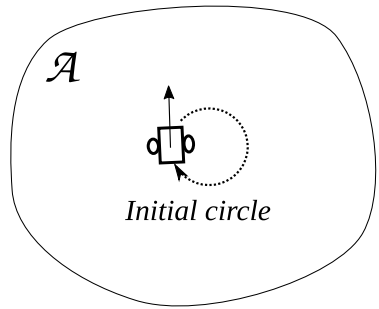,width=8cm}
\caption{
Map is built after a complete turn of an initial circle.
}
\label{c5_DOB3}
\end{figure}

In {\bf Case 2}, there is no   a tangent line connecting a tangent point on the boundary $\partial {\cal A}(d_0)$
with either a boundary $\partial D_i(d_0)$ or  an
initial circle since $\partial {\cal A}(d_0)$ does not have tangent points. In this case, the robot does not travel along $\partial {\cal A}(d_0)$. The algorithm {\bf A1--A4} includes switching
between tangent lines and the boundaries $\partial D_i(d_0)$ with some non-zero probabilities. Therefore, it follows from
Assumption \ref{c5_Aso7}  that with probability $1$ there exists a time $t_f\geq 0$ such that by this time the robot's trajectory includes all the points of the boundaries $\partial D_i(d_0)$ for all $i$.
Assumption \ref{c5_AsoN} immediately implies that
the robot by the time $t_f$ has seen every point of the boundaries of the obstacles. Moreover, this and the convexity
of the set ${\cal R}[{\cal A},d_0]$ imply that
the robot by the time $t_f$ has seen every point of the boundary of the region ${\cal A}$ as well. Indeed,
Let $P$ be an arbitrary point of the boundary of ${\cal A}$.
If there exists a straight line segment connecting $P$ with
a point $P_1$ of a boundary $\partial D_i(d_0)$ which does not cross any obstacle (see Fig. \ref{c5_DOB2}) than the robot
sensed the point $P$ when it was at $P_1$. If there is no such a straight line segment then the region  ${\cal A}$
does not contain any obstacles. In this case, 
the algorithm {\bf A1--A4} results in travelling along an initial circle and the map of the region will be  built after one complete turn (see Fig. \ref{c5_DOB3}).
 The proof of the statement that the algorithm {\bf A1--A4}  guarantees
that the distance between the robot and the obstacles and the boundary of the region is greater or equals $d_0$
is exactly the same as in {\bf Case 1}.
This completes the proof of Theorem \ref{c5_T2}.

\begin{remark}
\label{c5_Rem1}
It should be pointed out that the robot will stop
at a time $t_f\geq 0$ that is a map building completing time
(see Definition {\ref{c5_De2.2}}).
It immediately follows from Remark {\ref{c5_Rem}} and Theorem \ref{c5_T2} that trajectory $p(t)=(x(t),y(t)), t\in [0,t_f]$ of the robot (\ref{c5_1}) navigated by the 
algorithm {\bf A1--A4} is  complete map building
with obstacle avoidance. 
\end{remark}

\section{Computer simulations}

In this section, computer simulations of the collision free map building navigation algorithm {\bf A1--A4} are presented. In these simulations, the odometry errors, range finder noise and dynamic environments are discussed. Finally, our algorithm is compared with another environment exploration and map building algorithm.

To follow tangent lines and boundaries of the reduced region and the extended obstacles, a sliding mode control law which is a modification  of the law of \cite{Savkin2013} is used, that consists of switching between a boundary following approach proposed in \cite{Matveev2011a}, and the pure pursuit navigation approach; see e.g. \cite{Savkin2010}. Our sliding mode navigation  law can be expressed as follows:

\begin{equation}
\label{c5_eq:control}
u(t) = \left \{  \begin{array}[h]{cl}\pm u_M & R1 \\\Gamma sgn\left[ \phi _{tan}(t) \right]u_M & R2, \\ \Gamma sgn\left[\dot{d}_{min}(t) + X(d_{min}(t)-d_{0} )\right]u_M & R3 \\ \end{array} \right.
\end{equation}
where the function $sgn(x)$ is defined as follows:

\begin{equation}
\label{c5_eq:sgn}
sgn(x) = \left \{  \begin{array}[h]{cc}1 & x>0 \\0 & x=0  \\-1 & x<0 \end{array}. \right.
\end{equation}

This navigation law defined three control strategies corresponding to three separate modes $R1-R3$. There are three rules for switching the mode between three modes $R1-R3$. Initially mode $R1$ is active, and transitions to other modes are determined as follows:

\begin{enumerate}
\item
$R1 \rightarrow R2$: it occurs when the robot is at a tangent point on one of the initial circles and the exit direction at the tangent point coincides with the robot's current heading.

\item
$R2 \rightarrow R3$: it occurs when the robot is at a tangent point on a tangent segment and reaches the boundary $\partial D_i(d_0)$ or  
$\partial {\cal A}(d_0)$.

\item
$R3 \rightarrow R2$: with the probability $q_0$, it occurs when the robot is at a tangent point on the boundary $\partial D_i(d_0)$ or $\partial {\cal A}(d_0)$ and the exit direction at the tangent point coincides with the robot's current heading. 
\end{enumerate}

Mode $R1$ describes motion along the initial circle with maximal angular velocity. Mode $R2$ describes pursuit navigation, where $\phi_{tan}(t)$ is defined as the angle between the robot's heading and a line segment connecting the vehicle and the currently tracked tangent segment. Mode $R3$ describes boundary following navigation, where the control calculation is based on the minimum distance to the nearest obstacle, defined as $d_{min}(t)$. The control law (\ref{c5_eq:control}) consisting of switching constructs a hybrid control system; see \cite{Matveev2000Estimation,Savkin2002Hybrid,
SAVKIN199969Robust,SKAFIDAS1999553Stability}.

This control law is subject to some restrictions which are inherited from \cite{Matveev2011a}.
The variable $\Gamma$ is defined as $+1$ if the boundary followed is on the left of the tangent being tracked, $-1$ if it is on the right. A constant $d_{trig} > d_0$ is also introduced to determine when the control system transitions to boundary following mode  \cite{Matveev2011a}. The saturation function $X$ is defined as follows:
\begin{equation} 
\label{c5_eq:sat}
X(r) = \left \{  \begin{array}[h]{cc} lr & |r| < k\\ lk\ sgn(r) & otherwise\\\end{array} \right.
\end{equation}
where $l$ and $k$ are tunable constants. Because of any potential chattering in the robot's heading caused by the sliding mode control, once it is decided to not pursue a tangent line, there is a short pause until tangent following can potentially be engaged again.
It means, during this short pause, the mode $R3$ will not switch to $R2$ even if the robot's heading coincides with a tangent line again.

According to the system description, the robot is equipped with a range finder sensor and an odometry sensor. The range finder sensor can inform the distance from the mobile robot to the nearest obstacles in a finite number of directions. The measurements of the range finder sensor can be represented as a finite point set $\lbrace W_i(\omega_i,\varsigma_i) \vert i=1, 2\ldots n \rbrace$. $\varsigma_i$ is the distance from the robot to the nearest obstacle in the direction $\omega_i$. $\omega_i$ is the direction angle relative to the robot's heading. Moreover, the odometry sensor measures the robot's speed and angular velocity. The robot can calculate its position and orientation by using odometry.

Notice that the robot will stop
at a time $t_f\geq 0$ that is a map building completing time
(see Definition \ref{c5_De2.2}).
As it is pointed out in   Remark \ref{c5_Rem1}, in this case, the trajectory $p(t)=(x(t),y(t)), t\in [0,t_f]$ of the robot (\ref{c5_1}) navigated by the 
algorithm {\bf A1--A4} is  complete map building
with obstacle avoidance.

Now, the following approaches to implement the proposed algorithm {\bf A1--A4} are presented.

With a small angular resolution $\Delta \omega$ of the range finder sensor, the variable $d_{min}(t)$ can be approximated by the minimum value $\varsigma_{min}$. Similarly, $\dot{d}_{min}(t)$ can be approximated by $\frac{d_{min}(t+T)-d_{min}(t)}{T}$ with a short sampling interval $T$.

\begin{figure}[thpb]
\centering
\epsfig{figure=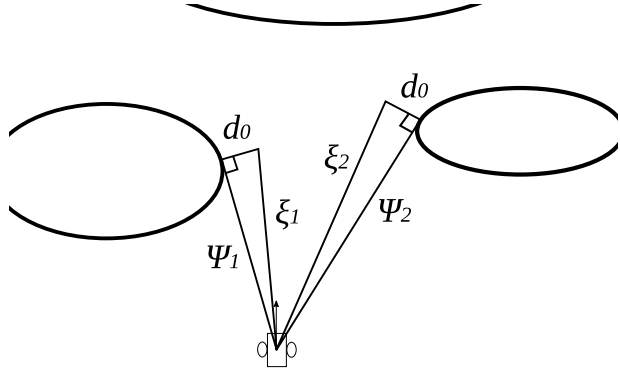,width=8cm}
\caption{
Calculation of the tangent line $\xi_k$.
}
\label{c5_H1}
\end{figure}

\begin{figure}[thpb]
\centering
\epsfig{figure=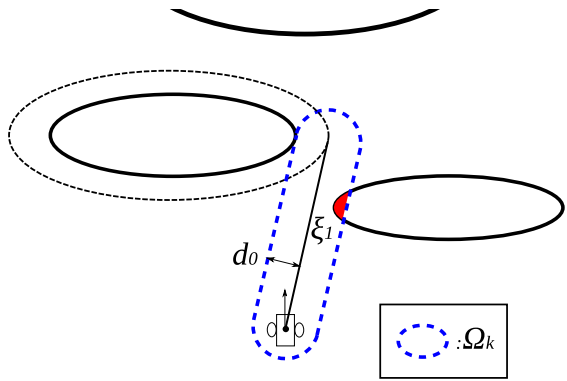,width=8cm}
\caption{
Definition of safe area $\Omega_k$.
}
\label{c5_H2}
\end{figure}

To determine when the mode $R2$ should be activated, the following steps are designed to check whether the robot's orientation coincides with a tangent line (see Fig. \ref{c5_H1}):

\begin{enumerate}

\item
Calculate $\vert \Delta \varsigma_i \vert$ between successive $\varsigma_i$ and $\varsigma_{i+1}$ for any $i=1,2\ldots n-1$. According to the Assumption 2.2, the minimum distance between any two obstacles is $>2\cdot d_0$. Therefore, the tangent lines on the boundary of obstacles are obtained by checking if $\vert \Delta \varsigma_i \vert >2 \cdot d_0$. When there exists any $\vert \Delta \varsigma_k \vert >2 \cdot d_0$, let $\varsigma'_k$ denote the smaller value between $\varsigma_k$ and $\varsigma_{k+1}$. Then, the line segment $\Psi_k$ from the pole to the point $(\omega'_k,\varsigma'_k)$ is a tangent line on an obstacle.

\item
According to the geometry, let $\Psi_k$ be a leg and construct a right triangle. The length of another leg which shares the common endpoint $(\omega'_k,\varsigma'_k)$ with $\Psi_k$ is $d_0$. After this, the hypotenuse $\xi_k$ can approximately represent the tangent line on the corresponding safe boundary $\partial D(d_0)$. Note that if $\varsigma_k<\rho_{k+1}$, the polar angle of $\xi_k$ is bigger than the polar angle of $\Psi_k$ (see $\xi_2$ in Fig. \ref{c5_H1}). If $\varsigma_k>\rho_{k+1}$, the situation is opposite (see $\xi_1$ in Fig. \ref{c5_H1}).

\item
Define a safe area $\Omega_k$ for each line segment $\xi_k$ (see Fig. \ref{c5_H2}). The set $\Omega_k$ is formed by all points at the distance $<d_0$ from the line segment $\xi_k$. If there is any point $W_i$ within the area $\Omega_k$, $\xi_k$ is not a real tangent line because in geometry, it intersects with other $\partial D(d_0)$ and can not be pursued safely by the robot. 

\item
If there exist a $\xi_k$ that the absolute value of its angle $\omega_k$ in the local polar coordinate system is smaller than a given threshold $\theta_{trig}$, the robot's orientation approximately coincides with the tangent line $\xi_k$.

\end{enumerate}

To calculate $\phi_{tan}(t)$ in the equation (\ref{c5_eq:control}), a intermediate target $P_T$ is defined. Assume that when the mode $R2$ is activated, the robot's orientation coincides with the tangent line $\xi_k$. Then, the temporary target $P_T$ is the endpoint of $\xi_k$. After this, $\phi_{tan}(t)$ is the angle between the robot's heading and the vector from the robot's position $(x(t),y(t))$ to the intermediate target $P_T$.

A raster graphics map is used to represent the environment. Each cell of the map has three optional states (unknown, unoccupied and occupied). Initially, all the cells are unknown. At each sampling time, the measurements $W_i(\omega_i,\varsigma_i)$ are transformed from the local polar coordinate system to the global Cartesian coordinate system $W_i(x_{W(i)},y_{W(i)})$. Then, the Bresenham's line algorithm \cite{Bresenham1965} is used to determine the state of each cell intersecting with the line segments between robot's position $(x(t),y(t))$ and each $W_i(x_{W(i)},y_{W(i)})$. The state of each cell is updated by the following strategy:

\begin{enumerate}

\item
If an unknown cell or an unoccupied cell is occupied by $W_i(x_{W(i)},y_{W(i)})$, the cell will be updated to occupied cell.

\item
If an unknown cell intersects with a line segment between $(x(t),y(t))$ and $W_i(x_{W(i)},y_{W(i)})$ and is not occupied by $W_i(x_{W(i)},y_{W(i)})$, the cell will be updated to unoccupied cell.

\end{enumerate}

In the implementations of the proposed algorithm, there are some parameters which should be determined as follows:

\begin{enumerate}

\item
The maximal angular velocity $u_M$ is a known physical parameter of a mobile robot.

\item
For a real robot, the safety margin $d_0$ should be bigger than the maximum distance from the robot's center to its boundary. Moreover, it should be smaller than a half of the minimal distance between any obstacles (see Assumption \ref{c5_Aso1}).

\item
The robot's speed $v$ should be $<u_M \cdot d_0$ because $R_{min}$ should be $<d_0$ (see Assumption \ref{c5_Aso2}).

\item
The two thresholds $\theta_{trig}$ and $d_{trig}$ should be tuned in different implementations to guarantee the correct mode transitions.

\item
If the precision of a range finder sensor is $\pm e$, the resolution of the map is recommended to be $>(2e)^2/cell$ because a higher resolution than $(2e)^2/cell$ is unnecessary.

\end{enumerate}

In the following parts, five groups of computer simulations are carried out to confirm and discuss the performance of the proposed algorithm.

\begin{figure}[thpb]
\centering
\subfigure[$t=125 \text{s}$]{
\epsfig{figure=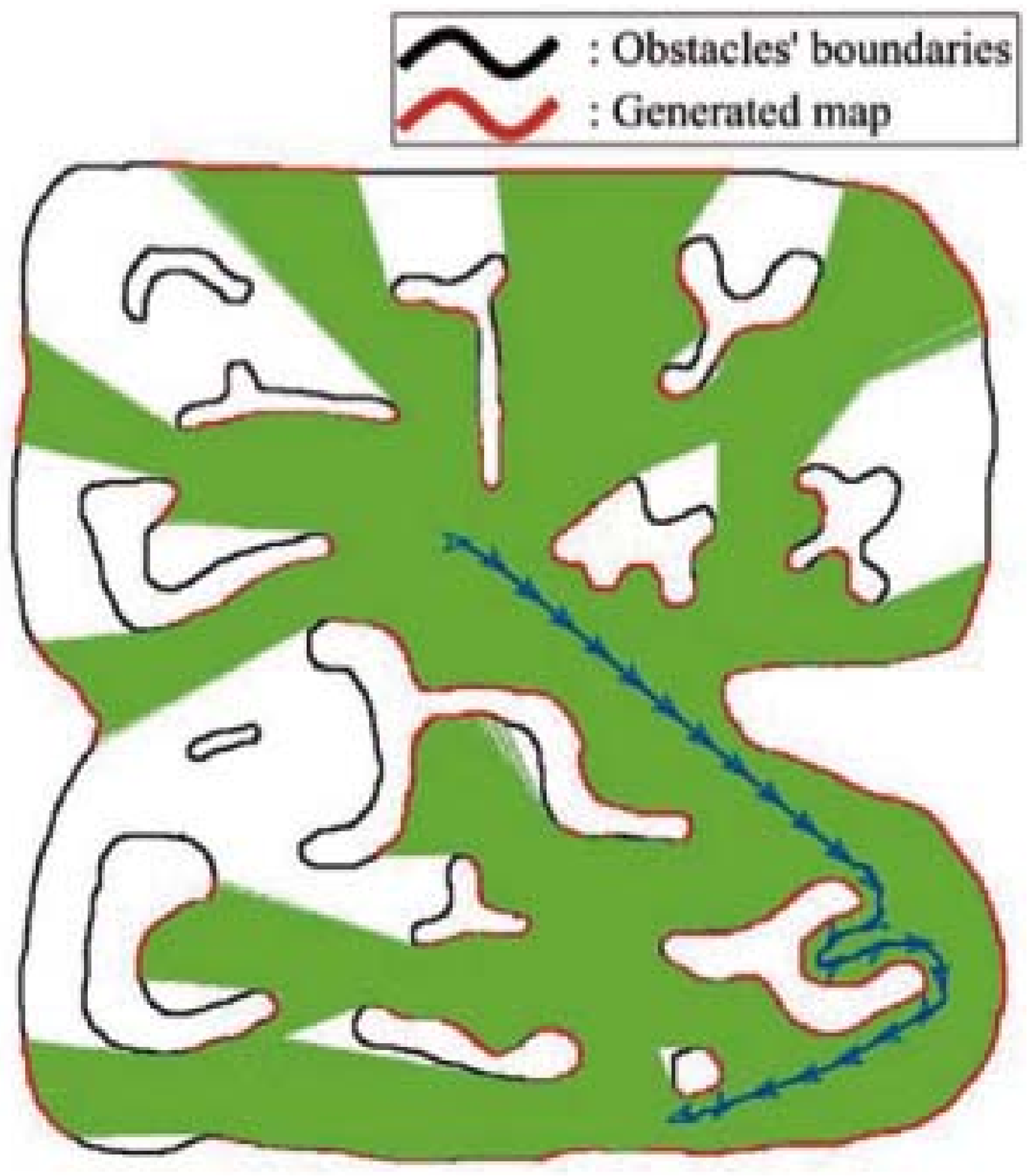,width=7cm}
}
\subfigure[$t=920.5 \text{s}$]{
\epsfig{figure=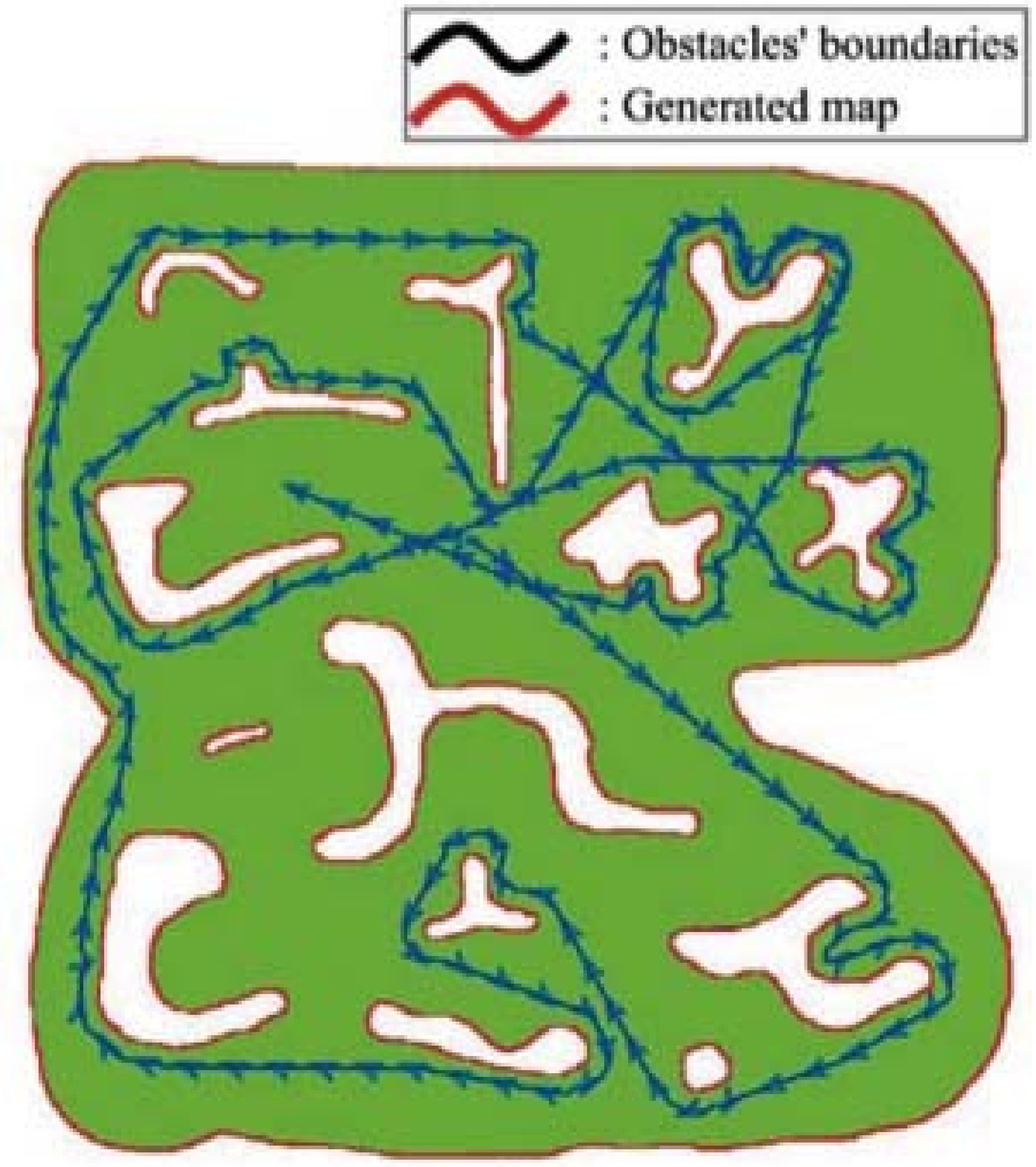,width=7cm}
}
\caption{
Result of the simulation in the first scene of Simulation 1.
}
\label{c5_FS1}
\end{figure}

{\bf Simulation 1}: The proposed algorithm was simulated in two $112.5\text{m}\times 112.5\text{m}$ cluttered environments with different types of obstacles.
In the first scene, there are some non-convex obstacles. The parameters used for the simulation are listed in Table. {\ref{c5_H3}}. The result of the simulation in the first scene is shown in Fig. {\ref{c5_FS1}}. It can be seen that the robot searched the whole area successfully and built a complete map of the environment in $920.5\text{s}$.

\begin{table}[!htb]
\caption{Parameters for Simulation 1}
\label{c5_H3}
\begin{center}
\begin{tabular}{c|c|c}
\hline
Speed of robot & $v$ & $1 \text{m/s}$\\
\hline
Maximum angular velocity & $u_{M}$ & $1.4 \text{rad/s}$\\
\hline
Minimum turning radius & $R_{min}$ & $0.71 \text{m}$\\
\hline
Safety margin & $d_0$ & $1.7 \text{m}$\\
\hline
Sampling interval & $T$ & $0.25 \text{s}$\\
\hline
Angle threshold & $\theta_{trig}$ & $0.08 \text{rad}$\\
\hline
Distance threshold & $d_{trig}$ & $0.3 \text{m}$\\
\hline
Probability & $q_0$ & $0.4$\\
\hline
Resolution of the map & & $20 \text{cm}^2/cell$\\
\hline
\end{tabular}
\end{center}
\end{table}

\begin{figure}[thpb]
\centering
\subfigure[$t=250 \text{s}$]{
\epsfig{figure=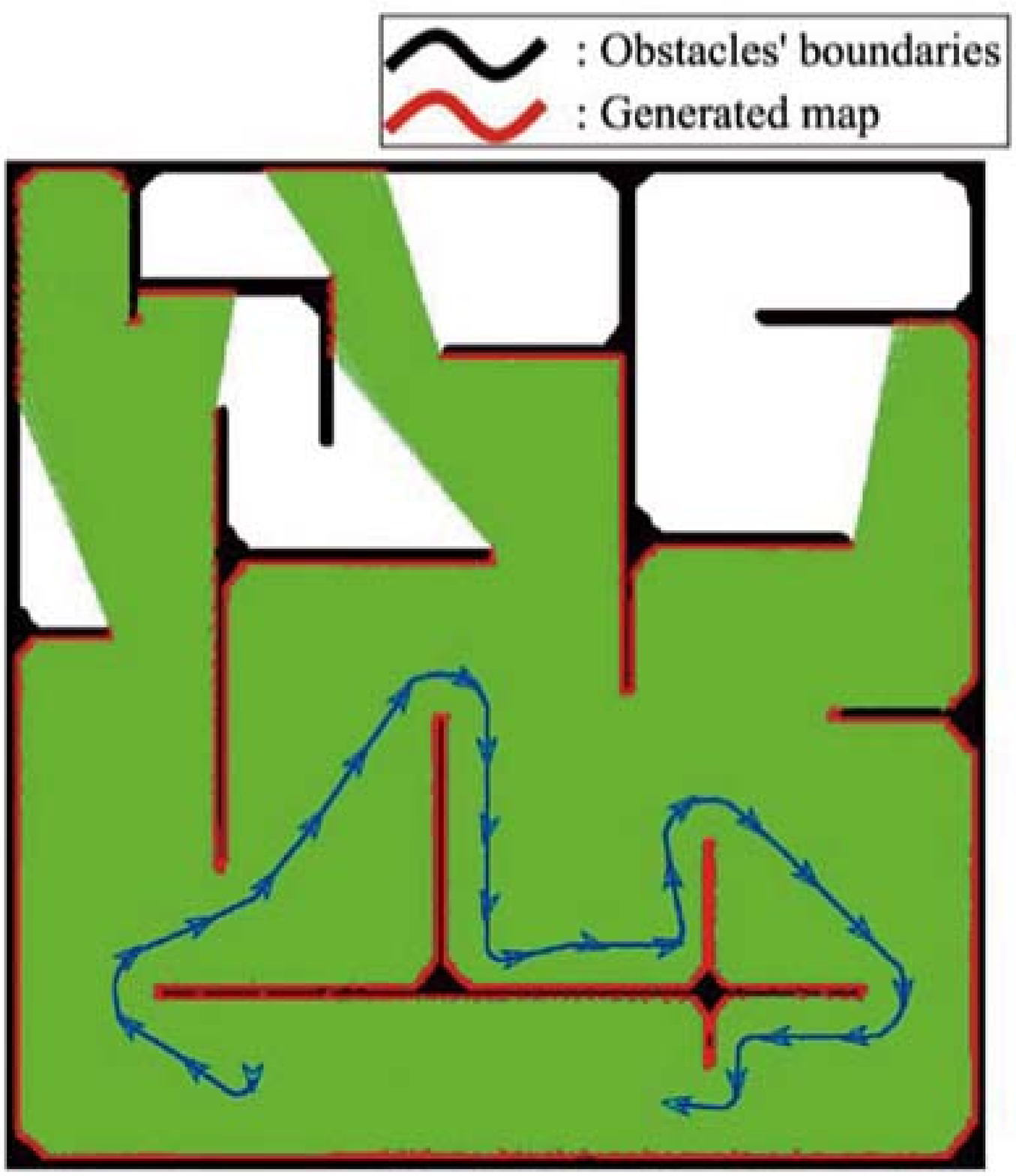,width=7cm}
}
\subfigure[$t=1186 \text{s}$]{
\epsfig{figure=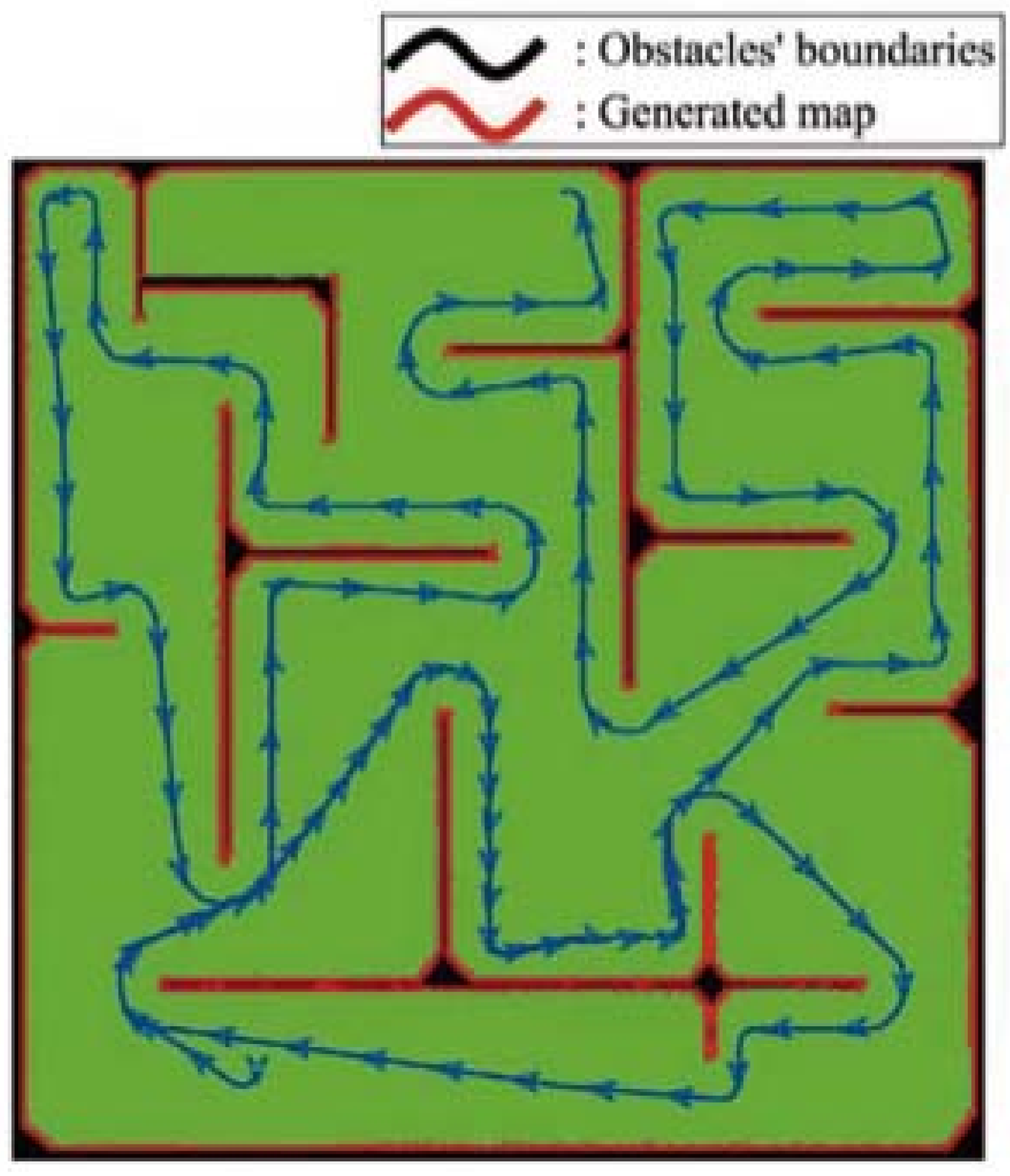,width=7cm}
}
\caption{
Result of the simulation in the second scene of Simulation 1.
}
\label{c5_FS1B}
\end{figure}

In the second scene (a maze), the safety margin was changed to $4\text{m}$ to make the result clear to see. Thus, the speed and angular velocity used for this simulation were changed to $0.8\text{m/s}$ and $0.5 \text{rad/s}$. As the result shows in Fig. {\ref{c5_FS1B}}, the robot started from a initial position and searched the whole area of this maze without any collision. Finally, it took $1168\text{s}$ to build a complete map of this environment.

\begin{figure}[thpb]
\centering
\epsfig{figure=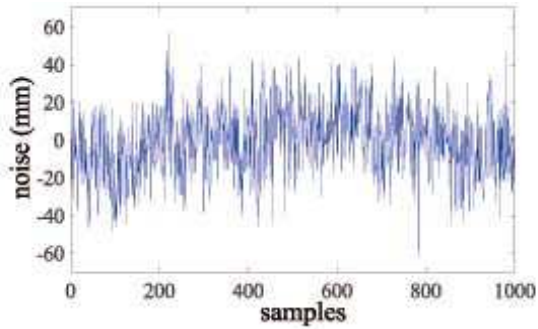,width=8cm}
\caption{
Modelled laser range finder noise.
}
\label{c5_NOISE}
\end{figure}

\begin{figure}[thpb]
\centering
\subfigure[Obstacles' boundaries]{
\epsfig{figure=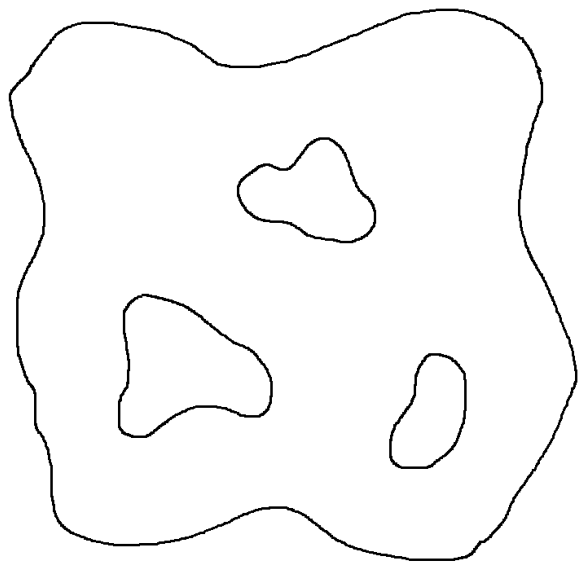,width=7cm}
}
\subfigure[Generated map]{
\epsfig{figure=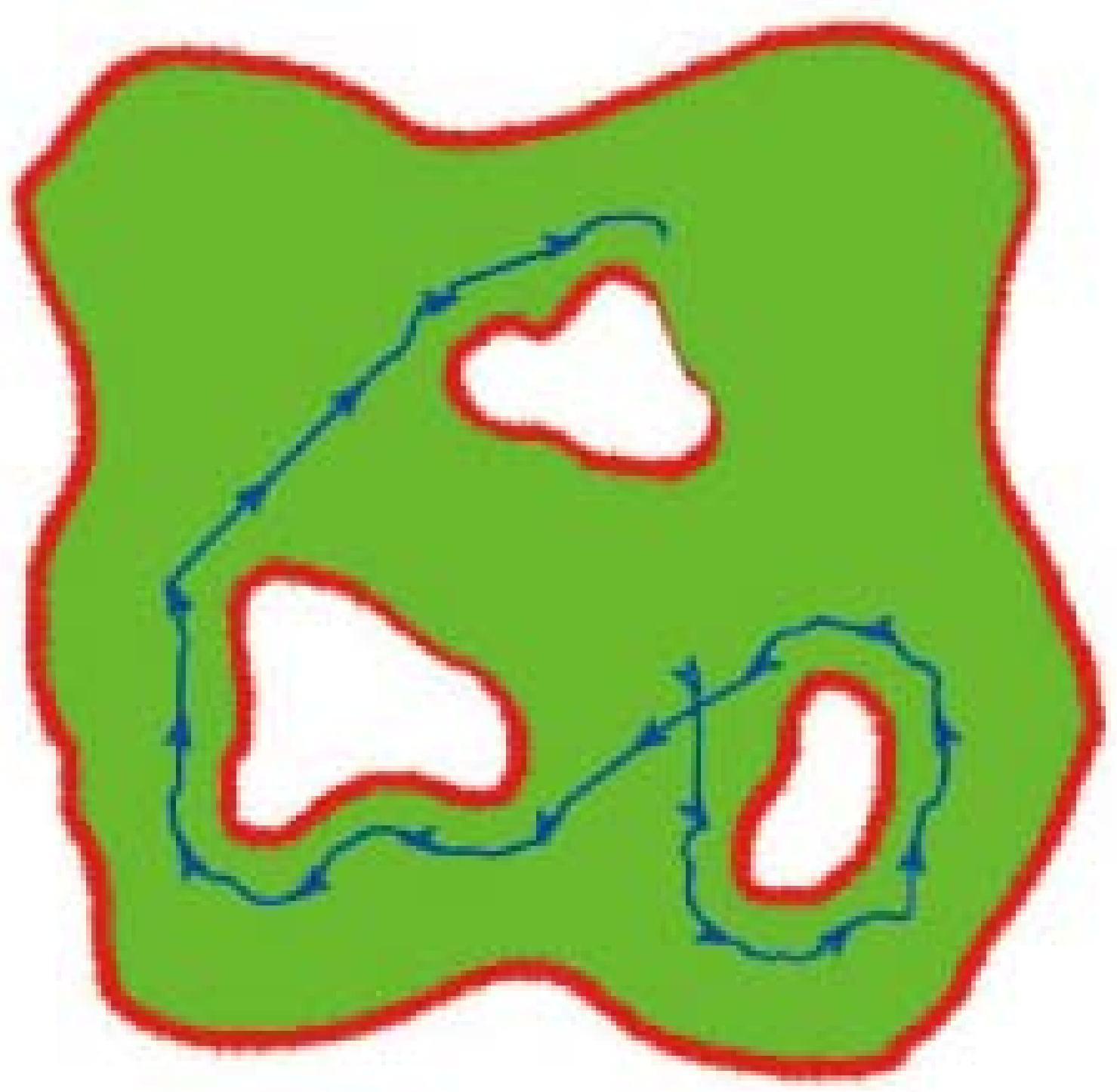,width=7cm}
}
\caption{
Result of Simulation 2 with range finder noise.
}
\label{c5_FS2A}
\end{figure}

{\bf Simulation 2}: In simulation 2, the range finder sensor is considered as a laser range finder and the noise model proposed in \cite{Jain2011} was used to simulate the laser range finder noise. The modelled noise is a combination of flicker noise and white noise with standard deviation of $1.77023 \text{mm}$ and $3.03856 \text{mm}$ (see Fig. {\ref{c5_NOISE}}). In this simulation, the modelled noise was increased by $5$ times to make the effects of the noise more obvious in the result of the simulation. The size of the environment is $11.25\text{m}\times 11.25\text{m}$ and the resolution of the map is $1 \text{cm}^2/cell$. Fig. {\ref{c5_FS2A}} shows the complete map in this simulation with the laser range finder noise.
It can be seen that the proposed algorithm is robust with the range finder errors, although the map quality is influenced (i.e. rough boundaries in generated maps).

\begin{figure}[thpb]
\centering
\epsfig{figure=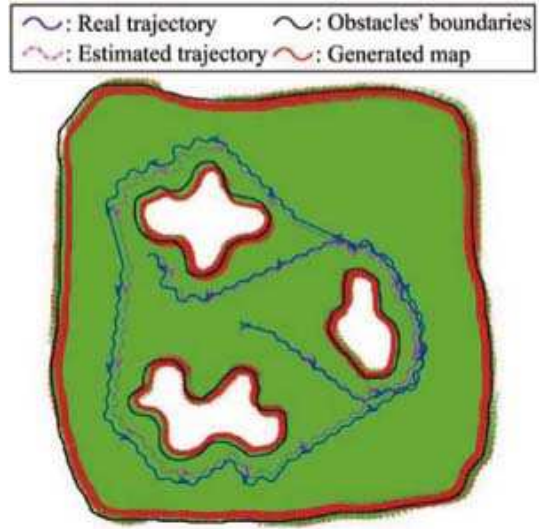,width=7cm}
\caption{
Result of Simulation 3 with odometry errors.
}
\label{c5_FS3A}
\end{figure}

{\bf Simulation 3}: In the proposed algorithm, odometry is used to estimate the position and orientation of the robot. The odometry errors are caused by many reasons \cite{Borenstein1996}.
In this simulation, only the odometry errors caused by the approximation in the odometry equations \cite{Borenstein1996} are considered. The sampling interval in this simulation is $0.5\text{s}$, which caused an obvious odometry errors. The size of the environment is $11.25\text{m}\times 11.25\text{m}$. The result of the simulation (see Fig. {\ref{c5_FS3A}}) shows that the estimation of the robot's position diverged from the real position gradually with the increase of time. As a result of odometry errors, there are obvious errors on the generated map.

\begin{figure}[thpb]
\centering
\epsfig{figure=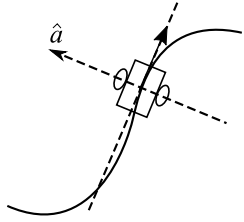,width=8cm}
\caption{
Direction of the acceleration measured by the accelerometer.
}
\label{c5_ACCELEROMETER}
\end{figure}

To alleviate the odometry errors, an improved estimation model with the help of an accelerometer is proposed here. The accelerometer measures the acceleration of the robot in the direction vertical to the robot's heading (see Fig. \ref{c5_ACCELEROMETER}). Let $\hat{v}$ and $\hat{u}$ be the speed and angular velocity measured by odometry sensors, respectively. Let $\hat{a}$ be the acceleration measured by the accelerometer. In the improved estimation model, $\hat{v}$, $\hat{u}$ and $\hat{a}$ are approximated as constant during the sampling interval $T$. Based on the approximation, the improved estimation model is proposed as follows:

\begin{equation}
\label{c5_EMo}
\begin{array}{l}
x(k+1) = x(k) + Q \cos \theta (k)+\frac{\hat{a}T^2}{2}\cos (\theta (k)+\frac{\pi}{2})
\\
y(k+1) = y(k) + Q \sin \theta (k)+\frac{\hat{a}T^2}{2}\sin (\theta (k)+\frac{\pi}{2})
\\
\theta(k+1) = \theta(k) + \hat{u}(k)
\end{array},
\end{equation}
where
\begin{equation}
Q=\int_0^T  \sqrt{{\hat{v}}^2-(\hat{a}t)^2} \, \mathrm{d}t=(\frac{T}{2}\sqrt{{\hat{v}}^2-(\hat{a}T)^2}+\frac{{\hat{v}}^2}{2\hat{a}}\arcsin (\frac{\hat{a}T}{{\hat{v}}^2})).
\end{equation}

\begin{figure}[thpb]
\centering
\epsfig{figure=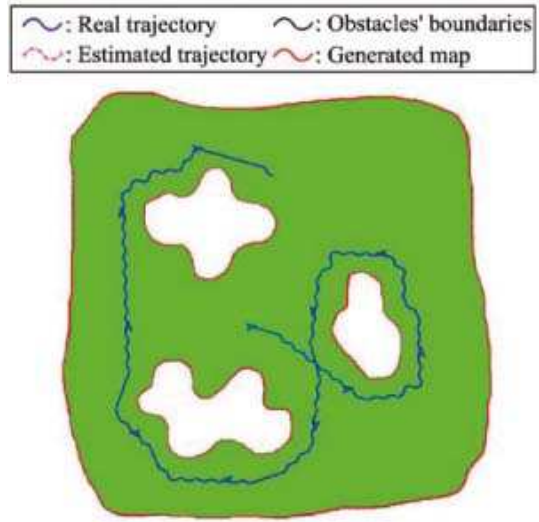,width=7cm}
\caption{
Result of the simulation with the improved estimation model (\ref{c5_EMo}).
Notice that the obstacles' boundaries nearly coincide with the generated map and the real trajectory of the robot nearly coincides with the estimated trajectory.
}
\label{c5_FS3}
\end{figure}

To confirm the performance of the improved estimation model (\ref{c5_EMo}), a new simulation is carried out with this improved model in a same environment. The result of the new simulation is shown in Fig. \ref{c5_FS3}. Comparing with Fig. \ref{c5_FS3A}, the odometry errors are alleviated significantly by the improved estimation model (\ref{c5_EMo}). Additionally, for other types of odometry errors, there are many methods focusing on the alleviation of odometry errors e.g. \cite{Doh2003,Doh2006,Sauer2001}.

{\bf Simulation 4}: In Simulation 4, a dynamic environment with both steady and  moving obstacles is considered. In the dynamic environment, the proposed algorithm should be modified to explore the environment safely and build the complete map correctly. In this simulation, the robot navigation algorithm contains  a simple motion detector block to distinguish moving  obstacles and avoid collisions with them. There are many methods focusing on the moving obstacle detection e.g. \cite{Mertz2013,Rebai2009,Yokoyama2005}. Moreover, in our future research, more advanced methods of avoiding collisions with moving obstacles can be integrated with the proposed algorithm (see e.g. \cite{Hoy2015,Matveev2015a}). With the motion detector block, the robot can avoid pursuing a tangent segment which is tangent to a moving obstacles (see algorithm {\bf A3}). With the update of the map during the exploration, the boundaries of moving obstacles on the map will be eliminated completely and it will end up with a correct map of the steady environment
(the region and all its steady obstacles).

\begin{figure}[thpb]
\centering
\epsfig{figure=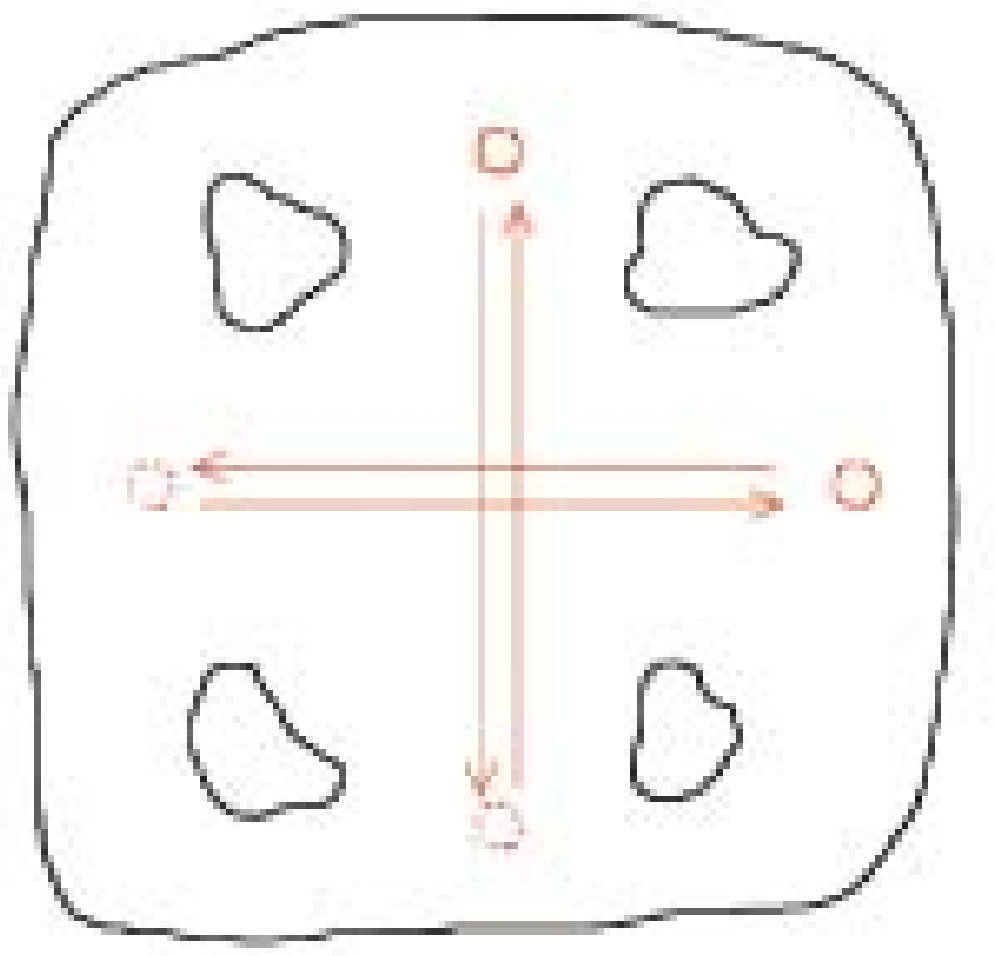,width=8cm}
\caption{
Dynamic environment in Simulation 4.
}
\label{c5_FS4A}
\end{figure}

\begin{figure}[thpb]
\centering
\subfigure[$t=6.1 \text{s}$]{
\epsfig{figure=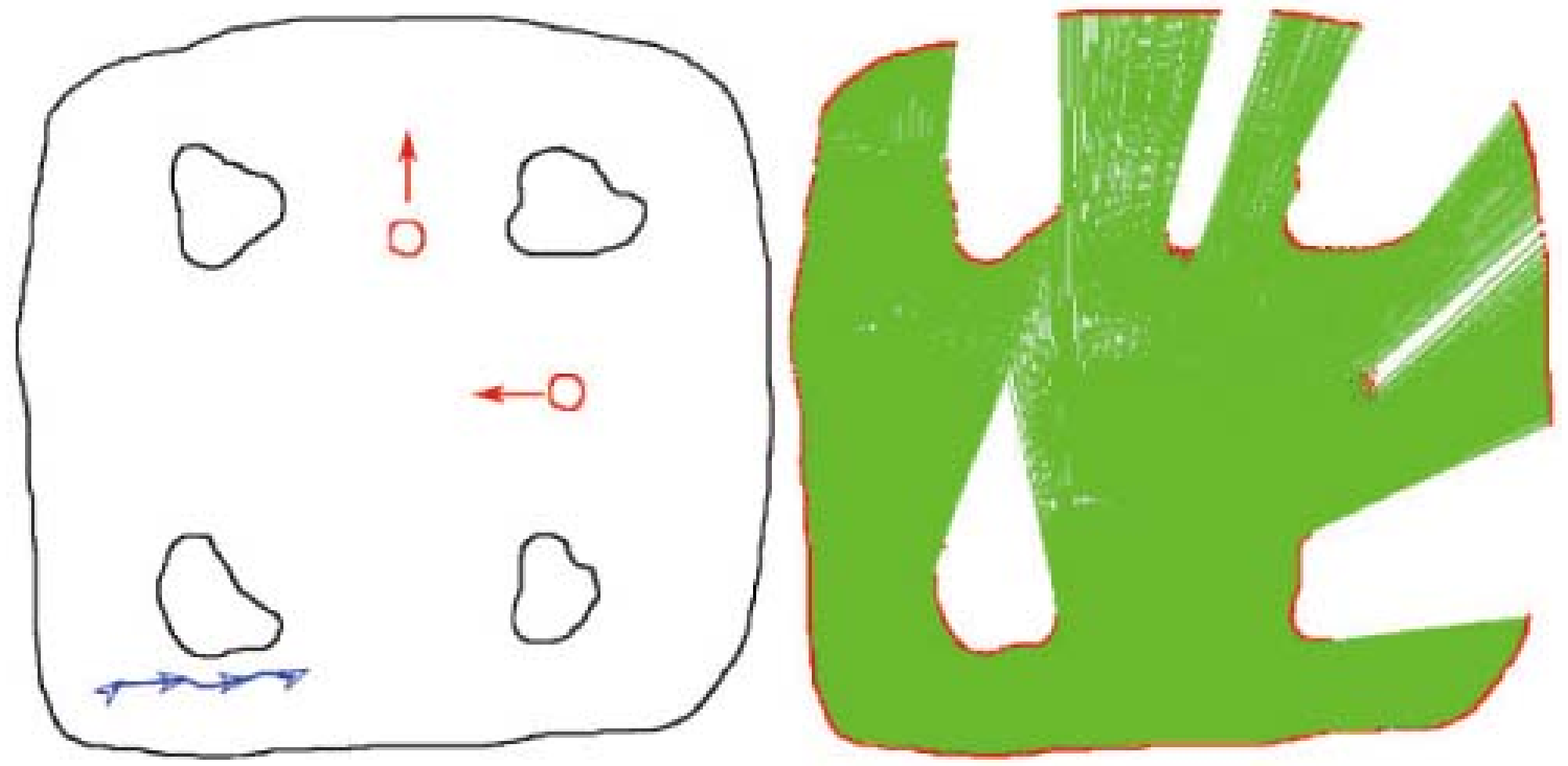,width=8cm}
}
\subfigure[$t=47.1 \text{s}$]{
\epsfig{figure=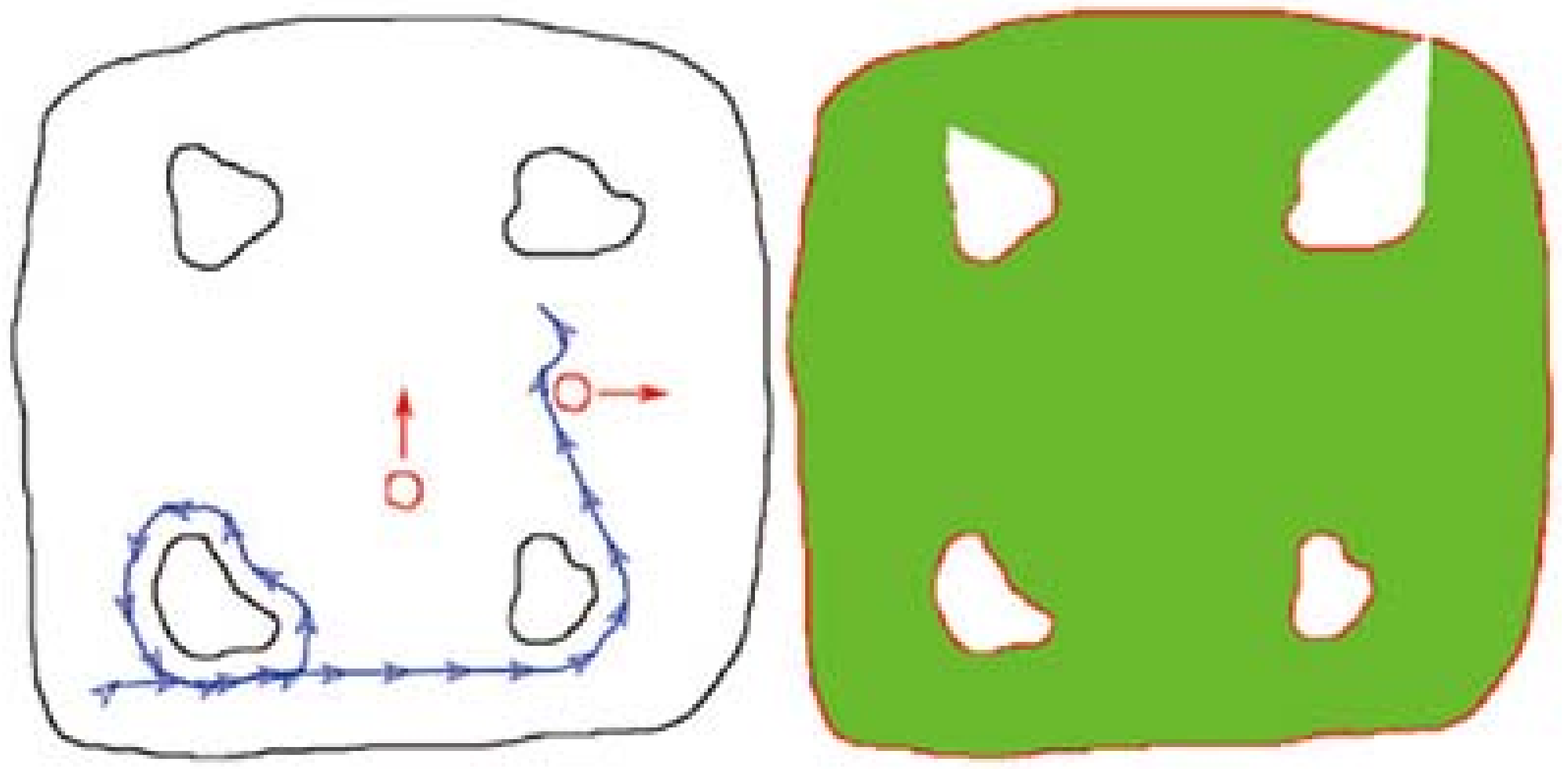,width=8cm}
}
\subfigure[$t=112.2 \text{s}$]{
\epsfig{figure=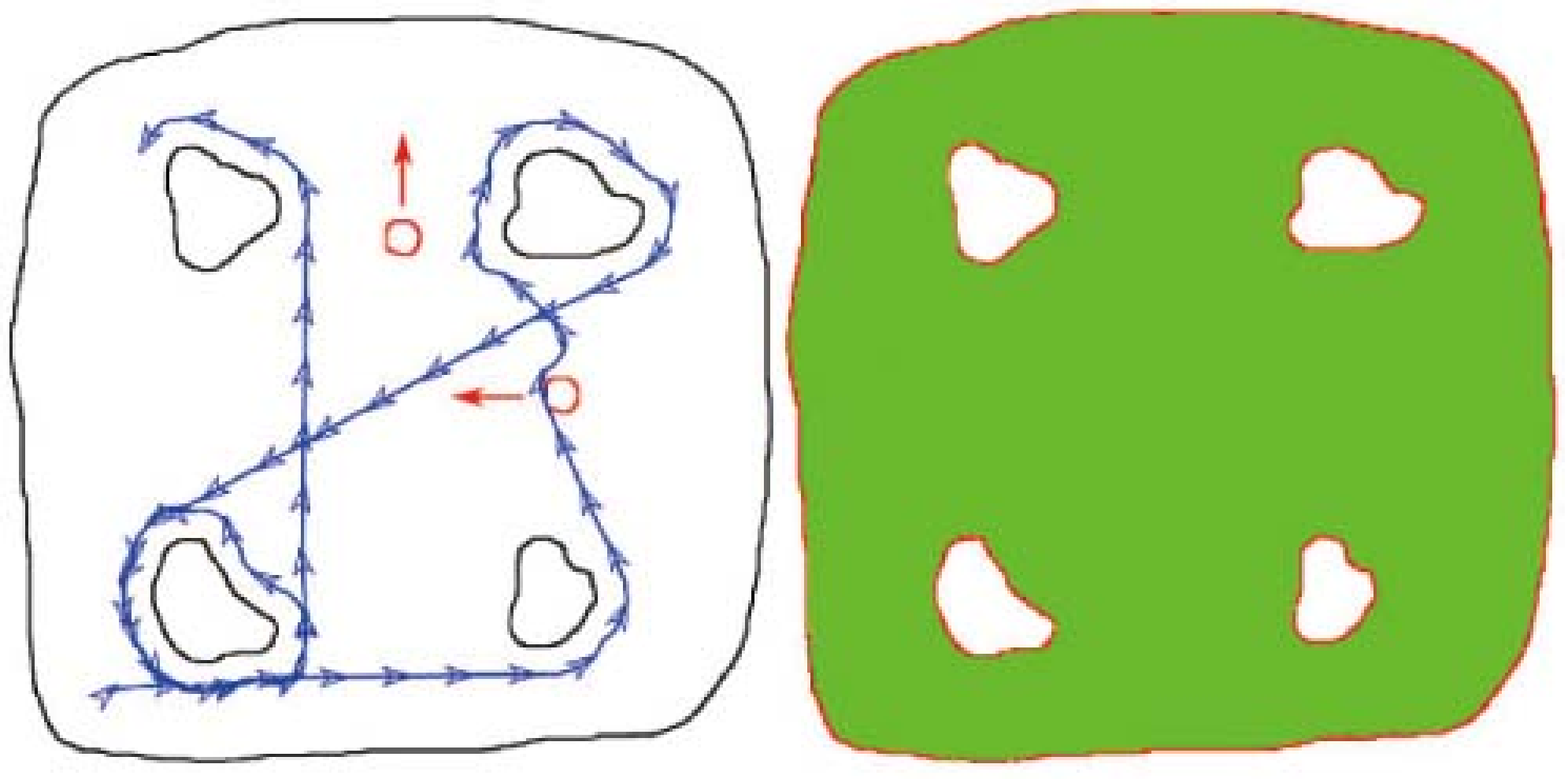,width=8cm}
}
\caption{
Result of Simulation 4 in the dynamic environment.
}
\label{c5_FS4}
\end{figure}

With the above modifications, Simulation 4 is carried out with two moving obstacles in a $5\text{m}\times 5\text{m}$ environment (see Fig. \ref{c5_FS4A}). The result of the simulation is shown in Fig. \ref{c5_FS4}. It can be seen that the robot successfully explored the area without pursuing any tangent line belonging to moving obstacles. Moreover, it can be seen that, at the time $47.1\text{s}$, the robot avoided the moving obstacle when pursing a tangent line. According to the update of the map, it can be seen that the boundaries of moving obstacles are eliminated gradually. Finally, the robot built a correct complete map of the region and all its steady  obstacles.

{\bf Simulation 5}: In recent researches, the frontier-based exploration algorithms are the most common exploration strategies. In this simulation, a frontier-based exploration algorithm proposed in \cite{Wang2015a} was selected to compare the exploration time with the proposed algorithm.
The selected algorithm has a better performance than current exploration methods. It keeps the map information integrated during exploration and uses a multi criteria decision method to overcome the drawbacks of the traditional weighted average method to improve the efficiency of the exploration.

Notice that, although the Growing Neural Gas network topological map is used to represent the environment, which is different from geometrical map, it does not affect the exploration time. In this simulation, the algorithm of \cite{Wang2015a} was simulated and run in some new examples to compare the exploration time with the proposed algorithm.

In these new examples, the exploration time of the two algorithms is compared in different environment with different number of obstacles. Notice that the algorithm of \cite{Wang2015a} refers to a front exploration area which is a semi-circle at the front of the robot. Therefore, different radius of the front exploration area was considered in the simulations.

\begin{figure}[thpb]
\centering
\epsfig{figure=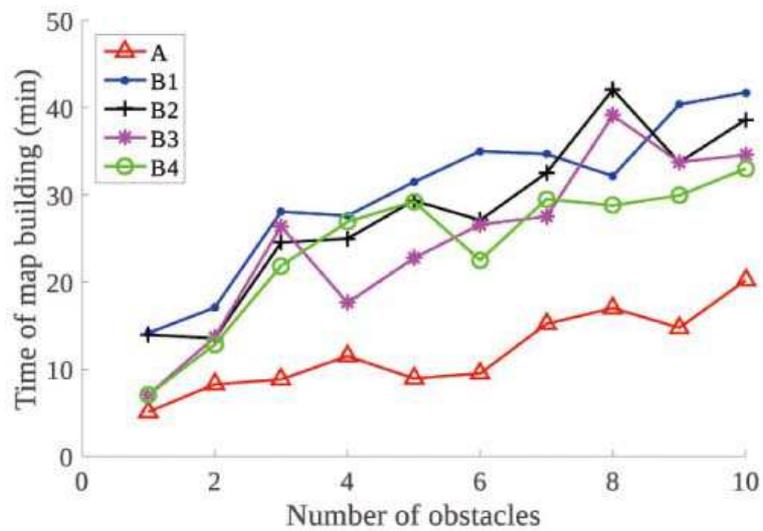,width=10cm}
\caption{
Exploration time of the two algorithms with different number of obstacles.
}
\label{c5_COM}
\end{figure}

In the comparison, the size of the environment is $37.5 \text{m} \times 37.5 \text{m}$ and the robot's speed is $0.2\text{m/s}$. Fig. \ref{c5_COM} shows the different exploration time of two algorithms with different number of obstacles. The performance of the proposed algorithm refers to Line A. The performance of the algorithm of \cite{Wang2015a} with different radius of $5\text{m}$, $6\text{m}$, $7\text{m}$ and $8\text{m}$ refers to Line B1 to B4, respectively.

According to the comparison, it can be seen that our randomized search algorithm has a strong advantage because our method builds a complete map of the environment nearly twice faster than the algorithm of \cite{Wang2015a}. Furthermore, our algorithm is designed for a non-holonomic model (see Equation (\ref{c5_1})), whereas the algorithm of 
 \cite{Wang2015a} and the most of works in this fields do not consider any non-holonomic constraint of the robot's motion. Moreover, the algorithm of \cite{Wang2015a} may fail to search some part of the area when the diameter of the front exploration area is smaller than the minimum distance between any obstacles. Additionally, the performance of our algorithm is verified by a mathematically rigorous proof. Finally, the developed randomized algorithm requires a smaller computational load than the exploration algorithm proposed in  \cite{Wang2015a}.

\section{Experiments with real mobile robot}

In this section, the real implementation of the proposed collision free map building algorithm {\bf A1--A4} are proposed with a Pioneer-3DX wheeled mobile robot. The robot has an on-board computer with a Linux operating system. It can be programmed to control the speed of each wheel separately and process the data of the mounted sensors. The Pioneer-3DX robot calculates its real-time position and orientation by odometry.

\begin{figure}[thpb]
\centering
\epsfig{figure=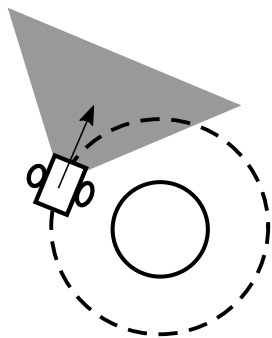,width=8cm}
\caption{
The case when scanning range is $<180^{\circ}$.
}
\label{c5_EA}
\end{figure}

In the experiments, the robot is equipped with a SICK LMS-200 laser range finder (LRF) and an odometry sensor. This LRF's angular resolution, scanning range and measurement range are $0.5^{\circ}$, $180^{\circ}$ and $80 \text{m}$ respectively. The LRF's precision is $\pm 15\text{mm}$.
Notice that, to guarantee that the minimum distance to obstacles can be measured correctly and any point on obstacles' boundaries can be detected by the range finder, the scanning range of the LRF should be $\geq180^{\circ}$. If the scanning range is $<180^{\circ}$, the minimum distance cannot be measured correctly and some parts of obstacles' boundaries may never be detected (see Fig. \ref{c5_EA}).
The change of the scanning range does not affect the final result of the map building and safe navigation. However, with the decrease of the scanning range, the map building completing time will increase.
Moreover, it is noticed that the scene we built in the experiments is not quite large. Thus, the odometry errors of the robot is quite small and the generated map is accurate. If the proposed algorithm is implemented in a larger scene, the approach proposed in Simulation 3 can be used to reduce odometry errors. Other technologies can also be used to  reduce the localization error of the robot such as the use of  Global Positioning System (GPS) in outdoor environments.

A closed area in an indoor environment was built by blocking the hallway and doors. Seven obstacles were placed on the floor. Each obstacle was higher than the robot to guarantee that the obstacle can be detected by the LRF. Considering the measurement range of the LRF, the maximum distance from any point to any boundary should be $<80 \text{m}$.

The robot was programmed with the proposed algorithm and put into the environment. When the robot started running, its mode R1 was activated and it moved along the initial circle firstly (see {\bf A1}). When its heading coincided with a tangent line $\xi$ (See Fig. \ref{c5_H1}), its mode was transitioned to R2 from R1 or R3 (see {\bf A2} and {\bf A3}). When the robot reached the endpoint of the line segment $\xi$, its mode was transitioned back to R3 (see {\bf A4}) and it started tracking $\partial D_i(d_0)$ or $\partial {\cal A}(d_0)$. During the period of the safe navigation, the robot was continuously updating the map and checked whether the map building is completed (see Definition \ref{c5_De2.2}). Finally, the robot stopped when the complete map was built.

\begin{remark}
The saturation function (\ref{c5_eq:sat}) is used to replace the sign function to reduce control chattering, which is undesired in experiments with a real robot \cite{Teimoori2010}.
\end{remark}

In the following parts, five experiments are carried out with different probability $q_0$ in the preset environment to confirm the performance of the proposed algorithm. The parameters used in these experiments are listed in Table. {\ref{c5_H5}}. The probability $q_0$ for the five experiments are $0.3$, $0.4$, $0.5$, $0.6$ and $0.7$.

\begin{table}[!htb]
\caption{Parameters for Experiment 1}
\label{c5_H5}
\begin{center}
\begin{tabular}{c|c|c}
\hline
Speed of robot & $v$ & $0.15 \text{m/s}$\\
\hline
Maximum angular velocity & $u_{M}$ & $0.4 \text{rad/s}$\\
\hline
Minimum turning radius & $R_{min}$ & $0.375 \text{m}$\\
\hline
Safety margin & $d_0$ & $0.55 \text{m}$\\
\hline
Sampling interval & $T$ & $0.2 \text{s}$\\
\hline
Angle threshold & $\theta_{trig}$ & $0.1 \text{rad}$\\
\hline
Distance threshold & $d_{trig}$ & $0.1 \text{m}$\\
\hline
Resolution of the map & & $3 \text{cm}^2/cell$\\
\hline
\end{tabular}
\end{center}
\end{table}

\begin{figure}[thpb]
\centering
\subfigure[$t=10 \text{s}$]{
\epsfig{figure=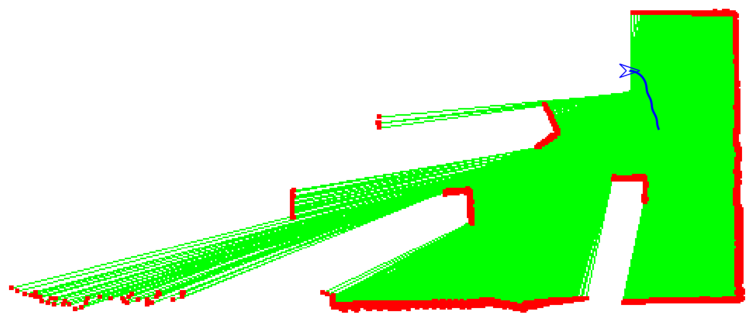,width=7cm}
}
\subfigure[$t=40 \text{s}$]{
\epsfig{figure=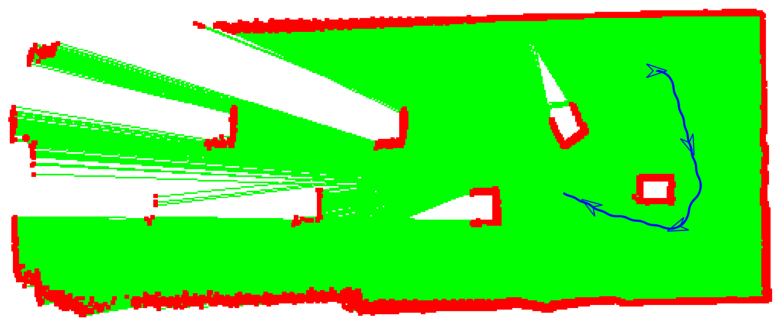,width=7cm}
}
\subfigure[$t=100 \text{s}$]{
\epsfig{figure=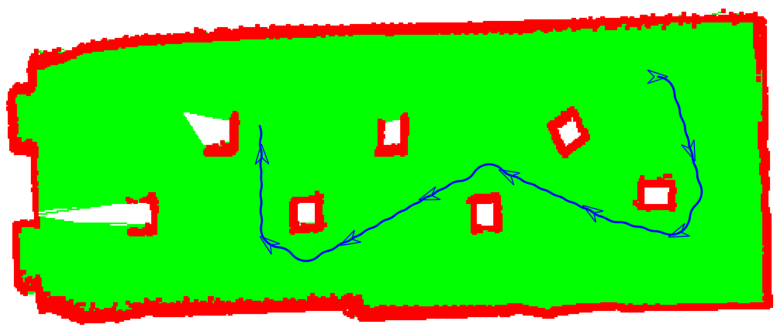,width=7cm}
}
\subfigure[$t=174.2 \text{s}$]{
\epsfig{figure=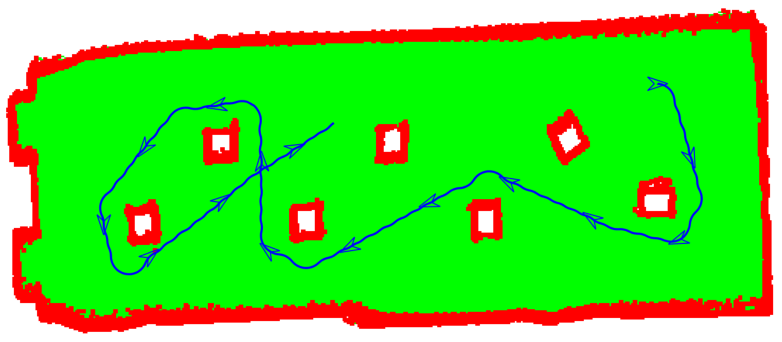,width=7cm}
}
\caption{
Generated maps at different time with probability $q_0=0.6$.
}
\label{c5_E2}
\end{figure}

\begin{figure}[thpb]
\centering
\subfigure[$t=51 \text{s}$]{
\epsfig{figure=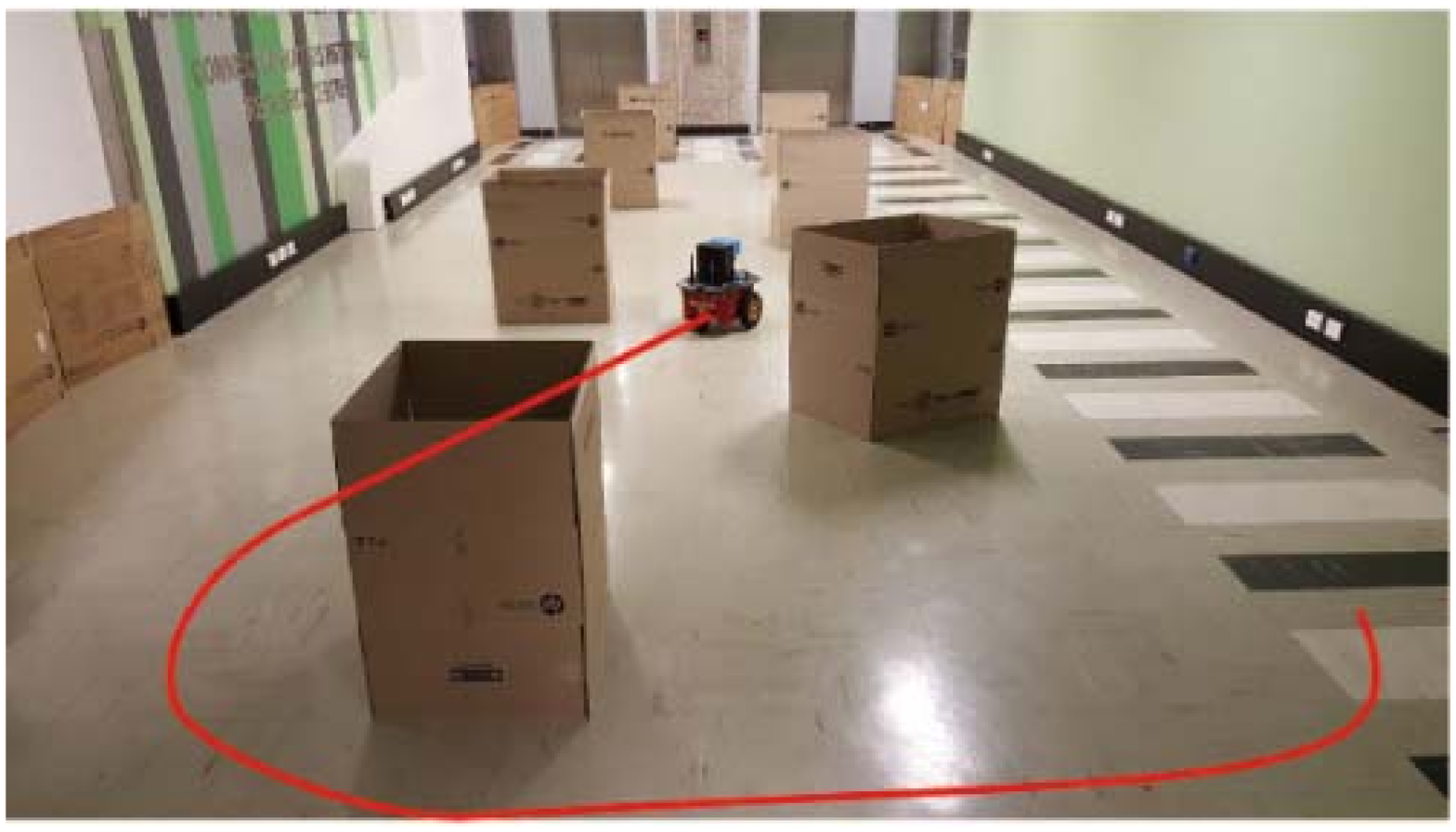,width=7cm}
}
\subfigure[$t=97 \text{s}$]{
\epsfig{figure=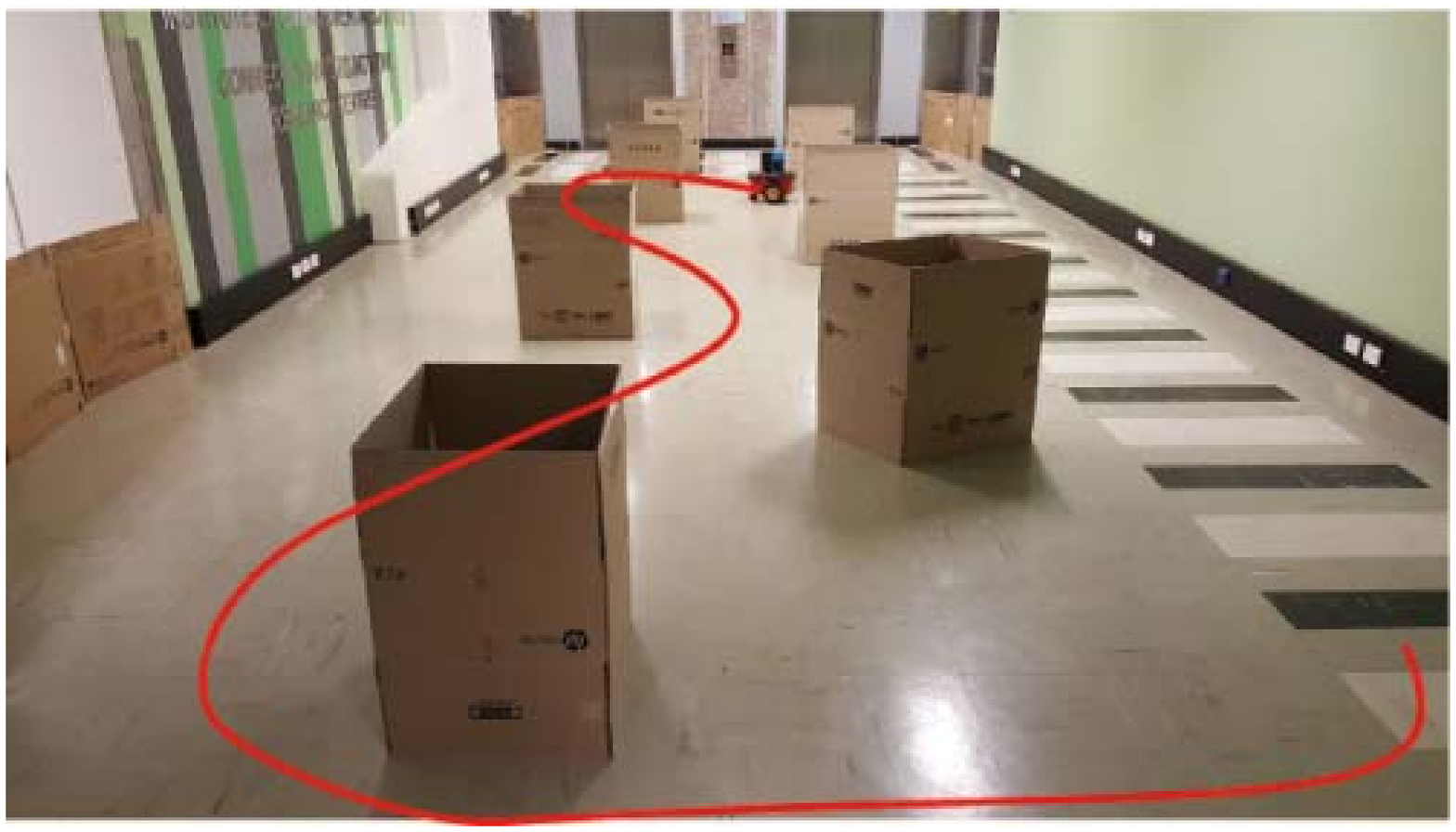,width=7cm}
}
\subfigure[$t=143 \text{s}$]{
\epsfig{figure=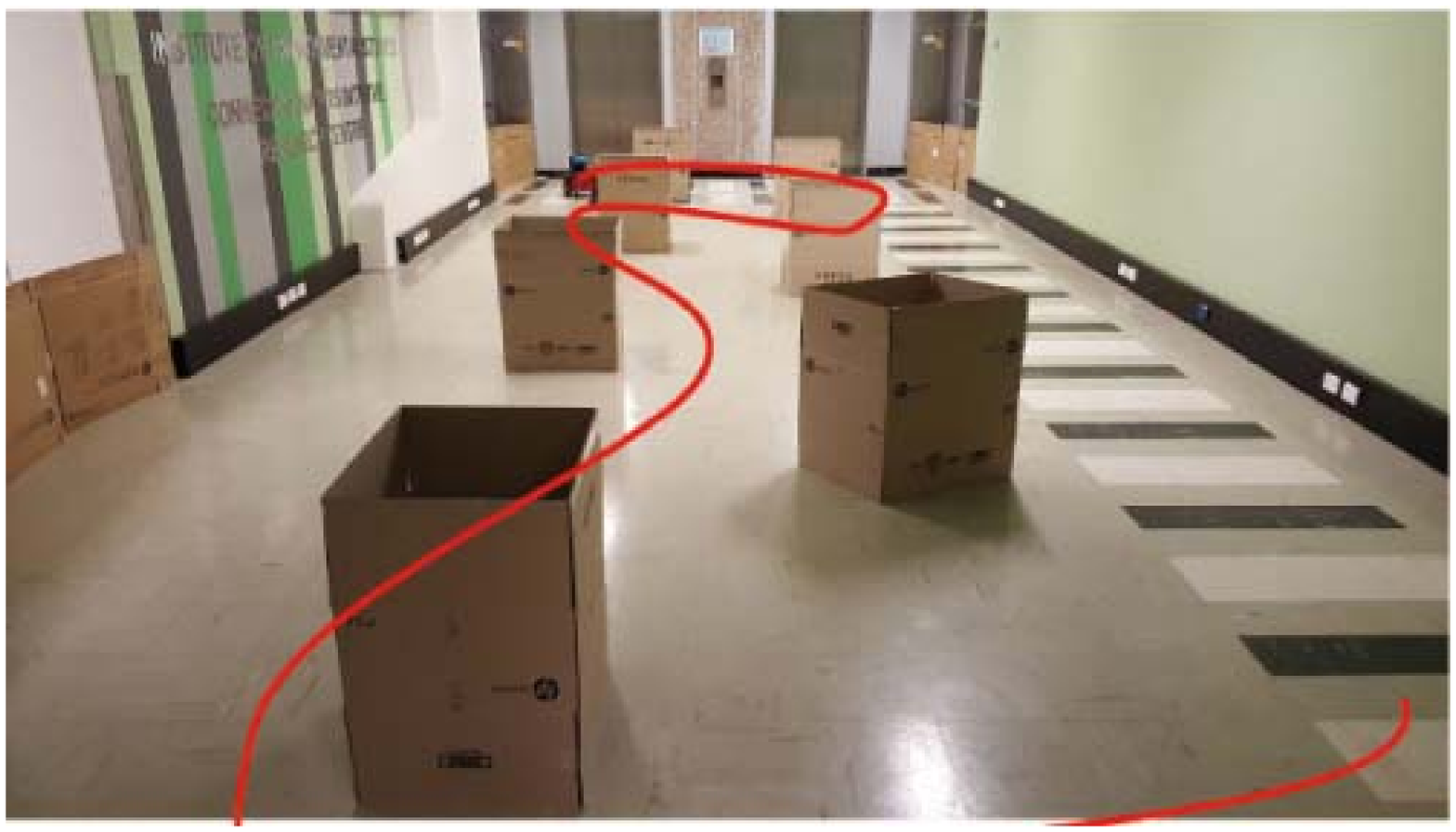,width=7cm}
}
\subfigure[$t=174.2 \text{s}$]{
\epsfig{figure=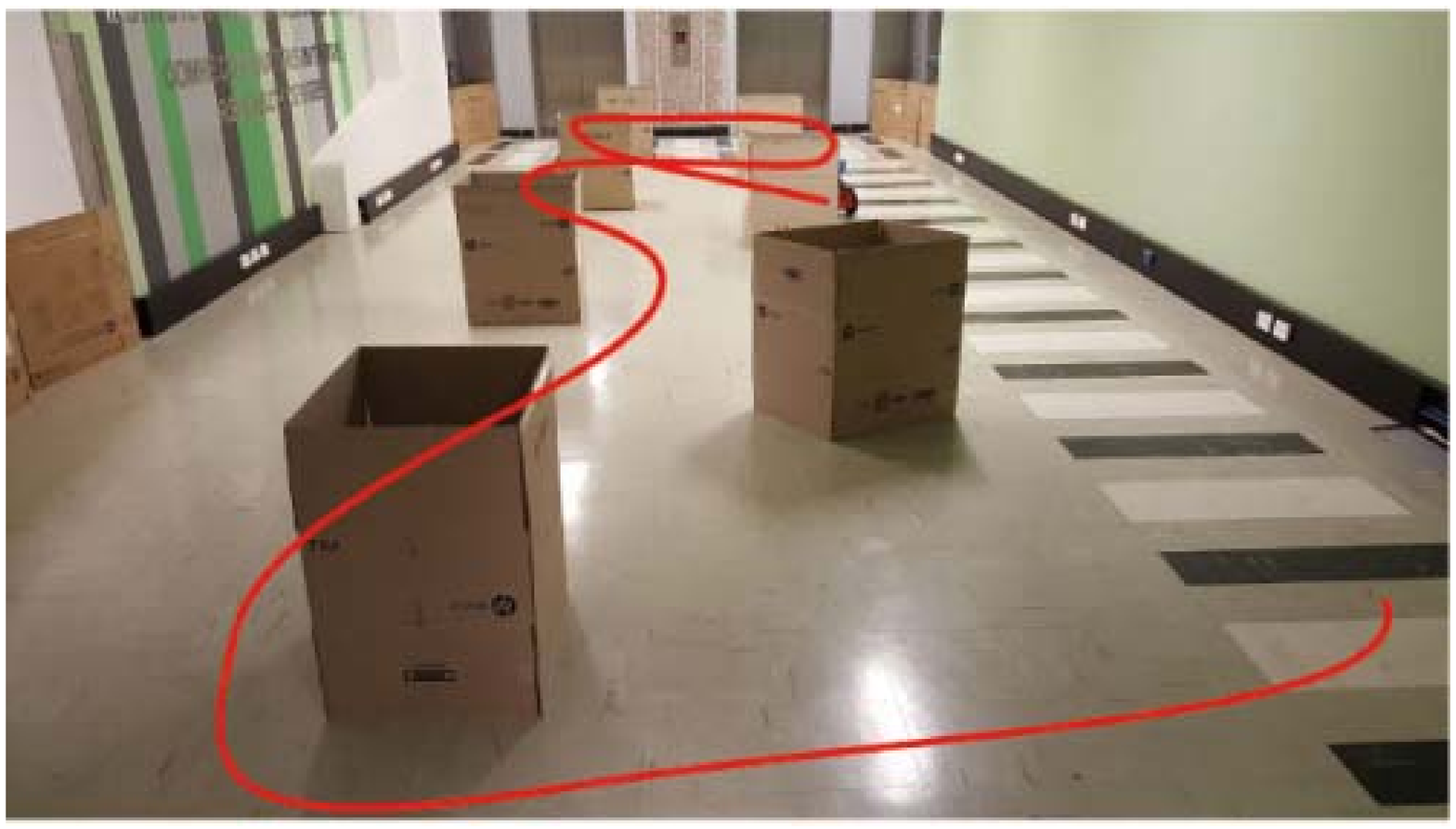,width=7cm}
}
\caption{
Experiment pictures at different time with probability $q_0=0.6$.
}
\label{c5_E2P}
\end{figure}

At the beginning, the robot selected one of two initial circles randomly to move. When the robot's heading coincided with a tangent segment, the robot left the initial circle and reached the enlarged boundary by pursuing this tangent segment. When reached the enlarged boundary, the robot started patrolling this boundary and randomly selected the next tangent segment to pursue. Finally the robot explored the closed area completely and built a correct complete map of this environment.
This procedure is indicated in Fig. \ref{c5_E2} and \ref{c5_E2P}. Fig. \ref{c5_E2} shows the generated maps and the robot's trajectory at different time in the experiment with probability $q_0=0.6$. Fig. \ref{c5_E2P} shows some pictures of the corresponding experiment.

\begin{figure}[thpb]
\centering
\subfigure[$q_0=0.3$ and $t_f=313.6 \text{s}$]{
\label{c5_E1-1}
\epsfig{figure=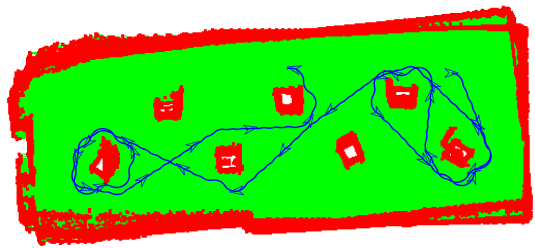,width=7cm}
}
\subfigure[$q_0=0.4$ and $t_f=298.6 \text{s}$]{
\epsfig{figure=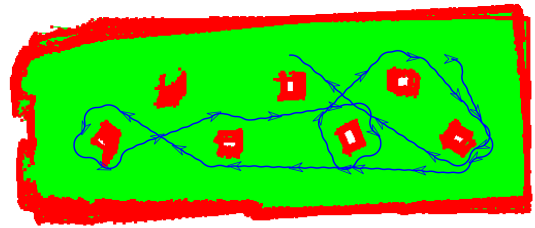,width=7cm}
}
\subfigure[$q_0=0.5$ and $t_f=336.2 \text{s}$]{
\epsfig{figure=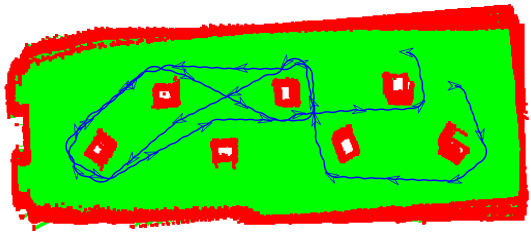,width=7cm}
}
\subfigure[$q_0=0.6$ and $t_f=174.2 \text{s}$]{
\epsfig{figure=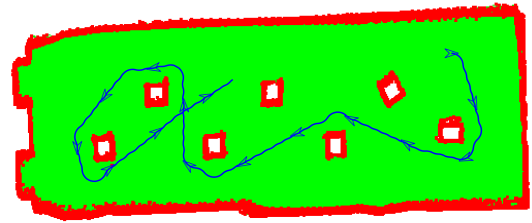,width=7cm}
}
\subfigure[$q_0=0.7$ and $t_f=130.2 \text{s}$]{
\epsfig{figure=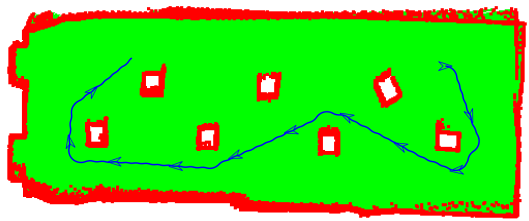,width=7cm}
}
\caption{
Generated maps with different probability $q_0$ and different map building completing time $t_f$.
}
\label{c5_E1}
\end{figure}

The results of the five experiments including the generated complete maps, map building completing time $t_f$, the robot trajectories and corresponding probability $q_0$ are indicated in Fig. \ref{c5_E1}. It can be seen that, in the five experiments, the robot successfully avoided all the obstacles and built the complete maps of the closed area in a finite time. Moreover, according to the comparison of the experimental results, it can be seen that the map building completing time $t_f$ and the length of the robot trajectory are influenced by different values of the parameter  $q_0$. Furthermore, it is easy to see that the accuracy of the generated map is influenced by the length of the robot trajectory because a longer robot trajectory involves bigger odometry errors and reduces the accuracy of the  map (e.g. Fig. \ref{c5_E1-1}).

\section{Summary}

In this chapter, a safe area search and map building algorithm for a mobile ground robot in a closed 
unknown 2D environment with obstacles was proposed. The obstacles are not required to be convex. The developed algorithm belongs to the class of randomized algorithms. It was proved that with probability 1 the robot finishes searching the area and builds a complete map of the environment in a finite time. Moreover, the odometry errors, range finder noise and dynamic environments are considered and discussed in the computer simulations. The computer simulations and experiments with a real Pioneer-3DX robot confirmed the expected performance of the proposed navigation algorithm. 

The proposed method has a number of  advantages. First, the area search and safe navigation were combined together in our method. Second, the proposed algorithm is relatively simple and computationally efficient. Moreover, unlike many other publications in the area, a non-holonomic robot model was considered in the proposed  method. Finally, the performance of the proposed algorithm  is proved by mathematically rigorous  analysis. The usefulness of the method in real applications is based on the fact that the proposed algorithm is computationally simple and easy to implement. Moreover, the proposed mathematically rigorous theoretical framework describes  cluttered environments in which a good performance of the algorithm will be guaranteed. 

While the simulations and experiments have confirmed the performance and advantages of the proposed algorithm, there are still some limitations on our method. In particular, in the proposed method, only the range finder sensor is considered. In the future work, the proposed methodology will be generalized and extended to different types of sensors. 
Moreover, the algorithm will be extended to 3D environments.  An interesting direction of our future research will be to find a method to select a better value of the parameter $q_0$ in the step {\bf A3} of the algorithm  depending on the environment to be explored.
Implementation of the proposed method in real applications will be another important direction of our future research. Potential real applications include 
environment map building in planetary development,
seabed map building in oceans, and 
surveillance and patrolling in sensitive and contaminated regions.
%%%%%%%%%%%%%%%%%%%%%%%%%%%%%%%%%%%%%%%%%%%%%%%%
\chapter{Safe Area Search and 3D Map Building of a Ground Mobile Robot in Indoor Environments}
\chaptermark{3D Map Building}
\label{Chapter6}

This chapter is based on the the publications \cite{Hang_uj_1} and \cite{Hang_robio2017_1}. In this chapter, we consider a non-holonomic ground mobile robot equipped with two 2D range finder sensors. One range finder sensor is used to detect the obstacles on the ground plane and another range finder sensor is used to capture and scan the 3D structure of an indoor environment above the ground plane. In our method, we propose a probabilistic navigation algorithm that performs a safe area search and exploration in the unknown indoor environment and builds a 3D map of the environment in a finite time. The 3D map can be presented by an octree and indicates the occupied area and unoccupied area in the indoor environment. The control law we present in this method consists of switching between different control strategies with different conditions to construct a hybrid system; see \cite{Matveev2000Estimation,Savkin2002Hybrid,
SAVKIN199969Robust,SKAFIDAS1999553Stability}.

\section{Problem description}
\begin{figure}[!htb]
\centering
\epsfig{figure=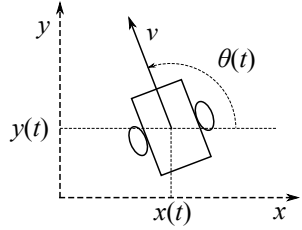,width=4cm}
\caption{
Unicycle model of the robot.
}
\label{c6_F:1}
\end{figure}

A mobile robot is modelled as a Dubins car \cite{Dubins1957} with a non-holonomic constraint in a planar environment, which is the flat ground of a 3D indoor environment. It is widely used to describe many ground robots, unmanned aerial vehicles and missile etc.
\cite{Manchester2006,
Matveev2011b,
Savkin2016a}.
The robot travels with a constant speed $v$ and is controlled by angular velocity $u$. The model of the vehicle is described as follows (see Fig. \ref{c6_F:1}):

\begin{equation}
\label{c6_E:1}
\left\{
\begin{array}{l}
\dot{x}(t) = v \cos \theta (t)
\\
\dot{y}(t) = v \sin \theta (t)
\\
\dot{\theta}(t) = u(t) \in [-u_M, u_M]
\end{array},
\begin{array}{l}
x(0) = x_{0}
\\
y(0) = y_{0}
\\
\theta(0) = \theta_{0}
\end{array}.
\right.
\end{equation}
In the robot model (\ref{c6_E:1}), $(x(t),y(t))$ is the Cartesian coordinates of the vehicle and $\theta(t)$ is the robot's heading at time $t$. The angular velocity $u(t)$ is control input which satisfies the following non-holonomic constraint:

\begin{equation}
\vert u(t) \vert\leq u_M.
\end{equation}
This implies that the robot's minimum turning radius is

\begin{equation}
\label{c6_E:2}
R_{\min}=\frac{v_r}{u_M}.
\end{equation}

\begin{figure}[!htb]
\centering
\epsfig{figure=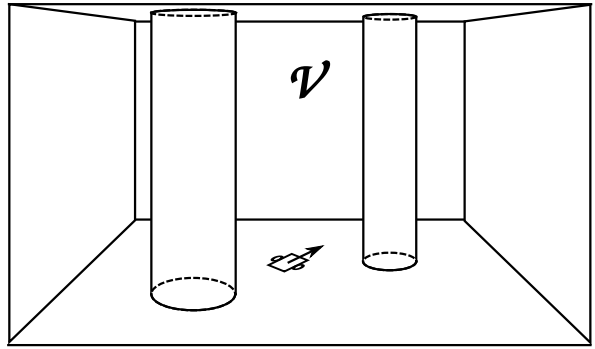,width=8cm}
\caption{
Closed indoor environment ${\cal V}$.
}
\label{c6_F:5}
\end{figure}

\begin{figure}[!htb]
\centering
\epsfig{figure=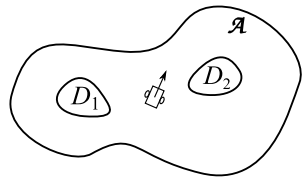,width=4cm}
\caption{
Planar area ${\cal A}$ with obstacles.
}
\label{c6_F:2}
\end{figure}

The wheeled robot travels on the flat ground of an unknown indoor environment ${\cal V}\subset \mathbb{R}^3$ (see Fig. \ref{c6_F:5}). The boundary $\partial {\cal V}$ of the set ${\cal V}$ consists of the ground and the boundaries of several obstacles, like furnitures and walls. The flat ground can be modelled as a closed and bounded planar unknown area ${\cal A}$ with several disjoint planar obstacles $D_1,\ldots,D_k$ (see Fig. \ref{c6_F:2}). Let $d_s>0$ be a given safety margin. The requirement is to drive the ground mobile robot through the collision-free part of the area ${\cal A}$ while keeping the safety margin $d_s$ from both the boundaries of area ${\cal A}$ and obstacles $D_1,\ldots,D_k$ on the plane of the ground.

\begin{definition}
Area ${\cal V}$ is a closed, connected and bounded point set in $\mathbb{R}^3$.
\end{definition}

\begin{remark}
The boundary $\partial{\cal V}$ of ${\cal V}$ is a closed surface that any point on $\partial{\cal V}$ belongs to an obstacle or the ground. Other points inside ${\cal V}$ are unoccupied points.
\end{remark}

\begin{definition}
The area ${\cal A}$ and obstacle $D_i$ for any $i$ are closed, connected and bounded point sets in $\mathbb{R}^2$.
\end{definition}

\begin{figure}[!htb]
\centering
\epsfig{figure=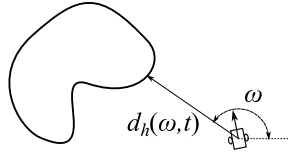,width=4cm}
\caption{
Distance $d_h(\omega,t)$ measurement by the first range finder sensor \textbf{S1}.
}
\label{c6_F:3}
\end{figure}

\begin{figure}[!htb]
\centering
\epsfig{figure=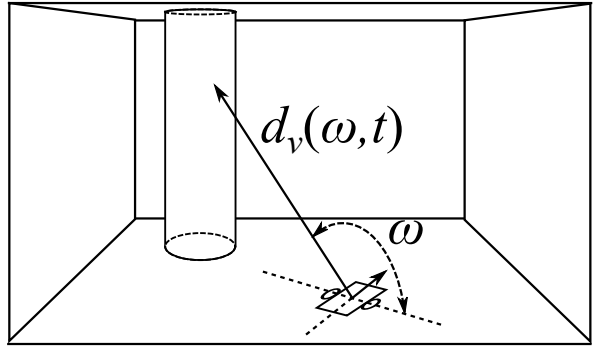,width=8cm}
\caption{
Distance $d_v(\omega,t)$ measurement by the second range finder sensor \textbf{S2}.
}
\label{c6_F:4}
\end{figure}

{\bf Available Measurements}: The robot is equipped with two 2D range finder sensors. The first range finder sensor \textbf{S1} has a scanning range of $360^\circ$ and is mounted on the robot's centre with the scanning plane parallel to the flat ground. At any time $t\geq 0$, it measures the distance to the nearest object in all directions on the ground; i.e. for any angle $0\leq\omega\leq 2\pi$, the sensor measures the distance $d_h(\omega,t)$ to the nearest boundary of an obstacle or the area ${\cal A}$ in the direction given by $\omega$ on the ground (see Fig. \ref{c6_F:3}). The second range finder sensor \textbf{S2} has a scanning range of $180^\circ$ and is mounted on the robot's centre with the scanning plane orthogonal to the flat ground and the robot's heading. At any time $t\geq 0$, it measures the distance to the nearest object on the sensor's scanning plane above the ground; i.e. for any angle $0\leq\omega\leq \pi$ on the scanning plane, the sensor measures the distance $d_v(\omega,t)$ to the nearest boundary of an obstacle in the direction given by $\omega$ above the ground (see Fig. \ref{c6_F:4}). Furthermore, the robot is equipped with an odometry type sensor which measures the robot's location and heading relative to the initial location and heading.

The objective of the robot is to travel through the collision-free part of the unknown area ${\cal A}$ and build a complete 3D map of the unknown indoor environment ${\cal V}$.

\begin{definition}
Let ${\cal M}$ be a point set in $\mathbb{R}^3$ and ${\cal F}$ be a binary map ${\cal M}\rightarrow\lbrace 0,1 \rbrace$. The pair $({\cal M},{\cal F})$ is said to be a complete map of the area ${\cal V}$ if ${\cal M}$ and ${\cal V}$ are congruent, i.e. one can be transformed into the other by an isometry, and for any point $p\in{\cal M}$, ${\cal F}(p)=1$ if and only if $p$ corresponds to a point of the boundary $\partial{\cal V}$ under such isometry.
\end{definition}

The complete map $({\cal M},{\cal F})$ indicates the unoccupied parts and the boundary of the closed indoor environment. To build the complete map, we are here to propose the following map building algorithm, which is a modification of a map building algorithm in 2D planar environment proposed in \cite{Savkin2016}.

\begin{figure}[!htb]
\centering
\epsfig{figure=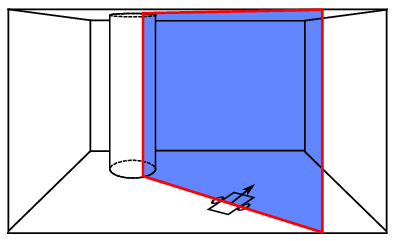,width=8cm}
\caption{
Instant map $({\cal M}_0(t),{\cal F}_0(t))$.
}
\label{c6_F:7}
\end{figure}

\begin{enumerate}
\item[\textbf{A1:}]
At any time $t\geq 0$, the robot builds a planar set ${\cal M}_0(t)$ and a binary map ${\cal F}_0(t)$ as follows: the set ${\cal M}_0(t)$ is formed by all the points $p$ that can be seen by the range finder sensor \textbf{S2} and ${\cal F}_0(t)(p)=1$ if $p$ belongs to either obstacle's boundary or the ground; otherwise, ${\cal F}_0(t)(p)=0$ (see Fig. \ref{c6_F:7}). The pair $({\cal M}_0(t),{\cal F}_0(t))$ is called instant map at time $t$.
\item[\textbf{A2:}]
At any time $t\geq 0$, the robot builds a set ${\cal M}(t)\subset\mathbb{R}^3$ and a binary map ${\cal F}(t)$ by fusing all the instant maps $( {\cal M}_0(\tau),{\cal F}_0(\tau) )$ for $\tau\in [0,t]$ with the help of odometry sensor. The pair $( {\cal M}(t),{\cal F}(t) )$ is called the total map at time $t$.
\end{enumerate}

Now, we are here to give some definitions as follows.

\begin{figure}[!htb]
\centering
\epsfig{figure=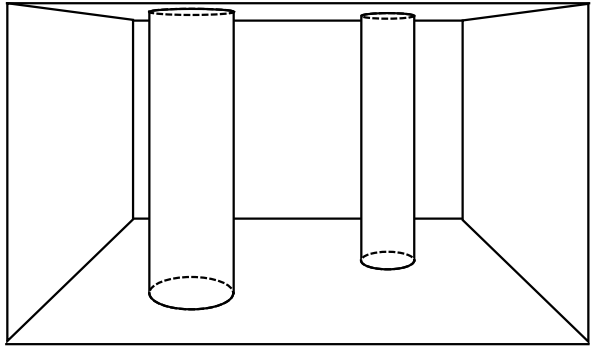,width=8cm}
\caption{
Complete map $({\cal M}(t_f),{\cal F}(t_f))$. For all points $p$ on the boundary of the set ${\cal M}(t_f)$, ${\cal F}(t_f)(p)=1$. For other points in the set ${\cal M}(t_f)$, ${\cal F}(t_f)(p)=0$.
}
\label{c6_F:8}
\end{figure}

\begin{definition}
Consider a robot trajectory $p(t)=(x(t),y(t)),t\in[0,t_f]$ with the map building algorithm \textbf{A1}--\textbf{A2}. The time $t_f>0$ is called a map building completing time if for all the points $p$ on the boundary of the set ${\cal M}(t_f)$, ${\cal F}(t_f)(p)=1$. It means the boundary of the set ${\cal M}(t_f)$ does not include any point $p$ such that ${\cal F}(t_f)(p)=0$ (see Fig. \ref{c6_F:8}).
\end{definition}

If a robot completes the map building at the map building completing time $t_f$, the map $({\cal M}(t_f),{\cal F}(t_f))$ is a complete map showing the whole unoccupied parts of the area ${\cal V}$ and the obstacles.

\begin{definition}
Let $t_f$ be a map building completing time for a robot trajectory $p(t)=(x(t),y(t)),t\in[0,t_f]$. The trajectory is said to be collision-free complete map building if the distance between the robot's position $p(t)$ and the boundary of ${\cal A}$ or $D_i$ is $\geq d_s$ for any $t\in[0,t_f]$.
\end{definition}

\begin{notation}
Let $p$ be an arbitrary point in the area ${\cal A}$. For any closed set $D\subset {\cal A}$, the minimum distance $\rho(D,p)$ between $p$ and $D$ is introduced as 
\begin{equation}
\rho(D,p):=\min_{q\in D}\Vert p-q\Vert.
\end{equation}
\end{notation}

\begin{definition}
For the safety margin $d_s>0$, the $d_s$-enlargement of a closed set $D\subset\mathbb{R}^2$ is a set ${\cal E}[D,d_s]$ defined as follows:

\begin{equation}
{\cal E}[D,d_s]:=\lbrace p\in \mathbb{R}^2 : \rho(D,p)\leq d_s  \rbrace.
\end{equation}
\end{definition}

\begin{definition}
For the safety margin $d_s>0$, the $d_s$-reduction of a closed set $D\subset\mathbb{R}^2$ is a set ${\cal R}[D,d_s]$ defined as follows:

\begin{equation}
{\cal R}[D,d_s]:=\lbrace p\in D : \rho(\partial D,p)\geq d_s  \rbrace,
\end{equation}
where $\partial D$ is the boundary of set $D$.
\end{definition}

\begin{assumption}
\label{c6_A:1}
The planar sets ${\cal R}[{\cal A},d_s]$ and ${\cal E}[D_i,d_s]$, $i=1,\ldots,k$ are closed, bounded, connected and linearly connected sets.
\end{assumption}

\begin{assumption}
\label{c6_A:2}
The sets ${\cal E}[D_i,d_s]$ and ${\cal E}[D_j,d_s]$ do not overlap for any $i\neq j$. Any set ${\cal E}[D_i,d_s]$ is a subset of ${\cal R}[{\cal A},d_s]$.
\end{assumption}

If Assumption \ref{c6_A:2} does not hold, the robot may not be able to reach some parts of the planar area ${\cal A}$ with obstacle avoidance and build the complete map of the area ${\cal V}$.

\begin{notation}
Let $\partial D_i(d_s)$ denote the boundary of the set ${\cal E}[D_i,d_s]$, and $\partial {\cal A}(d_s)$ denote the boundary of the set ${\cal R}[{\cal A},d_s]$.
\end{notation}

\begin{assumption}
\label{c6_A:3}
All boundaries $\partial D_i(d_s)$ and $\partial {\cal A}(d_s)$ are non-self-intersecting, closed smooth curve. The absolute value of the curvature at any point of $\partial D_i(d_s)$ or $\partial {\cal A}(d_s)$ is $\leq \frac{1}{R_{\min}}$.
\end{assumption}

\begin{figure}[!htb]
\centering
\epsfig{figure=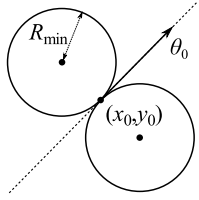,width=3.5cm}
\caption{
The two initial circles.
}
\label{c6_F:10}
\end{figure}

\begin{definition}
The robot has two initial circles at the robot's initial position. The initial circles are tangent to the robot's initial heading $\theta_0$ and cross the robot's initial position $(x_0,y_0)$. The radius of each initial circle is equal to $R_{\min}$ (see Fig. \ref{c6_F:10}).
\end{definition}

\begin{assumption}
\label{c6_A:4}
Both of the two initial circles lie in the set ${\cal R}[{\cal A},d_s]$ and do not intersect with any set ${\cal E}[D_i,d_s]$.
\end{assumption}

\section{Safe navigation algorithm}
Here, we propose a collision-free navigation algorithm for the mobile robot to explore the planar area ${\cal A}$ and build a complete map of area ${\cal V}$. Firstly, several definitions and assumptions are introduced.

\begin{definition}
\label{c6_D:1}
Let $L$ be a straight line segment which lies in the set ${\cal R}[{\cal A},d_s]$ and does not cross any set ${\cal E}[D_i,d_s]$, $\bar{L}$ be the straight line which coincides with $L$. The straight line segment $L$ is said to be a common tangent line segment if it satisfies any one of the following conditions:
\begin{enumerate}
\item
$\bar{L}$ is simultaneously tangent to $\partial D_i(d_s)$ and $\partial D_j(d_s)$, $i\neq j$ at the two distinct end points of $L$.
\item
$\bar{L}$ is tangent to $\partial D_i(d_s)$ at two distinct end points of $L$.
\item
$\bar{L}$ is simultaneously tangent to $\partial D_i(d_s)$ and $\partial {\cal A}(d_s)$ at the two distinct end points of $L$.
\item
$\bar{L}$ is simultaneously tangent to $\partial D_i(d_s)$ and an initial circle at the two distinct end points of $L$.
\item
$\bar{L}$ is simultaneously tangent to $\partial {\cal A}(d_s)$ and an initial circle at the two distinct end points of $L$.
\item
$\bar{L}$ is tangent to $\partial {\cal A}(d_s)$ at two distinct end points of $L$.
\end{enumerate}
\end{definition}

\begin{assumption}
\label{c6_A:5}
Any end point of a common tangent line segment does not belong to other distinct common tangent line segments.
\end{assumption}

%\begin{assumption}
%\label{c6_A:6}
%If the boundary $\partial {\cal A}(d_s)$ is non-convex, then there exists at least one common tangent line segment connecting $\partial {\cal A}(d_s)$ with $\partial D_i(d_s)$ or an initial circle.
%\end{assumption}

Now, we are here to introduce a planar graph ${\cal G}$ as follows.

\begin{figure}[!htb]
\centering
\epsfig{figure=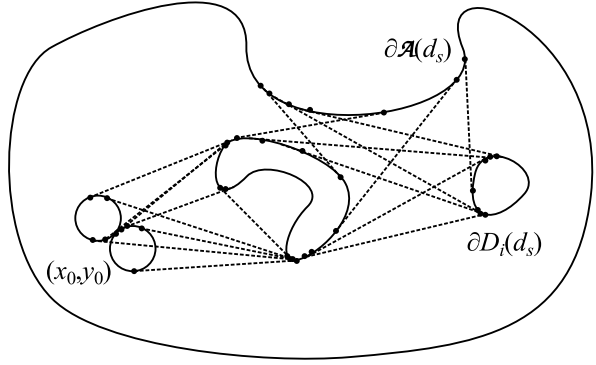,width=8cm}
\caption{
Graph ${\cal G}$ with vertices and edges.
}
\label{c6_F:6}
\end{figure}

\begin{definition}
The graph ${\cal G}$ is defined as follows. The vertices of ${\cal G}$ are the robot's initial position $(x_0,y_0)$ and the end points of all the common tangent line segments. The edges of ${\cal G}$ are all the common tangent line segments, arcs of the initial circles and the segments of all the boundaries $\partial D_i(d_s)$ and $\partial {\cal A}(d_s)$. The edges connects all the vertices (see Fig. \ref{c6_F:6}).
\end{definition}

With the presented graph ${\cal G}$, some definitions and assumptions are proposed as follows.

\begin{figure}[!htb]
\centering
\epsfig{figure=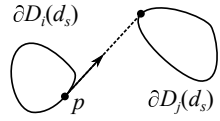,width=3.5cm}
\caption{
Exit direction of the vertex $p$.
}
\label{c6_F:9}
\end{figure}

\begin{definition}
Let $L$ be a common tangent line segment on graph ${\cal G}$. For any end point $p$ of $L$, the direction from $p$ to another end point of $L$ is called the exit direction of the vertex $p$ on graph ${\cal G}$ (see Fig. \ref{c6_F:9}).
\end{definition}

\begin{figure}[!htb]
\centering
\epsfig{figure=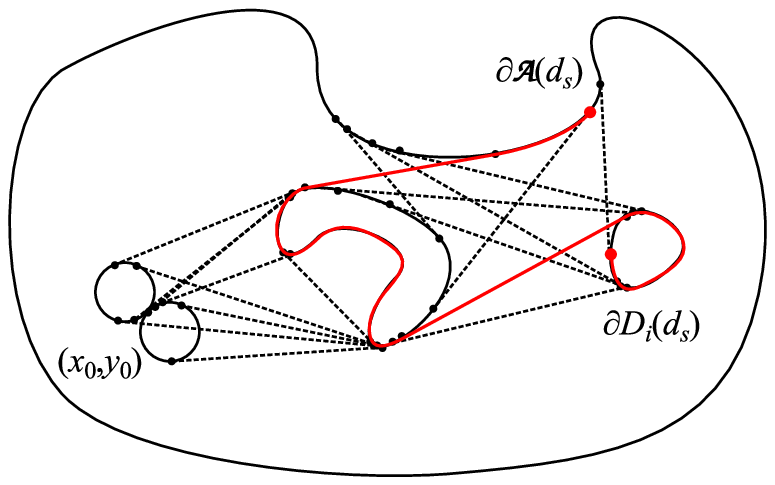,width=8cm}
\caption{
A viable path on graph ${\cal G}$.
}
\label{c6_F:11}
\end{figure}

\begin{definition}
A path on the graph ${\cal G}$ connecting two vertices is said to be viable if the heading at the end of each edge of the path coincides with the heading at the beginning of the next edge of the path (see Fig. \ref{c6_F:11}).
\end{definition}

\begin{assumption}
\label{c6_A:7}
Let $p_1$ and $p_2$ be any two distinct vertices on any boundary $\partial D_i(d_s)$ or $\partial {\cal A}(d_s)$. From the exit direction of $p_1$ or another direction opposite to the exit direction of $p_1$, there exists at least one viable path from $p_1$ and $p_2$ on the graph ${\cal G}$.
\end{assumption}

\begin{definition}
A vertex on the boundary $\partial D_i(d_s)$ or $\partial {\cal A}(d_s)$ is said to be non-circular if another vertex on the corresponding common tangent line segment does not belong to the initial circles. Otherwise, the vertex is said to be circular.
\end{definition}

\begin{figure}[!htb]
\centering
\epsfig{figure=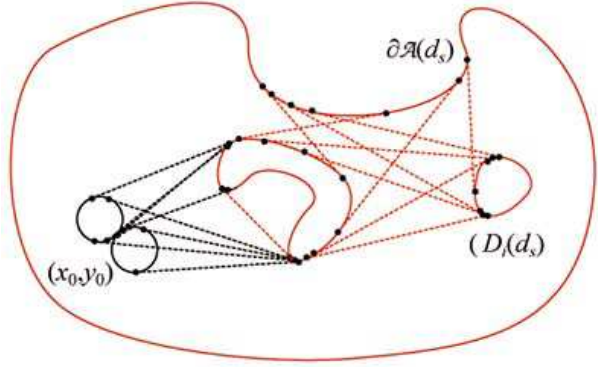,width=8cm}
\caption{
Map building edges on graph ${\cal G}$.
}
\label{c6_F:12}
\end{figure}

\begin{definition}
If the boundary $\partial {\cal A}(d_s)$ is convex, the edges $\partial D_i(d_s)$ and the common tangent line segments which do not include circular vertex are called map building edges. If $\partial {\cal A}(d_s)$ is non-convex, $\partial {\cal A}(d_s)$ is also called map building edge. See Fig. \ref{c6_F:12}.
\end{definition}

For the case of simplicity, the following assumption is introduced.

\begin{assumption}
\label{c6_A:8}
On the graph ${\cal G}$, the set of map building edges is not empty.
\end{assumption}

\begin{assumption}
\label{c6_A:9}
For any point $p\in\partial {\cal V}$, there exists at least a point $q$ on map building edges of the graph ${\cal G}$ such that the straight line segment between $p$ and $q$ belongs to the set ${\cal V}$, does not cross the boundary $\partial {\cal V}$ and is orthogonal to the edge on point $q$.
\end{assumption}

Assumption \ref{c6_A:8} and \ref{c6_A:9} guarantees that any point on the boundary $\partial {\cal V}$ of the environment ${\cal V}$ can be detected by the sensor \textbf{S2} when the robot searches all the map building edges on the graph ${\cal G}$.

Let $q_c\in (0,1)$ be a given probability. To explore the ground area ${\cal A}$ and build the complete map of the indoor environment ${\cal V}$, a probabilistic navigation algorithm is introduced as follows:

\begin{enumerate}
\item[\textbf{B1:}]
The robot starts to move along one of two initial circles.

\item[\textbf{B2:}]
When the robot moving along an initial circle reaches a vertex, which is not the robot's initial position, with the heading direction coinciding with the exit direction of the vertex, the robot starts to move along the corresponding common tangent line segment.

\item[\textbf{B3:}]
When the robot moving along a common tangent line segment reaches a vertex, the robot starts to move along the corresponding boundary $\partial D_i(d_s)$ or $\partial {\cal A}(d_s)$ which the vertex belongs to.

\item[\textbf{B4:}]
When the robot moving along a boundary $\partial D_i(d_s)$ or $\partial {\cal A}(d_s)$ reaches a non-circular vertex with the heading direction coinciding with the exit direction of the vertex, with the probability $q_c$ the robot starts to move along the corresponding common tangent line segment and with the probability $(1-q_c)$ the robot continues to move along the boundary.
\end{enumerate}

%According to the theoretical result in \cite{Savkin2016} corresponding to the navigation algorithm \textbf{B1}--\textbf{B4}, a lemma is proposed here, which was proved in \cite{Savkin2016}.
%
%\begin{lemma}
%\label{L:1}
%Suppose that Assumptions \ref{c6_A:1}--\ref{c6_A:4}, \ref{c6_A:5}--\ref{c6_A:7} hold and the robot is navigated by the algorithm \textbf{B1}--\textbf{B4} with probability $q_c\in (0,1)$. Then for any initial position and heading of the robot, with probability $1$ there exists a time $t_c > 0$ such that the trajectory $p(t)=(x(t),y(t)),t\in [0,t_c]$ of the robot (\ref{c6_E:1}) includes all the points on map building edges.
%\end{lemma}

%According to Lemma \ref{L:1} and Assumption \ref{c6_A:8}--\ref{c6_A:9}, the following theorem can be proposed and proved directly.

Now, we are here to propose the main theoretical result as follows.

\begin{theorem}
\label{c6_T:1}
Suppose that Assumptions \ref{c6_A:1}--\ref{c6_A:4}, \ref{c6_A:5}--\ref{c6_A:9} hold, the robot is navigated by the algorithm \textbf{B1}--\textbf{B4} with probability $q_c\in (0,1)$ and, simultaneously, builds the map by the algorithm \textbf{A1}--\textbf{A2}. Then for any initial position and heading of the robot on the ground of the indoor environment, with probability $1$ there exists a time $t_f > 0$ such that the trajectory $p(t)=(x(t),y(t)),t\in [0,t_f]$ of the robot (\ref{c6_E:1}) is collision-free complete map building.
\end{theorem}

\textbf{Proof of Theorem \ref{c6_T:1}:}
According to Assumption \ref{c6_A:8} and the algorithm \textbf{B1}--\textbf{B4}, the robot starts from its initial position, moves along an initial circle and a common tangent line segment to a vertex belonging to a map building edge. Then the robot travels along the map building edges on the graph ${\cal G}$. According to Assumption \ref{c6_A:5} and \ref{c6_A:7}, with a non-zero probability the robot can reach any vertex $b$ on the map building edges from any other vertex $a$ on the map building edges with any starting direction on vertex $a$. Therefore, it is easy to known that there exist a time $t_f>0$ such that the robot has travelled all the points on any map building edge.

According to Assumption \ref{c6_A:9}, it can be seen that the time $t_f$ is a map building completing time. According to Assumption \ref{c6_A:4}, the robot moves along any initial circle while keeping the given safety margin $d_s$. According to Definition \ref{c6_D:1}, the distance between any point on any common tangent line segment on the graph ${\cal G}$ and any obstacle $D_i$ or the boundary of ${\cal A}$ is $\geq d_s$. According to Assumption \ref{c6_A:1} and \ref{c6_A:2}, it is easy to known that the robot moving along the boundaries $\partial D_i(d_s)$ and $\partial {\cal A}(d_s)$ also keeps the given safety margin $d_s$. Therefore, with the navigation of the algorithm \textbf{B1}--\textbf{B4}, the distance between the robot and any obstacle $D_i$ or the boundary of ${\cal A}$ is $\geq d_s$ at any time. The Assumption \ref{c6_A:3} guarantees the trajectory of the robot satisfies the non-holonomic constraint (\ref{c6_E:2}).

According to the above proof. It can be indicated that the robot's trajectory $p(t)=(x(t),y(t)), t\in [0,t_f]$ of the robot model (\ref{c6_E:1}) is collision-free complete map building. This completes the proof of Theorem \ref{c6_T:1}.

\section{Computer simulations}
We carry out computer simulations to confirm the proposed map building algorithm \textbf{A1}--\textbf{A2} and navigation algorithm \textbf{B1}--\textbf{B4}.

To apply the proposed navigation algorithm \textbf{B1}--\textbf{B4} on the robot, a sliding mode control law which is a modification  of the law of \cite{Savkin2013} is used, that consists of switching between a boundary following approach proposed in \cite{Matveev2011a}, and the pure pursuit navigation approach; see e.g. \cite{Savkin2010}. The control law is described as follows with three separate modes $R1$, $R2$ and $R3$:
\begin{equation}
\label{c6_E:3}
u(t) = \left\{
\begin{array}[h]{cl}
\pm u_M & R1
\\
\text{sgn}\left[ \phi_m(t) \right]u_M & R2
\\
\Gamma \text{sgn}\left[\dot{d}_{m}(t) + H(d_{m}(t)-d_{s} )\right]u_M & R3
\end{array},
\right.
\end{equation}
where the sign function $\text{sgn}(x)$ is defined as follows:

\begin{equation}
\text{sgn}(x) = \left \{  
\begin{array}[h]{cl}1 & x>0
\\
0 & x=0  
\\
-1 & x<0 
\end{array}
\right.
\end{equation}
and the saturation function $H(x)$ is defined as follows with two tunable parameters $l$ and $r$:

\begin{equation} 
H(x) = \left \{  
\begin{array}[h]{cl}
lx & |x| < r
\\ 
lr\cdot\text{sgn}(x) & \textrm{otherwise}
\end{array} .
\right.
\end{equation}
When mode $R1$ is active, the robot is moving along an initial circle. Mode $R1$ is the initial mode. When mode $R2$ is active, the robot is moving along a common tangent line segment. When mode $R3$ is active, the robot is moving along a boundary $\partial{D_i}(d_s)$ or the boundary $\partial{\cal A}(d_s)$. The three modes $R1$--$R3$ corresponds to the above three separate control strategies and transitions to each other by the following mode transition rules:

$R1 \rightarrow R2$:
It occurs when the robot is at a vertex on one of the initial circles and the exit direction at the vertex coincides with the robot's current heading.

$R2 \rightarrow R3$:
It occurs when the robot is at a vertex on a common tangent line segment and reaches the boundary $\partial D_i(d_s)$ or  
$\partial {\cal A}(d_s)$.

$R3 \rightarrow R2$:
With the probability $q_c$, it occurs when the robot is at a vertex on the boundary $\partial D_i(d_s)$ or $\partial {\cal A}(d_s)$ and the exit direction at the tangent point coincides with the robot's current heading. 

In the control law (\ref{c6_E:3}), when mode $R3$ is active, the variable $d_m(t)$ is the minimum distance from the robot to the obstacles $D_1,\ldots,D_k$ and the area boundary $\partial{\cal A}$ on the ground plane. It is measured directly by the range finder sensor \textbf{S1}. When mode $R2$ is active, the variable $\phi_m(t)$ is the angle between the robot's heading $\theta(t)$ and the bearing of the end point which the robot is moving forwards from the robot. It can be calculated with the help of the odometry sensor and the range finder sensor \textbf{S1}. When mode $R3$ is activated at a vertex on boundary $\partial D_i(d_s)$ or $\partial {\cal A}(d_s)$, the variable $\Gamma$ is defined as $+1$ if the boundary is on the left of the robot, $-1$ if it is on the right. It can be calculated with the help of the range finder sensor \textbf{S1}. The robot stops at the map building completing time $t_f$. The control law (\ref{c6_E:3}) consisting of switching constructs a hybrid control system; see \cite{Matveev2000Estimation,Savkin2002Hybrid,
SAVKIN199969Robust,SKAFIDAS1999553Stability}.

\begin{figure}[!htb]
\centering
\epsfig{figure=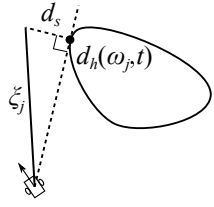,width=3cm}
\caption{
Tangent line segment $\xi_j$.
}
\label{c6_F:13}
\end{figure}

\begin{figure}[!htb]
\centering
\epsfig{figure=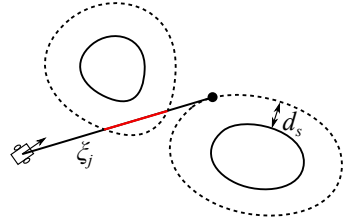,width=4.5cm}
\caption{
Unsafe tangent line segment $\xi_j$.
}
\label{c6_F:14}
\end{figure}

Here, an approach is proposed to determine if the robot's heading coincides with the exit direction of a vertex. Suppose the range finder sensor \textbf{S1} measures distances $d_h(\omega_i,t)$ in finite directions $\omega_i$, $i=0,\pm 1,\ldots,\pm n$, where $\omega_0$ is the robot's heading and the angular intervals between any two successive direction $\omega_i$ and $\omega_{i+1}$ are equal and small. Then the steps of calculating tangent line segment on boundary $\partial D_i(d_s)$ or $\partial {\cal A}(d_s)$ crossing the robot's position are described as follows:

\begin{enumerate}
\item
Search any direction pairs $\omega_i$ and $\omega_{i+1}$ satisfying $\vert (d_h(\omega_i,t)-d_h(\omega_{i+1},t)\vert\geq 2d_s$. Between the pairs $\omega_i$ and $\omega_{i+1}$ satisfying the inequality, denote the direction corresponding to the shorter distance by $\hat{\omega}_j$, $j=1,2,\ldots,m$.
\item
For any $j$, let $d_h(\hat{\omega}_j,t)$ be a leg and construct a right triangle. The length of another leg is $d_s$ sharing the common end point with $d_h(\hat{\omega}_j,t)$ on measurement point. Then the hypotenuse denoted by $\xi_j$ can approximately represent the tangent line segment on the corresponding boundary $\partial D_i(d_s)$ or $\partial {\cal A}(d_s)$ (see Fig. \ref{c6_F:13}). Notice that if there exist any other measurement point $p$ such that the minimum distance between point $p$ and line segment $\xi_j$ is $<d_s$, $\xi_j$ cannot represent the tangent line segment since it cannot be tracked with the safety margin $d_s$ (see Fig. \ref{c6_F:14}).
\end{enumerate}

If there exist a line segment $\xi_j$ such that the angle between direction $\omega_0$ and line segment $\xi_j$ is smaller than a threshold, $\xi_j$ is said to coincide with the robot's heading. When mode transition $R1 \rightarrow R2$ occurs, the end point of the corresponding tangent line segment $\xi_j$ is defined as a intermediate target and is used to calculated the variable $\phi_m(t)$.

In the following part, several computer simulations are carried out to confirm the performance of the proposed algorithms. To avoid the oscillation of the robot's heading involved by sign function $\text{sgn}(x)$, saturation function with upper bound $1$, lower bound $-1$ and a suitable gradient is used to replace the sign function $\text{sgn}(x)$. Moreover, when the robot is moving on mode $R3$ and decides not to activate mode $R2$, there is a following short time interval that the robot will not activate mode $R2$ in any case. It is used to avoid repeatedly checking the coinciding of the robot's heading with a same common tangent line segment. Moreover, the octree can be used to represent the 3D map of the environment.

\begin{figure}[!htb]
\centering
\subfigure[]{
\epsfig{figure=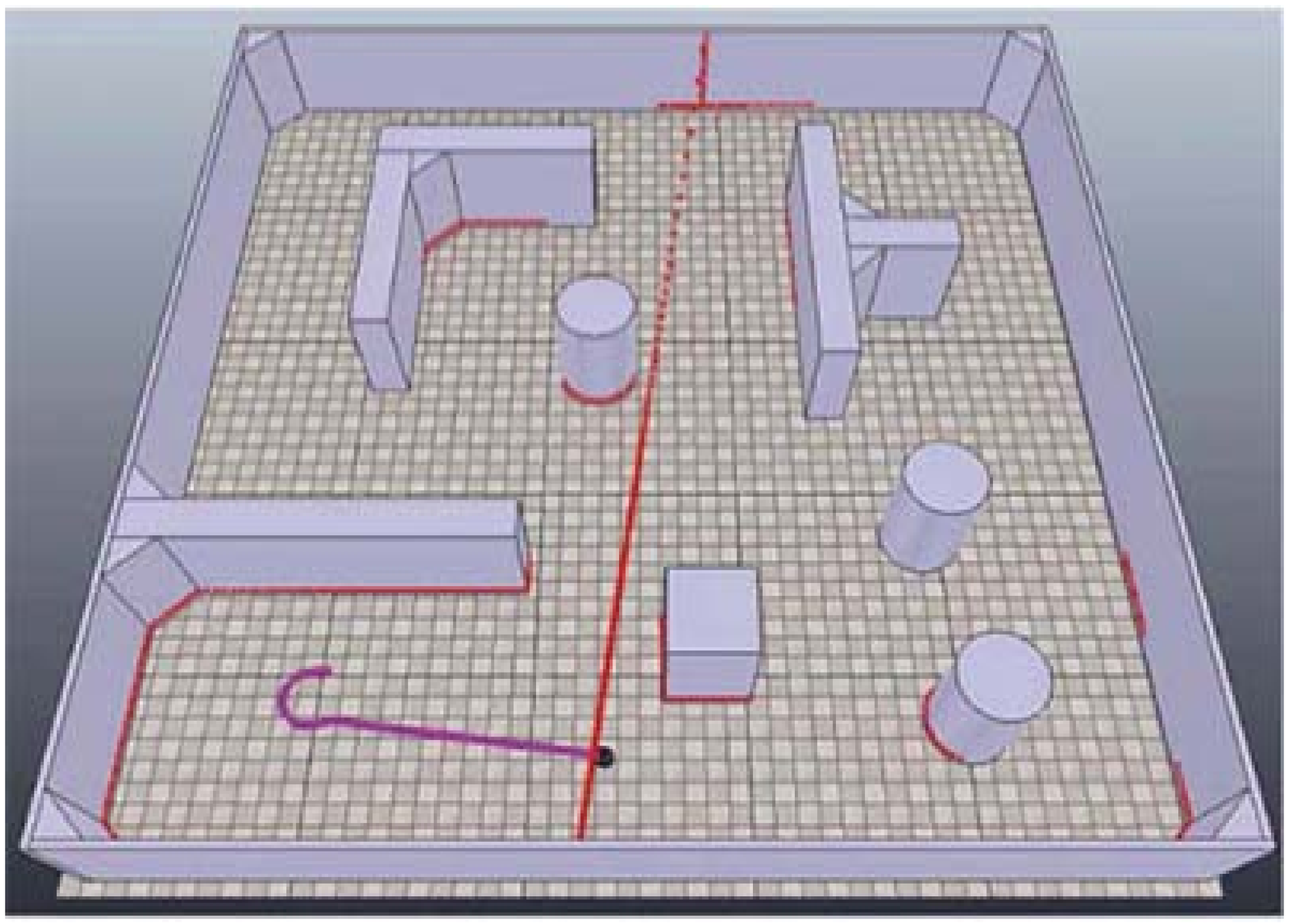,width=6cm}
}
\subfigure[]{
\epsfig{figure=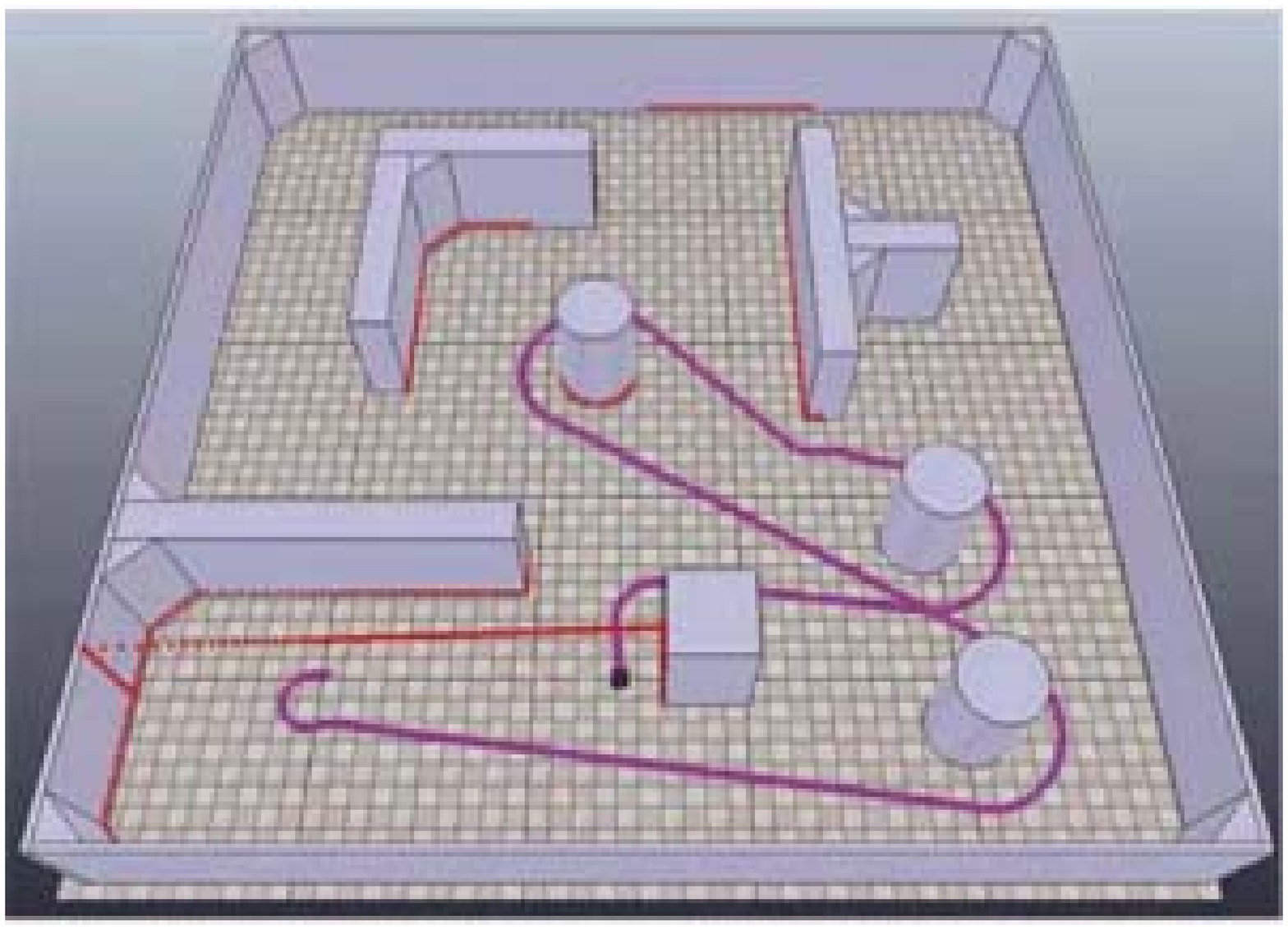,width=6cm}
}
\subfigure[]{
\epsfig{figure=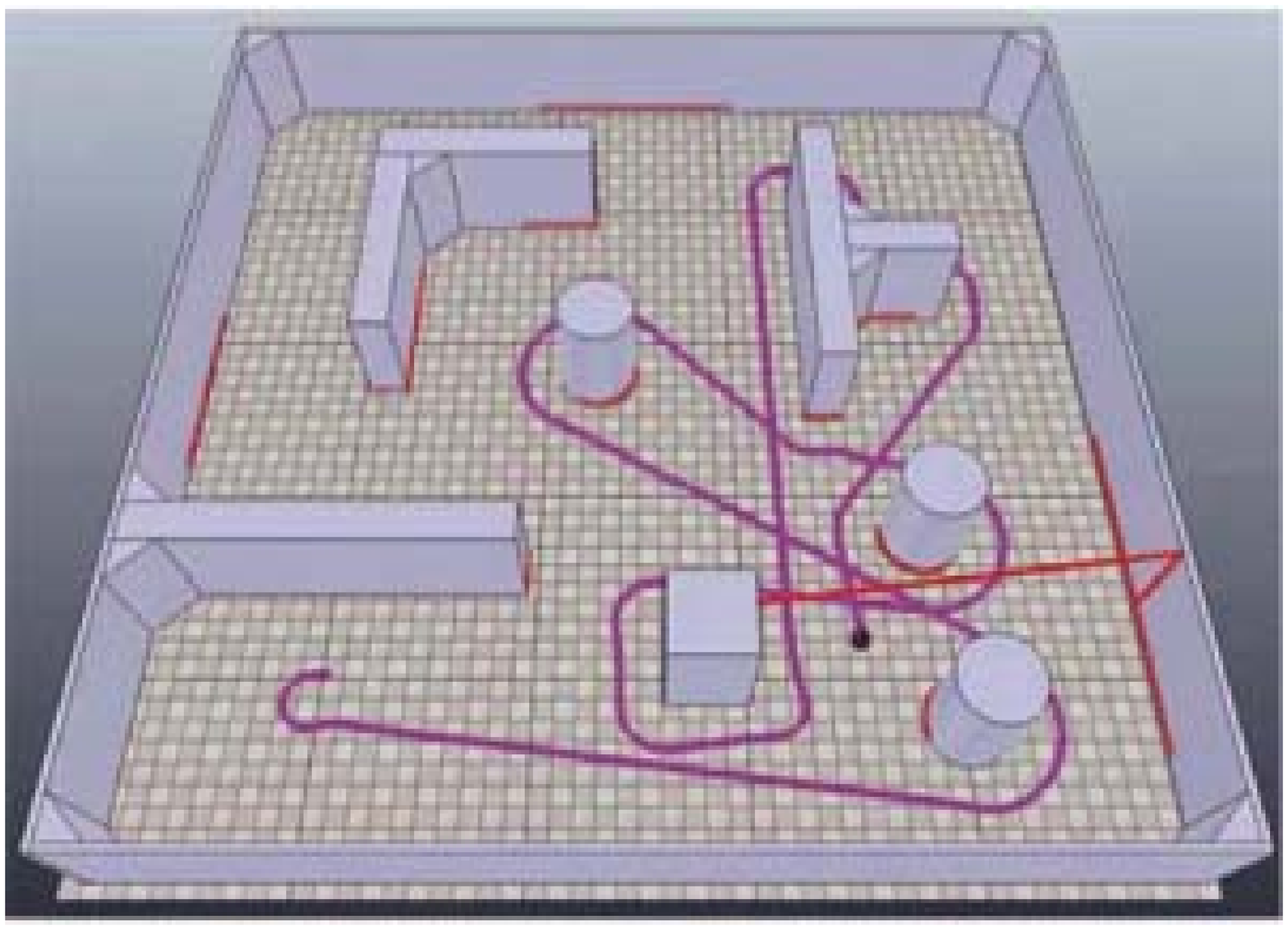,width=6cm}
}
\subfigure[]{
\epsfig{figure=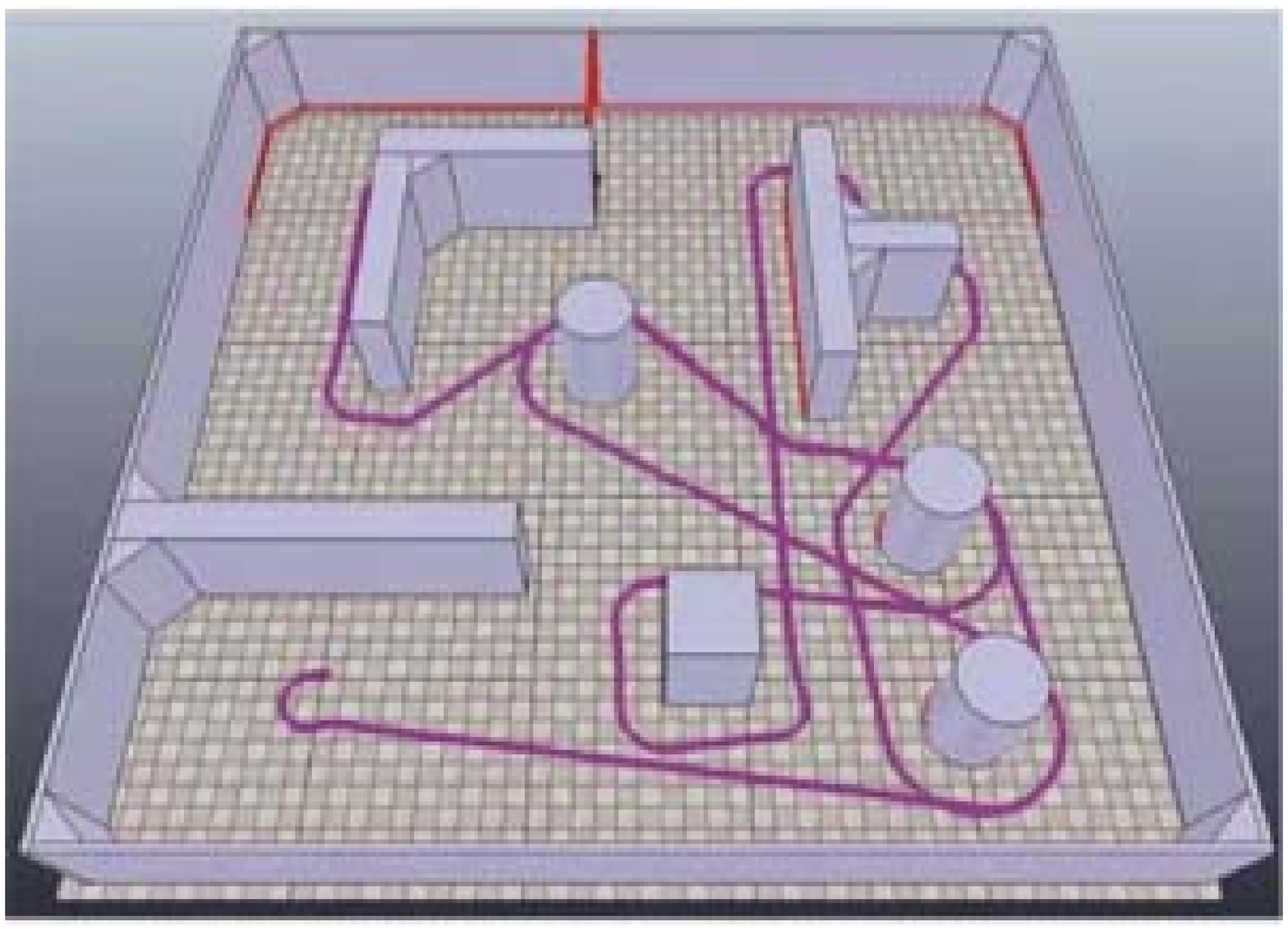,width=6cm}
}
\caption{
Robot trajectory in computer simulation.
}
\label{c6_F:15}
\end{figure}

\begin{figure}[!htb]
\centering
\subfigure[]{
\epsfig{figure=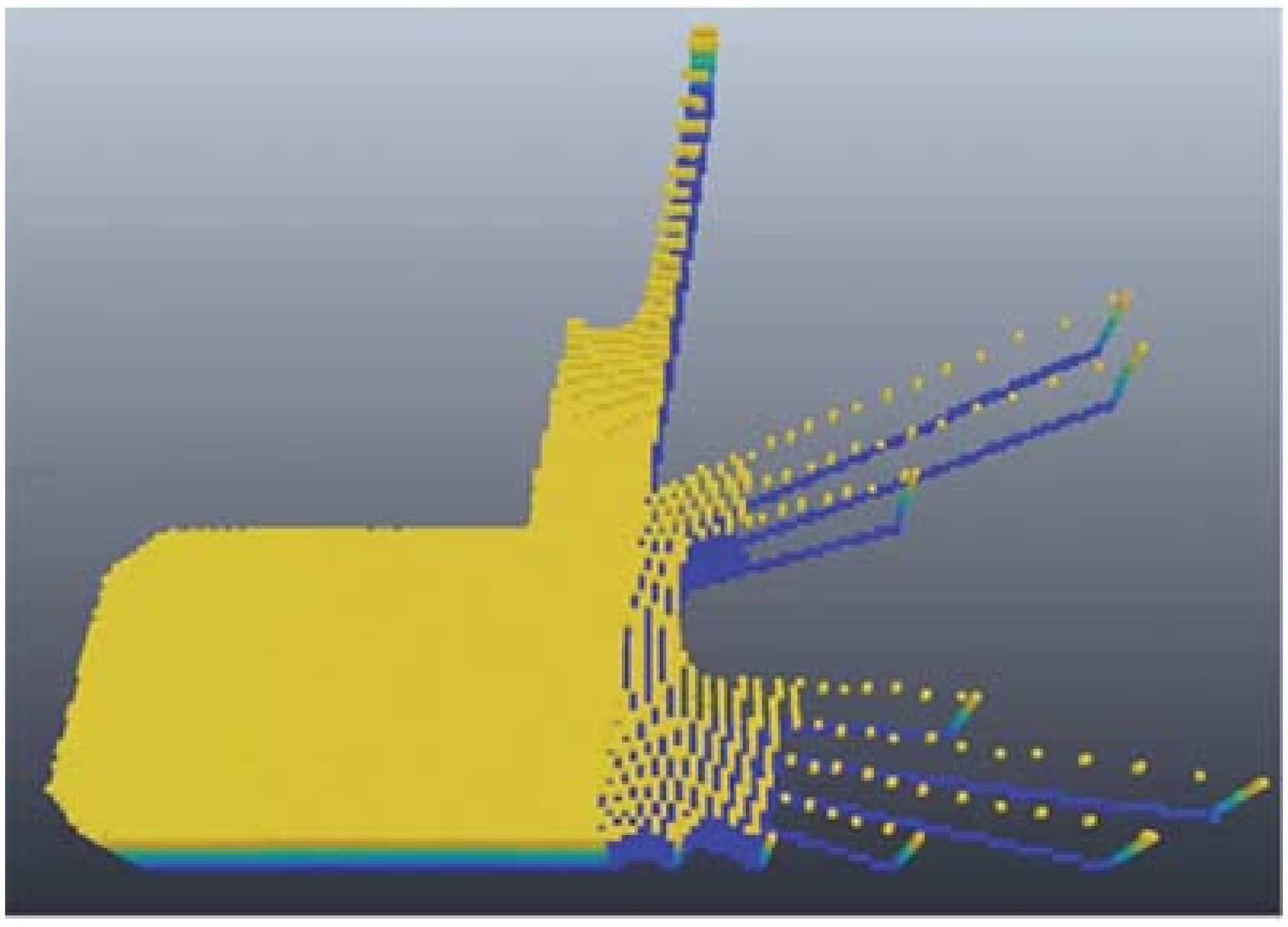,width=6cm}
}
\subfigure[]{
\epsfig{figure=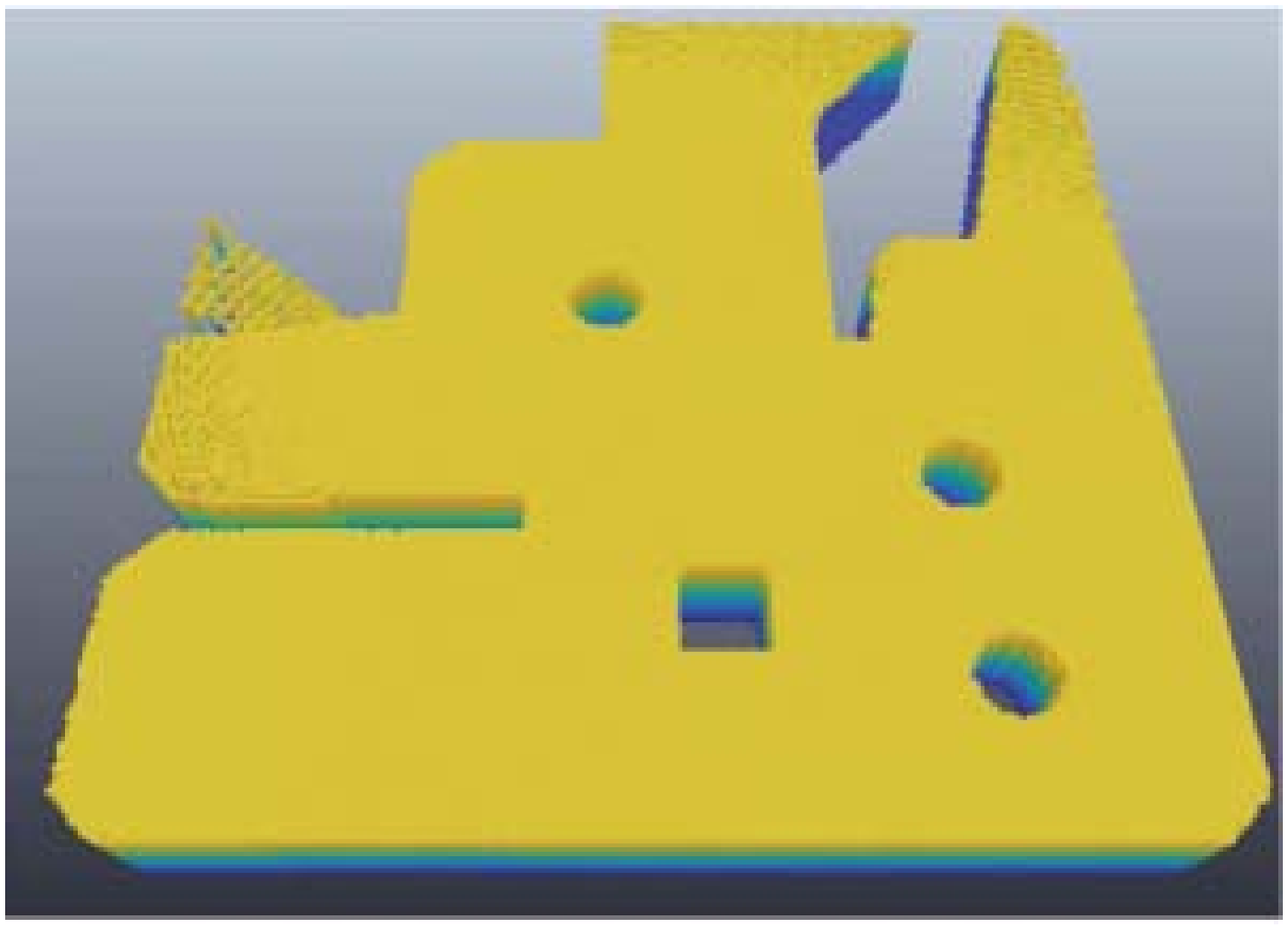,width=6cm}
}
\subfigure[]{
\epsfig{figure=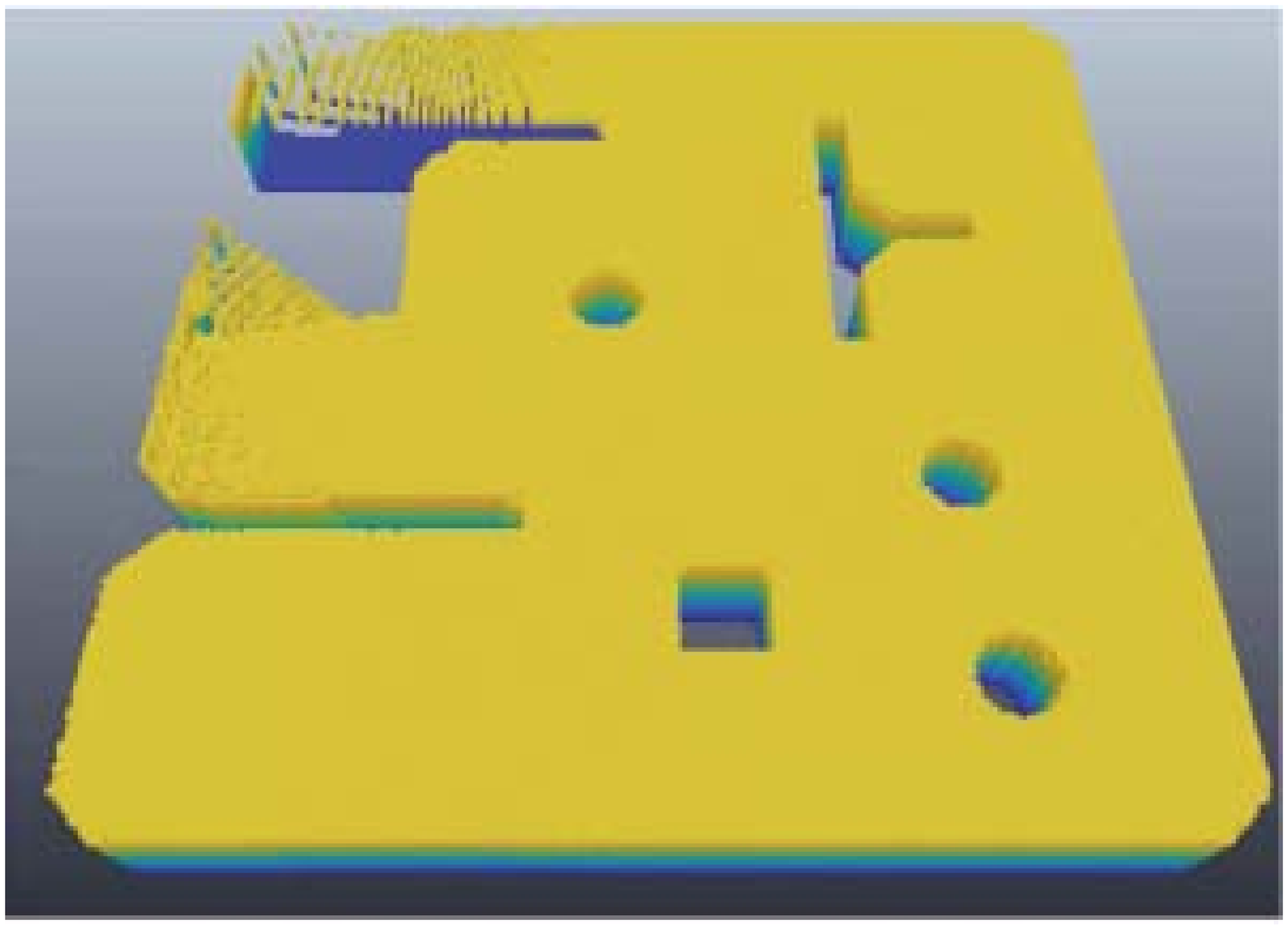,width=6cm}
}
\subfigure[]{
\epsfig{figure=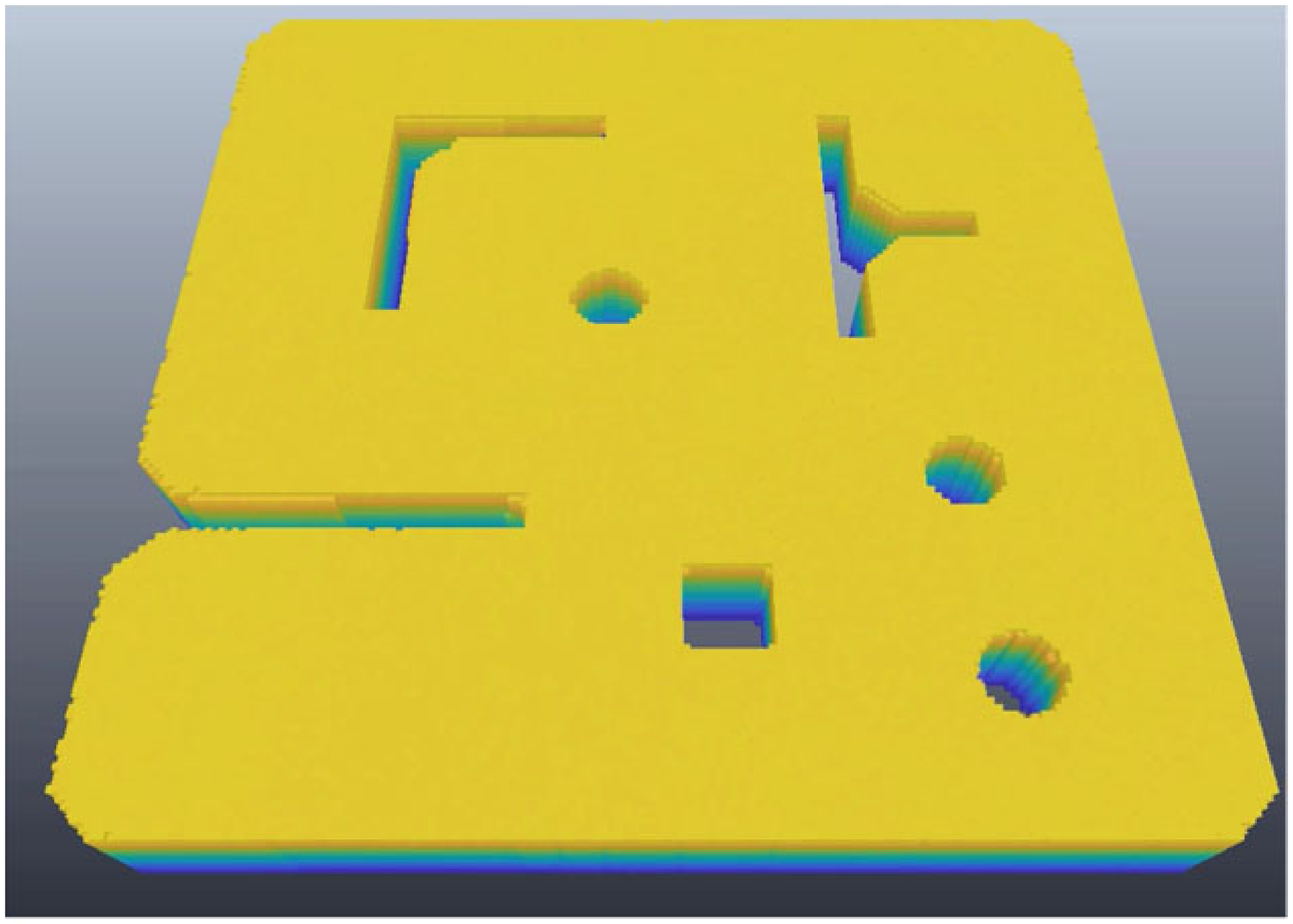,width=6cm}
}
\caption{
Map building in computer simulation. Different colors represent different height.
}
\label{c6_F:16}
\end{figure}

\begin{figure}[!htb]
\centering
\epsfig{figure=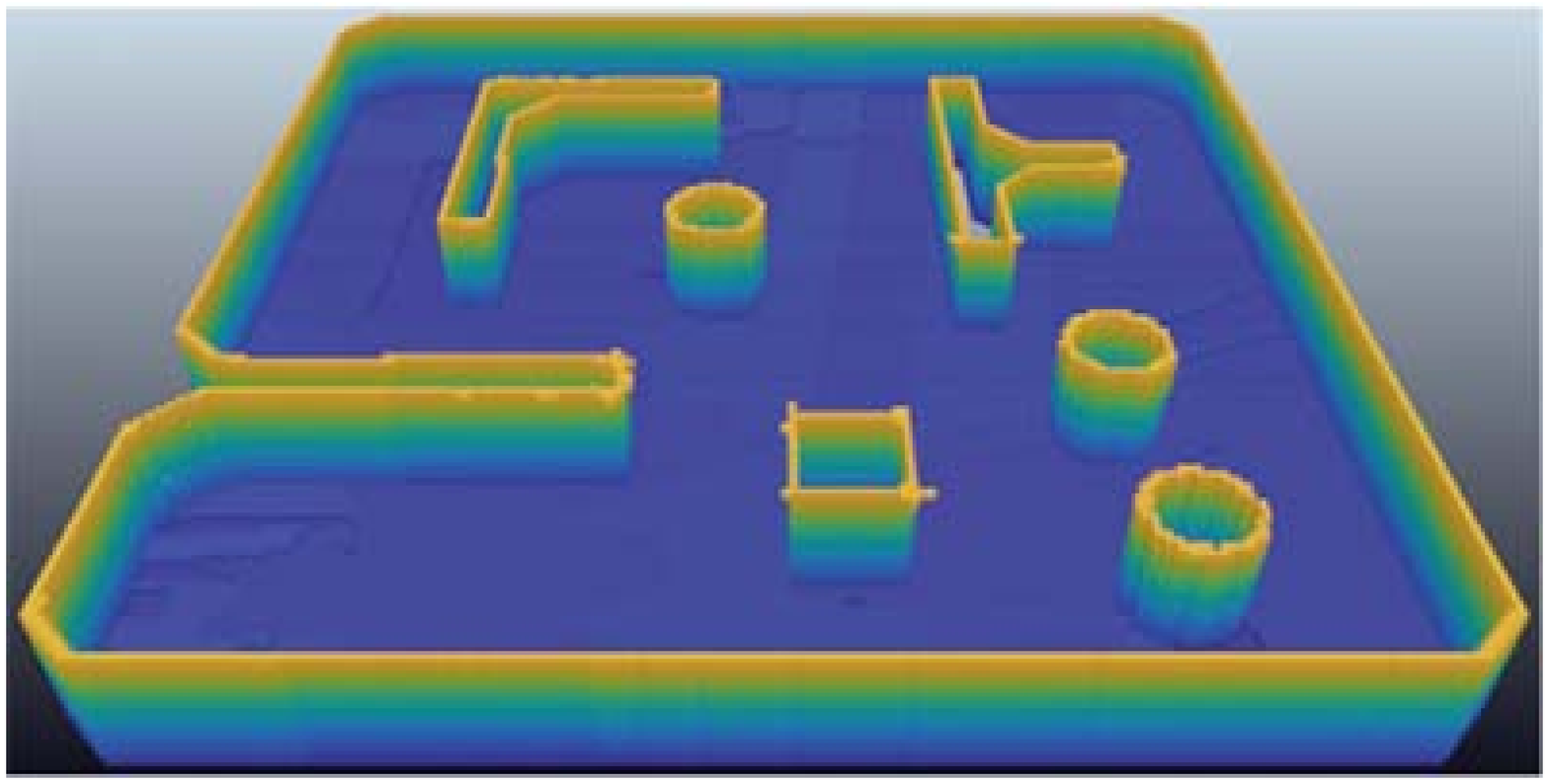,width=8cm}
\caption{
Complete 3D map of the environment in computer simulation.
}
\label{c6_F:17}
\end{figure}

In the presented computer simulation, an indoor environment was built with several obstacles with the size of $25\text{m}\times 25\text{m}\times 3\text{m}$. A ground mobile robot moved on the flat ground of the room and was equipped with two laser range finder sensors. The mobile robot was navigated by the proposed algorithm \textbf{B1}--\textbf{B4} in the indoor environment with obstacle avoidance. At a map building completing time, the robot stopped and completed the map building. The trajectory of the robot is shown in Fig. \ref{c6_F:15} that the robot kept the safety margin from the boundary of the room and any other obstacles. Fig. \ref{c6_F:16} shows the map building of the robot while the robot was moving. Finally, the robot stopped with a complete 3D map of the indoor environment indicated in Fig. \ref{c6_F:17}.

\section{Experiments with real mobile robot}
\begin{figure}[!htb]
\centering
\epsfig{figure=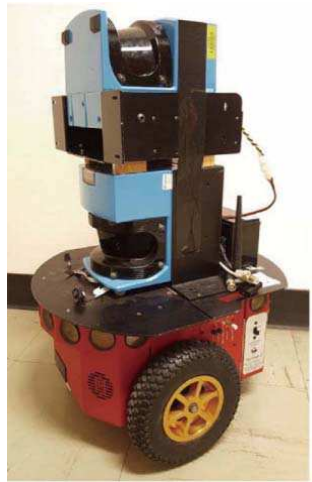,width=4cm}
\caption{
Pioneer3-DX robot with two SICK LMS-200 laser range finders. 
}
\label{c6_F:18}
\end{figure}

\begin{figure}[!htb]
\centering
\subfigure[]{
\epsfig{figure=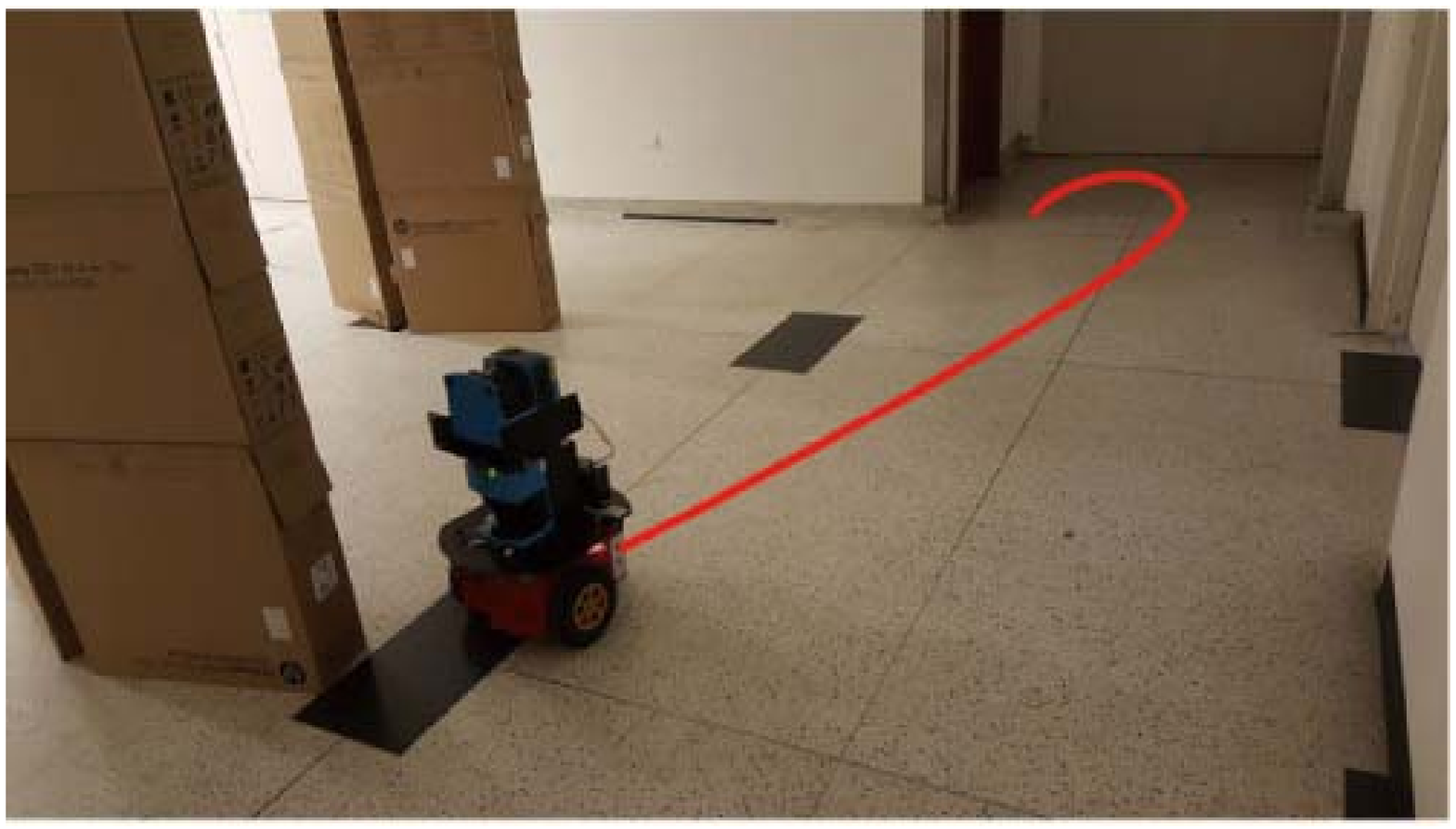,width=6cm}
}
\subfigure[]{
\epsfig{figure=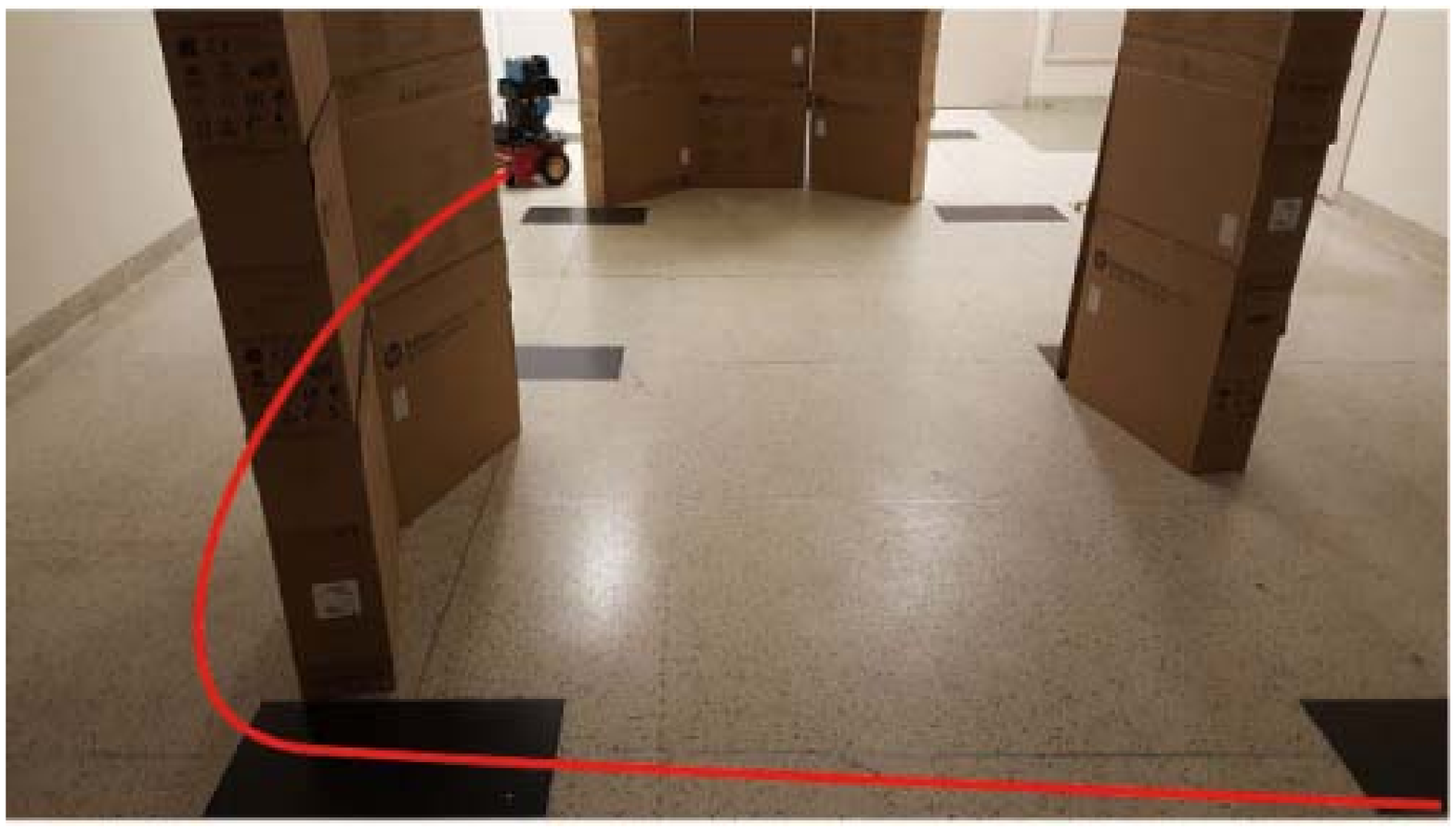,width=6cm}
}
\subfigure[]{
\epsfig{figure=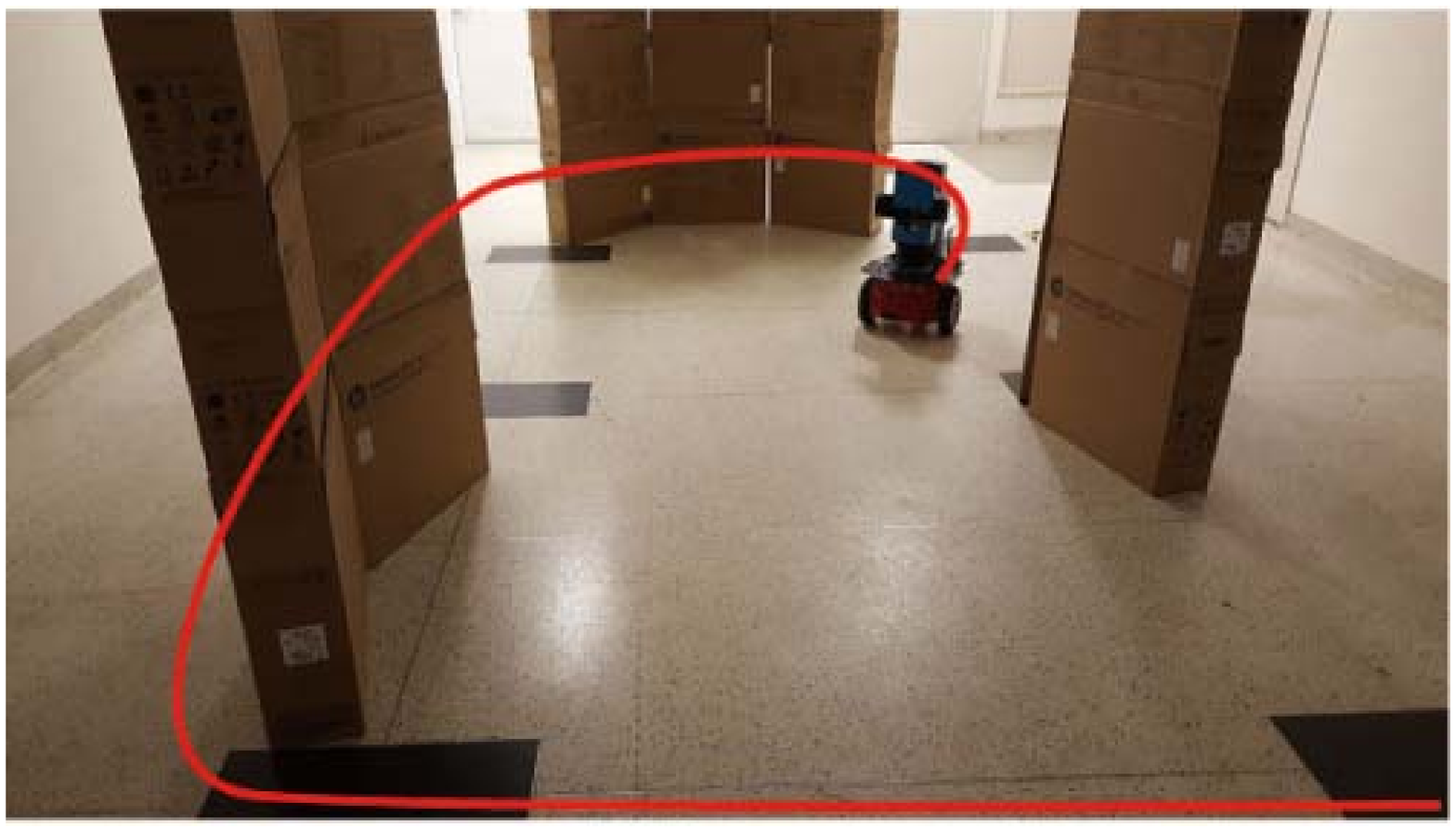,width=6cm}
}
\subfigure[]{
\epsfig{figure=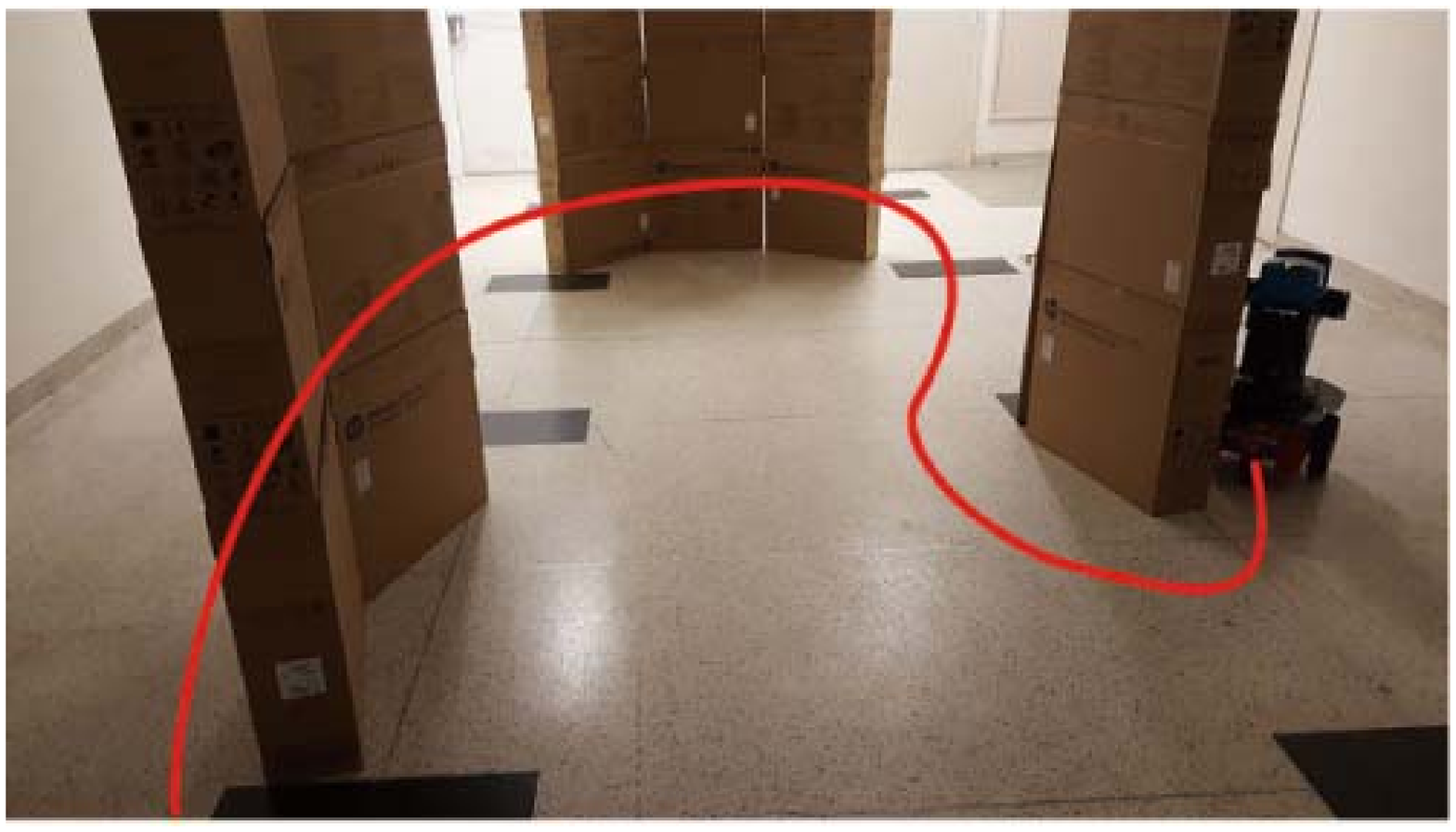,width=6cm}
}
\caption{
Pictures of the experiment.
}
\label{c6_F:19}
\end{figure}

\begin{figure}[!htb]
\centering
\subfigure[]{
\epsfig{figure=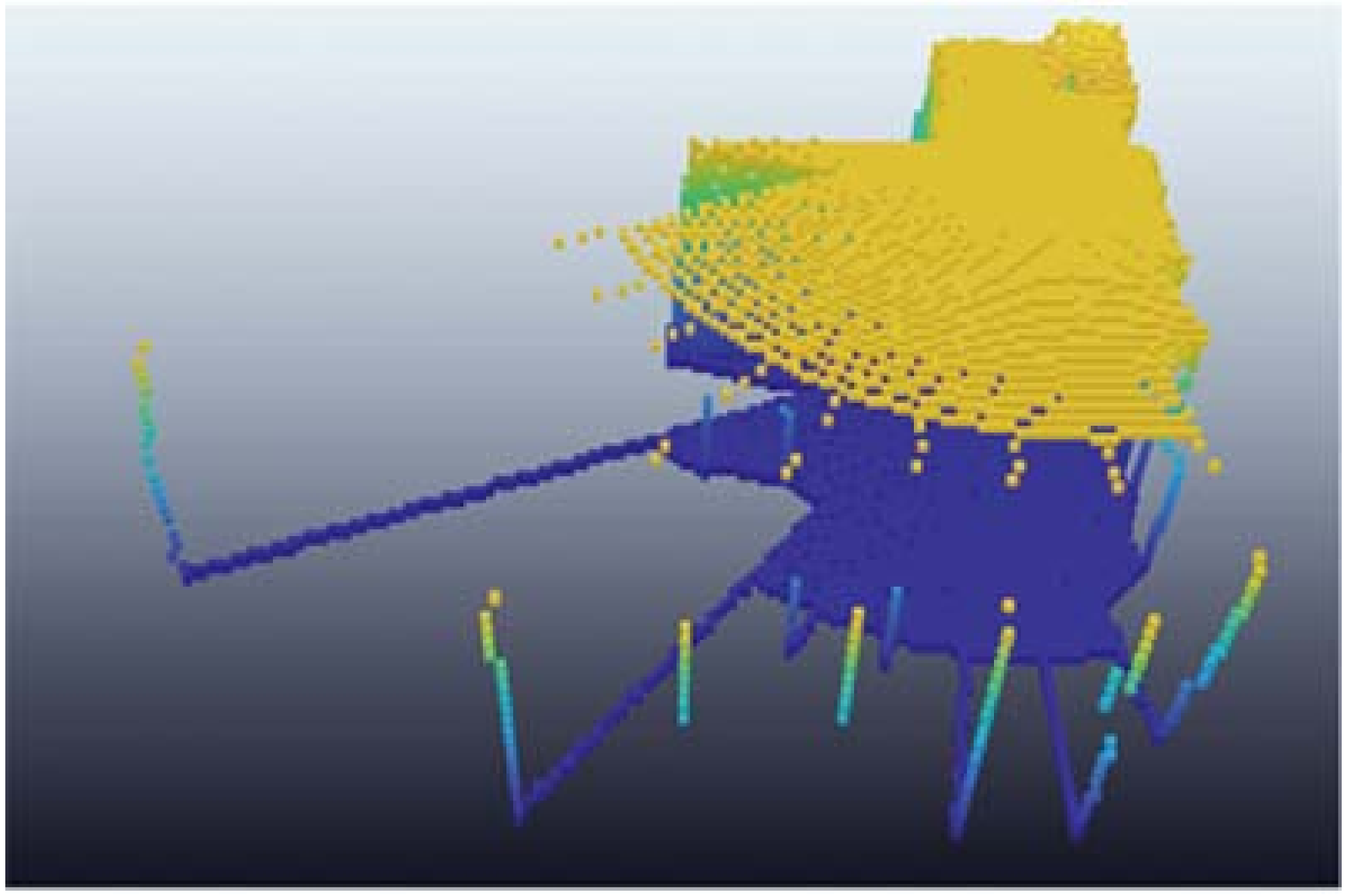,width=6cm}
}
\subfigure[]{
\epsfig{figure=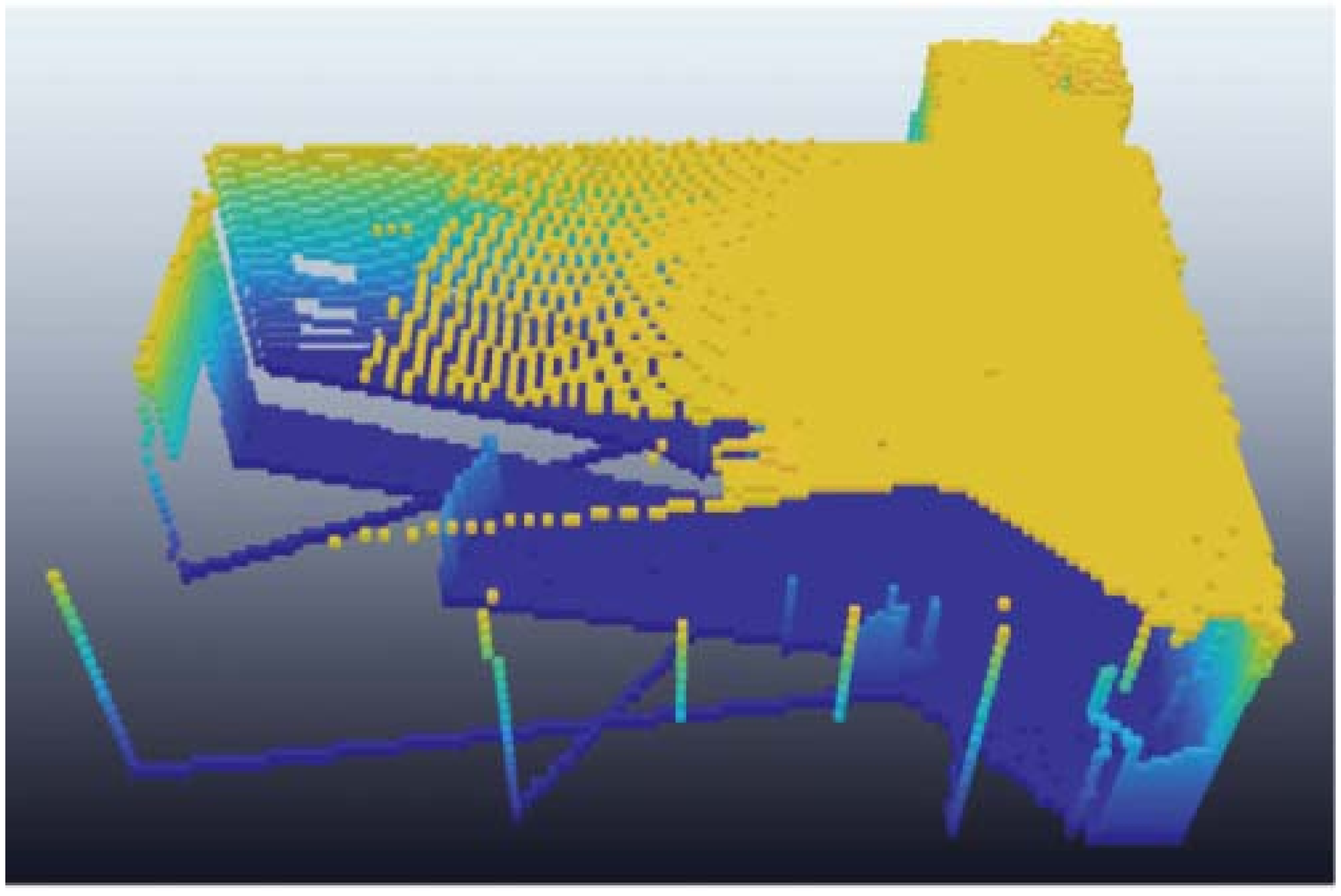,width=6cm}
}
\subfigure[]{
\epsfig{figure=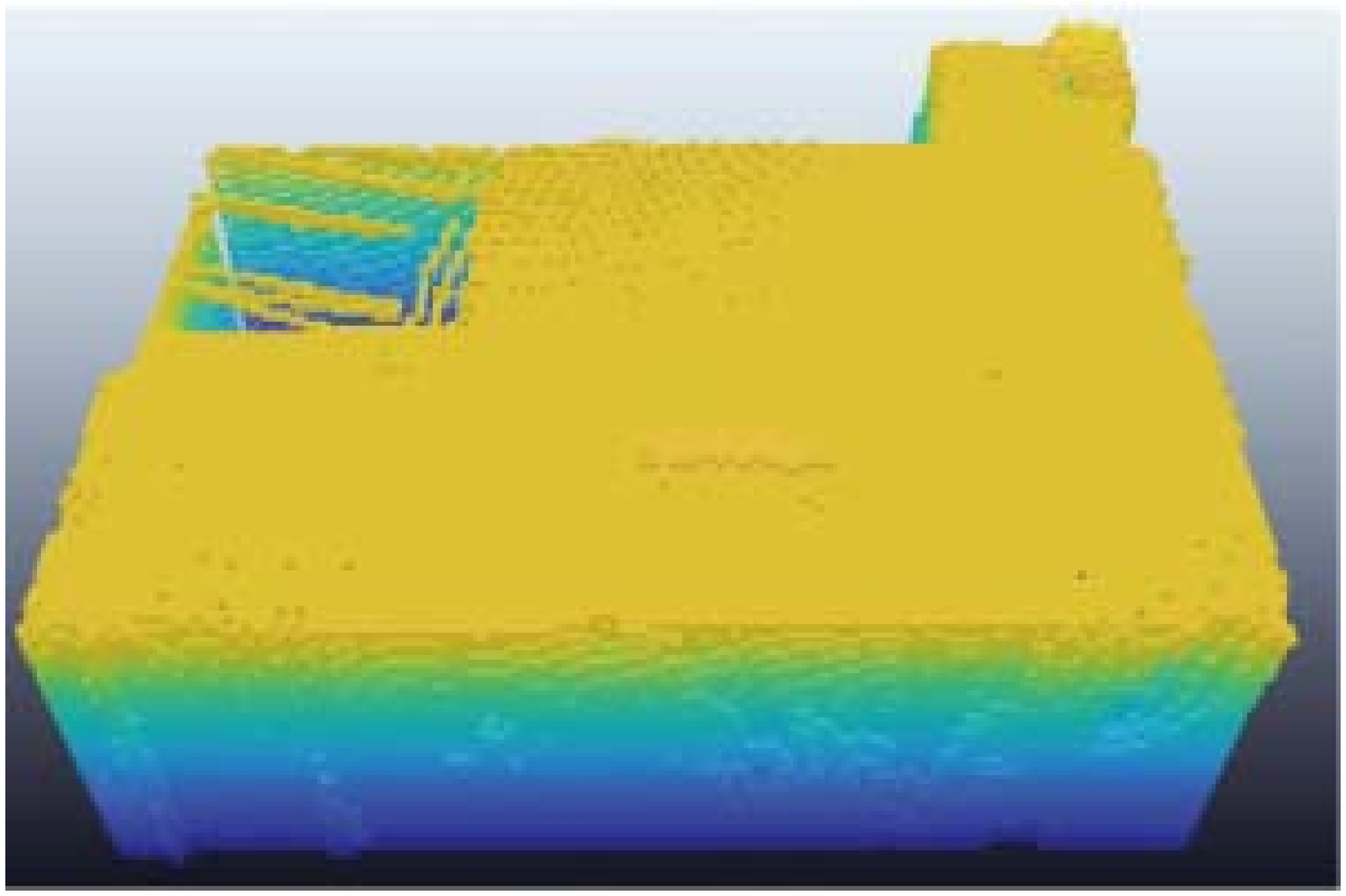,width=6cm}
}
\subfigure[]{
\epsfig{figure=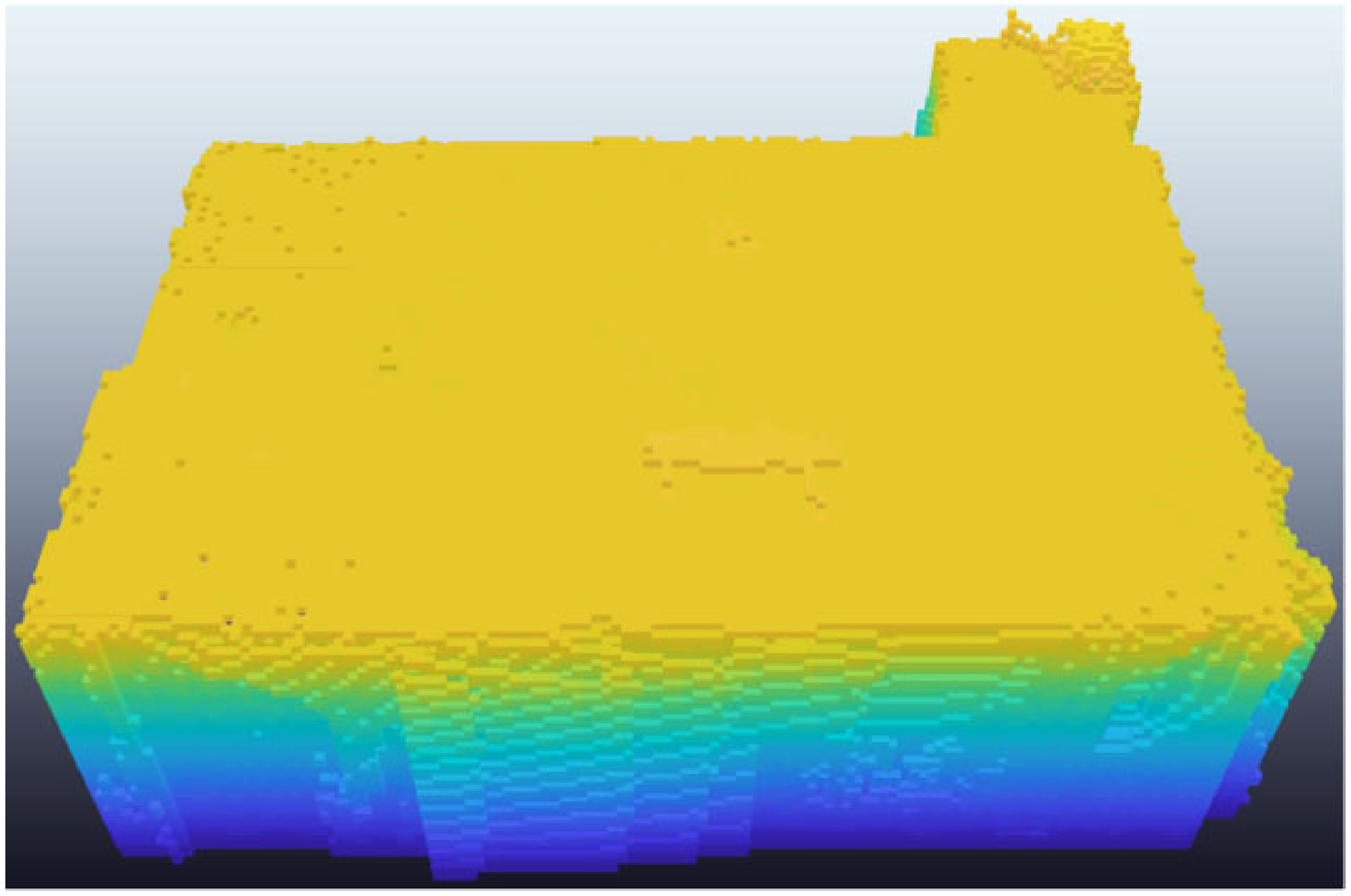,width=6cm}
}
\caption{
Map building of the environment in the experiment. Different colors represent different height.
}
\label{c6_F:20}
\end{figure}

\begin{figure}[!htb]
\centering
\epsfig{figure=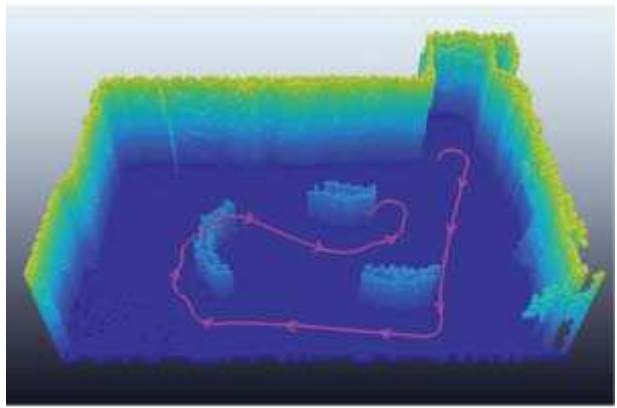,width=8cm}
\caption{
Complete 3D map of the environment and robot trajectory in the experiment. 
}
\label{c6_F:21}
\end{figure}

We carry out an experiment with a real ground mobile robot. In the experiment, a Pioneer3-DX robot is used and there are two SICK LMS-200 laser range finders mounted on the robot (see Fig. \ref{c6_F:18}). In a closed room, there were some folding cartons put on the ground as the obstacles. The robot started from an initial circle and was navigated by the proposed algorithm to avoid collision with any obstacle. Simultaneously, the robot scanned the 3D environment by the second laser range finder and used the measurements to build the 3D map of the environment (see Fig. \ref{c6_F:19} and \ref{c6_F:20}). Finally, the robot stopped when the complete map was built at the map building completing time (see Fig. \ref{c6_F:21}).

\section{Summary}
We consider a non-holonomic ground mobile robot equipped with two 2D range finder sensors and propose a probabilistic safe navigation algorithm to search and explore an unknown indoor environment and build a complete 3D map. The performance of the proposed method is confirmed by the computer simulation and the experiment with a real mobile robot.

The proposed method has several advantages. Firstly, the collision-free area search problem and complete 3D map building problem are solved simultaneously by the proposed algorithm. Secondly, comparing with the previously works, we consider the 3D map building of the environment, which is more useful than 2D map in practical implementations. Furthermore, the proposed navigation algorithm is relatively simple and fast. Therefore, it can be implemented in small-sized robots with low power supply and poor computer performance. Moreover, the non-holonomic constraint of the robot's motion is considered in the proposed algorithm, which is more practical than other works. According to the mathematical analysis, it can be proved that the complete 3D map can be built in a finite time by the ground mobile robot.

The proposed method can be used in several commercial and military tasks, like rescue and planetary exploration. In future work, flying robot can be considered for the 3D map building task.
%%%%%%%%%%%%%%%%%%%%%%%%%%%%%%%%%%%%%%%%%%%%%%%%
\chapter{Conclusion and Future Work}
\chaptermark{Conclusion and Future Work}
\label{Chapter7}

In this chapter, we give the conclusion of this report and present the future work as follows.
\section{Conclusion}
The main focus of this report is to design sensor network based navigation algorithm for non-holonomic mobile robots. Firstly, we present a path planning algorithm for ground mobile robots in dynamic environments. Then we consider the sensor network and develop the path planning algorithm to a sensor network based navigation algorithm for ground mobile robots. In the presented method, a sensor network consists of several 2D range finder sensors. The sensor network detects and monitors any obstacles in dynamic environments. With the measurements, the sensor network guides each ground mobile robot from its initial position to the target with a relatively short trajectory by the navigation algorithm. The computer simulations and experiments with a real mobile robot confirm the performance of the presented method. Furthermore, we consider the implementations of small-sized flying robots in industries and extend the presented 2D navigation algorithm into 3D environments. In the 3D environments, the sensor network consists of several time-of-flight cameras. Each of the sensor nodes detects the dynamic obstacles in the field of view. The modified algorithm navigates the micro flying robots to avoid any detected obstacles and undetected areas.

Comparing with other navigation algorithms, the main feature of the presented sensor network based navigation algorithms is that the navigation task for any  mobile robots in the workspace is transferred to the sensor network from the mobile robots. Each robot in the workspace with the presented sensor network is only required for a simple low-level path tracking controller. There is not any sensor for obstacle detection on the robots. It is a economical solution for multi-robot navigation in smart factories.

Furthermore, in this report, the safe area search and map building algorithms for single ground mobile robot is presented. Firstly, we consider a 2D map building and present a collision-free area search navigation algorithm for a non-holonomic ground mobile robot to explore a 2D bounded area. Simultaneously, the robot builds a complete 2D map of the unknown 2D area that indicates the boundary of the bounded area and the obstacles in the area. Then we consider an extra 2D range finder sensor mounted on the robot and develop the algorithm for 3D map building. The computer simulations and experiments confirm the performance of the presented area search and map building algorithms. The presented algorithms belong to probabilistic navigation algorithm and the mathematical analysis proves that with probability $1$ the robot can build a complete map in a finite time by the presented algorithms.
\section{Future work}
With the development of the sensor network and robotics, the mobile robots and sensor networks will be implemented widely in both industries and military. Therefore, the methods presented in this report can be applied in a variety of fields.

In Chapter \ref{Chapter2}, \ref{Chapter3} and \ref{Chapter4}, we developed a sensor network based navigation framework for both ground mobile robots and flying robots. In this navigation framework, different micro-sized ground mobile robots and flying robots can be navigated with obstacle avoidance by the sensor network consisting of some 2D and 3D range finder sensors. In the future work of this topic, firstly, the completely distributed navigation strategies should be focused on with the presented sensor network. In a distributed navigation framework, a central computer is not used and each sensor node works individually and equally to perform the navigation for the mobile robots within its coverage, like other distributed sensor network \cite{1234435,Prabhu2014,1038076,SAVKIN20061094}. The main advantage of a completely distributed navigation algorithm is that the system can keep working when a part of the sensor nodes are faulty. Another interesting problem is the robot formation with collision avoidance in the presented sensor network framework. Mobile robot formation is a significant type of the multi-robot cooperation \cite{Guzey2017,WANG2017Distributed,5400224,6606099,SAVKIN2016463}. It can be combined with the presented sensor network based navigation framework in dynamic smart factories or indoor environments to complete some multi-robot cooperation tasks like object transport \cite{Alonso-Mora2017} with collision avoidance. Moreover, the presented sensor network based navigation algorithm can be combined with the target or source search problem \cite {Ishida2006Mobile,Li2011Odor,Han2014Multiple} in dynamic indoor environments with moving obstacles. The target search combined with the safe indoor sensor network based navigation can be implemented in gas source or radioactive source searching in chemical plants or nuclear power plants with obstacle avoidance.

In Chapter \ref{Chapter5} and \ref{Chapter6}, we developed the collision-free area search and map building algorithms for a single ground mobile robot in 2D and 3D environments. With the presented algorithms, a non-holonomic ground mobile robot can be navigated with low computational cost to explore an unknown environment and build the complete 2D or 3D map. In the future work of this topic, collaborative multiple robot map building, like \cite{SHENG2006945Distributed,
Carpin2008Fast,Rekleitis2001Multi}, can be considered based on the presented single robot area search and map building algorithm by combining the presented algorithm with some multi-robot cooperation strategies, such as \cite{HOY20121253Collision,
Capitan2013Decentralized}. Another significant problem is the map building and area search on uneven terrain with obstacle avoidance. The path planning and navigation problem is more difficult than on uneven terrain than on a flat ground, such as \cite{RAJA2015295New,PUTZ20162123D,
SINGH2016156Feasible}. Therefore, it can be a challenge to solve the safe area search and complete map building problem by combining the presented area search strategy with other uneven terrain navigation algorithms, such as \cite{1570542,4650592,Matveev2017}. Moreover, the presented research works can be combined with robust Kalman filtering approach \cite{petersen1999robust} to build a more sophisticated map of the environment; see an example of the robust Kalman filtering approach in \cite{1413186} and another robust control design approach by using H-$\infty$ methods \cite{Robust2000Petersen}.
Furthermore, the presented area search and map building algorithm can be extended into 3D environments with flying robots; see \cite{Nex2014UAV}. A flying robot can be considered to search a 3D environment with complex structure safely and build a 3D map for the environment by a 2D or 3D range finder sensor. It can be implemented in many fields, such as indoor 3D structure scanning, cave exploration and tunnel rescue; see \cite{SURMANN2003181autonomous,6935206,7222733}.

%%%%%%%%%%%%%%%%%%%%%%%%%%%%%%%%%%%%%%%%%%%%%%%%%%%%%%%

\bibliographystyle{plain}
\bibliography{ms}

\begin{thebibliography}{100}

\bibitem{ABDESSEMED200431}
F.~Abdessemed, K.~Benmahammed, and E.~Monacelli.
\newblock A fuzzy-based reactive controller for a non-holonomic mobile robot.
\newblock {\em Robotics and Autonomous Systems}, 47(1):31--46, 2004.

\bibitem{AlKhawaldah2015}
M.~Al~Khawaldah and A.~N{\"u}chter.
\newblock Enhanced frontier-based exploration for indoor environment with
  multiple robots.
\newblock {\em Advanced Robotics}, 29(10):657--669, 2015.

\bibitem{Alam2015}
M.~S. Alam and M.~U. Rafique.
\newblock Mobile robot path planning in environments cluttered with non-convex
  obstacles using particle swarm optimization.
\newblock In {\em 2015 International Conference on Control, Automation and
  Robotics (ICCAR)}, pages 32--36, Singapore, Singapore, 2015. IEEE.

\bibitem{AlDahak2013}
A.~AlDahak, L.~Seneviratne, and J.~Dias.
\newblock Frontier-based exploration for unknown environments using incremental
  triangulation.
\newblock In {\em 2013 IEEE International Symposium on Safety, Security, and
  Rescue Robotics (SSRR)}, pages 1--6, Link{\"o}ping, Sweden, 2013. IEEE.

\bibitem{Alenya2014}
G.~Aleny{\`a}, S.~Foix, and C.~Torras.
\newblock Using {ToF} and {RGBD} cameras for {3D} robot perception and
  manipulation in human environments.
\newblock {\em Intelligent Service Robotics}, 7(4):211--220, 2014.

\bibitem{Almansa-Valverde2012}
S.~Almansa-Valverde, J.~C. Castillo, and A.~Fern{\'a}ndez-Caballero.
\newblock Mobile robot map building from time-of-flight camera.
\newblock {\em Expert Systems with Applications}, 39(10):8835--8843, 2012.

\bibitem{Alonso-Mora2017}
J.~Alonso-Mora, S.~Baker, and D.~Rus.
\newblock Multi-robot formation control and object transport in dynamic
  environments via constrained optimization.
\newblock {\em The International Journal of Robotics Research},
  36(9):1000--1021, 2017.

\bibitem{Babinec2014}
A.~Babinec, F.~Ducho{\v{n}}, M.~Dekan, P.~P{\'a}szt{\'o}, and M.~Kelemen.
\newblock {VFH}* {TDT} ({VFH}* with {Time Dependent Tree}): A new laser
  rangefinder based obstacle avoidance method designed for environment with
  non-static obstacles.
\newblock {\em Robotics and autonomous systems}, 62(8):1098--1115, 2014.

\bibitem{6878733}
D.~Bacciu, C.~Gallicchio, A.~Micheli, M.~Di Rocco, and A.~Saffiotti.
\newblock Learning context-aware mobile robot navigation in home environments.
\newblock In {\em IISA 2014, The 5th International Conference on Information,
  Intelligence, Systems and Applications}, pages 57--62, Chania Crete, Greece,
  2014.

\bibitem{Basilico2011}
N.~Basilico and F.~Amigoni.
\newblock Exploration strategies based on multi-criteria decision making for
  searching environments in rescue operations.
\newblock {\em Autonomous Robots}, 31(4):401--417, 2011.

\bibitem{Beeson2010428}
P.~Beeson, J.~Modayil, and B.~Kuipers.
\newblock Factoring the mapping problem: Mobile robot map-building in the
  hybrid spatial semantic hierarchy.
\newblock {\em The International Journal of Robotics Research}, 29(4):428--459,
  2010.

\bibitem{Bohd2015}
F.~Bohdanowicz, H.~Frey, R.~Funke, D.~Mosen, F.~Neumann, and
  I.~Stojmenovi\'{c}.
\newblock {RSSI}-based localization of a wireless sensor node with a flying
  robot.
\newblock In {\em Proceedings of the 30th Annual ACM Symposium on Applied
  Computing}, SAC '15, pages 708--715, New York, USA, 2015. ACM.

\bibitem{Borenstein1996}
J.~Borenstein and L.~Feng.
\newblock Measurement and correction of systematic odometry errors in mobile
  robots.
\newblock {\em IEEE Transactions on robotics and automation}, 12(6):869--880,
  1996.

\bibitem{Brass2015}
P.~Brass, I.~Vigan, and N.~Xu.
\newblock Shortest path planning for a tethered robot.
\newblock {\em Computational Geometry}, 48(9):732--742, 2015.

\bibitem{Bresenham1965}
J.~E. Bresenham.
\newblock Algorithm for computer control of a digital plotter.
\newblock {\em IBM Systems journal}, 4(1):25--30, 1965.

\bibitem{7558972}
H.~Cao, J.~Gao, F.~Zhao, J.~Zhao, C.~Liu, Y.~Liu, and X.~Shi.
\newblock The {3D} map building of the mobile robot.
\newblock In {\em 2016 IEEE International Conference on Mechatronics and
  Automation}, pages 2576--2581, Harbin, China, Aug 2016.

\bibitem{6303906}
Y.~Cao, W.~Yu, W.~Ren, and G.~Chen.
\newblock An overview of recent progress in the study of distributed
  multi-agent coordination.
\newblock {\em IEEE Transactions on Industrial Informatics}, 9(1):427--438,
  2013.

\bibitem{Capitan2013Decentralized}
J.~Capitan, M.~T.J. Spaan, L.~Merino, and A.~Ollero.
\newblock Decentralized multi-robot cooperation with auctioned {POMDPs}.
\newblock {\em The International Journal of Robotics Research}, 32(6):650--671,
  2013.

\bibitem{Carpin2008Fast}
S.~Carpin.
\newblock Fast and accurate map merging for multi-robot systems.
\newblock {\em Autonomous Robots}, 25(3):305--316, Oct 2008.

\bibitem{7027292}
P.~Chanak, I.~Banerjee, J.~Wang, and R.~S. Sherratt.
\newblock Obstacle avoidance routing scheme through optimal sink movement for
  home monitoring and mobile robotic consumer devices.
\newblock {\em IEEE Transactions on Consumer Electronics}, 60(4):596--604, Nov
  2014.

\bibitem{5649136}
C.~K. Chang, C.~Siagian, and L.~Itti.
\newblock Mobile robot vision navigation amp; localization using gist and
  saliency.
\newblock In {\em 2010 IEEE/RSJ International Conference on Intelligent Robots
  and Systems}, pages 4147--4154, Taipei, Taiwan, 2010.

\bibitem{Chang2013}
H.~Chang and T.~Jin.
\newblock Adaptive tracking controller based on the pid for mobile robot path
  tracking.
\newblock In {\em Intelligent Robotics and Applications: 6th International
  Conference (ICIRA 2013)}, pages 540--549, Berlin, Heidelberg, 2013. Springer.

\bibitem{cheng_jiang_hu_2014}
Y.~Cheng, P.~Jiang, and Y.~F. Hu.
\newblock A snake-based scheme for path planning and control with constraints
  by distributed visual sensors.
\newblock {\em Robotica}, 32(4):477--499, 2014.

\bibitem{5707420}
S.~H. Cho and S.~Hong.
\newblock Map based indoor robot navigation and localization using laser range
  finder.
\newblock In {\em 2010 11th International Conference on Control Automation
  Robotics Vision}, pages 1559--1564, Singapore, Singapore, 2010.

\bibitem{Colares2016}
R.~G. Colares and L.~Chaimowicz.
\newblock The next frontier: combining information gain and distance cost for
  decentralized multi-robot exploration.
\newblock In {\em Proceedings of the 31st Annual ACM Symposium on Applied
  Computing}, pages 268--274, Pisa, Italy, 2016. ACM.

\bibitem{Contreras-Cruz2015}
M.~A. Contreras-Cruz, V.~Ayala-Ramirez, and U.~H. Hernandez-Belmonte.
\newblock Mobile robot path planning using artificial bee colony and
  evolutionary programming.
\newblock {\em Applied Soft Computing}, 30:319--328, 2015.

\bibitem{Corke2005}
P.~Corke, R.~Peterson, and D.~Rus.
\newblock Networked robots: Flying robot navigation using a sensor net.
\newblock In {\em Robotics Research. The Eleventh International Symposium},
  pages 234--243, Berlin, Heidelberg, 2005. Springer Berlin Heidelberg.

\bibitem{7383260}
P.~J. Costa, N.~Moreira, D.~Campos, J.~Gon{\c c}alves, J.~Lima, and P.~L.
  Costa.
\newblock Localization and navigation of an omnidirectional mobile robot: The
  robot\@factory case study.
\newblock {\em IEEE Revista Iberoamericana de Tecnologias del Aprendizaje},
  11(1):1--9, Feb 2016.

\bibitem{6106751}
M.~S. Couceiro, R.~P. Rocha, and N.~M.~F. Ferreira.
\newblock A novel multi-robot exploration approach based on particle swarm
  optimization algorithms.
\newblock In {\em 2011 IEEE International Symposium on Safety, Security, and
  Rescue Robotics}, pages 327--332, Kyoto, Japan, Nov 2011.

\bibitem{6423636}
A.~R. de~Carvalho, A.~D. Ribas, V.~F. da~Camara~Neto, E.~F. Nakamura, and C.~M.
  Figueiredo.
\newblock An {RSSI}-based navigation algorithm for a mobile robot in wireless
  sensor networks.
\newblock In {\em 37th Annual IEEE Conference on Local Computer Networks},
  pages 308--311, Clearwater Beach, USA, Oct 2012.

\bibitem{6943217}
N.~Deshpande, E.~Grant, M.~Draelos, and T.~C. Henderson.
\newblock Received signal strength based bearing-only robot navigation in a
  sensor network field.
\newblock In {\em 2014 IEEE/RSJ International Conference on Intelligent Robots
  and Systems}, pages 4618--4623, Chicago, USA, Sept 2014.

\bibitem{6579651}
N.~Deshpande, E.~Grant, and T.~C. Henderson.
\newblock Target localization and autonomous navigation using wireless sensor
  networks--a pseudogradient algorithm approach.
\newblock {\em IEEE Systems Journal}, 8(1):93--103, March 2014.

\bibitem{Dewangan2016}
K.~Dewangan, A.~Saha, K.~Vaiapury, and R.~Dasgupta.
\newblock {3D} environment reconstruction using mobile robot platform and
  monocular vision.
\newblock In {\em Advanced Computing and Communication Technologies:
  Proceedings of the 9th ICACCT, 2015}, pages 213--221, Singapore, 2016.

\bibitem{Doh2003}
N.~Doh, H.~Choset, and W.~K. Chung.
\newblock Accurate relative localization using odometry.
\newblock In {\em IEEE International Conference on Robotics and Automation,
  2003}, pages 1606--1612, Taipei, Taiwan, 2003. IEEE.

\bibitem{Doh2006}
N.~L. Doh, H.~Choset, and W.~K. Chung.
\newblock Relative localization using path odometry information.
\newblock {\em Autonomous Robots}, 21(2):143--154, 2006.

\bibitem{6851896}
X.~Dong, J.~Xi, G.~Lu, and Y.~Zhong.
\newblock Formation control for high-order linear time-invariant multiagent
  systems with time delays.
\newblock {\em IEEE Transactions on Control of Network Systems}, 1(3):232--240,
  2014.

\bibitem{DONG201626Time}
X.~Dong, Y.~Zhou, Z.~Ren, and Y.~Zhong.
\newblock Time-varying formation control for unmanned aerial vehicles with
  switching interaction topologies.
\newblock {\em Control Engineering Practice}, 46:26--36, 2016.

\bibitem{Dubins1957}
L.~E. Dubins.
\newblock On curves of minimal length with a constraint on average curvature,
  and with prescribed initial and terminal positions and tangents.
\newblock {\em American Journal of Mathematics}, 79(3):497--516, 1957.

\bibitem{Enriquez2013}
G.~Enriquez, S.~Park, and S.~Hashimoto.
\newblock Wireless sensor network and {RFID} fusion approach for mobile robot
  navigation.
\newblock {\em ISRN Sensor Networks}, 2013, 2013.

\bibitem{1038076}
G.~L. Foresti and L.~Snidaro.
\newblock A distributed sensor network for video surveillance of outdoor
  environments.
\newblock In {\em 2002 International Conference on Image Processing}, volume~1,
  pages I--525--I--528, Rochester, USA, 2002.

\bibitem{4209183}
A.~Franchi, L.~Freda, G.~Oriolo, and M.~Vendittelli.
\newblock A randomized strategy for cooperative robot exploration.
\newblock In {\em Proceedings 2007 IEEE International Conference on Robotics
  and Automation}, pages 768--774, Roma, Italy, April 2007.

\bibitem{Freda2005}
L.~Freda and G.~Oriolo.
\newblock Frontier-based probabilistic strategies for sensor-based exploration.
\newblock In {\em 2005 IEEE International Conference on Robotics and Automation
  (ICRA)}, pages 3881--3887, Hong Kong, China, 2005. IEEE.

\bibitem{5434384}
S.~M. George, W.~Zhou, H.~Chenji, M.~Won, Y.~O. Lee, A.~Pazarloglou,
  R.~Stoleru, and P.~Barooah.
\newblock Distressnet: a wireless ad hoc and sensor network architecture for
  situation management in disaster response.
\newblock {\em IEEE Communications Magazine}, 48(3):128--136, March 2010.

\bibitem{Guerra2016}
M.~Guerra, D.~Efimov, G.~Zheng, and W.~Perruquetti.
\newblock Finite-time obstacle avoidance for unicycle-like robot subject to
  additive input disturbances.
\newblock {\em Autonomous Robots}, pages 1--12, 2016.

\bibitem{MWHMET2017}
M.~S. G\"{u}zel, M.~Kara, and M.~S. Beyazk{\i}l{\i}\c{c}.
\newblock An adaptive framework for mobile robot navigation.
\newblock {\em Adaptive Behavior}, 25(1):30--39, 2017.

\bibitem{Guzey2017}
H.~M. G{\"u}zey, T.~Dierks, S.~Jagannathan, and L.~Acar.
\newblock Hybrid consensus-based control of nonholonomic mobile robot
  formation.
\newblock {\em Journal of Intelligent {\&} Robotic Systems}, Apr 2017.

\bibitem{5658491}
D.~Hai, Y.~Li, H.~Zhang, and X.~Li.
\newblock Simultaneous localization and mapping of robot in wireless sensor
  network.
\newblock In {\em 2010 IEEE International Conference on Intelligent Computing
  and Intelligent Systems}, volume~3, pages 173--178, Xiamen, China, Oct 2010.

\bibitem{Han2014Multiple}
J.~Han and Y.~Chen.
\newblock Multiple uav formations for cooperative source seeking and contour
  mapping of a radiative signal field.
\newblock {\em Journal of Intelligent {\&} Robotic Systems}, 74(1):323--332,
  Apr 2014.

\bibitem{5509993}
A.~D. Haumann, K.~D. Listmann, and V.~Willert.
\newblock Discoverage: A new paradigm for multi-robot exploration.
\newblock In {\em 2010 IEEE International Conference on Robotics and
  Automation}, pages 929--934, Anchorage, USA, May 2010.

\bibitem{Held2016}
D.~Held, J.~Levinson, S.~Thrun, and S.~Savarese.
\newblock Robust real-time tracking combining {3D} shape, color, and motion.
\newblock {\em The International Journal of Robotics Research}, 35(1-3):30--49,
  2016.

\bibitem{HIREMATH201441Laser}
S.~A. Hiremath, G.~W. A.~M. van~der Heijden, F.~K. van Evert, A.~Stein, and
  C.~J.F. ter Braak.
\newblock Laser range finder model for autonomous navigation of a robot in a
  maize field using a particle filter.
\newblock {\em Computers and Electronics in Agriculture}, 100:41--50, 2014.

\bibitem{5350445}
G.~M. Hoffmann and C.~J. Tomlin.
\newblock Mobile sensor network control using mutual information methods and
  particle filters.
\newblock {\em IEEE Transactions on Automatic Control}, 55(1):32--47, Jan 2010.

\bibitem{Hoog2010}
J.~Hoog, S.~Cameron, and A.~Visser.
\newblock Autonomous multi-robot exploration in communication-limited
  environments.
\newblock In {\em Proceedings of the 11th Conference Towards Autonomous Robotic
  Systems (Taros 2010)}, pages 68--75, Plymouth, UK, 2010. University of
  Plymouth, School of Computing and Mathematics.

\bibitem{Hossain2015}
M.~A. Hossain and I.~Ferdous.
\newblock Autonomous robot path planning in dynamic environment using a new
  optimization technique inspired by bacterial foraging technique.
\newblock {\em Robotics and Autonomous Systems}, 64:137--141, 2015.

\bibitem{HOY20121253Collision}
M.~Hoy, A.~S. Matveev, and A.~V. Savkin.
\newblock Collision free cooperative navigation of multiple wheeled robots in
  unknown cluttered environments.
\newblock {\em Robotics and Autonomous Systems}, 60(10):1253--1266, 2012.

\bibitem{Hoy2015}
M.~Hoy, A.~S. Matveev, and A.~V. Savkin.
\newblock Algorithms for collision-free navigation of mobile robots in complex
  cluttered environments: a survey.
\newblock {\em Robotica}, 33(3):463--497, 2015.

\bibitem{Hsu2013}
C.~Hsu, H.~E. Chang, and Y.~Lu.
\newblock Map building of unknown environment using {PSO}-tuned enhanced
  iterative closest point algorithm.
\newblock In {\em 2013 International Conference on System Science and
  Engineering (ICSSE)}, pages 279--284, Wroclaw, Poland, 2013. IEEE.

\bibitem{6935206}
Y.~Hwang, D.~Lee, and J.~Lee.
\newblock {3D} map building for a moving based on mobile robot.
\newblock In {\em 2014 Proceedings of the SICE Annual Conference (SICE)}, pages
  474--479, Sapporo, Japan, Sept 2014.

\bibitem{7222733}
Y.~S. Hwang and J.~M. Lee.
\newblock Robust {3D} map building for a mobile robot moving on the floor.
\newblock In {\em 2015 IEEE International Conference on Advanced Intelligent
  Mechatronics (AIM)}, pages 1388--1393, Busan, Korea, July 2015.

\bibitem{4650592}
K.~Iagnemma, S.~Shimoda, and Z.~Shiller.
\newblock Near-optimal navigation of high speed mobile robots on uneven
  terrain.
\newblock In {\em 2008 IEEE/RSJ International Conference on Intelligent Robots
  and Systems}, pages 4098--4103, Nice, France, Sept 2008.

\bibitem{Ishida2006Mobile}
H.~Ishida, H.~Tanaka, H.~Taniguchi, and T.~Moriizumi.
\newblock Mobile robot navigation using vision and olfaction to search for a
  gas/odor source.
\newblock {\em Autonomous Robots}, 20(3):231--238, Jun 2006.

\bibitem{Jain2011}
S.~Jain, S.~Nandy, G.~Chakraborty, C.~S. Kumar, R.~Ray, and S.~N. Shome.
\newblock Error modeling of laser range finder for robotic application using
  time domain technique.
\newblock In {\em 2011 IEEE International Conference on Signal Processing,
  Communications and Computing (ICSPCC)}, pages 1--5, Xi'an, China, 2011. IEEE.

\bibitem{Ji2006Role}
M.~Ji, S.~Azuma, and M.~Egerstedt.
\newblock Role-assignment in multi-agent coordination,.
\newblock {\em International Journal of Assistive Robotics and Mechatronics},
  7(1):32--40, 2006.

\bibitem{7294063}
Y.~Ji, A.~Yamashita, and H.~Asama.
\newblock Automatic calibration and trajectory reconstruction of mobile robot
  in camera sensor network.
\newblock In {\em 2015 IEEE International Conference on Automation Science and
  Engineering (CASE)}, pages 206--211, Gothenburg, Sweden, Aug 2015.

\bibitem{7827337}
Q.~Jia, M.~Wang, S.~Liu, J.~Ge, and C.~Gu.
\newblock Research and development of mecanum-wheeled omnidirectional mobile
  robot implemented by multiple control methods.
\newblock In {\em 2016 23rd International Conference on Mechatronics and
  Machine Vision in Practice (M2VIP)}, pages 1--4, Nanjing, China, Nov 2016.

\bibitem{Jin2017}
J.~Jin, Y.~Kim, S.~Wee, D.~Lee, and N.~Gans.
\newblock A stable switched-system approach to collision-free wheeled mobile
  robot navigation.
\newblock {\em Journal of Intelligent {\&} Robotic Systems}, pages 1--18, 2017.

\bibitem{Karaboga2012Cluster}
D.~Karaboga, S.~Okdem, and C.~Ozturk.
\newblock Cluster based wireless sensor network routing using artificial bee
  colony algorithm.
\newblock {\em Wireless Networks}, 18(7):847--860, Oct 2012.

\bibitem{Kavraki1996}
L.~E. Kavraki, P.~Svestka, J.~C. Latombe, and M.~H. Overmars.
\newblock Probabilistic roadmaps for path planning in high-dimensional
  configuration spaces.
\newblock {\em IEEE Transactions on Robotics and Automation}, 12(4):566--580,
  Aug 1996.

\bibitem{Kim2015}
E.~K. Kim, H.~Cho, E.~Jang, M.~K. Park, and S.~Kim.
\newblock Map building of indoor environment using laser range finder and
  geometrics.
\newblock In {\em 2015 IEEE International Conference on Advanced Intelligent
  Mechatronics (AIM)}, pages 1259--1264, Busan, Korea, 2015. IEEE.

\bibitem{6290366}
S.~J. Kim and B.~K. Kim.
\newblock Dynamic ultrasonic hybrid localization system for indoor mobile
  robots.
\newblock {\em IEEE Transactions on Industrial Electronics}, 60(10):4562--4573,
  2013.

\bibitem{Kim2015a}
Y.~Kim and S.~Kwon.
\newblock A heuristic obstacle avoidance algorithm using vanishing point and
  obstacle angle.
\newblock {\em Intelligent Service Robotics}, 8(3):175--183, 2015.

\bibitem{Klingenberg2013}
W.~Klingenberg.
\newblock {\em A course in differential geometry}, volume~51.
\newblock Springer Science \& Business Media, New York, USA, 2013.

\bibitem{Ko2015}
C.~Ko, K.~Young, and Y.~Hsieh.
\newblock Optimized trajectory planning for mobile robot in the presence of
  moving obstacles.
\newblock In {\em 2015 IEEE International Conference on Mechatronics (ICM)},
  pages 70--75, Nagoya, Japan, 2015. IEEE.

\bibitem{1435479}
S.~Koenig and M.~Likhachev.
\newblock Fast replanning for navigation in unknown terrain.
\newblock {\em IEEE Transactions on Robotics}, 21(3):354--363, June 2005.

\bibitem{Kownacki2016}
C.~Kownacki.
\newblock A concept of laser scanner designed to realize {3D} obstacle
  avoidance for a fixed-wing {UAV}.
\newblock {\em Robotica}, 34(2):243--257, 002 2016.

\bibitem{4406749}
S.~Kubota, Y.~Ando, and M.~Mizukawa.
\newblock Navigation of the autonomous mobile robot using laser range finder
  based on the non quantity map.
\newblock In {\em 2007 International Conference on Control, Automation and
  Systems}, pages 2329--2333, Seoul, Korea, 2007.

\bibitem{7413791}
A.~S. Kundu, O.~Mazumder, A.~Dhar, and S.~Bhaumik.
\newblock Occupancy grid map generation using ${360}^\circ$ scanning xtion pro
  live for indoor mobile robot navigation.
\newblock In {\em 2016 IEEE First International Conference on Control,
  Measurement and Instrumentation (CMI)}, pages 464--468, Kolkata, India, Jan
  2016.

\bibitem{Lapierre2007}
L.~Lapierre, R.~Zapata, and P.~Lepinay.
\newblock Combined path-following and obstacle avoidance control of a wheeled
  robot.
\newblock {\em The International Journal of Robotics Research}, 26(4):361--375,
  2007.

\bibitem{1234435}
J.~H. Lee and H.~Hashimoto.
\newblock Controlling mobile robots in distributed intelligent sensor network.
\newblock {\em IEEE Transactions on Industrial Electronics}, 50(5):890--902,
  Oct 2003.

\bibitem{Li2015}
G.~Li, S.~Tong, F.~Cong, A.~Yamashita, and H.~Asama.
\newblock Improved artificial potential field-based simultaneous forward search
  method for robot path planning in complex environment.
\newblock In {\em 2015 IEEE/SICE International Symposium on System Integration
  (SII)}, pages 760--765, Nagoya, Japan, 2015. IEEE.

\bibitem{7554216}
H.~Li.
\newblock Global shortest path planning for a wheeled mobile robot navigated by
  a sensor network in dynamic cluttered environments.
\newblock In {\em 2016 35th Chinese Control Conference (CCC)}, pages
  5525--5530, Chengdu, China, July 2016.

\bibitem{Hang_uj_2}
H.~Li and A.~V. Savkin.
\newblock An algorithm for safe navigation of mobile robots by a sensor network
  in dynamic cluttered industrial environments.
\newblock {\em Robotics and Computer-Integrated Manufacturing}.
\newblock Submitted.

\bibitem{Hang_uj_3}
H.~Li and A.~V. Savkin.
\newblock Wireless sensor network based navigation of micro flying robots in
  the industrial internet of things.
\newblock {\em IEEE Transactions on Industrial Informatics}.
\newblock Submitted.

\bibitem{Hang_ccc2017_2}
H.~Li and A.~V. Savkin.
\newblock A method for collision free sensor network based navigation of flying
  robots among moving and steady obstacles.
\newblock In {\em 2017 37th Chinese Control Conference (CCC)}, Dalian, China,
  July 2017.
\newblock Accepted.

\bibitem{Hang_ccc2017_1}
H.~Li and A.~V. Savkin.
\newblock A navigation algorithm for a non-holonomic mobile robot navigated by
  a sensor network in dynamic cluttered environments.
\newblock In {\em 2017 37th Chinese Control Conference (CCC)}, Dalian, China,
  July 2017.
\newblock Accepted.

\bibitem{6935304}
J.~Li, S.~Liu, B.~Zhang, and X.~Zhao.
\newblock {RRT-A*} motion planning algorithm for non-holonomic mobile robot.
\newblock In {\em 2014 Proceedings of the SICE Annual Conference (SICE)}, pages
  1833--1838, Sapporo, Japan, Sept 2014.

\bibitem{Li2011Odor}
J.~G. Li, Q.~H. Meng, Y.~Wang, and M.~Zeng.
\newblock Odor source localization using a mobile robot in outdoor airflow
  environments with a particle filter algorithm.
\newblock {\em Autonomous Robots}, 30(3):281--292, Apr 2011.

\bibitem{7831932}
X.~Li, S.~Li, S.~Jia, and C.~Xu.
\newblock Mobile robot map building based on laser ranging and kinect.
\newblock In {\em 2016 IEEE International Conference on Information and
  Automation (ICIA)}, pages 819--824, Ningbo, China, Aug 2016.

\bibitem{Liao2016}
F.~Liao, S.~Lai, Y.~Hu, J.~Cui, J.~L. Wang, R.~Teo, and F.~Lin.
\newblock {3D} motion planning for {UAVs} in {GPS}-denied unknown forest
  environment.
\newblock In {\em 2016 IEEE Intelligent Vehicles Symposium (IV)}, pages
  246--251, Gothenburg, Sweden, June 2016.

\bibitem{6871831}
J.~Lim, S.~J. Lee, G.~Tewolde, and J.~Kwon.
\newblock Ultrasonic-sensor deployment strategies and use of smartphone sensors
  for mobile robot navigation in indoor environment.
\newblock In {\em 2014 IEEE International Conference on Electro/Information
  Technology}, pages 593--598, Milwaukee, USA, 2014.

\bibitem{Liu2015}
T.~Liu and D.~M. Lyons.
\newblock Leveraging area bounds information for autonomous decentralized
  multi-robot exploration.
\newblock {\em Robotics and Autonomous Systems}, 74(Part A):66--78, 2015.

\bibitem{Liu2012}
Y.~Liu and Y.~Sun.
\newblock Mobile robot instant indoor map building and localization using {2D}
  laser scanning data.
\newblock In {\em 2012 International Conference on System Science and
  Engineering (ICSSE)}, pages 339--344, Coventry, UK, 2012. IEEE.

\bibitem{Manchester2006}
I.~R. Manchester and A.~V. Savkin.
\newblock Circular-navigation-guidance law for precision missile/target
  engagements.
\newblock {\em Journal of Guidance, Control, and Dynamics}, 29(2):314--320,
  2006.

\bibitem{Masehian2015}
E.~Masehian and N.~Mohamadnejad.
\newblock Path planning of nonholonomic flying robots using a new virtual
  obstacle method.
\newblock In {\em 2015 3rd RSI International Conference on Robotics and
  Mechatronics (ICROM)}, pages 612--617, Tehran, Iran, Oct 2015.

\bibitem{Mastellone2008Formation}
S.~Mastellone, D.~M. Stipanovi{\'c}, C.~R. Graunke, K.~A. Intlekofer, and M.~W.
  Spong.
\newblock Formation control and collision avoidance for multi-agent
  non-holonomic systems: Theory and experiments.
\newblock {\em The International Journal of Robotics Research}, 27(1):107--126,
  2008.

\bibitem{Matveev2015}
A.~S. Matveev, M.~Hoy, and A.~V. Savkin.
\newblock A globally converging algorithm for reactive robot navigation among
  moving and deforming obstacles.
\newblock {\em Automatica}, 54:292--304, 2015.

\bibitem{MATVEEV2013312problem}
A.~S. Matveev, M.~C. Hoy, and A.~V. Savkin.
\newblock The problem of boundary following by a unicycle-like robot with
  rigidly mounted sensors.
\newblock {\em Robotics and Autonomous Systems}, 61(3):312--327, 2013.

\bibitem{Matveev2017}
A.~S. Matveev, K.~S. Ovchinnikov, and A.~V. Savkin.
\newblock A method of reactive {3D} navigation for a tight surface scan by a
  nonholonomic mobile robot.
\newblock {\em Automatica}, 75:119--126, 2017.

\bibitem{Matveev2000Estimation}
A.~S. Matveev and A.~V. Savkin.
\newblock {\em Qualitative Theory of Hybrid Dynamical Systems}.
\newblock Birkh{\"a}user, Boston, 2000.

\bibitem{1193753}
A.~S. Matveev and A.~V. Savkin.
\newblock The problem of state estimation via asynchronous communication
  channels with irregular transmission times.
\newblock {\em IEEE Transactions on Automatic Control}, 48(4):670--676, April
  2003.

\bibitem{MATVEEV200451problem}
A.~S. Matveev and A.~V. Savkin.
\newblock The problem of lqg optimal control via a limited capacity
  communication channel.
\newblock {\em Systems {\&} Control Letters}, 53(1):51--64, 2004.

\bibitem{Matveev2009Estimation}
A.~S. Matveev and A.~V. Savkin.
\newblock {\em Estimation and Control over Communication Networks}.
\newblock Birkh{\"a}user, Boston, 2009.

\bibitem{Matveev2015a}
A.~S. Matveev, A.~V. Savkin, M.~Hoy, and C.~Wang.
\newblock {\em Safe Robot Navigation among Moving and Steady Obstacles}.
\newblock Elsevier, Oxford, 2015.

\bibitem{Matveev2011a}
A.~S. Matveev, H.~Teimoori, and A.~V. Savkin.
\newblock A method for guidance and control of an autonomous vehicle in
  problems of border patrolling and obstacle avoidance.
\newblock {\em Automatica}, 47(3):515--524, 2011.

\bibitem{Matveev2011b}
A.~S. Matveev, H.~Teimoori, and A.~V. Savkin.
\newblock Navigation of a unicycle-like mobile robot for environmental extremum
  seeking.
\newblock {\em Automatica}, 47(1):85--91, 2011.

\bibitem{Matveev2011}
A.~S. Matveev, H.~Teimoori, and A.~V. Savkin.
\newblock Range-only measurements based target following for wheeled mobile
  robots.
\newblock {\em Automatica}, 47(1):177--184, 2011.

\bibitem{Matveev2012}
A.~S. Matveev, C.~Wang, and A.~V. Savkin.
\newblock Real-time navigation of mobile robots in problems of border
  patrolling and avoiding collisions with moving and deforming obstacles.
\newblock {\em Robotics and Autonomous systems}, 60(6):769--788, 2012.

\bibitem{Mertz2013}
C.~Mertz, L.~E. Navarro-Serment, R.~MacLachlan, P.~Rybski, A.~Steinfeld,
  A.~Supp\'{e}, C.~Urmson, N.~Vandapel, M.~Hebert, C.~Thorpe, D.~Duggins, and
  J.~Gowdy.
\newblock Moving object detection with laser scanners.
\newblock {\em Journal of Field Robotics}, 30(1):17--43, 2013.

\bibitem{4543197}
N.~Michael, M.~M. Zavlanos, V.~Kumar, and G.~J. Pappas.
\newblock Distributed multi-robot task assignment and formation control.
\newblock In {\em 2008 IEEE International Conference on Robotics and
  Automation}, pages 128--133, Pasadena, USA, May 2008.

\bibitem{Mo2015}
H.~Mo and L.~Xu.
\newblock Research of biogeography particle swarm optimization for robot path
  planning.
\newblock {\em Neurocomputing}, 148:91--99, 2015.

\bibitem{Montiel2015}
O.~Montiel, U.~Orozco-Rosas, and R.~Sep{\'u}lveda.
\newblock Path planning for mobile robots using bacterial potential field for
  avoiding static and dynamic obstacles.
\newblock {\em Expert Systems with Applications}, 42(12):5177--5191, 2015.

\bibitem{Murray2000Using}
D.~Murray and J.~J. Little.
\newblock Using real-time stereo vision for mobile robot navigation.
\newblock {\em Autonomous Robots}, 8(2):161--171, Apr 2000.

\bibitem{Nex2014UAV}
F.~Nex and F.~Remondino.
\newblock {UAV} for {3D} mapping applications: a review.
\newblock {\em Applied Geomatics}, 6(1):1--15, Mar 2014.

\bibitem{Nishitani2015}
I.~Nishitani, T.~Matsumura, M.~Ozawa, A.~Yorozu, and M.~Takahashi.
\newblock Human-centered {X}--{Y}--{T} space path planning for mobile robot in
  dynamic environments.
\newblock {\em Robotics and Autonomous Systems}, 66:18--26, 2015.

\bibitem{Noh2014}
S.~W. Noh, N.~Y. Ko, and J.~H. Han.
\newblock Integrating elementary functions for autonomous navigation of a
  mobile robot.
\newblock In {\em 2014 11th International Conference on Ubiquitous Robots and
  Ambient Intelligence (URAI)}, pages 591--593, Kuala Lumpur, Malaysia, 2014.
  IEEE.

\bibitem{OFlaherty2015}
R.~O'Flaherty and M.~Egerstedt.
\newblock Optimal exploration in unknown environments.
\newblock In {\em 2015 IEEE/RSJ International Conference on Intelligent Robots
  and Systems (IROS)}, pages 5796--5801, Hamburg, Germany, 2015. IEEE.

\bibitem{OH2015424survey}
K.~K. Oh, M.~C. Park, and H.~S. Ahn.
\newblock A survey of multi-agent formation control.
\newblock {\em Automatica}, 53:424--440, 2015.

\bibitem{1605401}
R.~Olfati-Saber.
\newblock Flocking for multi-agent dynamic systems: algorithms and theory.
\newblock {\em IEEE Transactions on Automatic Control}, 51(3):401--420, 2006.

\bibitem{4118472}
R.~Olfati-Saber, J.~A. Fax, and R.~M. Murray.
\newblock Consensus and cooperation in networked multi-agent systems.
\newblock {\em Proceedings of the IEEE}, 95(1):215--233, Jan 2007.

\bibitem{Olivka2016}
P.~Olivka, M.~Mihola, P.~Nov{\'a}k, T.~Kot, and J.~Babjak.
\newblock The design of {3D} laser range finder for robot navigation and
  mapping in industrial environment with point clouds preprocessing.
\newblock In {\em Modelling and Simulation for Autonomous Systems (MESAS
  2016)}, pages 371--383, Rome, Italy, 2016. Springer International Publishing.

\bibitem{Oswald2016}
S.~O{\ss}wald, M.~Bennewitz, W.~Burgard, and C.~Stachniss.
\newblock Speeding-up robot exploration by exploiting background information.
\newblock {\em IEEE Robotics and Automation Letters}, 1(2):716--723, 2016.

\bibitem{Park2016}
J.~Park, H.~Kwak, Y.~Kang, and D.~W. Kim.
\newblock Advanced fuzzy potential field method for mobile robot obstacle
  avoidance.
\newblock {\em Computational Intelligence and Neuroscience}, 2016:1--13, 2016.

\bibitem{1413186}
P.~N. Pathirana, N.~Bulusu, A.~V. Savkin, and S.~Jha.
\newblock Node localization using mobile robots in delay-tolerant sensor
  networks.
\newblock {\em IEEE Transactions on Mobile Computing}, 4(3):285--296, May 2005.

\bibitem{5509803}
Y.~Pei, M.~W. Mutka, and N.~Xi.
\newblock Coordinated multi-robot real-time exploration with connectivity and
  bandwidth awareness.
\newblock In {\em 2010 IEEE International Conference on Robotics and
  Automation}, pages 5460--5465, Anchorage, USA, May 2010.

\bibitem{petersen1999robust}
I.~R. Petersen and A.~V. Savkin.
\newblock {\em Robust Kalman filtering for signals and systems with large
  uncertainties}.
\newblock Birkh{\"a}user, Boston, 1999.

\bibitem{Robust2000Petersen}
I.~R. Petersen, V.~A. Ugrinovskii, and A.~V. Savkin.
\newblock {\em Robust Control Design Using H-$\infty$ Methods}.
\newblock Springer-Verlag, London, 2000.

\bibitem{1241772}
S.~T. Pfister, S.~I. Roumeliotis, and J.~W. Burdick.
\newblock Weighted line fitting algorithms for mobile robot map building and
  efficient data representation.
\newblock In {\em 2003 IEEE International Conference on Robotics and
  Automation}, volume~1, pages 1304--1311, Taipei, Taiwan, Sept 2003.

\bibitem{Prabhu2014}
S.~R.~B. Prabhu, C.~V. Dhasharathi, R.~Prabhakaran, M.~R. Kumar, S.~W. Feroze,
  and S.~Sophia.
\newblock Environmental monitoring and greenhouse control by distributed sensor
  network.
\newblock {\em International Journal of Advanced Networking and Applications},
  5(5):2060--2065, Mar 2014.

\bibitem{Pudics2015}
G.~Pudics, M.~Zsolt Szab{\'o}-Resch, and Z.~V{\'a}mossy.
\newblock Safe robot navigation using an omnidirectional camera.
\newblock In {\em 2015 16th IEEE International Symposium on Computational
  Intelligence and Informatics (CINTI)}, pages 227--231, Hangzhou, China, 2015.
  IEEE.

\bibitem{PUTZ20162123D}
S.~P{\"u}tz, T.~Wiemann, J.~Sprickerhof, and J.~Hertzberg.
\newblock 3d navigation mesh generation for path planning in uneven terrain.
\newblock {\em IFAC-PapersOnLine}, 49(15):212--217, 2016.
\newblock 9th IFAC Symposium on Intelligent Autonomous Vehicles IAV 2016.

\bibitem{RAJA2015295New}
R.~Raja, A.~Dutta, and K.S. Venkatesh.
\newblock New potential field method for rough terrain path planning using
  genetic algorithm for a 6-wheel rover.
\newblock {\em Robotics and Autonomous Systems}, 72:295--306, 2015.

\bibitem{Rao2017}
A.~M. Rao, K.~Ramji, B.~S.~K. Sundara Siva~Rao, V.~Vasu, and C.~Puneeth.
\newblock Navigation of non-holonomic mobile robot using neuro-fuzzy logic with
  integrated safe boundary algorithm.
\newblock {\em International Journal of Automation and Computing},
  14(3):285--294, Jun 2017.

\bibitem{Rastgoo201512}
M.~N. Rastgoo, B.~Nakisa, and M.~Z.~A. Nazri.
\newblock A hybrid of modified {PSO} and local search on a multi-robot search
  system.
\newblock {\em International Journal of Advanced Robotic Systems},
  12(86):1--15, 2015.

\bibitem{Rawat2014}
P.~Rawat, K.~D. Singh, H.~Chaouchi, and J.~M. Bonnin.
\newblock Wireless sensor networks: a survey on recent developments and
  potential synergies.
\newblock {\em The Journal of Supercomputing}, 68(1):1--48, 2014.

\bibitem{Ray2012}
R.~Ray, V.~Kumar, D.~Banerji, and S.~N. Shome.
\newblock Simultaneous localisation and image intensity based occupancy grid
  map building--a new approach.
\newblock In {\em 2012 Third International Conference on Intelligent Systems,
  Modelling and Simulation (ISMS)}, pages 143--148, Kingston, Canada, 2012.
  IEEE.

\bibitem{Rebai2009}
K.~Rebai, A.~Benabderrahmane, O.~Azouaoui, and N.~Ouadah.
\newblock Moving obstacles detection and tracking with laser range finder.
\newblock In {\em 2009 International Conference on Advanced Robotics (ICAR)},
  pages 1--6, Munich, Germany, 2009. IEEE.

\bibitem{Rekleitis2001Multi}
I.~Rekleitis, G.~Dudek, and E.~Milios.
\newblock Multi-robot collaboration for robust exploration.
\newblock {\em Annals of Mathematics and Artificial Intelligence}, 31(1):7--40,
  Oct 2001.

\bibitem{6027049}
H.~Rezaee and F.~Abdollahi.
\newblock Mobile robots cooperative control and obstacle avoidance using
  potential field.
\newblock In {\em 2011 IEEE/ASME International Conference on Advanced
  Intelligent Mechatronics (AIM)}, pages 61--66, Budapest, Hungary, July 2011.

\bibitem{6451251}
H.~Rezaee and F.~Abdollahi.
\newblock A decentralized cooperative control scheme with obstacle avoidance
  for a team of mobile robots.
\newblock {\em IEEE Transactions on Industrial Electronics}, 61(1):347--354,
  Jan 2014.

\bibitem{Roesmann2017}
C.~R{\"o}smann, F.~Hoffmann, and T.~Bertram.
\newblock Integrated online trajectory planning and optimization in distinctive
  topologies.
\newblock {\em Robotics and Autonomous Systems}, 88:142--153, 2017.

\bibitem{Royer2007Monocular}
E.~Royer, M.~Lhuillier, M.~Dhome, and J.~M. Lavest.
\newblock Monocular vision for mobile robot localization and autonomous
  navigation.
\newblock {\em International Journal of Computer Vision}, 74(3):237--260, Sep
  2007.

\bibitem{6461945}
J.~C. L.~Barreto S., A.~G.~S. Concei{\c c}{\~a}o, C.~E.~T. D{\'o}rea,
  L.~Martinez, and E.~R. de~Pieri.
\newblock Design and implementation of model-predictive control with friction
  compensation on an omnidirectional mobile robot.
\newblock {\em IEEE/ASME Transactions on Mechatronics}, 19(2):467--476, April
  2014.

\bibitem{Sauer2001}
C.~T. Sauer, H.~Brugger, E.~P. Hofer, and B.~Tibken.
\newblock Odometry error correction by sensor fusion for autonomous mobile
  robot navigation.
\newblock In {\em Proceedings of the 18th IEEE Instrumentation and Measurement
  Technology Conference (IMTC 2001)}, pages 1654--1658, Budapest, Hungary,
  2001. IEEE.

\bibitem{1304923}
A.~V. Savkin.
\newblock Coordinated collective motion of groups of autonomous mobile robots:
  analysis of vicsek's model.
\newblock {\em IEEE Transactions on Automatic Control}, 49(6):981--982, 2004.

\bibitem{SAVKIN200651Analysis}
A.~V. Savkin.
\newblock Analysis and synthesis of networked control systems: Topological
  entropy, observability, robustness and optimal control.
\newblock {\em Automatica}, 42(1):51--62, 2006.

\bibitem{SAVKIN20061094}
A.~V. Savkin.
\newblock The problem of coordination and consensus achievement in groups of
  autonomous mobile robots with limited communication.
\newblock {\em Nonlinear Analysis: Theory, Methods \& Applications},
  65(5):1094--1102, 2006.

\bibitem{Savkin2002Hybrid}
A.~V. Savkin and R.~J. Evans.
\newblock {\em Hybrid Dynamical Systems. Controller and Sensor Switching
  Problems}.
\newblock Birkh{\"a}user, Boston, 2002.

\bibitem{Savkin2013}
A.~V. Savkin and M.~Hoy.
\newblock Reactive and the shortest path navigation of a wheeled mobile robot
  in cluttered environments.
\newblock {\em Robotica}, 31(02):323--330, 2013.

\bibitem{Savkin2016a}
A.~V. Savkin and H.~Huang.
\newblock The problem of minimum risk path planning for flying robots in
  dangerous environments.
\newblock In {\em 2016 35th Chinese Control Conference (CCC)}, pages
  5404--5408, Chengdu, China, July 2016.

\bibitem{Hang_robio2017_1}
A.~V. Savkin and H.~Li.
\newblock A collision-free area search and {3D} map building algorithm for a
  ground mobile robot in unknown indoor environments.
\newblock In {\em 2017 IEEE International Conference on Robotics and
  Biomimetics (ROBIO)}.
\newblock Submitted.

\bibitem{Hang_uj_1}
A.~V. Savkin and H.~Li.
\newblock Simultaneous collision free area exploration and {3D} map building of
  unknown indoor environments by a ground mobile robot.
\newblock {\em Robotics and Autonomous Systems}.
\newblock Submitted.

\bibitem{Savkin2016}
A.~V. Savkin and H.~Li.
\newblock Collision free navigation of a non-holonomic ground robot for search
  and building maps of unknown areas with obstacles.
\newblock In {\em 2016 35th Chinese Control Conference (CCC)}, pages
  5409--5414, Chengdu, China, July 2016.

\bibitem{savkin_li_2017}
A.~V. Savkin and H.~Li.
\newblock A safe area search and map building algorithm for a wheeled mobile
  robot in complex unknown cluttered environments.
\newblock {\em Robotica}, 2017.
\newblock Advance online publication.

\bibitem{SAVKIN199969Robust}
A.~V. Savkin, E.~Skafidas, and R.~J. Evans.
\newblock Robust output feedback stabilizability via controller switching.
\newblock {\em Automatica}, 35(1):69--74, 1999.

\bibitem{5400224}
A.~V. Savkin and H.~Teimoori.
\newblock Decentralized formation flocking and stabilization for networks of
  unicycles.
\newblock In {\em Proceedings of the 48h IEEE Conference on Decision and
  Control (CDC) held jointly with 2009 28th Chinese Control Conference}, pages
  984--989, Shanghai, China, Dec 2009.

\bibitem{Savkin2010}
A.~V. Savkin and H.~Teimoori.
\newblock Bearings-only guidance of a unicycle-like vehicle following a moving
  target with a smaller minimum turning radius.
\newblock {\em IEEE Transactions on Automatic Control}, 55(10):2390--2395,
  2010.

\bibitem{Savkin2013a}
A.~V. Savkin and C.~Wang.
\newblock A simple biologically inspired algorithm for collision-free
  navigation of a unicycle-like robot in dynamic environments with moving
  obstacles.
\newblock {\em Robotica}, 31(06):993--1001, 2013.

\bibitem{Savkin2014}
A.~V. Savkin and C.~Wang.
\newblock Seeking a path through the crowd: Robot navigation in unknown dynamic
  environments with moving obstacles based on an integrated environment
  representation.
\newblock {\em Robotics and Autonomous Systems}, 62(10):1568--1580, 2014.

\bibitem{6606099}
A.~V. Savkin, C.~Wang, A.~Baranzadeh, Z.~Xi, and H.~T. Nguyen.
\newblock A method for decentralized formation building for unicycle-like
  mobile robots.
\newblock In {\em 2013 9th Asian Control Conference (ASCC)}, pages 1--5,
  Istanbul, Turkey, June 2013.

\bibitem{SAVKIN2016463}
A.~V. Savkin, C.~Wang, A.~Baranzadeh, Z.~Xi, and H.~T. Nguyen.
\newblock Distributed formation building algorithms for groups of wheeled
  mobile robots.
\newblock {\em Robotics and Autonomous Systems}, 75, Part B:463--474, 2016.

\bibitem{SCHMUCK2016230}
P.~Schmuck, S.~A. Scherer, and A.~Zell.
\newblock Hybrid metric-topological {3D} occupancy grid maps for large-scale
  mapping.
\newblock {\em IFAC-PapersOnLine}, 49(15):230--235, 2016.

\bibitem{6058518}
J.~Sfeir, M.~Saad, and H.~Saliah-Hassane.
\newblock An improved artificial potential field approach to real-time mobile
  robot path planning in an unknown environment.
\newblock In {\em 2011 IEEE International Symposium on Robotic and Sensors
  Environments (ROSE)}, pages 208--213, Montreal, Canada, Sept 2011.

\bibitem{SGORBISSA2012628Planning}
A.~Sgorbissa and R.~Zaccaria.
\newblock Planning and obstacle avoidance in mobile robotics.
\newblock {\em Robotics and Autonomous Systems}, 60(4):628--638, 2012.

\bibitem{SHENG2006945Distributed}
W.~Sheng, Q.~Yang, J.~Tan, and N.~Xi.
\newblock Distributed multi-robot coordination in area exploration.
\newblock {\em Robotics and Autonomous Systems}, 54(12):945--955, 2006.

\bibitem{7866323}
X.~Shi, G.~Junyao, Y.~Liu, J.~Zhao, F.~Zhao, H.~Cao, and C.~Liu.
\newblock Low-cost map building and obstacle avoidance of miniature
  reconnaissance robot.
\newblock In {\em 2016 IEEE International Conference on Robotics and
  Biomimetics (ROBIO)}, pages 208--213, Qingdao, China, Dec 2016.

\bibitem{1570542}
S.~Shimoda, Y.~Kuroda, and K.~Iagnemma.
\newblock Potential field navigation of high speed unmanned ground vehicles on
  uneven terrain.
\newblock In {\em Proceedings of the 2005 IEEE International Conference on
  Robotics and Automation}, pages 2828--2833, Barcelona, Spain, April 2005.

\bibitem{6630630}
C.~Siagian, C.~K. Chang, and L.~Itti.
\newblock Mobile robot navigation system in outdoor pedestrian environment
  using vision-based road recognition.
\newblock In {\em 2013 IEEE International Conference on Robotics and
  Automation}, pages 564--571, Karlsruhe, Germany, 2013.

\bibitem{SINGH2016156Feasible}
A.~K. Singh and K.~M. Krishna.
\newblock Feasible acceleration count: A novel dynamic stability metric and its
  use in incremental motion planning on uneven terrain.
\newblock {\em Robotics and Autonomous Systems}, 79:156--171, 2016.

\bibitem{SKAFIDAS1999553Stability}
E.~Skafidas, R.~J. Evans, A.~V. Savkin, and I.~R. Petersen.
\newblock Stability results for switched controller systems.
\newblock {\em Automatica}, 35(4):553--564, 1999.

\bibitem{Sun2010}
X.~Sun, X.~Zhao, E.~Li, H.~Li, and Z.~Liang.
\newblock Mobile robot navigation using {RSSI} potential field in wireless
  sensor network.
\newblock {\em Journal of Computational Information Systems}, 6(14):4751--4759,
  2010.

\bibitem{SURMANN2003181autonomous}
H.~Surmann, A.~N{\"u}chter, and J.~Hertzberg.
\newblock An autonomous mobile robot with a {3D} laser range finder for {3D}
  exploration and digitalization of indoor environments.
\newblock {\em Robotics and Autonomous Systems}, 45(3):181--198, 2003.

\bibitem{6122042}
N.~K. Suryadevara and S.~C. Mukhopadhyay.
\newblock Wireless sensor network based home monitoring system for wellness
  determination of elderly.
\newblock {\em IEEE Sensors Journal}, 12(6):1965--1972, June 2012.

\bibitem{5565069}
L.~Tang, S.~Dian, G.~Gu, K.~Zhou, S.~Wang, and X.~Feng.
\newblock A novel potential field method for obstacle avoidance and path
  planning of mobile robot.
\newblock In {\em 2010 3rd International Conference on Computer Science and
  Information Technology}, volume~9, pages 633--637, Chengdu, China, 2010.

\bibitem{Tatsukawa2015}
S.~Tatsukawa, T.~Nakanishi, R.~Nagao, T.~Wada, M.~Fujimoto, and K.~Mutsuura.
\newblock New moving control of mobile robot without collision with wall and
  obstacles by passive {RFID} system.
\newblock In {\em 2015 International Conference on Indoor Positioning and
  Indoor Navigation (IPIN)}, pages 1--7, Calcary, Canada, 2015. IEEE.

\bibitem{Teimoori2010}
H.~Teimoori and A.~V. Savkin.
\newblock A biologically inspired method for robot navigation in a cluttered
  environment.
\newblock {\em Robotica}, 28(05):637--648, 2010.

\bibitem{Teimoori2010a}
H.~Teimoori and A.~V. Savkin.
\newblock Equiangular navigation and guidance of a wheeled mobile robot based
  on range-only measurements.
\newblock {\em Robotics and Autonomous Systems}, 58(2):203--215, 2010.

\bibitem{Tian201688}
F.~Tian, K.~M. Chao, Z.~Feng, K.~Xing, and N.~Shah.
\newblock Using a wireless visual sensor network to harmonically navigate
  multiple low-cost wheelchairs in an indoor environment.
\newblock {\em Journal of Network and Computer Applications}, 62:88--99, 2016.

\bibitem{7125610}
J.~Ure{\~n}a, D.~Gualda, {\'A}.~Hern{\'a}ndez, E.~Garc{\'i}a, J.~M.
  Villadangos, M.~C. P{\'e}rez, J.~C. Garc{\'i}a, J.~J. Garc{\'i}a, and
  A.~Jim{\'e}nez.
\newblock Ultrasonic local positioning system for mobile robot navigation: From
  low to high level processing.
\newblock In {\em 2015 IEEE International Conference on Industrial Technology
  (ICIT)}, pages 3440--3445, Seville, Spain, 2015.

\bibitem{Vallve2015}
J.~Vallv{\'e} and J.~Andrade-Cetto.
\newblock Potential information fields for mobile robot exploration.
\newblock {\em Robotics and Autonomous Systems}, 69:68--79, 2015.

\bibitem{Vilca2015}
J.~Vilca, L.~Adouane, and Y.~Mezouar.
\newblock A novel safe and flexible control strategy based on target reaching
  for the navigation of urban vehicles.
\newblock {\em Robotics and Autonomous Systems}, 70:215--226, 2015.

\bibitem{Wang2016a}
C.~Wang, A.~V. Savkin, and M.~Garratt.
\newblock Collision free navigation of flying robots among moving obstacles.
\newblock In {\em 2016 35th Chinese Control Conference (CCC)}, pages
  4545--4549, Chengdu, China, 2016.

\bibitem{Wang2015a}
D.~Wang, Y.~Duan, and J.~Wang.
\newblock Environment exploration and map building of mobile robot in unknown
  environment.
\newblock {\em International Journal of Simulation and Process Modelling},
  10(3):241--252, 2015.

\bibitem{Wang2013}
M.~Wang, W.~Wang, J.~Xiong, and L.~Yan.
\newblock A consistent map building method based on surf loop closure
  detection.
\newblock In {\em 2013 IEEE 3rd Annual International Conference on Cyber
  Technology in Automation, Control and Intelligent Systems (CYBER)}, pages
  92--95, Nanjing, China, 2013. IEEE.

\bibitem{WANG2017Distributed}
Q.~Wang, Z.~Chen, P.~Liu, and Q.~Hua.
\newblock Distributed multi-robot formation control in switching networks.
\newblock {\em Neurocomputing}, 2017.
\newblock Advance online publication.

\bibitem{WANG2013858}
W.~T. Wang and K.~F. Ssu.
\newblock Obstacle detection and estimation in wireless sensor networks.
\newblock {\em Computer Networks}, 57(4):858--868, 2013.

\bibitem{WANG2014137}
Y.~Wang, T.~Mai, and J.~Mao.
\newblock Adaptive motion/force control strategy for non-holonomic mobile
  manipulator robot using recurrent fuzzy wavelet neural networks.
\newblock {\em Engineering Applications of Artificial Intelligence},
  34:137--153, 2014.

\bibitem{Wattanavekin2013}
T.~Wattanavekin, T.~Ogata, T.~Hara, and J.~Ota.
\newblock Mobile robot exploration by using environmental boundary information.
\newblock {\em ISRN Robotics}, 2013:1--11, 2013.

\bibitem{6596518}
G.~Wen, Z.~Duan, G.~Chen, and W.~Yu.
\newblock Consensus tracking of multi-agent systems with lipschitz-type node
  dynamics and switching topologies.
\newblock {\em IEEE Transactions on Circuits and Systems I: Regular Papers},
  61(2):499--511, 2014.

\bibitem{Wu2007Consensus}
Z.~Wu, Z.~Guan, X.~Wu, and T.~Li.
\newblock Consensus based formation control and trajectory tracing of
  multi-agent robot systems.
\newblock {\em Journal of Intelligent and Robotic Systems}, 48(3):397--410, Mar
  2007.

\bibitem{XIAO20092605}
F.~Xiao, L.~Wang, J.~Chen, and Y.~Gao.
\newblock Finite-time formation control for multi-agent systems.
\newblock {\em Automatica}, 45(11):2605--2611, 2009.

\bibitem{Xie2014}
S.~Xie and Y.~Wang.
\newblock Construction of tree network with limited delivery latency in
  homogeneous wireless sensor networks.
\newblock {\em Wireless Personal Communications}, 78(1):231--246, Sep 2014.

\bibitem{Yang2016}
H.~Yang, X.~Fan, P.~Shi, and C.~Hua.
\newblock Nonlinear control for tracking and obstacle avoidance of a wheeled
  mobile robot with nonholonomic constraint.
\newblock {\em IEEE Transactions on Control Systems Technology},
  24(2):741--746, March 2016.

\bibitem{YANG2016A}
J.~Yang, Y.~Shi, and H.~Rong.
\newblock Random neural {Q}-learning for obstacle avoidance of a mobile robot
  in unknown environments.
\newblock {\em Advances in Mechanical Engineering}, 8(7):1--15, 2016.

\bibitem{Yao2015}
Y.~Yao, Q.~Ni, Q.~Lv, and K.~Huang.
\newblock A novel heterogeneous feature ant colony optimization and its
  application on robot path planning.
\newblock In {\em 2015 IEEE Congress on Evolutionary Computation (CEC)}, pages
  522--528, Sendai, Japan, 2015. IEEE.

\bibitem{5766050}
Z.~Yao and K.~Gupta.
\newblock Distributed roadmaps for robot navigation in sensor networks.
\newblock {\em IEEE Transactions on Robotics}, 27(5):997--1004, Oct 2011.

\bibitem{Yokoyama2005}
M.~Yokoyama and T.~Poggio.
\newblock A contour-based moving object detection and tracking.
\newblock In {\em 2005 IEEE International Workshop on Visual Surveillance and
  Performance Evaluation of Tracking and Surveillance}, pages 271--276,
  Beijing, China, 2005. IEEE.

\bibitem{Yorozu2016}
A.~Yorozu and M.~Takahashi.
\newblock Obstacle avoidance with translational and efficient rotational motion
  control considering movable gaps and footprint for autonomous mobile robot.
\newblock {\em International Journal of Control, Automation and Systems},
  14(5):1352--1364, Oct 2016.

\bibitem{Yuan2014}
W.~Yuan, Z.~Cao, M.~Tan, and J.~Liu.
\newblock Topological mapping and navigation based on visual sensor network.
\newblock In {\em 2014 IEEE International Conference on Mechatronics and
  Automation}, pages 1686--1690, Tianjin, China, 2014.

\bibitem{savkin_2015Distributed}
A.~Zakhar'eva, A.~S. Matveev, M.~C. Hoy, and A.~V. Savkin.
\newblock Distributed control of multiple non-holonomic robots with sector
  vision and range-only measurements for target capturing with collision
  avoidance.
\newblock {\em Robotica}, 33(2):385--412, 2015.

\bibitem{ZhangY2013}
Y.~Zhang, S.~Wang, G.~Fan, and J.~Zhou.
\newblock Navigation for indoor mobile robot based on wireless sensor network.
\newblock In {\em Wireless Algorithms, Systems, and Applications: 8th
  International Conference (WASA 2013)}, pages 325--336, Zhangjiajie, China,
  Aug 2013. Springer Berlin Heidelberg.

\bibitem{ZhengZhang2013}
Z.~Zhang, Z.~Li, D.~Zhang, and J.~Chen.
\newblock Path planning and navigation for mobile robots in a hybrid sensor
  network without prior location information.
\newblock {\em International Journal of Advanced Robotic Systems}, 10(3):1--12,
  2013.

\bibitem{6775785}
N.~Zhou, X.~Zhao, and M.~Tan.
\newblock {RSSI}-based mobile robot navigation in grid-pattern wireless sensor
  network.
\newblock In {\em 2013 Chinese Automation Congress}, pages 497--501, Changsha,
  China, Nov 2013.

\bibitem{6885782}
Z.~Ziaei, R.~Oftadeh, and J.~Mattila.
\newblock Global path planning with obstacle avoidance for omnidirectional
  mobile robot using overhead camera.
\newblock In {\em 2014 IEEE International Conference on Mechatronics and
  Automation}, pages 697--704, Tianjin, China, Aug 2014.

\end{thebibliography}

\end{document}